# A comparative study of conformal prediction methods for valid uncertainty quantification in machine learning

Thesis submitted in partial fulfilment of the requirements for the degree of
*Doctor (Ph.D.) of Bioscience Engineering: Mathematical Modelling*

by

## Nicolas Dewolf

Department of Data Analysis and Mathematical Modelling

Faculty of Bioscience Engineering

Ghent University

Academic year 2023 – 2024

*Dutch translation of the title:*
Een vergelijkende studie van conformevoorspellingsmethoden voor geldige onzekerheidsquantificatie in machine learning.

*Acknowledgement of the financing institution:*
This research received funding from the Flemish Government under the "Onderzoeksprogramma Artificiële Intelligentie (AI) Vlaanderen" programme.

*Please refer to this work in the following way:*
N. Dewolf (**2024**). A comparative study of conformal prediction methods for valid uncertainty quantification in machine learning, Ph.D. Thesis, Faculty of Bioscience Engineering, Ghent University, Ghent, Belgium.

# Summary

In the past decades, most work in the area of data analysis and machine learning was focused on optimizing predictive models and getting better results than what was possible with existing models. To what extent the metrics with which such improvements were measured were accurately capturing the intended goal, whether the numerical differences in the resulting values were significant, or whether uncertainty played a role in this study and if it should have been taken into account, was of secondary importance. Whereas probability theory, be it frequentist or Bayesian, used to be the gold standard in science before the advent of the supercomputer, it was quickly replaced in favor of black box models and sheer computing power because of their ability to handle large data sets. This evolution sadly happened at the expense of interpretability and trustworthiness.[1] However, while people are still trying to improve the predictive power of their models, the community is starting to realize that for many applications it is not so much the exact prediction that is of importance, but rather the variability or uncertainty.

The work in this dissertation tries to further the quest for a world where everyone is aware of uncertainty, of how important it is and how to embrace it instead of fearing it. A specific, though general, framework that allows anyone to obtain accurate uncertainty estimates is singled out and analysed. Certain aspects and applications of the framework — dubbed 'conformal prediction' — are studied in detail. Whereas many approaches to uncertainty quantification make strong assumptions about the data, conformal prediction is, at the time of writing, the only framework that deserves the title 'distribution-free'. No parametric assumptions have to be made and the nonparametric results also hold without having to resort to the law of large numbers in the asymptotic regime.

After introducing the concept of uncertainty (estimation) and the general framework of interest, the specific problem of (univariate) regression is considered. Already in this basic setting some interesting features can be analysed. Building on these results, the problem of conditional uncertainty estimation is tackled next. Whereas the former problem is situated at a global or marginal level, where all features are considered equally important, the latter focuses on specific subsets of the data. The third and last part of this work is concerned with a middle way, where achieving conditional guarantees is balanced with using as much of the available data as possible. This will also be the ideal setting to treat the problems of extreme classification and multitarget prediction.

---

[1] Some brave people such as Pearl fought for retaining probabilistic approaches, giving birth to e.g. *Bayesian networks*, but even today, these researchers form a small group compared to the deep learning community.

# Samenvatting

Gedurende de afgelopen decennia was het leeuwendeel van het onderzoek binnen de data-analyse en machine learning gefocust op het verbeteren van predictieve modellen en het bekomen van betere voorspellingen dan wat mogelijk was met bestaande modellen. Of de metrieken die hiervoor gebruikt werden representatief waren voor de beoogde doelstellingen, de numerieke verschillen in de bekomen resultaten significant waren, of onzekerheid in beschouwing genomen moest worden, was van beperkt belang. Waar kansrekening, zij het frequentistisch of Bayesiaans, de gouden standaard vormde voor de wetenschap vóór de komst van de supercomputer, werd deze snel vervangen door zwarte dozen en pure rekenkracht om de alsmaar groter wordende datasets aan te kunnen. Deze evolutie ging echter ten koste van de interpreteerbaarheid en de betrouwbaarheid.[2] Gelukkig begint men te beseffen dat, ook al blijft het verbeteren van voorspellingen een belangrijke drijfveer, voor veel toepassingen het niet zozeer de exacte voorspelling is die er toe doet, maar eerder de onzekerheid of variabiliteit.

Het werk in dit proefschrift probeert iets bij te brengen aan de strijd voor een wereld waarin iedereen zich bewust is van de onzekerheid in data, hoe belangrijk het is om deze correct in te schatten en hoe deze te leren gebruiken in plaats van er bang van te zijn. Een specifiek, maar breed toepasbaar raamwerk voor het bekomen van nauwkeurige onzekerheidsschattingen wordt uitgelicht en geanalyseerd. Enkele aspecten en toepassingen van dit framework — dat de naam 'conform voorspellen' gekregen heeft — worden in detail bestudeerd. Waar veel andere paradigma's voor het schatten van onzekerheid sterke veronderstellingen over de data maken, is conform voorspellen voorlopig de enige kandidaat die de titel 'assumptievrij' verdient. Geen enkele parametrische veronderstelling is vereist en de niet-parametrische resultaten kunnen bekomen worden zonder gebruik te maken van asymptotische stellingen zoals de wet van de grote aantallen.

---

[2] Enkele moedige zielen zoals Pearl hebben gevochten voor het behouden van probabilistische technieken, zoals *Bayesiaanse netwerken*, maar zelfs tot op heden vormen zij een kleine groep in vergelijking met de machine-learninggemeenschap.

Nadat de algemene theorie van onzekerheid en conform voorspellen geïntroduceerd is, wordt een van de prototypische probleemstellingen uit de statistiek beschouwd, dat van (univariate) regressie. Zelfs in deze doorsneesituatie kunnen reeds enkele interessante eigenschappen bestudeerd worden. Hierop verderbouwend wordt vervolgens het probleem van conditionele onzekerheidsschatting aangepakt. Waar de eerste situatie eerder op een globaal niveau plaatsvindt, waarbij alle aspecten van de data evenwaardig geacht worden, ligt de focus bij het tweede probleem eerder op specifieke deelverzamelingen van de data. In het derde luik van dit werk wordt een mogelijke middenweg onderzocht, waarbij conditionele garanties worden afgewogen tegen het gebruik van zo veel mogelijk beschikbare data. Dit zal ook de ideale kans bieden om extreme classificatieproblemen en problemen met meerdere variabelen te behandelen.

# Acknowledgements

It should go without mention that many people contributed to this dissertation and to my journey as a PhD student, and that I received a lot of support from both friends, family and colleagues.

First and foremost, I want to express my sincere gratitude to my supervisors Willem Waegeman and Bernard De Baets. During the past four years they were always willing to help, made ample time for meetings and always came up with ideas to further improve our research. Given that my roots lie in a completely different field of science, their help was also indispensable in learning to navigate the waters of machine learning and data science. I would also like to thank the members of the examination board — Ivo Couckuyt, Oliver Dukes, Sébastien Destercke and Stijn Luca — for the interesting discussions and useful suggestions which surely improved the final version of this thesis and, not in the least, improved my understanding of the subject and how it relates to other areas of (data) science.

I also want to thank Pieter, a fellow physics graduate, for introducing me to the department. After working on a PhD at the physics department on a temporary contract, I was looking for a new position. Pieter, whom I studied with for five years, informed me that there was a position available at the department. After some positive meetings with my current supervisors, I embarked on this new adventure. Funny enough, a year later I took over Pieter's role and introduced yet another fellow physics graduate, Bastiaan, to the department.

This opens up the opportunity for thanking the colleagues in my office. First of all Thomas, whom I have shared an office with for almost the entire length of my PhD (he sadly moved to another office just before I left). His knowledge and enthusiasm were indispensable. Another indispensable colleague would be Joris, one of the key members of the department, not only for his wisdom, but also for his kindness and generosity (and impeccable taste of music)! As mentioned, my apprentice Bastiaan, whom I introduced to the wonderful world of computational statistics, also had the opportunity to share the office with me ever since he joined. During the past four years, we also welcomed Jelle and Max, and the office would not have been the same without them. Every day started with the mystery of what the composition of our team would be. My gratitude of course extends to all the colleagues in the research unit and the department as a whole. The discussion groups, the drinks at Hal 16, the occasional indoor football match, the summer school in Italy. None of it would have been the same without them. Besides the social events, I also want to thank all the colleagues for the academic support and inspiration. I especially want to thank Thomas and Dimitrios, without whom Chapter 5 would not have achieved its final form.

During the final year of my PhD something happened for which my former Master thesis supervisor, Frank Verstraete of the UGent Quantum Group, and, by extension, the entire research unit also deserve a word of appreciation. As mentioned above, I had a very short-lived PhD experience at the physics department. There, my research consisted of extending the work in my Master thesis. After leaving the group, I could not suppress the feeling that my research was unsatisfactory (sadly a feeling that seems to follow me in whatever I do). However, in 2022, the group informed me that they were writing a paper about this research and in early 2023 I became a coauthor of said paper, thereby fulfilling one of my academic dreams: officially becoming part of the physics society.

At last, I would like to thank the people closest to me: my family, my friends and my girlfriend. My family has always been there for me and has always supported me in everything I do, and they will, without a doubt, continue to do so. Their support and motivation has brought me to where I am now. The activities with my friends and girlfriend, whether they were movie nights, boulder sessions, classy jazz nights or barbecues in the wild, were also a welcome and necessary diversion. I hope that I will have more time for them after the completion of my PhD.

# Symbols & Abbreviations

The following symbols and abbreviations are used throughout this dissertation:

**Abbreviations**

| | |
|---|---|
| a.e. | almost everywhere |
| APS | adaptive prediction sets |
| a.s. | almost surely |
| CCP | cross-conformal prediction |
| CDF | cumulative distribution function |
| CPS | conformal predictive system |
| CTM | conformal test martingale |
| ELBO | evidence lower bound |
| GP | Gaussian process |
| HQ | high quality |
| HSM | hierarchical softmax |
| ICP | inductive conformal prediction |
| KL | Kullback–Leibler |
| KS | Kolmogorov–Smirnov |
| LUBE | lower-upper bound estimation |
| MAE | mean absolute error |
| MC | Monte Carlo |
| MCP | Mondrian conformal prediction |
| MPIW | mean prediction interval width |
| MSE | mean squared error |

| MTP  | multitarget prediction |
| MVE  | mean-variance ensemble |
| MV   | mean-variance estimator |
| NCP  | normalized conformal prediction |
| NF   | normalizing flow |
| NN   | neural network |
| OOB  | out-of-bag |
| OVR  | one-versus-rest classifier |
| PDF  | probability density function |
| PI   | prediction interval |
| PLT  | probabilistic label tree |
| QRF  | quantile regression forest |
| QR   | quantile regressor |
| RAPS | regularized adaptive prediction sets |
| RBF  | radial basis function |
| RF   | random forest |
| RMSE | root-mean-square error |
| TCP  | transductive conformal prediction |
| TV   | total variation |
| VI   | variational inference |

**Symbols**

| $a^*$ | critical nonconformity score |
| $\mathcal{C}(\Gamma^\alpha, P)$ | coverage of a confidence predictor $\Gamma^\alpha$ with respect to a distribution $P$ |
| $\mathrm{Conv}(S)$ | convex hull of a set $S$ |
| $\mathcal{D}$ | generic data set |
| $\Delta^k$ | $k$-simplex |
| $\overset{d}{=}$ | equality in distribution |
| $g \triangleright x$ | group element $g$ acting on an element $x$ |
| $\{x_i\}_{i \in \mathfrak{I}}$ | (indexed) set with index set $\mathfrak{I}$, equivalent to the set-builder notation $\{x_i \mid i \in \mathfrak{I}\}$ |
| $\mathrm{Met}(\mathcal{X})$ | set of all metrics on a set $\mathcal{X}$ |
| $\mathbb{N}$ | set of natural numbers |
| $\mathbb{N}_0$ | set of natural numbers excluding $0$ |
| $\mathcal{N}(\mu, \sigma^2)$ | normal distribution with mean $\mu$ and standard deviation $\sigma$ |
| $2^{\mathcal{X}}$ | power set of a set $\mathcal{X}$ |
| $\mathcal{P}_2(\mathcal{X})$ | set of probability distributions characterized by the mean and variance |
| $\mathbb{P}_2(\mathcal{X})$ | set of probability distributions with finite variance |
| $\mathbb{R}$ | real line |
| $\overline{\mathbb{R}}$ | extended real line, i.e. $\mathbb{R} \cup \{-\infty, \infty\}$ |
| $\mathcal{T}$ | generic training set |
| $\mathcal{U}(S)$ | uniform distribution on a set $S$ |
| $A \cup B$ | union of the sets $A$ and $B$, where the context will tell if it is a disjoint union or not |
| $\mathcal{V}$ | generic calibration (or validation) set |
| $\mathcal{W}(\Gamma^\alpha, P)$ | (in)efficiency of a confidence predictor $\Gamma^\alpha$ with respect to a distribution $P$ |
| $x_{(k)}$ | $k^{\text{th}}$ element in the sorted version of a set $\{x_i\}_{i \in \mathfrak{I}}$ |
| $\mathcal{X}^*$ | set of all multisubsets of a set $\mathcal{X}$ |
| $x, y, \ldots$ | elements of a set / instantiations of random variables |
| $(x^*, y^*)$ | data point corresponding to the critical nonconformity score |
| $\mathcal{X}, \mathcal{Y}, \ldots$ | sets or spaces ($\mathcal{X}$ usually denotes the feature space and $\mathcal{Y}$ the target space) |
| $X, Y, \ldots$ | random variables |

# Reading Guide

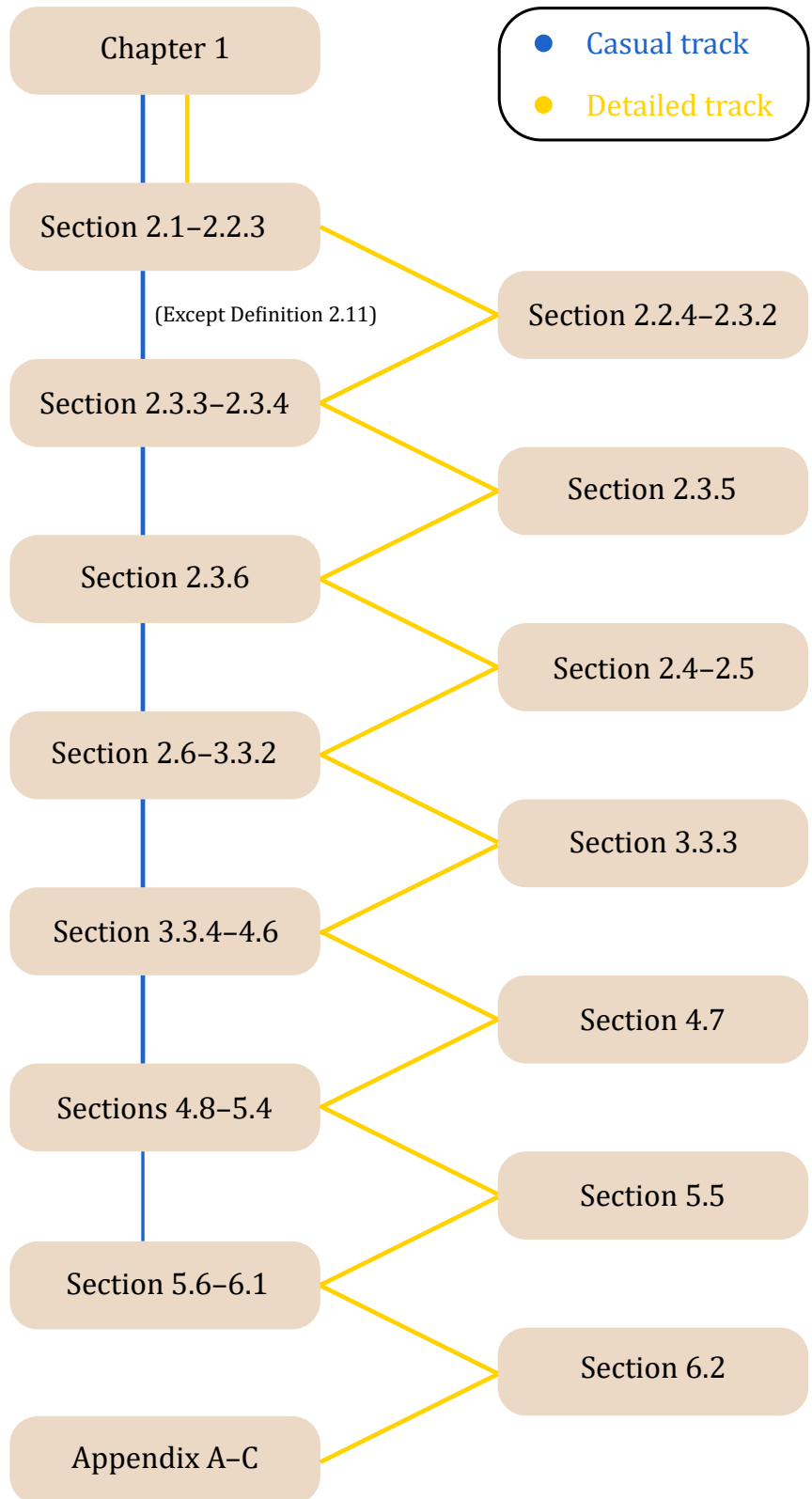

Chapter 1

Casual track
Detailed track

Section 2.1–2.2.3

(Except Definition 2.11)

Section 2.2.4–2.3.2

Section 2.3.3–2.3.4

Section 2.3.5

Section 2.3.6

Section 2.4–2.5

Section 2.6–3.3.2

Section 3.3.3

Section 3.3.4–4.6

Section 4.7

Sections 4.8–5.4

Section 5.5

Section 5.6–6.1

Section 6.2

Appendix A–C

# TABLE OF CONTENTS

# INTRODUCTION 1

## 1.1   Machine learning & Uncertainty

Although the work in this dissertation is not in itself about machine learning, since, as will be made clear throughout the text, the techniques can be applied to any predictive model, the eventual field of application will be very much that of machine learning and artificial intelligence. As such, a short review is in order. Most of this material can be found in e.g. Goodfellow, Bengio, and Courville (2016).

Ever since the dawn of humanity, people have tried to create things themselves. Whether it were works of art, practical inventions or scientific tools, the quest for creation has always been a fierce one. However, never have we — humanity as a whole — ever been so close to creating[1] true intelligence as in the current era of big data and machine learning. In the early days of this era (the second half of last century), this mostly involved 'teaching' machines or computers to apply formal rules in cases where the computational complexity of the problem transcends human capabilities. When the rules are, however, more difficult to formalize — these are called 'intuitive' rules in Goodfellow et al. (2016) — or, even worse, when human experts themselves do not know the rules, the situation changes. This is where the community finds itself right now, in the era of black-box models (although the rise of buzzwords such as 'interpretability', 'explainability' and 'trustworthiness' might indicate that this era is coming to an end and that white box models might reclaim their throne as the true vessels of scientific power).

Most people nowadays agree that the machine learning movement began around the middle of last century with the work of McCulloch and Pitts (1943) and Rosenblatt (1958). Based on a basic understanding of the biological structure of the (human) brain, the foundational structure behind neural networks was introduced (see Fig. 1.1). The true breakthrough, aside from the increase in computational power coming from advances in engineering, came from the introduction of the backpropagation algorithm (Rumelhart, Hinton, & Williams, 1986). This algorithm, which is basically an application of the chain rule from calculus, allows to efficiently update the weights of a (differentiable) model (such as the multilayer perceptron or neural network in Fig. 1.2) with a gradient descent optimization scheme.

---

[1]   Evolution and procreation are left aside here.

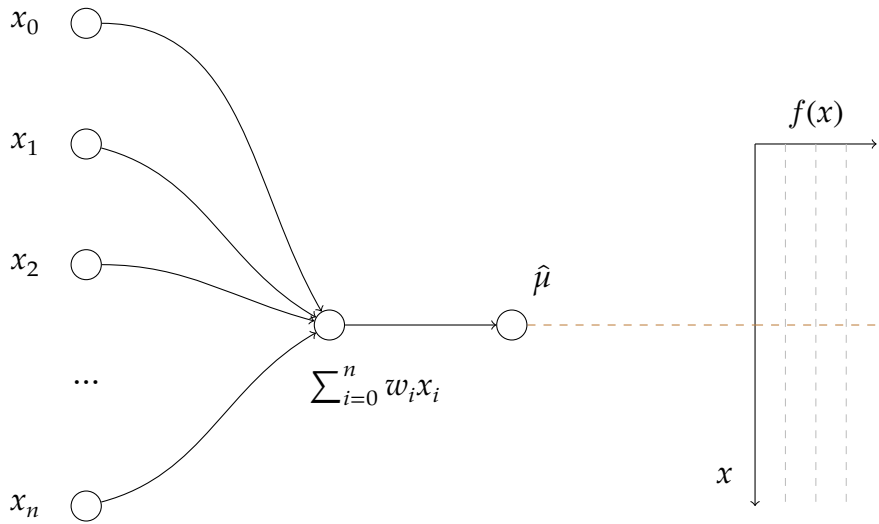

Figure 1.1: Schematic view of the McCullogh–Pitts neuron, the basic building block of modern neural networks. This is essentially a one-layer neural network with a nonlinear activation at the end (a *Heaviside step function* in the original version).

What all these predictive models have in common is that they give approximations

$$\hat{f} : \mathcal{X} \to \mathcal{Y} \tag{1.1}$$

to a 'ground truth' function

$$f : \mathcal{X} \to \mathcal{Y}. \tag{1.2}$$

If this ground truth function had been deterministic, as in the case of Figs. 1.1 and 1.2, the problem would be solved by standard results from approximation theory through various universal approximation theorems (Cybenko, 1989; Hornik, Stinchcombe, & White, 1989; Z. Lu, Pu, Wang, Hu, & Wang, 2017). However, in practice, the ground truth is obfuscated by stochastic effects coming from the fact that the quantities or observables that we can actually measure generally only constitute an effective theory. The macroscopic degrees of freedom — those that we can measure — are obtained by averaging out the microscopic ones — those that actually determine nature. By only having access to this effective set of degrees of freedom, be it due to a lack of resolution or by fundamental laws of nature (quantum mechanics enters the stage), we are forced to work with stochastic functions[2]

$$\widehat{f} : \mathcal{X} \to \mathbb{P}(\mathcal{Y}) \,, \tag{1.3}$$

where $\mathbb{P}(\mathcal{X})$ denotes the set of probability distributions on $\mathcal{X}$ (cf. Definition A.14). Even if we repeatedly measure the output of the function with seemingly identical inputs, the result might differ. An illustrative example is the theory of statistical mechanics as introduced by Boltzmann. Here, a large amount of particles is considered (about the same order of magnitude as Avogadro's number, i.e. $\sim 6.02 \times 10^{23}$). Since it is impossible to measure the position and velocity of every single particle, a probabilistic model has to be considered and only statistics such as the average velocity make sense as macroscopic quantities. An example of how this can be approximated is shown in Fig. 1.3, where the model outputs not only a single value, but directly predicts the mean $\hat{\mu}$ and standard deviation $\hat{\sigma}$ of a distribution (this approach is explained in more detail in Section 3.2.2).

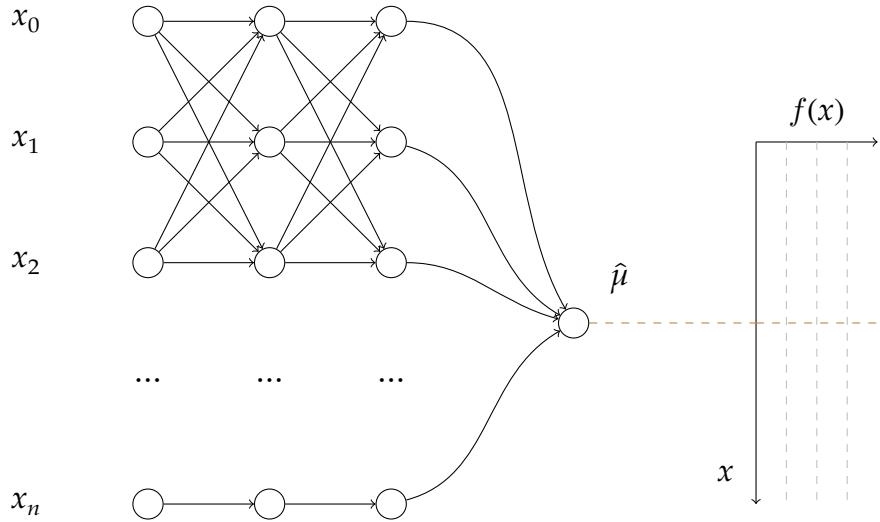

Figure 1.2: Schematic view of a basic neural network. By composing many simple functions, a more complex transformation can be obtained.

Characterizing how much these stochastic functions differ from ordinary functions or, in the language of probability theory, how much the probability distributions differ from point masses, will essentially be the content of this dissertation. To achieve this goal, we will, however, not consider the usual probabilistic frameworks. Instead, we will use what might look like a more primitive or rudimentary set-prediction framework. It will turn out that this could not be further from the truth, since conformal prediction will turn out to be (almost entirely) distribution-free!

---

[2] Here modelled as a Markov kernel for simplicity (Definition A.38).

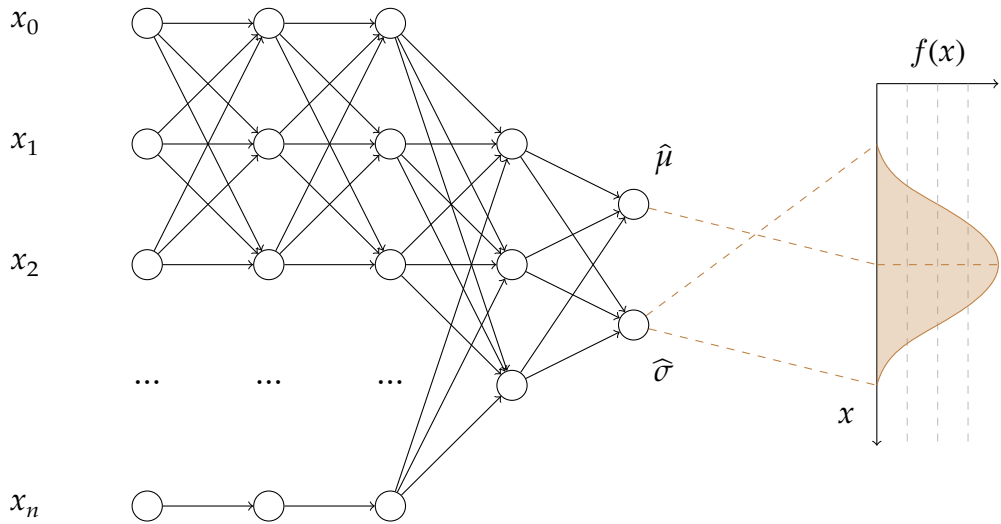

Figure 1.3: Schematic view of a neural network that predicts both the mean and variance of a probability distribution (see Section 3.2.2).

## 1.2 Personal history

My story in statistics and probability theory started quite a while ago. After high school, I enrolled for a Bachelor and Master degree in Physics and Astronomy also at Ghent University. Like most physicists, I got acquainted with these subjects early on during courses such as experimental physics, statistical and computational physics, and quantum mechanics. All of these subjects apply the same fundamental ideas and techniques that are also being used in modern-day statistics and data analysis. The true start of my adventures in the land of (theoretical) statistics, however, started during a trip organized by the physics student society to Oxford in 2016. There, we visited the famous Blackwell bookstore and, at that point not yet realizing the financial and mental consequences, I started my collection of books on mathematics. I left the store with two of these typical yellow Springer books. One was, and still is, a rather incomprehensible volume on *operator algebras* (Blackadar, 2013) — very relevant to theoretical physics, but as far as I know unrelated to probability theory — while the other was an introductory text on measure and probability theory (Capiński & Kopp, 2004). To this day, the latter is still one of the few mathematics books in my collection that I have completely read (and that I would genuinely recommend to others). I was so intrigued by the mathematics underlying this subject and, although statistics was not among my favourite courses, it is still a subject that I like to read about since many of the ideas permeate countless other areas of mathematics and physics (and science in general).

Aside from statistics courses, physicists are also quickly confronted with the idea of uncertainty in various ways: inherent uncertainty in quantum mechanics, numerical uncertainty in computational physics and stochastic uncertainty in statistical physics. (Preferably all of these together to enjoy a real challenge.) So, when I was looking for a possible Ph.D. position after leaving the physics department and heard from a fellow physics student that there was a position available at the statistics and data analysis department, I was immediately interested. The final goal of the Ph.D. was not immediately clear, but after working on a few topics, the choice to focus on uncertainty quantification became an obvious one. In fact, the framework of conformal prediction, which will form the backbone of this dissertation, was already present in the first project that I started with. The most promising method in De Clercq (2019) was a conformal prediction-based method and I was quickly taken aback by its simplicity, especially given how powerful its guarantees are.

## 1.3   Structure

This dissertation is structured as follows. In Chapter 2, the general notion of uncertainty is treated: what it means, how we can express it and what we can hope or expect to achieve in practice. This chapter will also give the general definition of the most important tool in this dissertation: conformal prediction.

Once the basic concepts and ideas have been introduced, Chapter 3 focuses on regression tasks, where one of the primary tools to express uncertainty is that of prediction intervals (PIs). First, some general classes of (machine learning) methods are introduced. Then, in a follow-up section, these methods are fine-tuned or enhanced with conformal prediction. These methods are further compared and analysed using some real-world data sets. The motivation for this chapter was, at the time of writing, the lack of a comprehensive review of uncertainty quantification methods, especially in comparison to or when combined with conformal prediction. Chapter 3 concludes with an exposition of some existing extensions from the literature to show that just the tip of the iceberg was unveiled and many more fascinating aspects exist. Although the chapter is mainly based on the paper *Valid prediction intervals for regression problems* (Dewolf, De Baets, & Waegeman, 2023b), there are some modifications: the part about conformal prediction has been moved to Chapter 2, a section about jackknife estimators has been added to tie it to the content of Section 2.5.1 and a section with code excerpts was added to give an idea of how a practical implementation looks like (various other minor additions or changes have been made to improve the text).

Although the results of Chapter 3 are promising, an important problem remains untreated: not all parts of the instance space share the same level of stochasticity, i.e. the data is inherently heteroskedastic. The downside of 'naive' conformal prediction methods is that they cheat in some sense by only focusing on the easier points. This conditional aspect is covered in Chapter 4. The chapter again starts with the introduction of some methods that allow to emphasize the importance of specific regions of the instance space. The main result in this chapter is a (partial) characterization of probability distributions for which certain 'naive' methods still satisfy conditional statistical guarantees. An important motivation for this chapter was the study of uncertainty quantification in the context of time series, a subject that eventually did not make its way into this dissertation. Two important aspects are characteristic for time series: *autocorrelation* and *nonstationarity*. Both properties violate exchangeability, the crucial property for conformal prediction, by either introducing a strong dependence between data points or by simply changing the data distribution. In certain cases, this problem can be alleviated, e.g. when the autocorrelation vanishes at a finite number of time steps, an *autoregressive model* can reinstate exchangeability.[3] One important consequence of these properties is that the variance of the response variable is not constant. This heteroskedastic behaviour is observed more generally and, accordingly, studying conformal prediction in this setting seemed like an interesting path forward. As with the preceding chapter, although it is mainly based on the preprint *Conditional validity of heteroskedastic conformal*

---

[3]  Some other solutions for time series are treated in Section 2.5.3.

*regression* (Dewolf, De Baets, & Waegeman, 2023a), there are some modifications to Chapter 4. The main addition is the inclusion of normalizing flows as an additional confidence predictor.

The closing chapter of this dissertation, Chapter 5, tries to treat the middle ground between marginal methods — the 'naive' methods in the foregoing paragraphs — and conditional methods — those that treat the different subsets of the instance space separately. The former have weaker guarantees, whereas the latter require more data. By exploring a clustering-based approach, the best of both worlds can be obtained. This is also the ideal situation to move to a slightly different problem setting and exchange the simple world of (univariate) regression for the murky waters of multitarget prediction and, in particular, extreme classification. Multiple motivations were at work for this chapter. On the one hand, a collaboration with Dimitrios Iliadis on the Flanders AI project involved the use of multiclass classification and multitarget prediction methods. On the other hand, recent work on cluster-wise conformal prediction sparked interest in how the theoretical properties developed in Chapter 4 could be generalized to the type of clustering that is relevant when side information is available.

The appendix to this dissertation consists of three parts. The first part is reserved for a 'concise' treatment of the mathematical tools used throughout the main text. Tables containing numerical results of experimental are located in the second part, since providing them throughout the main text would needlessly interrupt the flow and make this work harder to read. The third part of the appendix consists of the explanations for the Mondrian-like painting on the cover and for the 'quotes' at the beginning of every chapter. Many dissertations contain short literary quotes that inspired or were enjoyed by the author. However, since literature has never been my cup of tea, I opted to replace these by formulas that I think are fascinating and in some way[4] related to the chapter in which they appear.

An overview of the contributions in this dissertation is given below:

1. Common uncertainty quantification methods in the regression setting, with and without conformal prediction, are reviewed and compared. (Chapter 3)

2. The conditional validity of conformal prediction methods is analysed

---

[4] Possibly a very convoluted way.

in the case where data splitting is not desirable.  General conditions
for such conditional guarantees are derived, which tie this study to
Fisher's pivotal quantities. (Chapter 4)

3. The effects of misspecification on the conditional validity of general
   conformal predictors are studied and some diagnostic tools are pre-
   sented. (Chapter 4)

4. The usefulness of clustering in the setting of Mondrian conformal pre-
   diction is analysed using the notion of Lipschitz continuity.  (Chap-
   ter 5)

5. The use of domain knowledge and side information for improving clus-
   tering of conformal predictors is considered.  On the one hand this
   allows to introduce inductive biases, but on the other hand it also be-
   comes apparent that this is sometimes the only way to move forward.
   (Chapter 5)

6. The application of clusterwise conformal prediction (score-based and
   hierarchy-based) to extreme classification is considered. (Chapter 5)

## 1.4    Style guide

I chose to write this dissertation (or at least most of it) in the first person
plural.  The 'we' in this form should, however, not be read as representing
multiple authors, notwithstanding that this is the culmination of the work of
many people, but as the 'we' consisting of me (the author) and you (the au-
dience). This work is meant to take both of us on an adventure through the
wonderful land of mathematics, statistics, data science and machine learn-
ing.

Scattered throughout this dissertation the attentive reader might find some
text boxes indicated with colourful decorations.  Four kinds of boxes can be
found.

> **Theorem** 1.1**.** Boxes of this form indicate important or foundational re-
> sults and theorems.

**Property** 1.2. Although not different from theorems in a logical sense, properties (and corollaries) will often contain statements or results of lesser importance.

**Extra** 1.3. Extras are short properties or statements that can be ignored by the casual reader, but that I found particularly interesting. They are merely the ramblings of a mad physicist with too many academic interests.

**Definition** 1.4. Boxes of this type are reserved for definitions, methods and examples. This makes it easier to see where these blocks end and where the body of the text recommences.

Theorems and properties will come in two variants. Those for which a clear reference containing a proof exists, for which only the reference will be given, and those which for which no reference could be found, in which case an explicit proof is provided.

Just as the styling of boxes indicates the importance or meaning of those blocks of text, so will the style of individual words. When a term is defined for the first time, it will be indicated in **bold**. If a word having a specific technical meaning is being used without it being defined explicitly, it will be indicated in *italics*.

Some remarks should also be made about the terminology used in this work as to avoid any confusion (as much as possible). One of the important examples of a word where confusion might arise, especially in Chapter 4, is the notion of independence. Two kinds of independence are used throughout this dissertation. The first one is well known in statistics and probability theory and corresponds to the situation where a joint probability distribution factorizes as a product of individual probability distributions (Definition A.39):

$$P_{X,Y} = P_X P_Y.$$ 
$$(1.4)$$

For this reason, we will call this **probabilistic independence** whenever necessary. The second type of independence is of a more trivial kind and simply means that a function does not depend on a particular argument, such as in the following example where a function $f : \mathcal{X} \times \mathcal{Y} \to \mathcal{Z}$ is independent of $\mathcal{Y}$:

$$f(x,y) = f(x,y')$$ 
$$(1.5)$$

for all $x \in \mathcal{X}$ and $y, y' \in \mathcal{Y}$. We will call this **functional independence** (not to be confused with the eponymous notion from *database theory*). Another example where confusion might arise is that of probability distributions. These will also be called **probability measures** or, simply, **distributions** (cf. Section A.2). The distinction between estimators and estimates will also be left aside due to the inconsistencies in the literature. (The former are algorithms to obtain the latter.) However, a great effort has been made to be as consistent as possible, while still retaining the common terminology from the literature.

## 1.5   Notation

For clarity, the main notations used throughout the text will be introduced here. General sets, or 'spaces'[5] as they will often be called, will be denoted by capital calligraphic letters: $\mathcal{X}, \mathcal{Y}, \mathcal{Z}, \ldots$ This is to distinguish them from random variables, which are denoted by ordinary capital symbols: $X, Y, Z, \ldots$ Elements of a set will be denoted by corresponding lowercase symbols: $x \in \mathcal{X}$. When a set has a product structure, such that its elements can be expressed as tuples, and when this structure is explicitly used, the elements will be denoted by boldfont lowercase symbols as is often done in computer science: $\boldsymbol{x} \in \mathcal{X}$. When this structure is not of major importance, an ordinary symbol will be used: $x \in \mathcal{X}$. Entries of a tuple $\boldsymbol{x}$ will be indicated by upper indices, e.g. $\boldsymbol{x} \equiv (x^1, \ldots, x^n)$, since lower indices will be reserved for labeling elements in a set or sequence, e.g. $\{x_1, \ldots, x_n\}$.

Functions will always be denoted in the usual way as $f : \mathcal{X} \to \mathcal{Y}$, clearly stating both their domain and codomain (unless this is clear from the context). When a function (or some other object) depends on parameters that are of relevance, the parameters will be denoted by Greek symbols: $\alpha, \theta, \ldots$ If we want to put emphasis on the fact that a given function or parameter is actually only an estimate of the 'true' value, it will be written with a caret, e.g. $\widehat{\theta}$ is an estimate of $\theta \in \mathbb{R}$ and $\widehat{f}$ is an estimate of $f : \mathcal{X} \to \mathcal{Y}$.

The **indicator function** of a set $\mathcal{X}$ will be denoted by $\mathbb{1}_{\mathcal{X}}$:

$$\mathbb{1}_{\mathcal{X}}(x) := \begin{cases} 1 & \text{if } x \in \mathcal{X}, \\ 0 & \text{if } x \notin \mathcal{X}. \end{cases} \tag{1.6}$$

---

[5]  The reason for this terminology is clarified in Section A.2.

The **power set** of $\mathcal{X}$, i.e. the set of all subsets of $\mathcal{X}$, will be denoted by $2^{\mathcal{X}}$. To construct elements of the power set $2^{\mathcal{X}}$, the following 'set-builder notation' will be adopted:

$$\left\{ x \in \mathcal{X} \mid \Phi(x) \right\}, \tag{1.7}$$

where $\Phi$ is a *predicate* depending on $x$ (Lavrov & Maksimova, 2012). To denote the set of the first $n \in \mathbb{N}_0$ positive integers, we will use

$$[n] := \{1, \dots, n\}. \tag{1.8}$$

If two sets $\mathcal{X}, \mathcal{Y}$ are **isomorphic**, i.e. when there exists an invertible function between them, we will generally indulge in some abuse of notation and use an equals sign, unless the distinction is of importance, in which case the congruence symbol $\cong$ will be used.

Turning to the more statistical side of things, there is again plenty of notation to be introduced (see also Section A.2). The set of probability distributions on a set $\mathcal{X}$ will be denoted by $\mathbb{P}(\mathcal{X})$. Given a probability distribution $P \in \mathbb{P}(\mathcal{X})$, the associated expectation will be denoted by $\mathsf{E}_P$ or $\mathsf{E}_{X \sim P}$ if the distribution of the random variable cannot be deduced from the context. Whilst probability distributions are formally set functions, we will often, for convenience, work with expressions like

$$P(X \in A). \tag{1.9}$$

In such an expression, the event $X \in A$ should be read as the set $\{x \in \mathcal{X} \mid x \in A\}$. More generally, when $P$ is a probability distribution on $\mathcal{X}$ and $\Phi$ is a predicate about elements of $\mathcal{X}$, then

$$P\big(\Phi(X)\big) := P\big(\{x \in \mathcal{X} \mid \Phi(x)\}\big). \tag{1.10}$$

To not overload the notation and when no confusion can arise, the probability of an event $A$ with respect to some generic probability distribution (which can usually be deduced from the context) will be denoted by

$$\mathsf{Prob}(A). \tag{1.11}$$

A last important point is the notation for intervals. For open intervals, the Bourbaki-style notation

$$]a, b[ := \left\{ x \in \mathbb{R} \mid x > a \land x < b \right\} \tag{1.12}$$

will be used. Any (other) notations that are defined throughout the text can also be found in the symbols list just before the table of contents.

# Uncertainty Estimation 2

$$\triangle \hat{p} \, \triangle \hat{q} \geq \frac{\hbar}{2}$$

$$\triangle \widehat{A} \, \triangle \widehat{B} \geq \left| \tfrac{1}{2} \langle [\widehat{A}, \widehat{B}] \rangle \right|^2$$

Heisenberg–Robertson (1927–1929)

## 2.1 Introduction

The central idea of this chapter is to explain why uncertainty is important and how it can be characterized. To this end, the notion of 'confidence predictors' will be of paramount importance. This chapter is mainly a review of the literature and we will explore various related ideas and methods, ranging from what uncertainty means to how conformal prediction can help us quantify uncertainty in a meaningful way.

This chapter is structured as follows. In Section 2.2, the gist of uncertainty quantification and some associated ideas are covered. Concepts such as epistemic and aleatoric uncertainty, confidence and prediction regions, and validity will be treated. Once the basis is covered, Section 2.3 introduces the centrepiece of this dissertation, conformal prediction, in all its glory. Although the content of Section 2.3 suffices for the remainder of this work, Sections 2.4 and 2.5 provide some more information on related topics. The former rephrases the conformal prediction methodology in the context of hypothesis testing, whereas the latter covers various extensions of conformal prediction beyond the basic setting in which it was first formulated.

## 2.2 What is uncertainty?

Uncertainty as a whole is a very broad concept. Consequently, throughout history, scientists have come up with many different definitions, explana-tions and measures to express or quantify uncertainty. As a general defini-
tion, we can take uncertainty to reflect "the state of being in which a given
value or property cannot completely be described". Some examples are: "to-
morrow, the temperature will be $-1^{\circ}$C in Ghent" vs. "the weather tomor-
row will be warm in East Flanders" or "the number of cows in Paal is 35"
vs. "there are somewhere between 0 and 500 cows in Paal". In both exam-
ples, the first sentence gives a clear and definite statement, whereas the sec-
ond sentence gives a vague and imprecise statement. Already in these sim-
ple examples it is clear that for critical applications (and even for ordinary
interactions), there will be a need for clear and meaningful uncertainty quan-
tification.

### 2.2.1   Representations

The most relevant metric to characterize or quantify uncertainty strongly de-
pends on what value or property is being considered and even in what con-
text it is being considered. On the one hand, in the example about weather
forecasting, a spatiotemporal probability distribution would certainly be a
sensible choice.[1] On the other hand, when informing a farmer about the
weight of the next cow to be born on the pastures of Paal, we would be bet-
ter off having a simple measure, such as an interval covering most probable
cases, than a distribution over all cow breeds and weights. The choice of how
to express the uncertainty in a situation not only depends on the problem it-
self, but also on how we want to convey this information. A balance should
be achieved between statistical precision, expressivity and ease of use.

In this work, we will focus on an approach that tries to optimize this bal-
ance and is situated somewhere in between providing mere (scalar) statis-
tics and full distributional models. It provides more information and gives
more insight into the structure of the underlying problem than the former,
but it does not have the complexity of a complete probability distribution (or
worse, a set of probability distributions). As mentioned in the introductory
chapter, we will consider models that provide the user with a set of possible
outputs. Moreover, these 'prediction sets' will be constructed in such a way
that they have a high probability of containing the ground truth.

---

[1] We will not touch upon the interpretation of such distributions. The various wild ap-
proaches to weather forecasting are baffling.

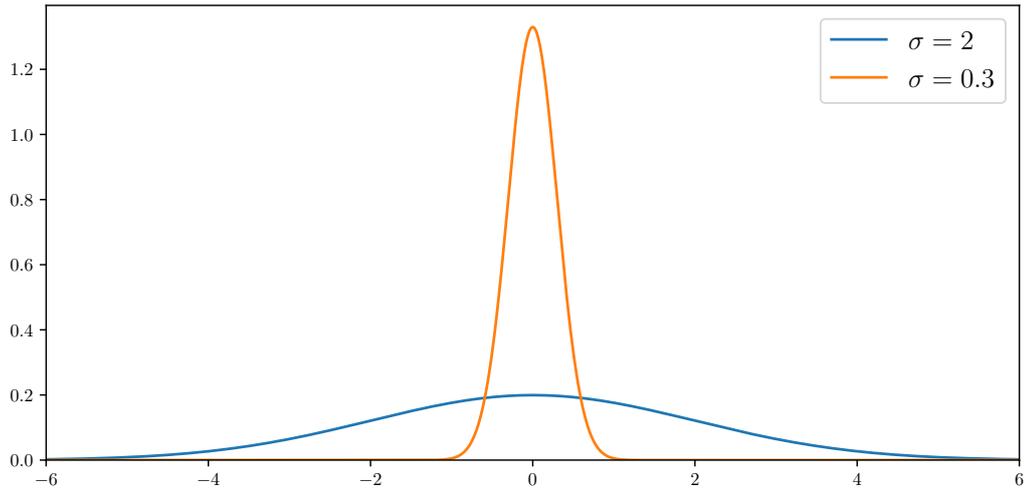

Figure 2.1: Probability density functions of two normal distributions with mean 0, but with different standard deviations. The higher the standard deviation (and, hence, the variance), the higher the probability of obtaining extreme values.

Let us consider some concrete examples of the different approaches to uncertainty quantification. In the simplest case, the scalar approach, our *epistemic state* — everything we know or do not know — is mapped onto a single number.[2] A good example would be the case where epistemic states are described by probability distributions $P_X \in \mathbb{P}(\mathcal{X})$ on a probability space (Section A.2). In this case, two common measures of uncertainty are the variance (see also Definition A.26)

$$\mathsf{Var}[X] := \int_{\mathbb{R}} \big(x - \mathsf{E}[X]\big)^2 \, \mathrm{d}P_X \tag{2.1}$$

and the **entropy**[3]

$$H_{\mathrm{cont}}[X] := - \int_{\mathbb{R}} f_X(x) \ln f_X(x) \, \mathrm{d}x. \tag{2.2}$$

The variance is a simple and straightforward measure of the spread of a distribution: the higher the variance, the wider the distribution (see Fig. 2.1). This agrees with our naive interpretation of uncertainty, since wider distributions, where the probability of having more diverse outcomes is greater,

---

[2] We will not try to formalize this in full generality, since this would require to define the space of epistemic states with suitable notions of continuity, smoothness, etc.

[3] The differential entropy for absolutely continuous distributions, i.e. those with a density function, is used for simplicity.

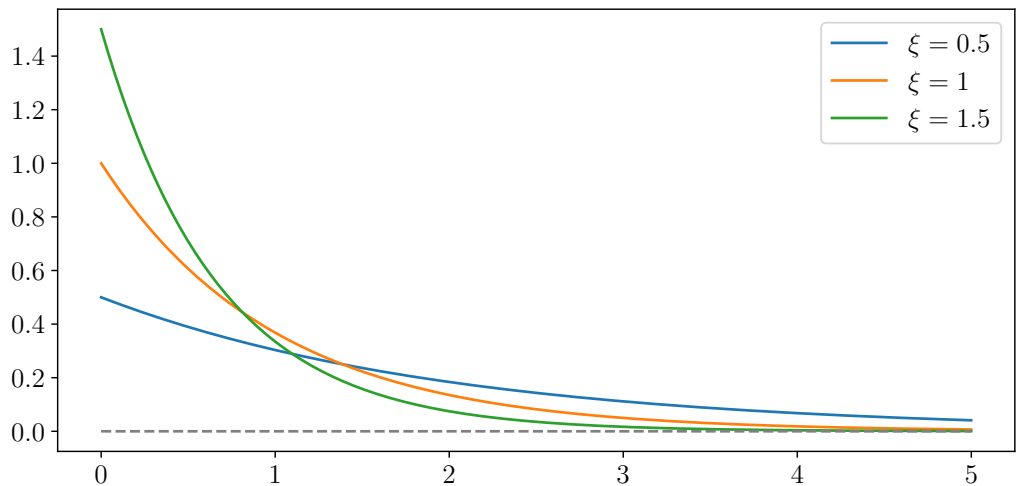

(a) Exponential distributions with different values of the rate parameter $\xi > 0$. The greater the rate, the steeper the slope of the density function.

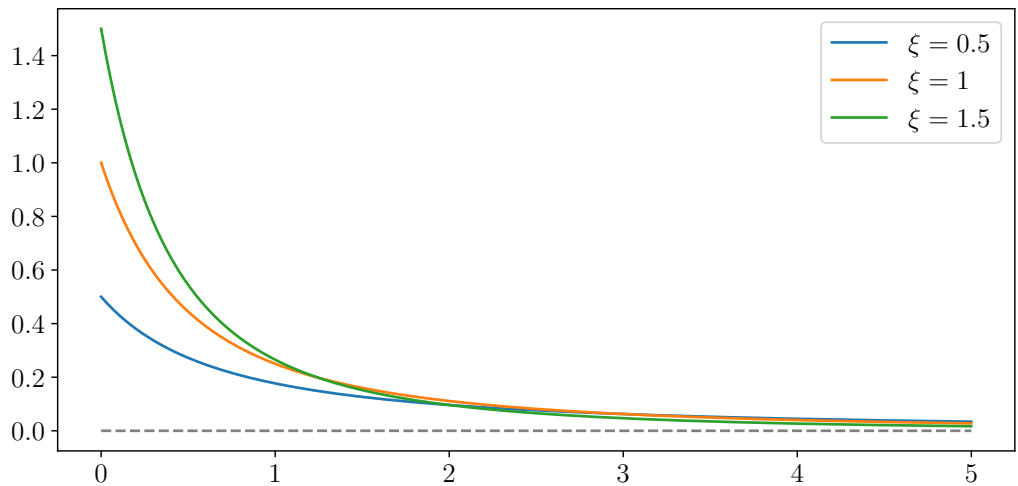

(b) (Shifted) Pareto distributions (Example A.34) with different values of the shape parameter $\xi > 0$ and fixed scale parameter $\lambda = 1$. The greater the shape parameter, the thinner the tail of the density function.

Figure 2.2: Probability density functions of exponential (a) and Pareto distributions (b).

are usually seen as representing states of higher uncertainty. The entropy on the other hand looks more at the complexity of the distribution or, as it is sometimes stated, how it compares to a uniform distribution.[4] The use of entropy to characterize uncertainty is also the basis of the *principle of maximum*

---

[4]  It can, for example, be shown that the maximizer of the entropy on a bounded interval is given by the uniform distribution.

*entropy* (Jaynes, 1957), which states that the distribution that best represents an epistemic state is the one with the greatest entropy. The downside of these measures is that they only tell us something about how uncertain our epistemic state is relative to other possible states. However, they give no indication of how this translates to actual samples, e.g. all uniform distributions with the same width have the exact same variance and entropy, even though samples from a uniform distribution on $[0, 1]$ or on $[100, 101]$ will clearly be very different.

On the other side of the spectrum, sidestepping higher-order uncertainty representations such as *credal sets* (Augustin, Coolen, De Cooman, & Troffaes, 2014), we could choose to simply work with the entire probability distribution. This entity contains or encodes all the information we have. However, in general, this approach can be rather unwieldy for multiple reasons: the distribution does not always exist in a closed or parametric form, exact inference might be NP hard (see Section 3.2.1), etc. Another issue is that 'visual inspection', i.e. looking at the shape of the distribution to infer properties, can be dangerous and misleading. Consider for example the exponential distribution in Fig. 2.2a and the Pareto distribution[5] in Fig. 2.2b. The corresponding curves in these figures look rather similar and, without further investigation, we might expect that they behave very similarly.

However, nothing could be further from the truth. Pareto distributions (Example A.34) belong to the class of so-called heavy-tailed distributions. These have a much higher probability for extreme events than exponentially decaying distributions such as the exponential or normal distributions. For example, for Pareto distributions with $\xi < 2$, the variance diverges. As a consequence, making predictions for future events becomes much harder. To get a better feeling for this issue, compare Fig. 2.3a to Fig. 2.3b. The shaded areas indicate the intervals, starting at 0, that contain 70% and 80% of the probability mass, respectively (the intervals are also indicated for clarity). It is immediately clear that for Pareto distributions these intervals are much wider than for exponential distributions, even though a simple inspection of the density functions might not have given the same impression.

For these reasons, the approach adopted in this thesis is that of **prediction**

---

[5] These are named after the Italian scientist Pareto who introduced the 80/20 rule, which says that in many situations 80% of the consequences results from 20% of the causes (Pareto, 1964).

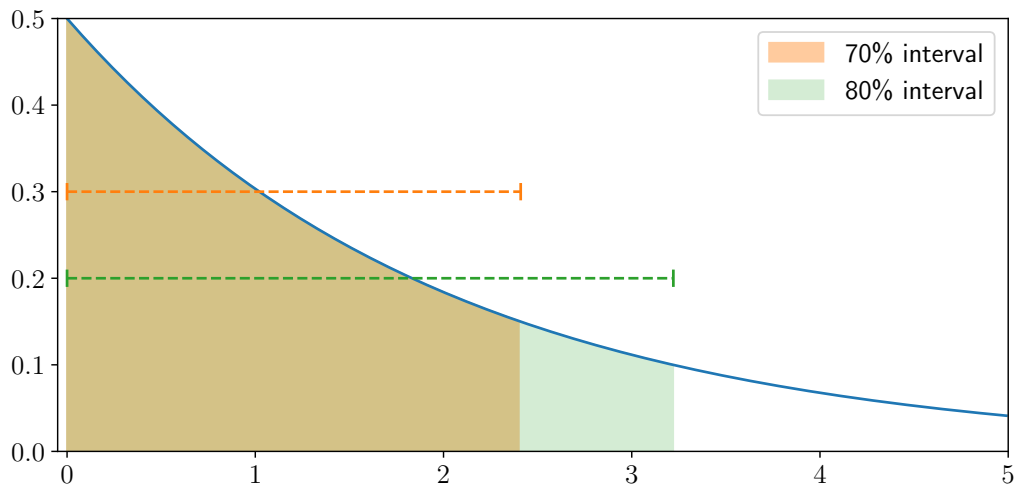

(a) Exponential distribution with rate parameter $\tilde{\xi} = 0.5$.

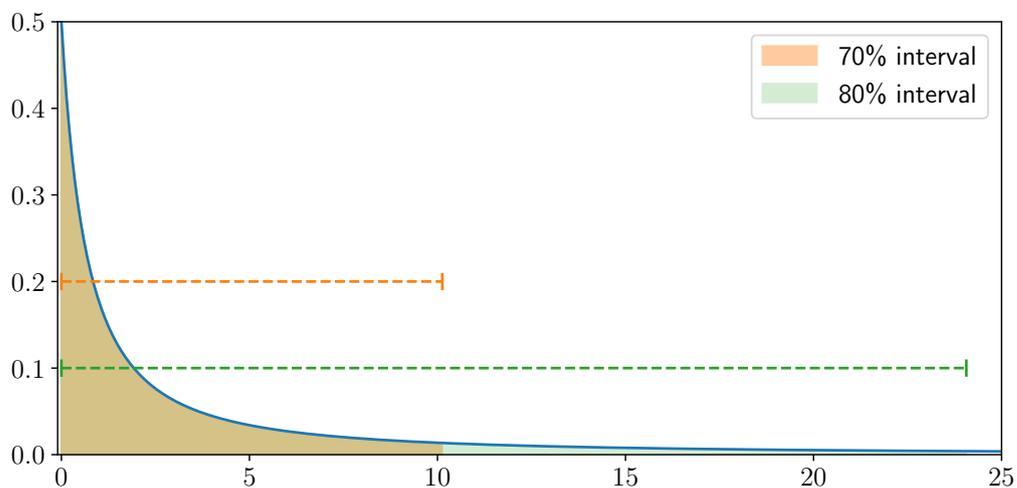

(b) (Shifted) Pareto distribution (Example A.34) with shape parameter $\tilde{\xi} = 0.5$ and scale parameter $\lambda = 1$. The interval $\Gamma^{0.3}$ reaches up to almost 25, making the difference between $\Gamma^{0.2}$ and $\Gamma^{0.3}$ much greater than in the exponential case (Fig. 2.3a).

Figure 2.3: Prediction intervals at significance levels $\alpha = 0.2, 0.3$. Shaded areas cover 70% and 80% of the distributions, respectively.

**regions** (see Section 2.2.3 below). These are sets containing the most likely values of a new event or point sampled from a distribution. They avoid the excessive simplicity of scalar quantities, can be related to the variables of interest in a straightforward manner and are easy to interpret. The main downside is the freedom in their definition. Consider for example the case of the uniform distribution $\mathcal{U}([0, 1])$ on the unit interval. Prediction intervals with a coverage (to be formally defined below) of 90% are not unique since any interval of the form $[u, u + 0.9]$ with $u \in [0, 0.1]$ will satisfy our requirement and contain 90% of the probability mass. This nonuniqueness holds in general. The only distributions for which prediction regions are unique are the Dirac measures (A.27), i.e. the point masses.

> **Remark** 2.1 (**Nonuniqueness**). The nonuniqueness is even more dire for uniform distributions than for most other distributions. Not only are the prediction intervals nonunique, their size is even invariant under many transformations, such as translations. (Since all points in the support have the exact same density, all prediction regions for a fixed confidence level will have the exact same 'size'.) For some distributions, such as symmetric or unimodal ones, we can make convenient choices, such as central intervals, to break the symmetry, but these are, as stated, mere conveniences.

### 2.2.2 Decomposition of uncertainty

Naively, or at least as a first impression, we could consider uncertainty, no matter what metric is used to express it, to be a single entity: the aspects of the data or the data-generating process that we are not certain about. However, there are generally multiple contributions to the overall uncertainty. Consider for example the situation where we try to model the relation between the daily temperature $T_t \in \mathbb{R}$ ($t \in \mathbb{Z}$ indicates the time or the day) and the amount of precipitation $\rho_t \in \mathbb{R}^+$. Given some historical data, we can try to fit a parametric model:

$$\hat{\rho}_t = f_\theta(T_t). \qquad (2.3)$$

However, two aspects play a role here. The most obvious contribution to the total uncertainty is a possible lack of data. Extreme events, for example, where unknown processes might result in chaotic or unexpected behaviour, are often less recorded. Another contribution is the inherent uncertainty in the measurements. All measurement devices have a finite precision and this measurement uncertainty gets propagated to an uncertainty in the final prediction. There is, however, a fundamental difference between these two contributions. The former can be reduced or even completely negated by collecting more data, whereas the latter is irreducible.[6]

---

[6] Some authors, such as Malinin and Gales (2018), consider the distribution shift between training and test data to be a separate contribution, called the **distributional uncertainty**.

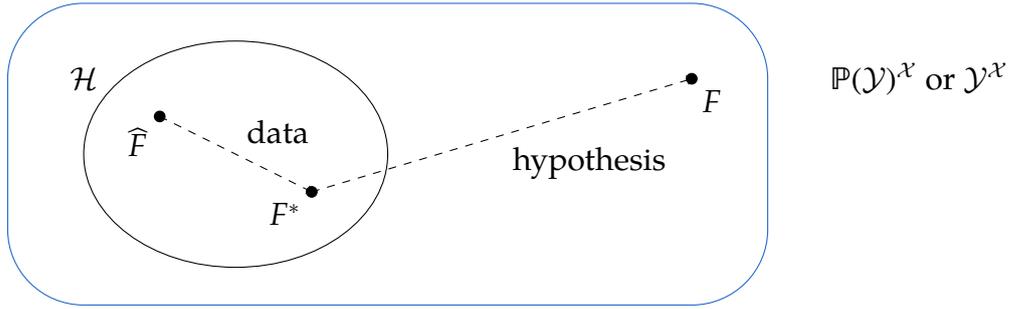

Figure 2.4: Decomposition of the (epistemic) uncertainty. Both the choice of hypothesis space $\mathcal{H}$ and the approximate modelling capabilities contribute to the uncertainty.

For this reason, we should make a clear distinction between these contributions. The reducible uncertainty is often called the **data** or **epistemic uncertainty** (Hüllermeier & Waegeman, 2021; Kendall & Gal, 2017) and results from our lack of understanding. It can be reduced by observing more data or by modifying the hypothesis space $\mathcal{H}$, since, even in the infinite data setting, a bad choice of model architecture might lead to incorrect models (**misspecification**). As shown in Fig. 2.4, when estimating a probability distribution $F \in \mathbb{P}(\mathcal{X})$ or a function $F : \mathcal{X} \to \mathcal{Y}$, we usually first select a hypothesis space $\mathcal{H}$. For example, an exponential family in the probabilistic case or a specific choice of neural network architecture in the case of function spaces. One part of the epistemic uncertainty is given by the difference between the optimal estimate $F^*$ within the hypothesis space and the ground truth $F$. However, since in general we only have access to a limited amount of (often noisy) data, the result of the estimation procedure $\widehat{F}$ is also different from the optimal estimate $F^*$. This is a second source of epistemic uncertainty.

The irreducible part of the total uncertainty is known as the **aleatoric uncertainty**. It is a form of uncertainty that cannot be mitigated at the level of data analysis. No matter how much data is acquired or how well-chosen the hypothesis space is, this type of uncertainty will not vanish. It stems purely from the stochasticity of the data-generating process and, hence, only appears if the object to be estimated is something more general than an ordinary function $F : \mathcal{X} \to \mathcal{Y}$, such as a probability distribution $F_{Y|X}$. This distribution would then be called the aleatoric probability or, in some circles, the *ontic* probability. For example, when the data is generated from a standard normal distribution, even if we have perfect estimates, i.e. $\widehat{\mu} = 0$

and $\widehat{\sigma} = 1$, it is impossible to predict the next data point with 100% accuracy.

> **Remark** 2.2 (**Representational uncertainty**). Whether the choice of feature space $\mathcal{X}$ is part of the choice of hypothesis space $\mathcal{H}$ is a difficult question. If the answer is yes, then, assuming a deterministic universe[a], aleatoric uncertainty coming from coarse-graining does not even exist and choosing the wrong feature space would by definition lead to epistemic uncertainty. However, in practice, it is impossible to measure the state of every object in the universe and, hence, the feature space available to us will always be a coarse-grained version, thereby inducing a certain level of uncertainty. Since avoiding this issue is *morally impossible*, adopting the language of Shafer (2007); Shafer and Vovk (2006), it begs the question whether aleatoric uncertainty should be an absolute notion or defined relative to a choice of feature (or data) representation.
>
> ___________
>
> [a] We ignore (the mainstream interpretations of) quantum mechanics for simplicity, since typical data sets in machine learning are often not of such a microscopic scale.

### 2.2.3 Confidence predictors

As mentioned in the introduction, given a data set with features and responses, it is not only relevant to obtain a point estimate of the response, but often we are also interested in a larger set of possibilities that will contain the ground truth with a predetermined probability. Especially for safety-critical applications such as self-driving cars (Chen & Huang, 2017; Michelmore et al., 2020) or precision medicine (Fang, Xu, Yang, & Qin, 2018; Jiang, Osl, Kim, & Ohno-Machado, 2012), the need for strong probabilistic guarantees or constraints is high. Many state-of-the-art models are optimized with respect to predictive power, even in present-day competitions, but often this is not the same as optimizing probabilistic accuracy.

To this end, we will consider functions of the following type.

> **Definition** 2.3 (**Confidence predictor**). Let $(\mathcal{X}, \Sigma_{\mathcal{X}})$ and $(\mathcal{Y}, \Sigma_{\mathcal{Y}})$ be two measurable spaces. A confidence predictor is[a] a function $\Gamma^\alpha : \mathcal{X} \to \Sigma_{\mathcal{Y}}$ that takes feature values as argument and returns a set of possible responses. The superscript $\alpha \in [0, 1]$, called the **significance level**[b], indicates that the true response should be contained in the predicted set with (at least) the predetermined probability $1 - \alpha$. The sets $\Gamma^\alpha(x)$ them-selves are called **confidence sets** or **confidence regions**. However, to avoid confusion (see also the next section), we will prefer the terminology **prediction set** or **prediction region**. Often, the desired probability is also made an argument of the function:

$$\Gamma : \mathcal{X} \times [0, 1] \to \Sigma_{\mathcal{Y}} : (x, \alpha) \mapsto \Gamma^{\alpha}(x) \,. \tag{2.4}$$

---

[a] Vovk, Gammerman, and Shafer (2022) use a slightly different definition since there the natural setting is that of sequences.

[b] The value $1 - \alpha$ is also called the **confidence level**.

In practice, confidence predictors, which will from now on be denoted by $\Gamma^{\alpha} : \mathcal{X} \to 2^{\mathcal{Y}}$ to make the choice of $\sigma$-algebra on the target space $\mathcal{Y}$ implicit, are usually required to satisfy a monotonicity constraint as a sanity check:

$$\alpha_1 \leq \alpha_2 \implies \Gamma^{\alpha_2}(x) \subseteq \Gamma^{\alpha_1}(x) \tag{2.5}$$

for all $x \in \mathcal{X}$. Increasing the probability of covering the ground truth should not result in a loss of possible responses. The conformal prediction framework, to be introduced in Section 2.3, will trivially satisfy this condition.

To formalize the probabilistic condition in Definition 2.3, we need the following notion, which will play an important and guiding role throughout this work.

**Definition 2.4 (Coverage).** Consider a confidence predictor $\Gamma^{\alpha} : \mathcal{X} \to 2^{\mathcal{Y}}$. Its coverage under the distribution $P \in \mathbb{P}(\mathcal{X} \times \mathcal{Y})$ is defined as follows:

$$\mathcal{C}(\Gamma^{\alpha}, P) := \mathbb{E}_P\big[\mathbb{1}_{\Gamma^{\alpha}(X)}(Y)\big] = P\big(Y \in \Gamma^{\alpha}(X)\big) \,, \tag{2.6}$$

where $\mathbb{1}$ denotes the indicator function as defined in Eq. (1.6). The probability should be read as in Eq. (1.10), i.e. as

$$P\big(\{(x, y) \in \mathcal{X} \times \mathcal{Y} \mid y \in \Gamma^{\alpha}(x)\}\big) \,. \tag{2.7}$$

It is the probability, over both features and responses, that the response will be captured by the confidence set constructed given the feature value. When $P$ is the empirical distribution of a data set $\mathcal{D}$, the notation $\mathcal{C}(\Gamma^{\alpha}, \mathcal{D})$ is also used.

**Remark 2.5 (Nuisance parameters).** When a confidence predictor $\Gamma^{\alpha} : \mathcal{X} \to 2^{\mathcal{Y}}$ is also a function of nuisance parameters[a] $\theta \in \Theta$, the above definition should be modified as follows:

$$\mathcal{C}(\Gamma^\alpha, P_{X,Y,\theta}) := P_{X,Y,\theta}\big(Y \in \Gamma^\alpha(X \mid \theta)\big), \qquad (2.8)$$

where the data-generating distribution $P$ in (2.6) can be obtained as a marginal of the joint distribution $P_{X,Y,\theta}$ over the instances and nuisance parameters.

The fact that $(X, Y)$ is not necessarily independent of $\theta$ will play an important role in the section on conformal prediction. To clarify this remark, consider the classical problem of constructing prediction intervals (James, Witten, Hastie, & Tibshirani, 2013), which serves as a parametric precursor of conformal prediction. For normal distributions, we would proceed by calculating a sample mean $\hat{\mu}$ and sample standard deviation $\hat{\sigma}$ on a data sample $\{x_1, \dots, x_n\}$ and construct the interval (see also Section 3.2.2)

$$\left[\hat{\mu} - z\hat{\sigma}, \hat{\mu} + z\hat{\sigma}\right] \qquad (2.9)$$

for some suitable value of $z \in \mathbb{R}^+$. However, for a fixed sample $\{x_1, \dots, x_n\}$, there is no guarantee that a new point $x_{n+1}$ will fall inside this interval with the predetermined probability. A simple example would be the (arguably very unlikely) case of a sample like $\{0, 0.01, -0.02, 0.02\}$ of a normal distribution with mean 0 and standard deviation 1. The sample standard deviation ($\hat{\sigma} \approx 0.02$) is much too small to be able to account for the stochasticity in future observations. Only when we also take the probability over the sample, i.e. resample the data every time, can we obtain statistical guarantees.

----

[a] Model parameters are not nuisance parameters since these fundamentally determine the model and the resulting distribution.

As stated before, when constructing prediction regions we have to specify the probability with which the estimator is allowed to make errors, since, in general, no realistic model — a model with a finite number of parameters inferred from a finite number of data points — can be perfect. Given a choice of significance level $\alpha \in [0, 1]$, the estimator $\Gamma^\alpha$ should satisfy the following consistency condition (Faulkenberry, 1973; Fraser & Guttman, 1956):

$$\mathcal{C}(\Gamma^\alpha, P) \geq 1 - \alpha. \qquad (2.10)$$

This condition will, however, not be satisfied in some cases. To distinguish between such situations, estimators satisfying the inequality are said to be (**conservatively**) **valid** (Vovk et al., 2022) or **calibrated**, in accordance with the classification literature (Guo, Pleiss, Sun, & Weinberger, 2017). (See also Definition 2.7 further on.)

It should be clear that the best models are those that get as close as possible to saturating the inequality. Of course, some could argue that a perfect model would have a coverage of 100% and should always be completely confident. However, this cannot be a realistic requirement, since this would mean that there is no epistemic or aleatoric uncertainty, something that is hardly ever the case. The only way for a confidence predictor to be completely confident about its prediction and satisfy Eq. (2.10) is to predict the whole target space $\mathcal{Y}$. Such a predictor clearly satisfies the validity condition, but it is hard to extract any meaning from the result. This indicates that simply having an estimate of the uncertainty that satisfies condition (2.10) is not sufficient. To this end, another competing measure that quantifies the 'meaningfulness' has to be considered, similar to how regularization terms are added to loss functions to control the model complexity (see e.g. Chapter 7 in Goodfellow et al. (2016)). The simplest such measure will be a size-based metric. For example, in the case of discrete distributions, the cardinality of the confidence set can be used:

$$\mathcal{W}(\Gamma^\alpha, P) := \mathsf{E}_P \big[ |\Gamma^\alpha(X)| \big] .$$

In the case of interval estimators, where $\mathcal{Y} = \mathbb{R}$, the average length (or width) is the default choice:

$$\mathcal{W}(\Gamma^\alpha, P) := \mathsf{E}_P \big[ |\hat{y}_+(X) - \hat{y}_-(X)| \big] , \tag{2.11}$$

where the functions $\hat{y}_\pm : \mathcal{X} \to \mathbb{R}$ denote the upper and lower bounds of the prediction intervals. As with the coverage of a confidence predictor, when $P$ is the empirical distribution of a data set $\mathcal{D}$, the average size is also denoted by $\mathcal{W}(\Gamma^\alpha, \mathcal{D})$. For more general metric spaces (Section A.3), the width $|\hat{y}_+(X) - \hat{y}_-(X)|$ could be replaced by, for example, the diameter (Definition A.57) of $\Gamma^\alpha(X)$. The average size of prediction sets is often called the (**in**)**efficiency** of the confidence predictor.

### 2.2.4   Confidence regions

Before proceeding with the study of these predictors, it is interesting to ponder the foregoing terminology. The notions of confidence predictor and con-fidence set might remind the reader of the notion of **confidence region** from parametric statistics (James et al., 2013). Consider a family of probability distributions $\{P_\theta \in \mathbb{P}(\mathcal{X}^*) \mid \theta \in \Theta\}$ parametrized by a set $\Theta$. A confidence region (predictor) $\Gamma^\alpha : \mathcal{X}^* \to 2^\Theta$ is said to be valid at the significance level $\alpha \in [0, 1]$ if

$$P_\theta\big(\theta \in \Gamma^\alpha(X)\big) \geq 1 - \alpha \qquad (2.12)$$

for all $\theta \in \Theta$. In frequentist terms, this means that if we repeatedly observe data points $x \in \mathcal{X}^*$ sampled from the probability distribution $P_\theta$ with parameter $\theta \in \Theta$ and, every time, construct the set $\Gamma^\alpha(x) \subseteq \Theta$, then this set will contain the true parameter $\theta$ a fraction $1 - \alpha$ of the time.

However, the confidence sets studied in this work are of a slightly different kind. They correspond to **prediction regions** (or **sets**) of the data-generating distribution $P_{Y|X}$, which is why this terminology will be adopted in the remainder of this dissertation. Though this distinction might seem crucial, it is partly of an artificial nature. This is best seen when looking at the classical definition of these two kinds of sets in parametric statistics:

1. Confidence region for a parameter $\theta$ of the distribution $P_X$:

$$P_X\big(\theta \in \Gamma^\alpha(X)\big) \geq 1 - \alpha. \qquad (2.13)$$

2. Prediction region for the predictive distribution $P_{Y|X}$:

$$P_{X,Y}\big(Y \in \Gamma^\alpha(X)\big) \geq 1 - \alpha. \qquad (2.14)$$

The two types of regions are defined by virtually the same condition, but, whereas for confidence regions the quantity of interest is a fixed parameter of the sampling distribution, the quantity of interest for prediction regions is the sample itself! Note that if the ground truth is a deterministic function $f : \mathcal{X} \to \mathcal{Y}$, i.e. when there is no aleatoric uncertainty, then $y$ can be seen as a parameter of $P_{Y|X}$ and these two notions coincide.

That this resemblance is, however, only superficial is also reflected in the no-go results that exist for nonparametric confidence regions. As will be made clear throughout this dissertation, the nonparametric construction of prediction regions is not only possible, it is even reasonably simple to do so (see e.g. Section 2.3). The same, however, cannot be said about confidence regions (Bahadur & Savage, 1956). Recent work has also led to new bounds on how sharp nonparametric confidence regions can be (Barber, 2020; Y. Lee & Barber, 2021; Medarametla & Candès, 2021).

## 2.2.5   **Strong validity**

This section is concerned with the study of some related (and stronger) notions of coverage that are sometimes used in the literature, especially in the 'older' conformal prediction literature (Vovk et al., 2022)[7]. The reason is that what we called a valid confidence predictor just below Eq. (2.10) is actually a rather weak notion. The right setting for this story and for what is yet to come is that of data sequences $((x_n, y_n))_{n \in \mathbb{N}}$ sampled from a distribution $P \in \mathbb{P}((\mathcal{X} \times \mathcal{Y})^{\mathbb{N}})$. The **coverage error** at index $i \in \mathbb{N}$ is defined as follows:[8]

$$\text{err}_i(\Gamma^\alpha, P) := \begin{cases} 1 & \text{if } y_i \notin \Gamma^\alpha\big(x_i \mid \{(x_1, y_1), \dots, (x_{i-1}, y_{i-1})\}\big), \\ 0 & \text{otherwise}, \end{cases} \qquad (2.15)$$

where the confidence predictor $\Gamma^\alpha$ now explicitly depends on the already observed instances $(x_1, y_1), \dots, (x_{i-1}, y_{i-1})$. The total number of coverage errors up to index $n \in \mathbb{N}$ is denoted by

$$\text{Err}_n(\Gamma^\alpha, P) := \sum_{i=1}^{n} \text{err}_i(\Gamma^\alpha, P). \qquad (2.16)$$

Using these random variables, two notions of validity can be introduced, both of which will be relevant in the future.

**Definition** 2.6 (**Exact validity**). The confidence predictor $\Gamma^\alpha$ is said to be exactly valid (in the strong sense) if the sequence of coverage errors $(\text{err}_n(\Gamma^\alpha, P))_{n \in \mathbb{N}}$ is a sequence of i.i.d. Bernoulli random variables with parameter $\alpha \in [0, 1]$.

**Definition** 2.7 (**Conservative validity**). The confidence predictor $\Gamma^\alpha$ is said to be conservatively valid (in the strong sense) if the sequence of coverage errors $(\text{err}_n(\Gamma^\alpha, P))_{n \in \mathbb{N}}$ is stochastically dominated (Definition A.41) by a sequence of i.i.d. Bernoulli random variables with parameter $\alpha \in [0, 1]$.

The definition above can be shown to be equivalent to the notion of conservative validity in the sense of Vovk et al. (2022).

---

[7]  This is the revision of a book that appeared in the earlier days of conformal prediction (Vovk, Gammerman, & Shafer, 2005).

[8]  This definite expression becomes a random variable if we plug in the random variables $X_j$ and $Y_j$ for $j \le i$.

**Property** 2.8 (**Equivalence with literature**). The confidence predictor $\Gamma^\alpha : \mathcal{X} \to 2^{\mathcal{Y}}$ is conservatively valid if and only if there exist sequences of binary random variables $(\xi_n)_{n \in \mathbb{N}}$ and $(\eta_n)_{n \in \mathbb{N}}$ such that:

1. $(\xi_n)_{n \in \mathbb{N}}$ is a sequence of i.i.d. Bernoulli random variables with parameter $\alpha \in [0, 1]$.

2. $(\eta_n)_{n \in \mathbb{N}}$ is bounded from above by $(\xi_n)_{n \in \mathbb{N}}$, i.e. $\forall n \in \mathbb{N} : \eta_n \leq \xi_n$.

3. Equality in distribution (Definition A.40):

$$(\mathrm{err}_n(\Gamma^\alpha, P))_{n \in \mathbb{N}} \overset{d}{=} (\eta_n)_{n \in \mathbb{N}}. \tag{2.17}$$

*Proof*. Only two probabilities have to be considered for stochastic dominance, since the variables are binary:

- $\mathrm{Prob}\big(\mathrm{err}_n(\Gamma^\alpha, P) \geq 0\big) = 1 = \mathrm{Prob}(\xi_n \geq 0)$, and

- $\mathrm{Prob}\big(\mathrm{err}_n(\Gamma^\alpha, P) = 1\big) = \mathrm{Prob}(\eta_n = 1) \leq \mathrm{Prob}(\xi_n = 1)$.

$\square$

The two foregoing definitions also have asymptotic counterparts. In the upcoming sections and chapters, we will see that, in practice, these are the appropriate incarnations of validity.

**Definition** 2.9 (**Asymptotic validity**). If

$$\mathrm{Prob}\left(\limsup_{n \to \infty} \frac{\mathrm{Err}_n(\Gamma^\alpha, P)}{n} = \alpha\right) = 1, \tag{2.18}$$

$\Gamma^\alpha$ is said to be **asymptotically exactly valid**. If the equality is replaced by an inequality, i.e.

$$\mathrm{Prob}\left(\limsup_{n \to \infty} \frac{\mathrm{Err}_n(\Gamma^\alpha, P)}{n} \leq \alpha\right) = 1, \tag{2.19}$$

$\Gamma^\alpha$ is said to be **asymptotically conservatively valid**.

Note that exactly (resp. conservatively) valid confidence predictors in this sense (Definitions 2.6 and 2.7) are also exactly (resp. conservatively) valid according to the definitions from the previous section. However, the conditions here are strictly stronger. In the previous section, we only talked about a single prediction set. So, even though for every single observation the pre-diction region might have guaranteed coverage, applying the procedure to a sequence of data points might result in a confidence predictor that is not valid and, in fact, not even asymptotically valid, since the events are not necessarily independent. (This will become more apparent when talking about inductive conformal prediction in the next section.)

## 2.3    Conformal prediction

In this section, we will give a concise, yet technical treatment of exchangeable models and, more importantly, our beloved framework conformal prediction.[9]

### 2.3.1    Exchangeability

Suppose for a second that we take the role of the data-generating process and, as the almighty ruler of the universe, we have a huge bag of possible events (huge as in containing absolutely everything). If we generate events by simply drawing them from the bag, the order is clearly arbitrary and the process takes on the following form.

**Example** 2.10 (**Pólya's urn**). Consider an urn with two[a] types of marbles: red and green. Let $r, g \in \mathbb{N}$ denote the number of red and green marbles, respectively, such that that $n := r + g$ is the total number of marbles in the urn. The distributions of consequent draws are clearly not independent, since the total number of marbles decreases after every draw. However, the joint distribution can be seen to be 'exchangeable':

$$\begin{aligned}
\mathsf{Prob}(r, g) &= \frac{r}{n}\frac{g}{n-1} \\
&= \frac{g}{n}\frac{r}{n-1} \\
&= \mathsf{Prob}(g, r).
\end{aligned} \tag{2.20}$$

[a]  This example can easily be generalized to any number of colours.

Because the order is irrelevant, any event is as likely as the other and we can simply look at how 'weird' past events were to get an idea of how future events will be. This is the general idea behind conformal prediction (to be formalized in the remainder of this section) and this notion of irrelevant ordering (to be called 'exchangeability') will be the only assumption that we have to make.

[9]  This might trigger some anxiety in fellow physicists, but do not worry, there is no connection at all with *conformal field theories*.

The academical bible on this subject is Vovk et al. (2022) or its predecessor Vovk et al. (2005). Any theorem or property in this section that is not proven explicitly should be found in these references. For a more gentle introduction, see Angelopoulos and Bates (2023). The content of this section is mainly based on Lehrer and Shaiderman (2020); Vovk et al. (2022).

**Definition** 2.11 (**Exchangeability**). Let $S_n$ denote the symmetric group on $n$ elements (Example A.5). An exchangeable distribution on $n \in \mathbb{N}$ random variables is the same as an $S_n$-invariant probability measure $P \in \mathbb{P}(\mathcal{X}^n)$. More explicitly, this is a probability measure that is invariant under any permutation $\sigma \in S_n$ of its arguments (Definition A.21):

$$P_{X_1,\dots,X_n} = P_{X_{\sigma(1)},\dots,X_{\sigma(n)}}. \tag{2.21}$$

More generally, a probability measure $P \in \mathbb{P}(\mathcal{X}^*)$ is said to be exchangeable if any of its finite-dimensional marginals is exchangeable, where the Kleene star $\mathcal{X}^*$ is defined in Eq. (A.1).

**Extra** 2.12 ($S_\infty$-**invariance**). Using Definition A.7, we can succinctly define an exchangeable probability measure as an $S_\infty$-invariant probability measure, where $S_\infty := \varinjlim S_n$ is the direct limit of (symmetric) groups under inclusions.

Note that exchangeability is a strictly weaker condition on a sequence of random variables than the usual i.i.d. assumption. However, exchangeable random variables are always identically distributed and being i.i.d. implies exchangeability.

**Property** 2.13 (**Exchangeability vs. i.i.d.**). The random variables in an exchangeable sequence are identically distributed.

*Proof*. For every integer $n \in \mathbb{N}$ and any two finite subsets $\{i_1, \dots, i_n\}$, $\{j_1, \dots, j_n\} \subset \mathbb{N}$, there exists a permutation $\sigma \in S_\infty$ of $\mathbb{N}$ mapping these two subsets onto each other and leaving all other integers unchanged. (Because of exchangeability, we can assume the indices to be ordered in increasing order.)  By definition of exchangeability, the joint distributions are the same:[a]

$$\mathsf{Prob}(X_{i_1}, \dots, X_{i_n}, \dots, X_{j_1}, \dots, X_{j_n}) = \mathsf{Prob}(X_{j_1}, \dots, X_{j_n}, \dots, X_{i_1}, \dots, X_{i_n})$$

and, hence, after marginalizing out all indices except the first $n$ or, equivalently, the last $n$, we obtain

$$\mathsf{Prob}(X_{i_1}, \dots, X_{i_n}) = \mathsf{Prob}(X_{j_1}, \dots, X_{j_n}).$$

Restricting to the case $n = 1$ gives the required result.          □

---

[a]  All indices below $i_1$ or above $j_n$ have been left out for notational simplicity.

I.i.d. sequences are easily seen to be exchangeable. Moreover, by the following theorem, exchangeable sequences are conditionally i.i.d. (although the necessary conditioning need not be straightforward). This shows that, although more general, they are not as esoteric as the definition might lead us to believe.

**Theorem** 2.14 (**de Finetti**). Let $P \in \mathbb{P}(\mathcal{X}^*)$ be an exchangeable probability measure. There exists a unique probability measure $\mathfrak{p} \in \mathbb{P}^2(\mathcal{X})$ over the set of probability measures $\mathbb{P}(\mathcal{X})$ such that

$$P(A_1 \times \cdots \times A_n) = \int_{\mathbb{P}(\mathcal{X})} Q(A_1) \cdots Q(A_n) \, \mathrm{d}\mathfrak{p}(Q) \qquad (2.22)$$

for all integers $n \in \mathbb{N}$ and all measurable subsets $A_1, \dots, A_n \subseteq \mathcal{X}$.

*Proof*. Originally by de Finetti (1929, 1937), the theorem only covered Bernoulli distributions and distributions over real-valued random variables. Hewitt and Savage (1955) extended the statement to more general distributions.[a]          □

---

[a]  These extensions require some *topological* restrictions that are beyond the scope of this work (Dubins & Freedman, 1979).

In particular, exchangeable distributions on finite probability spaces can always be reduced to urn models as in Example 2.10.

Now that the basic concept of exchangeability has been introduced, it is time to study some relevant properties. The following property shows that pushforwards (Definition A.19) preserve exchangeability. This result will be important in the coming sections to relate the exchangeability of data sets to the exchangeability as needed for conformal prediction. The property simply says that as long as the data is exchangeable and we apply a function that does not depend on the order of the data points, the result will still be exchangeable.

**Property** 2.15 (**Exchangeability-preserving function**). Let $P \in \mathbb{P}(\mathcal{X}^m)$ be an exchangeable probability measure. For every measurable function $f : (\mathcal{X}^m, \Sigma_{\mathcal{X}}) \to (\mathcal{Y}^n, \Sigma_{\mathcal{Y}})$, the pushforward $f_* P$ is also exchangeable if for every permutation $\sigma \in S_n$ there exists a permutation $\pi \in S_m$ such that

$$f \circ \pi = \sigma \circ f. \tag{2.23}$$

In particular, if $m = n$ and $f$ acts pointwise by a function $g : \mathcal{X} \to \mathcal{Y}$, the pushforward $f_* P$ is exchangeable whenever $P$ is exchangeable. The converse, however, is not true: exchangeability of $f_* P$ does not imply the exchangeability of $P$.

*Proof*. Assume that for every permutation $\sigma \in S_n$, there exists a permutation $\pi \in S_m$ such that Eq. (2.23) is satisfied.

$$\begin{aligned}
f_* P\big(\sigma(A)\big) &= P\big(f^{-1}(\sigma(A))\big) \\
&= P\big(\pi^{-1}\big(f^{-1}(\sigma(A))\big)\big) \\
&= P\big(f^{-1}(A)\big) \\
&= f_* P(A)
\end{aligned}$$

for all $A \in \Sigma_{\mathcal{Y}}$, where $\sigma(A)$ denotes the action of the permutation $\sigma$ on the set $A$ (this would be denoted by $\sigma \rhd A$ in Definition A.6). For the second part, consider as a counterexample the constant function

$$!_y : \mathcal{X} \to \mathcal{Y} : x \mapsto y$$

for some fixed element $y \in \mathcal{Y}$. For any probability measure $P$, the pushforward along $!_y$ is simply the Dirac measure at $y$:

$$!_{y,*} P(A) = \begin{cases} 1 & \text{if } (y, \dots, y) \in A, \\ 0 & \text{if } (y, \dots, y) \notin A. \end{cases}$$

It is easy to see that this measure is exchangeable. However, any push-forward along $!_y$ has this exact expression, independent of the exchangeability of $P$. □

**Extra** 2.16 (**Intertwiners**). This property can be reformulated entirely in terms of group theory. A function $f : \mathcal{X}^m \to \mathcal{Y}^n$ is exchangeability-preserving if the $S_m$-action can be *lifted*[a] to an $S_n$-action such that $f$ becomes an **intertwiner** of $S_n$-actions, i.e. a function satisfying Eq. (2.23) for all group elements.

---

[a] Lift is used here in the general sense where $f$ can also reduce the dimensionality ($n < m$).

**Remark** 2.17 (**Invertible functions**). The noninjectivity of $!_y$ is the crux of the counterexample in Property 2.15. If the intertwiner $f$ were invertible, the proof can easily be reversed to show that the exchangeability of $f_*P$ also implies that of $P$.

This result will be important for the construction of conformal predictors, since we can use all properties of exchangeable sequences after first transforming the data into a manageable format, i.e. mapping the data points to scalar quantities. Another property that will also be crucial for conformal prediction is that exchangeability implies uniform rank statistics.

**Property** 2.18 (**Uniformity**). Consider a set of random variables

$$\{X_1, \dots, X_n\},$$

taking values in a (totally) ordered set $\mathcal{X}$ (Definition A.2), such that any two-dimensional marginal vanishes on the diagonal $\Delta_{\mathcal{X}} \subset \mathcal{X} \times \mathcal{X}$.[a] If these random variables are exchangeable, the rank statistic $R_i$ of any of the variables $X_i$ is uniformly distributed over $\{1, \dots, n\}$. More generally, if there are ties, the rank distribution dominates the uniform distribution (Definition A.41).

*Proof.* Due to the invariance of exchangeable distributions, probabilities can be obtained from combinatorial arguments. Here, the probability of obtaining a given rank is of interest, i.e. $\text{Prob}(X_i = X_{(k)})$ for any $i, k \in \{1, \dots, n\}$. When no ties arise, the total number of orderings is given by $n!$, while the number of orderings with $X_i = X_{(k)}$ is given by

$(n-1)!$. This immediately gives

$$\mathsf{Prob}\big(X_i = X_{(k)}\big) = \frac{1}{n} \qquad \forall i,k \in \{1,\dots,n\}\,.$$

Now, consider the case where there are ties. If $m_k$ denotes the multiplicity of $X_{(k)}$, a combinatorial argument as above gives

$$\mathsf{Prob}\big(X_i = X_{(k)}\big) = \frac{m_k}{n} \qquad \forall i,k \in \{1,\dots,n\}\,,$$

which concludes the proof. $\qquad\qquad\qquad\qquad\qquad\qquad\qquad\qquad\square$

---

[a] This just means that the probability of having ties is 0.

The crux of conformal prediction will be the following result that gives a concentration inequality for rank statistics.

**Corollary** 2.19. Let $\widehat{Q}_n : [0,1] \to \mathcal{X}$ denote the empirical quantile function (A.66) of the random variables $\{X_1,\dots,X_n\}$. If the random variables are exchangeable, then

$$\mathsf{Prob}\big(X_i \leq \widehat{Q}_n(\alpha)\big) \geq \alpha\,, \qquad\qquad (2.24)$$

for any given $\alpha \in [0,1]$ and any $i \in \{1,\dots,n\}$. If the joint distribution is nonatomic (Section A.2.2), there is also an upper bound:

$$\mathsf{Prob}\big(X_i \leq \widehat{Q}_n(\alpha)\big) \leq \alpha + \frac{1}{n}\,. \qquad\qquad (2.25)$$

In fact, an explicit expression for the probability can be given:

$$\mathsf{Prob}\big(X_i \leq \widehat{Q}_n(\alpha)\big) = \frac{\lceil n\alpha \rceil}{n}\,. \qquad\qquad (2.26)$$

*Proof*. As noted in Eq. (A.66), the empirical quantile function is given by

$$\widehat{Q}_n(\alpha) = X_{(\lceil n\alpha \rceil)}\,.$$

For nonatomic distributions, the probability is then equal to

$$\begin{aligned}
\mathsf{Prob}\big(X_i \leq \widehat{Q}_n(\alpha)\big) &= \mathsf{Prob}\big(X_i \leq X_{(\lceil n\alpha \rceil)}\big) \\
&= \frac{\lceil n\alpha \rceil}{n}
\end{aligned}$$

as promised. Moreover, it is not hard to see that this value satisfies the inequalities

$$\alpha \leq \frac{\lceil n\alpha \rceil}{n} \leq \alpha + \frac{1}{n},$$

proving all statements at once.                                                    □

## 2.3.2  Transductive framework

The original approach to conformal prediction was introduced at the end of last century in Gammerman, Vovk, and Vapnik (1998); Saunders, Gammerman, and Vovk (1999); Vovk, Gammerman, and Saunders (1999). The central object of interest in this framework (except for the data or distribution themselves) is given by a function that measures or indicates how much a given data point differs from the elements of some hold-out data set, where the latter is often called the **validation** or **calibration set**: it measures the 'nonconformity' of data points.

**Definition** 2.20 (**Nonconformity measure**).  A nonconformity measure on a set $\mathcal{X} \times \mathcal{Y}$ is a measurable function [a]

$$A : (\mathcal{X} \times \mathcal{Y})^* \times \mathcal{X} \times \mathcal{Y} \to \mathbb{R}. \tag{2.27}$$

We will often write the evaluation of a nonconformity measure as follows:

$$A_{\mathcal{D}}(x, y) := A(\mathcal{D}, x, y). \tag{2.28}$$

Given a data set $\mathcal{D} \in (\mathcal{X} \times \mathcal{Y})^*$ and an element $(x, y) \in \mathcal{X} \times \mathcal{Y}$, the value $A_{\mathcal{D}}(x, y)$ is called the **nonconformity score** of $(x, y)$ (with respect to $\mathcal{D}$).

[a]  For a $\sigma$-algebra on the set of multisets, see Example A.13.

Note that this definition is very general and does not refer to any concrete idea of nonconformity yet. In Chapter 3 and onwards, more explicit examples will be given. This definition should, however, already give an idea of how widely applicable the methods of this chapter are.

As mentioned before, the cornerstone of the conformal prediction algorithm is the robustness of rank statistics. Following the theory of the previous section, Property 2.18 and Corollary 2.19 in particular, the algorithm is built around the use of quantiles of an exchangeable sequence of nonconformity scores. The full **transductive conformal prediction** algorithm, presented in pseudocode in Algorithm 1, goes as follows. Fix a data set $\mathcal{D} \in (\mathcal{X} \times \mathcal{Y})^*$, a significance level $\alpha \in [0, 1]$ and a new feature tuple $x \in \mathcal{X}$. For every possible response $y \in \mathcal{Y}$, we initialize a list of nonconformity scores $\mathcal{A}$ and consider the combined data set $\overline{\mathcal{D}} := \mathcal{D} \cup \{(x, y)\}$. Then, for each element $(x_i, y_i) \in \overline{\mathcal{D}}$ of the enhanced data set, we add the nonconformity score $A_{\overline{\mathcal{D}} \setminus \{(x_i, y_i)\}}(x_i, y_i)$ to $\mathcal{A}$. Finally, the response $y$ is included in the prediction set $\Gamma^\alpha(x)$ if $A_{\mathcal{D}}(x, y)$ is less than or equal to the empirical $(1 - \alpha)$-quantile $q_{1-\alpha}(\mathcal{A})$ as defined in Eq. (A.65). Equivalently, $y$ is included in $\Gamma^\alpha(x)$ if the rank of the nonconformity score $A_{\mathcal{D}}(x, y)$ in $\mathcal{A}$ is smaller than or equal to $(1 - \alpha)|\mathcal{A}|$. Whilst the nonconformity measure is the central object of the theory, these ranks are the determining characteristics and, therefore, deserve a proper name. (An explanation for this terminology will be given towards the end of this chapter.)

---

**Definition 2.21 (Conformal $p$-value).** Let $A$ be a nonconformity measure on $\mathcal{X} \times \mathcal{Y}$. Given a data set $\mathcal{D} \in (\mathcal{X} \times \mathcal{Y})^*$, write $\mathcal{D}_i$ for the reduced multiset $\mathcal{D} \setminus \{(x_i, y_i)\}$. The (conformal) $p$-value of an element $(x_i, y_i) \in \mathcal{D}$ is defined as follows:

$$p_i := \frac{\left| \left\{ (x_j, y_j) \in \mathcal{D} \mid A_{\mathcal{D}_j}(x_j, y_j) \geq A_{\mathcal{D}_i}(x_i, y_i) \right\} \right|}{|\mathcal{D}|}. \qquad (2.29)$$

---

Note that in this construction, all functions $A_{\mathcal{D}_i} : \mathcal{X} \times \mathcal{Y} \to \mathbb{R}$ can be aggregated into one function $\overline{A} : (\mathcal{X} \times \mathcal{Y})^{|\mathcal{D}|} \to \mathbb{R}^{|\mathcal{D}|}$. It is not hard to check that this function satisfies the conditions of Property 2.15 and, as such, the nonconformity scores $A_{\mathcal{D}_i}(x_i, y_i)$ are exchangeable whenever $\mathcal{D}$ is sampled from an exchangeable distribution. The theory from the previous section then implies the validity of this approach.

---

**Theorem 2.22 (Marginal validity).** Let $\Gamma^\alpha : \mathcal{X} \times (\mathcal{X} \times \mathcal{Y})^* \to 2^{\mathcal{Y}}$ be a transductive conformal predictor with nonconformity measure $A$ at significance level $\alpha \in [0, 1]$. If the distribution of nonconformity scores is exchangeable, then $\Gamma^\alpha$ is conservatively valid:

$$\mathsf{Prob}\big(Y \in \Gamma^\alpha(X \mid D)\big) \geq 1 - \alpha, \qquad (2.30)$$

where the probability is taken over both $(X, Y)$ and $D$, the latter denoting the random variable corresponding to the auxiliary calibration set. In

---

**Algorithm 1:** Transductive Conformal Prediction

---

**Input**   : Significance level $\alpha \in [0, 1]$, nonconformity measure
$A : (\mathcal{X} \times \mathcal{Y})^* \times \mathcal{X} \times \mathcal{Y} \to \mathbb{R}$, data point $x \in \mathcal{X}$, calibration set
$\mathcal{D} \in (\mathcal{X} \times \mathcal{Y})^*$

**Output** : Conformal prediction set $\Gamma^\alpha(x)$

**1** Initialize an empty list $\mathcal{A}$

**2 foreach** $y \in \mathcal{Y}$ **do**

**3**  $\quad$ Construct the enhanced data set $\overline{\mathcal{D}} \leftarrow \mathcal{D} \cup \{(x, y)\}$

**4**  $\quad$ **foreach** $(x_i, y_i) \in \overline{\mathcal{D}}$ **do**

**5**  $\quad\quad$ (Optional) Train the underlying model $A(\overline{\mathcal{D}} \backslash \{(x_i, y_i)\}, \cdot)$

**6**  $\quad\quad$ Calculate the nonconformity score $A_i \leftarrow A\overline{\mathcal{D}} \backslash \{(x_i, y_i)\}(x_i, y_i)$

**7**  $\quad\quad$ Add the score $A_i$ to $\mathcal{A}$

**8**  $\quad$ **end**

**9**  $\quad$ Calculate the conformal $p$-value $p_y$ using Eq. (2.29).

**10 end**

**11 return** $\{y \in \mathcal{Y} \mid p_y \geq \alpha\}$

---

fact, transductive conformal predictors are also conservatively valid in the stronger sense of Section 2.2.5.

From Property 2.18 we know that if the distribution of the nonconformity scores is atomic, i.e. when ties in the nonconformity scores can arise with nonzero probability, the coverage probability is dominated by a Bernoulli random variable (with parameter $\alpha$). In other words, the model will show overcoverage. However, if ties do not arise, the validity is exact.

**Corollary** 2.23 (**Exact validity**). Suppose that the assumptions of the previous theorem hold. If the probability measure of the nonconformity scores is nonatomic, i.e. if the nonconformity scores are almost surely distinct, the conformal predictor is exactly valid.

With this property in mind, Algorithm 1 can be strengthened in a simple way. To prevent ties from arising, we simple have to inject some (independent) noise and break the ties at random. Instead of using Eq. (2.29), we use the following **smoothed $p$-value**, where ties are uniformly broken:

$$
p_i := \frac{|\{(x_j, y_j) \in \mathcal{D} \mid A_{\mathcal{D}_j}(x_j, y_j) > A_{\mathcal{D}_i}(x_i, y_i)\}|}{|\mathcal{D}|} \tag{2.31}
$$
$$
+ \frac{\tau_i |\{(x_j, y_j) \in \mathcal{D} \mid A_{\mathcal{D}_j}(x_j, y_j) = A_{\mathcal{D}_i}(x_i, y_i)\}|}{|\mathcal{D}|},
$$

with $\tau_i \sim \mathcal{U}([0,1])$ for all $i \leq |\mathcal{D}|$. By the foregoing results, these smoothed TCPs are exactly valid in the strong sense.

**Remark** 2.24 (**Exchangeability**). In practice, the nonconformity scores are often calculated as follows. First, a predictive model $\widehat{f} : \mathcal{X} \to \mathcal{Y}$ is trained on $\mathcal{D}$ and, then, a numerical value is obtained by comparing $\widehat{f}(x)$ to $y$ using some discrepancy measure. However, the training step might violate the assumption of exchangeability since most modern machine learning algorithms inject some randomness related to the order of data points. For example, the result of stochastic gradient descent usually depends on the order in which data points are fed to the algorithm. For this reason it is important in the setting of transductive conformal prediction that the algorithms respect the exchangeability of the data-generating distribution.

### 2.3.3 Inductive framework

Although the properties of the transductive conformal prediction algorithm are very strong and promising, it has a major drawback. For every new observation and every calibration data point, we have to calculate a nonconformity score which might involve (re)training a machine learning model. For finite spaces $\mathcal{Y}$, this is computationally expensive, but still possible. However, whenever $\mathcal{Y}$ is infinite, this algorithm cannot generally terminate in finite time. Moreover, Remark 2.24 also implies that we need to be careful of the choice of hypothesis space in the transductive setting.

The class of inductive conformal predictors (Papadopoulos, Proedrou, Vovk, & Gammerman, 2002) tries to overcome this obstacle and improve on the computational efficiency. Instead of retraining the underlying model for every data point, the data set is divided into two parts: a proper **training** set $\mathcal{T}$ and a proper **calibration** (or **validation**) set $\mathcal{V}$. The underlying model of the nonconformity measure is trained on the training set, resulting in an (**inductive**) **nonconformity measure**

$$A : \mathcal{X} \times \mathcal{Y} \to \mathbb{R}, \tag{2.32}$$

and the nonconformity scores are calculated on the calibration set. This modification is shown in Algorithm 2. The costly retraining step is completely omitted in this approach. Once the model has been trained, it can be used for all future predictions. For further convenience, we note that this algorithm gives rise to two related confidence predictors: a model

$$\Gamma^\alpha(\cdot \mid \cdot) : \mathcal{X} \times (\mathcal{X} \times \mathcal{Y})^* \to 2^{\mathcal{Y}} : (x, \mathcal{V}) \mapsto \Gamma^\alpha(x \mid \mathcal{V}) \tag{2.33}$$

and, when the calibration set $\mathcal{V}$ is fixed and left implicit, a model

$$\Gamma^\alpha : \mathcal{X} \to 2^{\mathcal{Y}} : x \mapsto \Gamma^\alpha(x). \tag{2.34}$$

In general, we will use the latter form, unless the choice of calibration set $\mathcal{V}$ is of importance, since only one calibration set is provided in practice. Moreover, note that the distribution $P_A$ of the nonconformity scores $A(\mathcal{D})$, from here on simply called the **nonconformity distribution**, is simply the pushforward $A_* P$, where $P \in \mathbb{P}\big((\mathcal{X} \times \mathcal{Y})^{|\mathcal{D}|}\big)$ is the data-generating distribution.

**Remark** 2.25 (**Nonconformity measures**). Comparing the inductive approach, with its fixed inductive nonconformity measure, to the transductive approach, it might seem that they are different. However, the former is actually a specific case of the latter. Whenever a general nonconformity measure

$$A : (\mathcal{X} \times \mathcal{Y})^* \times \mathcal{X} \times \mathcal{Y} \to \mathbb{R} \tag{2.35}$$

only depends on $\mathcal{X} \times \mathcal{Y}$, i.e. when it is functionally independent of the choice of calibration set, it reduces to an inductive nonconformity measure and TCP Algorithm 1 reduces to ICP Algorithm 2.

The ICP algorithm makes essential use of the following property which allows to extract concentration inequalities from the empirical quantiles of an existing data set without having to take into account the new element.

---

**Algorithm 2:** Inductive Conformal Prediction

**Input** : Significance level $\alpha \in [0, 1]$, nonconformity measure
$A : \mathcal{X} \times \mathcal{Y} \to \mathbb{R}$, training set $\mathcal{T} \in (\mathcal{X} \times \mathcal{Y})^*$, calibration set
$\mathcal{V} \in (\mathcal{X} \times \mathcal{Y})^*$

**Output:** Conformal predictor $\Gamma^\alpha$

**1** (Optional) Train the underlying model of $A$ on $\mathcal{T}$

**2** Initialize an empty list $\mathcal{A}$

**3** **foreach** $(x_i, y_i) \in \mathcal{V}$ **do**

**4** $\quad$ Calculate the nonconformity score $A_i \leftarrow A(x_i, y_i)$

**5** $\quad$ Add the score $A_i$ to $\mathcal{A}$

**6** **end**

**7** Determine the critical score $a^* \leftarrow q_{(1-\alpha)(1+1/|\mathcal{V}|)}(\mathcal{A})$

**8** **procedure** $\Gamma^\alpha(x : \mathcal{X})$

**9** $\quad$ **return** $\{y \in \mathcal{Y} \mid A(x, y) \leq a^*\}$

**10** **return** $\Gamma^\alpha$

---

**Property** 2.26 (**Inflated quantile**). If $\{X_1, \dots, X_{n+1}\}$ is a collection of exchangeable random variables, then

$$\mathsf{Prob}\left( X_{n+1} \leq \widehat{Q}_n\left( \left(1 + \frac{1}{n}\right)\alpha \right) \right) = \mathsf{Prob}\left( X_{n+1} \leq \widehat{Q}_{n+1}(\alpha) \right), \qquad (2.36)$$

where $\widehat{Q}_n$ denotes the empirical quantile function (A.66).

*Proof*. A simple proof can be found in Y. Romano, Patterson, and Candès (2019). $\qquad\qquad\square$

Because of this property, we do not have to determine the critical score for every new data point. We only have to determine the (inflated) empirical quantile once and compare new nonconformity scores to this value. (Note that this does not remove all computational issues with the algorithm. We come back to this issue in Section 3.3.)

As before in the previous section, the prediction sets satisfy two validity the-orems. The first one states, without further assumptions, that any ICP model is conservatively valid and the second one states that, with the condition on tied scores, the validity is asymptotically exact.

**Theorem** 2.27 (**Marginal validity**). Let $\Gamma^\alpha : \mathcal{X} \to 2^{\mathcal{Y}}$ be an inductive conformal predictor at significance level $\alpha \in [0, 1]$. If the nonconformity scores are exchangeable for any calibration set and any new observation, then $\Gamma^\alpha$ is conservatively valid:

$$\mathsf{Prob}\big(Y \in \Gamma^\alpha(X \mid V)\big) \geq 1 - \alpha\,, \tag{2.37}$$

where the probability is taken over both $(X, Y)$ and $V$.[a]

---

[a]  The latter should be interpreted in light of Remark 2.5.

**Theorem** 2.28 (**Asymptotically exact validity**). As before, assume that the nonconformity scores are exchangeable for any calibration set and any new observation. If the nonconformity distribution is nonatomic, the conformal predictor is asymptotically exactly valid:

$$\mathsf{Prob}\big(Y \in \Gamma^\alpha(X \mid V) \mid |V| = n\big) \leq 1 - \alpha + \frac{1}{n+1}\,. \tag{2.38}$$

If the nonconformity distribution is atomic, we can slightly modify the conformal prediction procedure as in Eq. (2.31). Instead of the conformal $p$-value Eq. (2.29), the smoothed $p$-value is used:

$$
p := \frac{|\{(x', y') \in \mathcal{V} \mid A(x', y') > A(x, y)\}|}{|\mathcal{V}| + 1} \tag{2.39}
$$
$$
+ \frac{\tau\big(|\{(x', y') \in \mathcal{V} \mid A(x', y') = A(x, y)\}| + 1\big)}{|\mathcal{V}| + 1}\,,
$$

where $\tau \sim \mathcal{U}([0, 1])$.

Note that replacing the (inflated) sample quantile by the true (population) quantile $Q_A(1-\alpha)$ of the nonconformity distribution $P_A$ would give the same result. Moreover, if the nonconformity distribution has a density that does not vanish at $Q_A(1-\alpha)$, the sample quantile is a consistent estimator by Property A.45, i.e. it converges in probability to $Q_A(1-\alpha)$. Consequently, in the asymptotic limit, there is even no need to resample calibration sets. This is how (inductive) conformal prediction is applied in practice. It is, for example, reasonable when the data is i.i.d. and $P_A$ has unique quantiles. There-fore, from here on, the dependence on the calibration set will be left implicit.

Away from the asymptotic limit, the remaining issue is, however, that the critical quantile of the calibration set is either large enough, i.e. greater than the $(1-\alpha)$-quantile of $P_A$, or not and, consequently, the model gives intervals that are either too wide or too small. There is no expectation to be taken over all possible calibration sets. Luckily, there are three ways out (two of which are genuine solutions):

1. Turn a blind eye (hopefully with lots of data).

2. Use bootstrap samples for calibration sets (again, when there is sufficient data available).

3. Adopt an online training paradigm.

The first option amounts to having faith in Theorem 2.28 and hoping for the best. This approach, where we assume consistency of the sample quantiles as explained above, is further formalized by the following property (Angelopoulos & Bates, 2023; Vovk, 2012).

---

**Property** 2.29 (**Asymptotic validity**). Let $\Gamma^\alpha : \mathcal{X} \times (\mathcal{X} \times \mathcal{Y})^* \to 2^{\mathcal{Y}}$ be an inductive conformal predictor at significance level $\alpha \in [0, 1]$. If the nonconformity scores are exchangeable for any new observation and any calibration set $\mathcal{V}$, then $\Gamma^\alpha$ is asymptotically (conservatively) valid:

$$\lim_{|\mathcal{V}| \to \infty} P_{X,Y}\big(Y \in \Gamma^\alpha(X \mid \mathcal{V})\big) \geq 1 - \alpha \,, \qquad (2.40)$$

where we explicitly used the probability measure $P_{X,Y}$ to indicate that, here, the probability is only taken over new instances and not calibration sets as in Theorem 2.27. More specifically, the coverage probability follows a Beta distribution (Example A.35):

$$P_{X,Y}\big(Y \in \Gamma^\alpha(X \mid V)\big) \sim \mathsf{Beta}(k_\alpha, |V| + 1 - k_\alpha) \,, \qquad (2.41)$$

where $k_\alpha := \lceil (1 - \alpha)(1 + |V|) \rceil$.

---

It follows, that for suitably large calibration sets, we can expect the inductive conformal predictor to give approximately valid results. However, in practice, the available data sets might not always be very large. Figure 2.5 shows some scatter plots of the empirical coverage (2.6) as a function of the calibration set size for different synthetic experiments. In all these experiments,

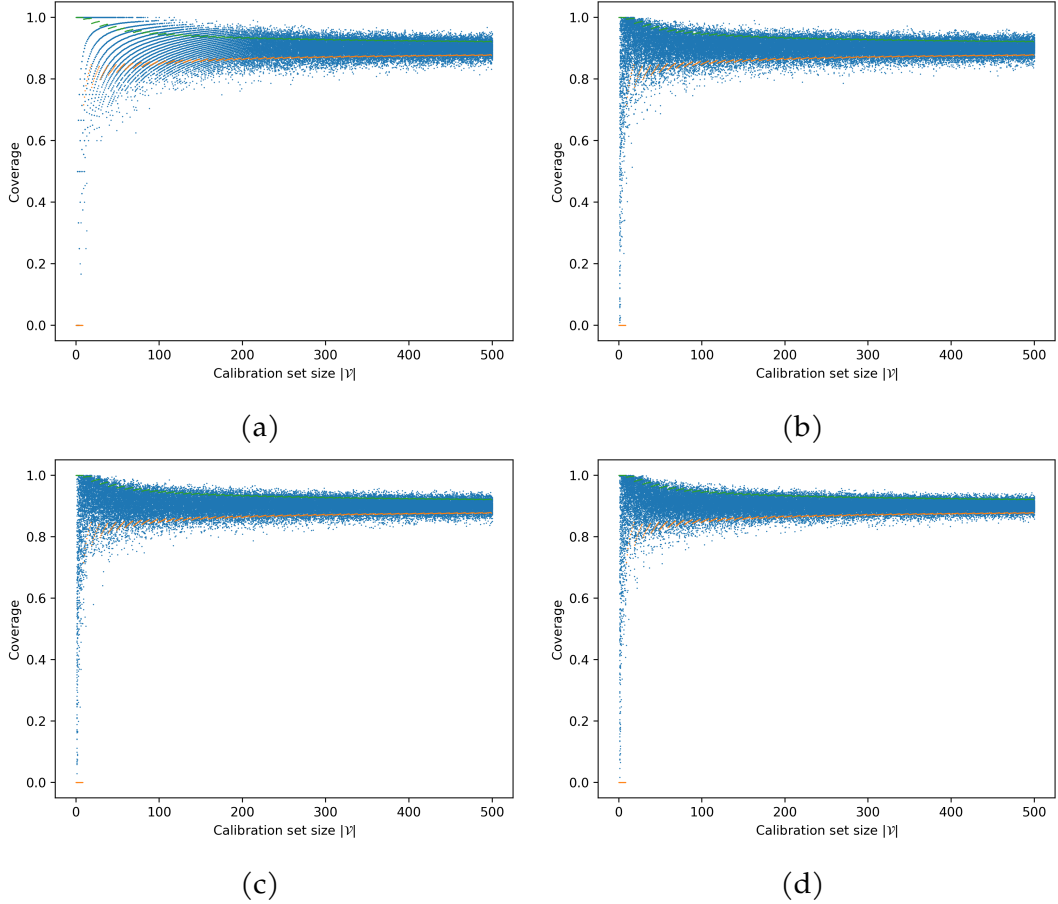

Figure 2.5: Empirical coverage of inductive conformal predictors in function of calibration set size $|\mathcal{V}|$ for synthetic experiments with different test set sizes $|\mathcal{T}|$. The values of $|\mathcal{T}|$ vary as follows: $|\mathcal{V}|, 500, 1000$ and $10000$.

the data sets are sampled as follows:

$$
\begin{aligned}
X &\sim \mathcal{N}\big((0,2,4,6,8),(2,3,4,5,6)\big), \\
Y &\sim \mathsf{Exp}\big(|\mathsf{mean}(X)|\big),
\end{aligned}
\tag{2.42}
$$

where $\mathsf{mean} : \mathbb{R}^d \to \mathbb{R}$ is applied in a rowwise manner. The mean was then estimated using linear regression on a training set of size 500. For every integer $k \in \{1, \dots, 500\}$, 100 calibration sets of size $|\mathcal{V}| = k$ were sampled (hence, $5 \times 10^4$ data points in total). The four subfigures together with Table 2.1 show the results for different sizes of test sets: $|\mathcal{T}| \in \{|\mathcal{V}|, 500, 1000, 10000\}$. The orange and green dots in every figure indicate the 5% and 95% quantiles of the beta distribution from Eq. (2.41), respectively. The fraction of data points that are captured by the 90% confidence intervals of the beta distribution is shown in Table 2.1.

Table 2.1: Percentage of data points, representing the empirical coverage of an ICP model, captured by the 90% confidence interval of the beta distribution in Eq. (2.41). The first column indicates the size of the training sets.

|            | Set size $|\mathcal{T}|$ | Coverage |
|------------|:------------------------:|:--------:|
| Figure 2.5a | $|\mathcal{V}|$          | 0.749    |
| Figure 2.5b | $5 \times 10^2$          | 0.810    |
| Figure 2.5c | $1 \times 10^3$          | 0.852    |
| Figure 2.5d | $1 \times 10^4$          | 0.893    |

The second option tries to improve on this naive application of consistency guarantees. If enough data is available, bootstrap samples will give good approximations to real data samples and we can expect the confidence predictor to be valid in the general sense. The last solution, given by the online learning paradigm (Vovk et al., 2022), is applicable in every situation. As for ordinary online learning methods, the main idea consists of adding data points to the calibration set after they have been processed.

**Method** 2.30 (**Online ICP**). Consider a sequence of integers $(m_n)_{n \in \mathbb{N}} \subseteq \mathbb{N}_0$. For every integer $l > m_0$, the prediction sets are constructed as follows:

1. Find the smallest $k \in \mathbb{N}$ such that $m_k < l \leq m_{k+1}$.

2. If $l = m_k + 1$, construct the nonconformity measure $A_k : \mathcal{X} \times \mathcal{Y} \to \mathbb{R}$ using the data set $\{(x_1, y_1), \ldots (x_{m_k}, y_{m_k})\}$.

3. Construct the prediction set $\Gamma^\alpha(x)$ using the inductive conformal predictor derived from $A_k : \mathcal{X} \to \mathbb{R}$ and the calibration set $\mathcal{V}_k := \{(x_{m_k+1}, y_{m_k+1}), \ldots (x_{l-1}, y_{l-1})\}$.

Online ICP models are conservatively valid in the strong sense.

**Property** 2.31. Let $P \in \mathbb{P}((\mathcal{X} \times \mathcal{Y})^{\mathbb{N}})$ be an exchangeable probability distribution and let $err_i$ be the event that $y_i \notin \Gamma^\alpha(x_i)$ for $\Gamma^\alpha$ an online ICP. The sequence $err_1(\Gamma^\alpha, P)$, $err_2(\Gamma^\alpha, P)$, ... is dominated by a sequence of independent Bernoulli random variables with parameter $\alpha$. Moreover, smoothed online ICPs are exactly valid in the strong sense in that the errors $err_i(\Gamma^\alpha, P)$ are independently sampled from a Bernoulli distribution

| with parameter $\alpha$.

Note that the asymptotic validity result 2.29 above can also be seen as an approximation of this property, where the calibration set is much larger than the set of data points for which a prediction region is constructed and, hence, the empirical nonconformity distribution is not expected to shift much in each step.

### 2.3.4    Illustration

Given the rather technical treatment of conformal prediction in the preceding sections, a nontechnical, illustrative example is in order. We will also already make use of this situation to introduce some motivations for Chapters 3 and 4. To this end, consider a situation where we are presented with a set of objects, each described by their weight, and we are asked to predict the weight of the next object that will be observed. The graph in the left panel of Fig. 2.6 shows the histogram of a sample of 20 objects with (integer) weights between 1 and 20.

If we would like to construct a prediction set for the weight of the next object that we would observe, the first thing (inductive) conformal prediction (Algorithm 2) tells us to do, is to find the inflated $(1-\alpha)$-quantile of the weights. Given the twenty objects in Fig. 2.6, we first sort the weights and, choosing $\alpha = 0.1$ for the significance level, the critical score is given by $a^* = 17$. The dashed line in the right panel of Fig. 2.6 indicates a (cumulative) probability of 0.9. The algorithm then tells us that the prediction set consists of all objects for which the weight is smaller than this critical value. Accordingly, we predict that the weight of the next object will lie between 1 and 17 (boundaries included).

Now, if we would find out that the data is sampled from a (discrete) uniform distribution on $\{1, \dots, 20\}$ (which is indeed how this sample was generated), we also immediately see Property 2.29 at play. If we would repeatedly use the obtained prediction set $\Gamma^\alpha(x) \equiv \{1, \dots, 17\}$, it would only cover the observed weights 85% of the time. However, it is already a reasonable approximation.

Of course, some could object at this point that this is a very conservative approach, its validity notwithstanding. The constructed prediction set covers most of the instance space, which could give a sense of meaninglessness.

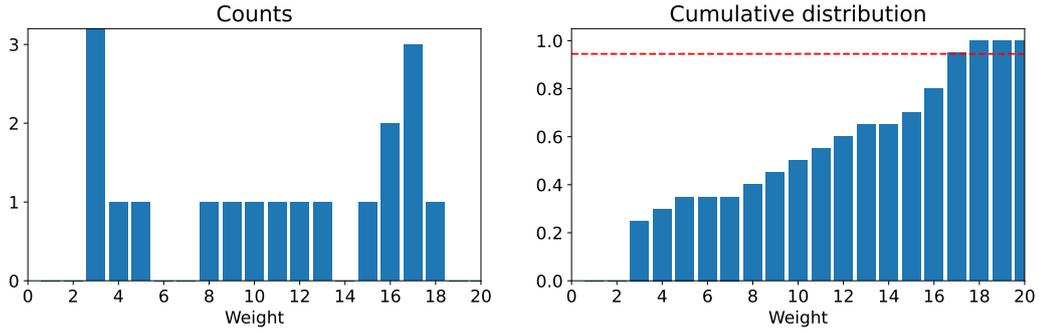

Figure 2.6: Sample of object weights. A histogram of the weights is shown in the left panel and a plot of the cumulative distribution is shown in the right panel.

Note, however, that without additional information about the distribution of the objects' weights, the oracle prediction set of the uniform distribution would also cover as much of the instance space. If we would have access to more information, for example where $x$ might contain information about the size, the composition or the origin, we could try to do better. In this case, the overall distribution over the weights $P(\text{weight})$ might be uniform, but the conditional distribution $P(\text{weight} \mid x)$ might be something completely different. Given this additional information, we could try to build a weight estimate $\widehat{\omega} : \mathcal{X} \rightarrow \{1, \dots, 20\}$ and apply the conformal prediction method to, for example, the residuals (this will be the content of Chapter 3). These are shown in Fig. 2.7. The plot in the right panel shows that the critical quantile is now given by 4 and, consequently, the resulting prediction sets are given by

$$\Gamma^{\alpha}(x) = \{\widehat{\omega}(x) - 4, \widehat{\omega}(x) - 3, \dots, \widehat{\omega}(x) + 3, \widehat{\omega}(x) + 4\}. \tag{2.43}$$

Figure 2.8 shows a scatter plot of the estimated weights in function of the true weights, where some noise has been added in the horizontal direction as to make the different points easier to distinguish. The bars indicate the prediction sets produced by the conformal prediction method. It can be seen that, except for one object, all prediction sets cover the true weight, i.e. the empirical coverage is 95%. However, now that we have access to more information, we could also look at the conditional behaviour (this will be the content of Chapter 4). For example, if we would find out that the conditional distribution $P(\text{weight} \mid x)$ depends on whether $x \leq 15$ or $x > 15$, we could look at the coverage for these two regions. In this case, we immediately see that the confidence level of 90% is not attained for the group with $x > 15$. In fact, if we would be told that the residuals for this subgroup follow a Poisson distribution[10] with mean parameter $\lambda = 3$, we could even conclude that the critical score 4 would lead to undercoverage at the population level, since the 90th percentile of this distribution is 5.

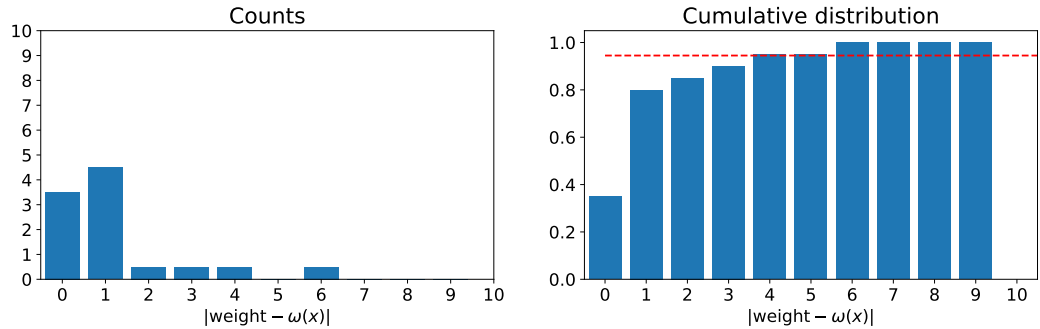

Figure 2.7: A histogram of the model residuals is shown in the left panel and a plot of the cumulative distribution is shown in the right panel.

### 2.3.5   Decomposition of uncertainty

In Section 2.2.2, we considered the different types of uncertainty that can arise in practice: aleatoric and epistemic uncertainty. Whereas the distinction is fairly straightforward from a philosophical or definitional point of view, Remark 2.2 left aside, finding an actual decomposition of a given uncertainty estimate turns out to be much harder.

The general purpose of conformal prediction is simply to aggregate all uncertainty and construct a prediction set satisfying a probabilistic condition. Nowhere in the definition of conformal prediction did we see a distinction between uncertainty coming from the data-generating process, the aleatoric part, and uncertainty coming from inaccurate estimates, the epistemic part. The reason is that at the level of nonconformity measures, the level at which conformal prediction operates, these uncertainties have already been aggregated.

Consider Fig. 2.9 as an example. Here, both the true density function and the (normalized) histogram of a sample (of size $10^5$) are shown for the nonconformity measure $A(x, y) = |\hat{\mu}(x) - y|$, where the point predictor is given

---

[10] If we would know that the residuals are necessarily positive, we could also make a better informed choice of nonconformity measure, which would further improve the inefficiency (such a choice will be covered in Section 3.3.2).

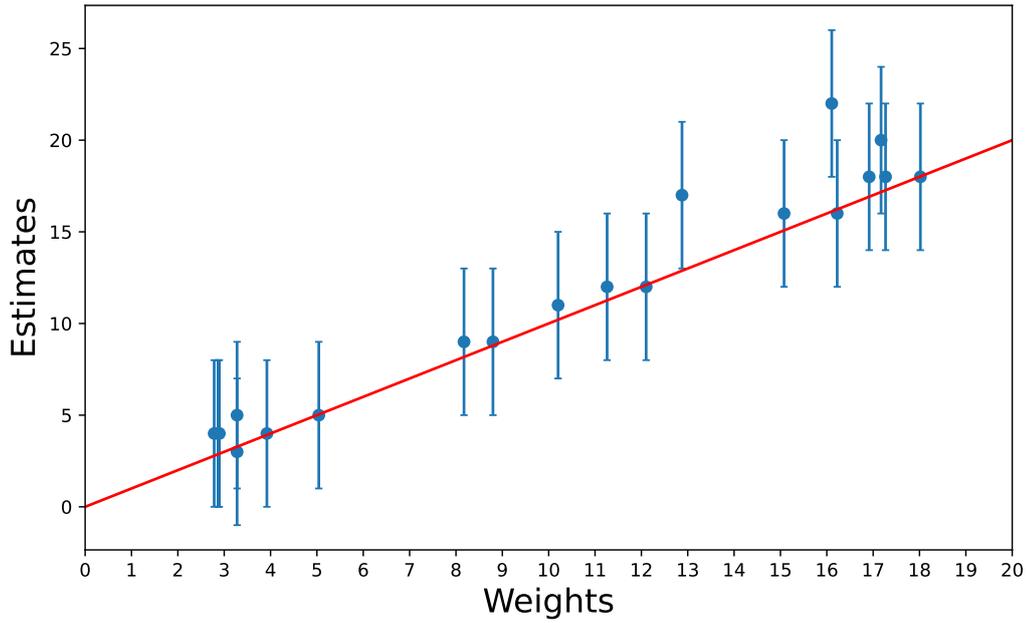

Figure 2.8: Estimates of the object weights in function of the true weights. The vertical bars indicate the prediction intervals. (Some noise has been added along the horizontal axis to improve readability.)

by projection on the first component:

$$\hat{\mu}(x) = x^1 \,, \tag{2.44}$$

and the data-generating process is chosen as follows:

$$X \equiv (X^1, X^2) \sim \mathcal{U}\big([0,1]^2\big) \,, \tag{2.45}$$

$$Y \mid X^1, X^2 \sim \mathcal{N}(X^1 + X^2, 1) \,. \tag{2.46}$$

If we were able to fully decompose the uncertainty, the normal distribution would be the aleatoric uncertainty and the difference $\mathsf{E}\big[Y \mid X^1, X^2\big] - \hat{\mu}(X) = X^2$ would be the epistemic uncertainty. An explicit expression for the density function is easy to find using the change-of-variables formula (Corollary A.31):

$$\begin{aligned}
f_A(a) &= \int_0^1 \int_0^1 \frac{1}{\sqrt{2\pi}} \exp\!\left(-\frac{\left|(x^1 - a) - (x^1 + x^2)\right|^2}{2}\right) \mathrm{d}x^1 \, \mathrm{d}x^2 \\
&\quad + \int_0^1 \int_0^1 \frac{1}{\sqrt{2\pi}} \exp\!\left(-\frac{\left|(x^1 + a) - (x^1 + x^2)\right|^2}{2}\right) \mathrm{d}x^1 \, \mathrm{d}x^2 \\
&= \int_0^1 \int_0^1 \frac{1}{\sqrt{2\pi}} \exp\!\left(-\frac{\left|a + x^2\right|^2}{2}\right) \mathrm{d}x^1 \, \mathrm{d}x^2
\end{aligned}$$

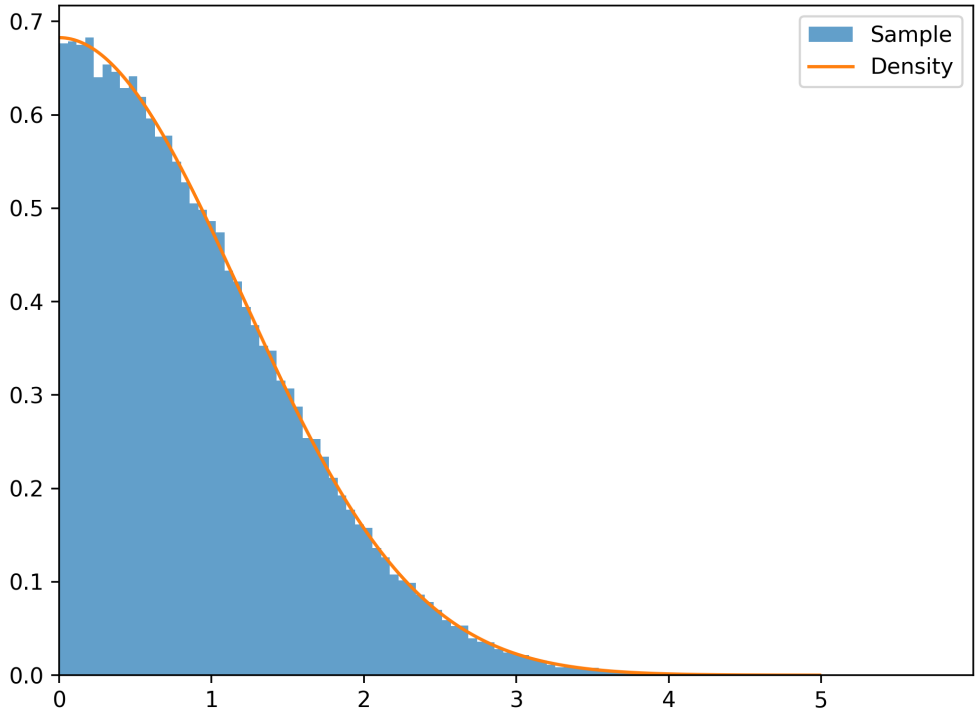

Figure 2.9: Probability density function and sample histogram of the non-conformity scores for the model $\hat{\mu}(x) = x^1$ with data-generating process $Y \mid X^1, X^2 \sim \mathcal{N}(X^1 + X^2, 1)$ and $(X^1, X^2) \sim \mathcal{U}([0, 1]^2)$.

$$+ \int_0^1 \int_0^1 \frac{1}{\sqrt{2\pi}} \exp\left(-\frac{|a - x^2|^2}{2}\right) \mathrm{d}x^1 \, \mathrm{d}x^2$$

$$= \frac{1}{2}\left[\operatorname{erf}\left(\frac{1 - a}{\sqrt{2}}\right) + \operatorname{erf}\left(\frac{1 + a}{\sqrt{2}}\right)\right], \tag{2.47}$$

where the **error function** erf is defined as follows:[11]

$$\operatorname{erf}(x) := \frac{2}{\sqrt{\pi}} \int_0^x e^{-x^2} \, \mathrm{d}x. \tag{2.48}$$

This derivation should make it clear that there is no hope of decomposing the resulting uncertainty in a sensible way (without further information). There are many different combinations of distributions that give rise to the same result.

---

[11] It is equal to the probability that a point sampled from a normal distribution with mean 0 and standard deviation $1/\sqrt{2}$ lies in the interval $]-x, x[$.

### 2.3.6 Applications

In this section, we will cover some applications of conformal prediction to other fields of data science and real-world settings.

- Active learning and Bayesian optimization: Corrigan, Hopcroft, Narvaez, and Bendtsen (2020); Matiz and Barner (2020); Romero (2019); Stanton, Maddox, and Wilson (2023).

- Biology and medical applications: Alvarsson, McShane, Norinder, and Spjuth (2021); Fannjiang, Bates, Angelopoulos, Listgarten, and Jordan (2022); C. Lu, Lemay, Chang, Höbel, and Kalpathy-Cramer (2022); Sun et al. (2017); Vazquez and Facelli (2022); Wieslander et al. (2020).

- Autonomous systems and safety-critical applications: de Grancey et al. (2022); Lindemann, Cleaveland, Shim, and Pappas (2023); Muthali et al. (2023); Sun et al. (2023).

- Finance: Bastos (2024); J. V. Romano (2022); Wisniewski, Lindsay, and Lindsay (2020).

It should go without saying that many more applications for conformal prediction exist and that the above list is not exhaustive.

## 2.4 Hypothesis testing

### 2.4.1 Conformal *p*-values

In the preceding sections, conformal prediction has been introduced as a constructive approach for quantifying uncertainty given the basic premise of exchangeability. In this section, an alternative interpretation of conformal prediction is given. In a sense, this is actually how conformal prediction was first introduced. The construction of prediction regions using conformal prediction can be reformulated in the language of hypothesis testing. To this end, we first quickly recall the basics of hypothesis testing à la Fisher–Neyman–Pearson (Fisher, 1925; Lehmann, 1993; Neyman & Pearson, 1933).

Assume we are given a null hypothesis $H_0$ that states that the data is sampled from a distribution $P$, which belongs to some set $\Omega \subseteq \mathbb{P}(\mathcal{X})$. (The alternative hypothesis $H_1$ states that $P \notin \Omega$.) The procedure consists of determining a 'critical region' $C \subseteq \mathcal{X}$ such that if an element from $C$ is observed, the null hypothesis $H_0$ is rejected. This region is chosen such that, under the null hypothesis $H_0$,

$$P(X \in C) = \alpha \qquad (2.49)$$

for some predetermined significance level $\alpha \in [0, 1]$, where lower values of $\alpha$ correspond to more powerful or more stringent tests. If $P$ assigns a sufficiently small probability to an event (and it is the task of the statistician to determine what sufficiently means here), but it still occurs, it gives evidence against $P$ being the true distribution. This idea is based on Cournot's principle (Shafer, 2021; Shafer & Vovk, 2006) which states that events with small enough probabilities should be regarded as *morally impossible*. In the style of Fisher, such a test is usually performed by calculating a certain test statistic for which the distribution is known under the null hypothesis and the value of the test statistic is then compared to a certain quantile of this distribution.

In the case of conformal prediction, the following hypothesis test is of interest:

- Null hypothesis: "The data set $\mathcal{D} \cup \{(x, y)\}$ is distributed exchangeably."

- Test statistic: the nonconformity score $A_\mathcal{D}(x, y)$.

By the results of the previous sections, Corollary 2.19 in particular, the ranks of the nonconformity scores should be distributed uniformly under the null hypothesis. Consequently, a conformal $p$-value (2.29) below the chosen significance level gives evidence against the null hypothesis, i.e. against exchangeability.

This interpretation of conformal prediction also allows to rephrase confidence prediction as a *decidability* problem instead of a *computability* problem. In critical situations, domain experts or end users often have a good idea of critical thresholds because of domain knowledge, e.g. the maximum height of waves that a dyke can withstand or the maximum pressure a machine can handle. In these cases, it suffices to decide whether this threshold is probable or not and the entire prediction set need not even be constructed, avoiding the entire computability problem of some conformal predictors altogether.

## 2.4.2 Anomaly detection

The hypothesis testing interpretation of conformal prediction also allows to check whether the predictive model and associated conformal predictor are actually to be trusted. Every time a new prediction is made and the ground truth is observed, the evidence against the exchangeability hypothesis can be added to an evidence pool and once a predetermined threshold is exceeded, the model should be discarded or retrained. This leads to the notion of exchangeability martingales (Vovk et al., 2022, 2021). In fact, this is the same idea as used by Wald (1945) in his *sequential probability ratio test*.

> **Definition** 2.32 (**Exchangeability martingale**). If a stochastic process is a supermartingale (Definition A.49) for all exchangeable probability distributions, it is called an exchangeability supermartingale. If exchangeable distributions are replaced by power distributions — probability distributions corresponding to i.i.d. sequences — the notion of **randomness supermartingale** is obtained.

The exchangeability supermartingales that are relevant in the context of conformal prediction are called **conformal test martingales** (Vovk et al., 2022, 2021). These are real-valued stochastic processes $(S_n)_{n\in\mathbb{N}}$ generated in the following way:[12]

$$S_n = F(p_1, \dots, p_n),\tag{2.50}$$

where $\{p_i\}_{i\le n}$ are the conformal $p$-values given by a fixed conformal predictor, for a measurable function $F : [0,1]^* \to [0, +\infty]$ satisfying:

1. **Normalization**: $F(\emptyset) = 1$, and

2. **Marginalization**:

$$\int_0^1 F(p_1, \dots, p_{n-1}, p)\,\mathrm{d}p = F(p_1, \dots, p_{n-1})\tag{2.51}$$

   for all $(p_1, \dots, p_{n-1}) \in [0,1]^{n-1}$.

These conditions say that $F$ can be constructed from a martingale (Definition A.49) with initial condition $F_0 = 1$. Constructing a conformal test martingale (CTM) is the same as constructing such a 'betting martingale' $F$. The

---

[12] As with conformal predictors, a randomized extension can be considered by allowing for smoothed predictors. This would also lead to martingales instead of supermartingales.

most straightforward way is the following choice:

$$F(p_1, \dots, p_n) = f_1(p_1) \cdots f_n(p_n) \,, \tag{2.52}$$

where $f_n : [0, 1] \to [0, +\infty[$ is a measurable function for all $n \in \mathbb{N}$ satisfying the normalization condition

$$\int_0^1 f_n(p) \, \mathrm{d}p = 1 \,, \tag{2.53}$$

i.e. $\mathsf{E}_{\mathcal{U}}[f_n] = 1$ for all $n \in \mathbb{N}$. It is this uniform expectation that gives the relation to conformal prediction. Namely, recalling Property 2.18, the $p$-values of a conformal predictor will be distributed uniformly under the null hypothesis of exchangeability and, accordingly, the $\mathcal{U}$-martingale property will be satisfied. Two simple examples for betting functions are given in the following example.

**Example** 2.33 (**Power martingale**).  Let $\varepsilon \in [0, 1]$ be fixed.  The $\varepsilon$-power martingale is defined as follows:

$$S_n^{\varepsilon}(p_1, \dots, p_n) := \prod_{i=1}^n \varepsilon p_i^{\varepsilon - 1} \,. \tag{2.54}$$

By integrating over $\varepsilon$, a mixture of power martingales is obtained:

$$S_n(p_1, \dots, p_n) := \int_0^1 S_n^{\varepsilon}(p_1, \dots, p_n) \, \mathrm{d}\varepsilon \,. \tag{2.55}$$

Now, given the choice of a CTM, the idea is as follows.  Data is observed in an online fashion, i.e. as a sequence $((x_n, y_n))_{n \in \mathbb{N}}$, and, after each data point $(x_i, y_i)$ is observed, its conformal $p$-value is calculated and the CTM is updated.  Given a predetermined threshold $\lambda \in \mathbb{R}^+$, an alert that exchangeability might be violated is issued once the martingale exceeds $\lambda$.  This approach is based on the Doob–Ville inequality (Property A.50).  This procedure is summarized in Algorithm 3.

## 2.5    Extensions

In the final section of this chapter, we consider some extensions of the standard conformal prediction framework as introduced above.  Although these extensions are applicable in a wide variety of situations, it might be better to return to this section only after reading Chapter 3.

---

**Algorithm 3:** Conformal test martingale

---

**Input** : Threshold $\lambda \in \mathbb{R}^+$, nonconformity measure $A : \mathcal{X} \times \mathcal{Y} \to \mathbb{R}$,
betting martingale $F : [0,1]^* \to [0, +\infty]$, training set
$\mathcal{T} \in (\mathcal{X} \times \mathcal{Y})^*$ and calibration set $\mathcal{V} \in (\mathcal{X} \times \mathcal{Y})^*$

**Output :** Sequential exchangeability test $\mathcal{E}$

**1** (Optional) Train the underlying model of $A$ on $\mathcal{T}$
**2** Initialize an empty list $\mathcal{A}$

**3** **foreach** $(x_i, y_i) \in \mathcal{V}$ **do**
**4**  $\quad$ Calculate the nonconformity score $A_i \leftarrow A(x_i, y_i)$
**5**  $\quad$ Add the score $A_i$ to $\mathcal{A}$
**6** **end**

**7** **procedure** $\mathcal{E}\big((x,y)_{n \in \mathbb{N}} : (\mathcal{X} \times \mathcal{Y})^{\mathbb{N}}\big)$
**8**  $\quad$ **foreach** $n \in \mathbb{N}$ **do**
**9**  $\quad\quad$ Calculate the conformal $p$-value $p_n$ from $\mathcal{A}$
**10**  $\quad\quad$ Update the CTM $M_n \leftarrow F(p_0, \ldots, p_n)$

**11**  $\quad\quad$ **if** $M_n \geq \lambda$ **then**
**12**  $\quad\quad\quad$ Issue an alert: **"Nonexchangeability detected"**
**13**  $\quad\quad$ **else**
**14**  $\quad\quad\quad$ Move to the next data point
**15**  $\quad\quad\quad$ (Optional) Retrain the underlying model of $A$
**16**  $\quad\quad$ **end**
**17**  $\quad$ **end**

**18** **return** $\mathcal{E}$

---

**Remark** 2.34 (**Omissions**). This section tries to cover some relevant settings beyond the basic framework of conformal prediction, but it is in no way exhaustive. The main omissions are the 'Prediction Powered Inference' framework by Angelopoulos, Bates, Fannjiang, Jordan, and Zrnic (2023) and the 'Risk Control' framework (Angelopoulos, Bates, Candès, Jordan, & Lei, 2022; Angelopoulos, Bates, Fisch, Lei, & Schuster, 2024). The former ties in with the discussion in Section 2.2.4 and allows to per-form inference instead of prediction. In this approach, an estimator is applied to not necessarily correctly labeled data and, in a second step, the result is calibrated on crisp data, similar to how conformal prediction works[a]. The risk control framework allows to obtain prediction sets satisfying validity guarantees with respect to a more general class of metrics besides the Type-1 error rate.

---

[a]  Note, however, that, strictly speaking, it is not a conformal prediction method.

### 2.5.1  Cross-validation

One approach to obtaining heteroskedastic prediction regions is to start from a collection of models such as ensemble methods (Section A.4.2). Ensembles that make use of bagging, such as random forests, give rise to their own natural choice of both homoskedastic and heteroskedastic nonconformity measures. As explained in Definition A.70, every individual submodel in a bagged ensemble only uses a subset of the training set and, therefore, we can use the associated out-of-bag (OOB) estimate for every training instance, i.e. the subensemble for which this instance was not used during training, to produce an independent prediction. This allows the full data set to act as both a training and calibration set in stark contrast to the situation of ordinary ICP methods, which are restricted by the exchangeability assumption. To this end, consider an inductive nonconformity measure of the form

$$A(x, y) = \Delta\big(y, \hat{y}(x)\big) \tag{2.56}$$

for some discrepancy measure $\Delta : \mathcal{Y} \times \mathcal{Y} \to \mathbb{R}$. The following asymmetric approach was introduced in Johansson, Boström, Löfström, and Linusson (2014) for the specific case of univariate regression problems (as treated in the next chapter), where $\Delta$ is given by the RMSE:

$$\begin{aligned} A_{\mathrm{cal}}(x_i, y_i) &:= \Delta\big(y_i, \hat{y}_{(i)}(x_i)\big) \\ A_{\mathrm{test}}(x, y) &:= \Delta\big(y, \hat{y}(x)\big), \end{aligned} \tag{2.57}$$

where $\hat{y}_{(i)}$ denotes the OOB predictor for the data point $(x_i, y_i) \in \mathcal{D}$. Although it cannot be formally proven that this procedure satisfies the same validity properties as ordinary ICP, the fact that OOB estimates often overestimate the uncertainty makes it plausible that the above measures will lead to (conservatively) valid prediction intervals. To obtain formal guarantees, the authors later introduced a small modification (Boström, Linusson, Löfström,

& Johansson, 2017). Instead of using the asymmetric definition above, they proposed the following scheme for every new observation $x \in \mathcal{X}$:

1. Choose a random training instance $(x_i, y_i) \in \mathcal{Y}$.

2. Use the OOB predictor $\hat{y}_{(i)}$ to calculate the nonconformity score of a candidate $(x, y)$.

3. Use the (complementary) calibration set $\mathcal{D} \setminus \{(x_i, y_i)\}$ to construct a prediction interval for $x$.

Although this method is theoretically valid, it does require the recalculation of the critical value for every new data point, thereby introducing some extra computational overhead. Furthermore, if this procedure is also used to extract a point prediction in the process of constructing a prediction region, the fact that the prediction is only made using a fraction of the total data set might lead to a deterioration of the predictive performance.

This approach to conformal prediction can also be extended to a framework in which the full data set is used to calibrate the model through a $k$-fold or leave-one-out approach (Vovk, 2015; Vovk, Nouretdinov, Manokhin, & Gammerman, 2018). For this reason, these methods are called **cross-conformal predictors** (CCPs). The splitting of the data set $\mathcal{D} \in (\mathcal{X} \times \mathcal{Y})^*$ is based on a clustering function

$$\mathcal{S} : \mathcal{D} \rightarrow [k] \tag{2.58}$$

for some $k \in \mathbb{N}_0$. The different folds are then given by the fibres $\mathcal{D}_s := \mathcal{S}^{-1}(s)$ for all $s \in [k]$. This approach is shown schematically in Algorithm 4. Here, the conformal $p$-value is defined by

$$p_y(x) = \frac{\sum_{s \in [k]} \left| \left\{ (x', y') \in \mathcal{D}_s \mid A_{\mathcal{D} \setminus \mathcal{D}_s}(x', y') \geq A_{\mathcal{D} \setminus \mathcal{D}_s}(x, y) \right\} \right| + 1}{|\mathcal{D}| + 1} \tag{2.59}$$

instead of

$$p_y(x) = \frac{|\{(x', y') \in \mathcal{V} \mid A(x', y') \geq A(x, y)\}| + 1}{|\mathcal{V}| + 1} \tag{2.60}$$

in the inductive case with Eq. (2.29).

Although this method tries to interpolate between transductive and inductive conformal prediction and might lead us to believe that it combines the best of both worlds, no direct coverage guarantees exist. The best we can do is the following property.

---

**Algorithm 4:** Cross-Conformal Prediction

---

**Input** : Significance level $\alpha \in [0, 1]$, nonconformity measure
$A : (\mathcal{X} \times \mathcal{Y})^* \times \mathcal{X} \times \mathcal{Y} \to \mathbb{R}$, data set $\mathcal{D} \in (\mathcal{X} \times \mathcal{Y})^*$, splitting
strategy $\mathcal{S}$

**Output:** Cross-conformal predictor $\Gamma^\alpha$

---

**1  foreach** $s \in [k]$ **do**
**2**      Select the data set $\mathcal{D}_s \leftarrow \mathcal{D} \cap \mathcal{S}^{-1}(s)$
**3**      (Optional) Train the underlying model of $A$ on $\mathcal{D} \backslash \mathcal{D}_s$

**4**      Initialize an empty list $\mathcal{A}_s$
**5**      **foreach** $(x_i, y_i) \in \mathcal{D}_s$ **do**
**6**          Calculate the nonconformity score $A_i \leftarrow A_{\mathcal{D} \backslash \mathcal{D}_s}(x_i, y_i)$
**7**          Add the score $A_i$ to $\mathcal{A}_s$
**8**      **end**
**9  end**

**10  procedure** $\Gamma^\alpha(x : \mathcal{X})$
**11**      Initialize an empty list $\mathcal{R}$
**12**      **foreach** $y \in \mathcal{Y}$ **do**
**13**          Calculate the conformal $p$-value $p_y$ using Eq. (2.59)
**14**          Add $y$ to $\mathcal{R}$ if $p_y \geq \alpha$
**15**      **end**
**16**      **return** $\mathcal{R}$

**17  return** $\Gamma^\alpha$

---

**Property** 2.35. All cross-conformal predictors at significance level $\alpha \in [0, 1]$ are conservatively valid at the $(1 - 2\alpha)$-level for $k \ll |\mathcal{V}|$:

$$\mathsf{Prob}\Big(Y \in \Gamma^\alpha_{\mathrm{CCP}}(X \mid V) \,\Big|\, |V| = n\Big) \geq 1 - 2\alpha - 2(1 - \alpha)\frac{1 - 1/k}{1 + n/k}. \quad (2.61)$$

*Proof.* Noted in Vovk et al. (2018) and proven in Barber, Candès, Ramdas, and Tibshirani (2021b). The gist of the proof, following a result in Vovk and Wang (2020), is that the (weighted) average of $p$-values is again a $p$-value up to a factor of 2. The correction term is due to the

> cross-conformal $p$-value only being a (weighted) average up to the term $\frac{k-1}{k+n}$. $\qquad\square$

However, as shown in Barber et al. (2021b), the result by Vovk and Wang (2020) can also be leveraged to modify the CCP algorithm in such a way as to obtain a nonasymptotic validity result at the $2\alpha$-level. The cross-conformal $p$-value can be replaced by the true (weighted) average of the $p$-values per fold:

$$\tilde{p}_y(x) := \sum_{s \in [k]} \frac{\left| \left\{ (x', y') \in \mathcal{D}_s \mid A_{\mathcal{D} \setminus \mathcal{D}_s}(x', y') > A_{\mathcal{D} \setminus \mathcal{D}_s}(x, y) \right\} \right| + 1}{|\mathcal{D}_s| + 1} . \qquad (2.62)$$

### 2.5.2 Conformal predictive systems

In Section 2.3, we discussed two possible approaches to conformal prediction. First, the transductive approach, with its strong statistical properties but equally strong computational inefficiency. Second, the inductive approach, which is computationally much easier to handle but, as a consequence, has less powerful guarantees. However, both of these approaches only led to prediction regions for a predetermined significance level $\alpha \in [0, 1]$.

Building upon prior research (Schweder & Hjort, 2016; Shen, Liu, & Xie, 2018), where the general notion of 'predictive distributions' was studied, Vovk, Shen, Manokhin, and Xie (2017) used the framework of conformal prediction to construct such distribution functions with statistical guarantees. We give the general definition for completeness' sake.[13]

> **Definition** 2.36 (**Predictive distribution**). Let $(\mathcal{X}, \Sigma, P)$ be a totally ordered probability space. A predictive distribution function on this space is a function $Q : \mathcal{X}^n \times \mathcal{X} \to [0, 1]$ satisfying:
>
> 1. **Càdlàg**: For any $x \in \mathcal{X}^n$, $Q(x, \cdot)$ is a cumulative distribution function (cf. Extra A.16).

---

[13] As before, a randomized variant should be considered to obtain exact validity. This is the main difference between the definitions in Vovk et al. (2017) and Shen et al. (2018).

2. **Uniformity**: For all $\alpha \in [0, 1]$, the following inequality holds:

$$P^{n+1}\big(Q(X_1, \ldots, X_n, X') \leq \alpha\big) = \alpha. \tag{2.63}$$

The first condition simply says that no matter what the calibration data set $\mathcal{V}$ used to construct the predictive distribution function is, the resulting function actually gives a bona fide probability distribution. The second condition in turn states that the resulting probability distribution is actually consistent with the ground truth distribution $P \in \mathbb{P}(\mathcal{X})$.

Given the above definition, a **conformal predictive system** (CPS) with non-conformity measure $A : (\mathcal{X} \times \mathcal{Y})^* \times \mathcal{X} \times \mathcal{Y} \to \mathbb{R}$ is defined as a predictive distribution function that spits out conformal $p$-values (2.29) or, in the randomized case, smoothed $p$-values (2.31).

> **Remark** 2.37 (**Transducers**). If $p$-values are constructed as in Eqs. (2.29) or (2.31), a model is often called a **conformal transducer** in the conformal prediction literature.[a] It follows that CPSs are predictive distributions that are also conformal transducers.
>
> ───────────
>
> [a] 'Transducer' here refers to the transductive approach.

Although standard conformal transducers can use any nonconformity measure, CPSs require stronger restrictions. The simplest one would be requiring that $A$ is a monotone function in $\mathcal{Y}$, which means that it is increasing in the test response $y'$ and decreasing in any (and, by $S_n$-invariance, every) calibration response $y_i$. It should be clear that this already excludes the most common nonconformity measure in the regression setting: the (absolute) residual measure (3.38), which will be introduced in the next chapter. However, this issue can easily be resolved by dropping the absolute value:

$$A_{\text{res}}(x, y) := y - \hat{y}(x), \tag{2.64}$$

where $\hat{y} : \mathcal{X} \to \mathbb{R}$ is a point predictor.

## 2.5.3 Beyond exchangeability

Conformal prediction is often said to have distribution-free guarantees, but this is not entirely true. The main theorems in this chapter strongly depend on the exchangeability condition, which, although more general than i.i.d., is still a strong condition. For this reason, the past years have seen a rise in research into nonexchangeable extensions of conformal prediction. Although these results will not be used throughout this dissertation, they are certainly relevant and deserve acknowledgement.

One of the earliest approaches, specifically targeted at times series data — this type of data is notorious for not being exchangeable due to the inherent (auto)correlation between different points in time — was developed in Chernozhukov, Wüthrich, and Yinchu (2018). Here, the symmetric group of permutations in Definition 2.11 is replaced by the subset of (block) permutations that preserve the dependency structure in the data. Another method aimed at time series was presented in Xu and Xie (2021). Here, the authors use a bootstrap method to correct possible deviations from exchangeability.

Aside from times series data, another example of nonexchangeable data is given by the problem of **covariate shift**, where only the feature distribution $P_X$ changes between calibration data and test data, but not the conditional response distribution $P_{Y|X}$. An interesting approach to tackling this problem was introduced in Tibshirani, Barber, Candès, and Ramdas (2019), where a weighting scheme is used that takes into account the *likelihood ratio* when calculating conformal $p$-values. However, this requires that the likelihood ratio is known a priori, which is often not the case in practice. An extension of this idea was presented in L. Lei and Candès (2021), where the likelihood ratio is estimated. A 'nonconformal' method with similar intents, also situated in the area of *causal inference*, was introduced in Qiu, Dobriban, and Tchetgen Tchetgen (2023).

Going beyond mere covariate shift, Cauchois, Gupta, and Duchi (2021) introduced a method for protecting against *distribution shift* in case the shift is bounded in terms of some suitable divergence measure (Definition A.60). The critical quantile in Algorithm 2 is replaced by the supremum of the quantiles among all distributions that lie close enough to the empirical one in terms of the chosen divergence measure.

### 2.5.4 Imprecise probability theory

Section 2.3 showed that conformal predictors already come with some powerful statistical guarantees (at least within the realm of exchangeable probability distributions cf. Section 2.5.3). Moreover, Section 2.5.2 on conformal predictive systems showed how the coverage guarantee can be used to obtain a calibrated probability distribution. A motivation for using distributions instead of 'simple' confidence predictors is that the former can be used to make statements about more general assertions, i.e. events more general than $Y \in \Gamma^\alpha(X)$. Cella and Martin (2022) investigated how validity can be generalized to this extended setting.

Without going into too much detail, a small primer on imprecise probabilities is in order (for more information, see e.g. Augustin et al. (2014)). Consider a function $\underline{\pi} : \mathcal{X}^* \to [0,1]^{2^{\mathcal{X}}}$ that takes a sequence of data points $x \in \mathcal{X}^*$ to a function $\underline{\pi}_x : 2^{\mathcal{X}} \to [0,1]$ satisfying the properties of a (*Choquet*) *capacity* for almost all $x \in \mathcal{X}^*$:[14]

1. **Emptiness**: $\underline{\pi}_x(\emptyset) = 0$ and $\underline{\pi}_x(\mathcal{X}) = 1$,

2. **Monotonicity**: $A \subseteq B \implies \underline{\pi}_x(A) \leq \underline{\pi}_x(B)$,

3. **Superadditivity**: $\underline{\pi}_x(A \cup B) \geq \underline{\pi}_x(A) + \underline{\pi}_x(B)$ for all disjoint $A, B \in 2^{\mathcal{X}}$, and

4. **Continuity**: For infinite spaces $\mathcal{X}$, suitable continuity conditions have to be imposed, but these would lead us too far astray.

Given a capacity $\underline{\pi}_x$, it is common to define its dual capacity as follows:

$$\overline{\pi}_x(A) := 1 - \underline{\pi}_x(\mathcal{X} \setminus A).  \tag{2.65}$$

A pair of dual capacities $\pi \equiv (\underline{\pi}_x, \overline{\pi}_x)$ will be called a **probabilistic predictor**. The components of such a predictor are sometimes called the **lower** and **upper probability**, respectively. Note that when $\underline{\pi}_x$ is a proper probability distribution, it coincides with its dual capacity.

The reason why imprecise probability theory is needed for this section is the following. Marginal Validity Theorem 2.27 characterizes what Cella and Martin (2022) call the **Type-1 validity** of conformal predictors (with respect to exchangeable distributions). However, to characterize the calibration of a probabilistic predictor with respect to more general assertions, a generalization is required.

---

[14] These should be compared to the Kolmogorov axioms in Definition A.14.

**Definition** 2.38 (**Type-2 validity**). A probabilistic predictor $\pi$ is said to be Type-2 valid with respect to a collection of probability distributions $\mathcal{P} \subset \mathbb{P}(\mathcal{X}^*)$ if

$$P\left(\overline{\pi}_X(A) \leq \alpha \wedge X \in A\right) \leq \alpha \qquad (2.66)$$

for all $A \in 2^{\mathcal{X}}$, $\alpha \in [0, 1]$ and $P \in \mathcal{P}$.

If this does not only hold pointwise in $A \in 2^{\mathcal{X}}$, but actually uniformly in $A$, i.e.

$$P\left(\overline{\pi}_X(A) \leq \alpha \wedge X \in A \text{ for some } A \in 2^{\mathcal{X}}\right) \leq \alpha \qquad (2.67)$$

holds for all $\alpha \in [0, 1]$ and $P \in \mathcal{P}$, the probabilistic predictor is said to be **strongly** Type-2 valid (with respect to $\mathcal{P}$).

The important point is that a probabilistic predictor is strongly Type-2 valid if and only if

$$P\left(\overline{\pi}_X(\{X\}) \leq \alpha\right) \leq \alpha \qquad (2.68)$$

for all $\alpha \in [0, 1]$ and $P \in \mathcal{P}$. For nonatomic probability distributions, this is, however, impossible if $\overline{\pi}_X$ is additive. Accordingly, ordinary probability theory is not sufficient when we want guarantees for general assertions. Luckily, conformal prediction again comes to the rescue. The following result from Cella and Martin (2022) shows that conformal transducers (Remark 2.37) give rise to strongly Type-2 valid predictors.

**Property** 2.39. Consider a conformal transducer $p_x : \mathcal{X} \rightarrow [0, 1]$ and define the *possibility measure* (or *consonant plausibility measure*)

$$\overline{\pi}_x(A) = \sup_{x \in A} p_x(x). \qquad (2.69)$$

The probabilistic predictor induced by this capacity is strongly Type-2 valid (with respect to exchangeable distributions).

## 2.6   Discussion

The main purpose of this chapter was to give a general, yet detailed overview of the conformal prediction framework. Since this framework fits in the broader setting of uncertainty quantification, an introduction to the fundamental notion of uncertainty, at least our current understanding of it, could not be missing. Conformal prediction has a strong relation to many other topics in statistics and probability theory, Sections 2.5.3 and 2.5.4 serve as an indication, and we can expect it to lead to many breakthroughs in the future.

In the subsequent chapters, we will delve a bit deeper into the methodology and (formal) applications of conformal prediction. This will cover both regression and classification problems (Chapters 3 and 5, respectively), conditional validity (Chapter 4) and the use of side information and inductive biases (Chapters 4 and 5).

# Regression <span style="float:right">3</span>

$$\langle F \rangle = \frac{\displaystyle\int F(\phi) e^{iS[\phi]/\hbar} \, \mathcal{D}\phi}{\displaystyle\int e^{iS[\phi]/\hbar} \, \mathcal{D}\phi}$$

<div style="text-align:right">Feynman (1948)</div>

The content of this chapter is based on the article *Valid prediction intervals for regression problems* (Dewolf et al., 2023b), jointly written with my promotors W. Waegeman and B. De Baets.

## 3.1   Introduction

In statistical learning theory, most problems can be divided into two general classes: classification and regression. In the former, the goal is to classify instances into a set of categories, whereas in the latter, the goal is to model or approximate a (real-valued) function as well as possible. These can of course be considered one and the same from an abstract point of view and the distinction between these problems is generally based on the structure of the target space $\mathcal{Y}$. In the case of classification, $\mathcal{Y}$ is discrete and, usually, even finite. The problem is often phrased as finding the probability of observing a given class: $\widehat{f} : \mathcal{X} \to \mathbb{P}(\mathcal{Y})$. In the case of regression, the problem is usually formulated for target spaces with more structure such as $\mathbb{N}$ — often called *ordinal regression* — or $\mathbb{R}$ — simply called regression — and consists of finding a function $\widehat{f} : \mathcal{X} \to \mathcal{Y}$ (with $\mathcal{Y} = \mathbb{N}, \mathbb{R}$) satisfying a certain optimization criterion.

Since for regression problems the response variable is of the ordinal, interval or even ratio type, we can use more powerful properties and techniques. For example, by providing two distinct elements of $\mathbb{R}$, a lower and an upper bound, we can construct a nontrivial confidence predictor whose output is the interval determined by the given bounds. This would not be possible for target spaces that do not admit the structure of a totally ordered set (Definition A.2). (Binary classification problems can be reformulated as regression problems where the linear structure on $[0, 1]$ can be used, but this does not extend to multiclass settings in a useful way, since the product of total orders is not necessarily total.)

The majority of techniques considered in this chapter serve two purposes. First, they do exactly what they are supposed to do. They solve a regression problem by producing a point estimate for the response variable given some features, i.e. they construct a function $\widehat{f} : \mathcal{X} \to \mathcal{Y}$. However, in light of the previous chapter, they also serve as confidence predictors by providing a region that should contain the true response with a predetermined probability. Although it is not necessarily required that these regions include the point estimate constructed in the first step, it seems reasonable to restrict to models that satisfy this property. Otherwise, we are predicting points that we know (or expect) to not be realized with high probability.

This chapter is structured as follows. In Section 3.2, different classes of confidence predictors for regression problems are discussed: Bayesian methods, mean-variance estimators, ensemble methods and direct interval estimators (a fifth class will be covered in Chapter 4). To continue the method overview, conformal regression is treated in Section 3.3. Since the main contribution of this chapter is giving a comprehensive review of existing uncertainty quantification methods, an experimental comparison could not be lacking. Examples of the different classes, with and without conformal prediction, are compared across a range of data sets in terms of coverage, efficiency and predictive power in Section 3.4.

## 3.2    Confidence predictors

In this section, some general classes of confidence predictors in the regression setting are discussed. The first class, that of Bayesian methods, has its roots in probability theory and, therefore, can be expected to have better theoretical guarantees with respect to the validity of the models. The second class, that of ensemble methods, consists of methods that are built from a collection of models and generally exhibit superior predictive performance when compared to individual models. The third class covers the models that are specifically trained to yield a prediction interval, while the last class incorporates conformal prediction to turn any given point predictor into a valid interval estimator. Many different methods are nowadays being used to produce uncertainty estimates. However, in our opinion, all methods that construct prediction intervals in a regression setting can be assigned to one of these four classes. (A fifth class, that of generative models, will be introduced in the next chapter in Section 4.7.)

### 3.2.1 Bayesian methods

In Bayesian inference, we try to model the distribution of interest by updating our prior belief using a collection of observed data. The conditional distribution $P(Y \mid X, \mathcal{D})$ is inferred from a given parametric model or likelihood function $f(Y \mid X, \theta)$, a prior distribution $P(\theta)$ over the model parameters and a data set $\mathcal{D} \equiv (X, y)$. The first step is to update our prior belief using Bayes' rule, where from now on we assume that all distributions $P$ have an associated density $f$:

$$f(\theta \mid \mathcal{D}) = \frac{f(y, \theta \mid X)}{f(y \mid X)} = \frac{f(y \mid X, \theta) f(\theta)}{\int_{\Theta} f(y \mid X, \theta) f(\theta) \, d\theta}, \tag{3.1}$$

where $\Theta$ denotes the parameter or hypothesis space. Note that we used the shorthand $f(y, \theta \mid X)$ to denote the joint density $\prod_{(x,y) \in \mathcal{D}} f(y, \theta \mid x)$, thereby assuming that the data is i.i.d. (This assumption is notably violated in the setting of time series, but this is also a problem with 'standard' conformal prediction, so we will ignore this issue for the time being.) After the posterior distribution over the parameter space is computed, the posterior predictive distribution is calculated by marginalizing over the parameters:

$$f(Y \mid X, \mathcal{D}) = \int_{\Theta} f(Y \mid X, \theta) f(\theta \mid \mathcal{D}) \, d\theta. \tag{3.2}$$

This process is summarized in Algorithm 5. Note that the algorithm can simply be repeated in an online fashion when more data becomes available. We simply have to take the 'old' posterior distribution $P(\theta \mid \mathcal{D})$ as the new prior distribution.

To obtain a point estimate for future predictions, the most popular choice is the conditional mean $\mathsf{E}[Y \mid X, \mathcal{D}]$. However, the correct choice depends on the metric that is used to score the predictions. For instance, the conditional mean would be the choice that minimizes the MSE score ($\ell^2$-loss), whereas the conditional median would be the correct choice for minimizing the MAE

---

**Algorithm 5:** Bayesian modelling

**Input**  : Model architecture $\mathcal{A}$, likelihood function $\mathcal{L}(Y \mid X, \theta)$, prior
              distribution $P_\Theta$ and data set $\mathcal{D}$

**Output:** Predictive distribution $f(Y \mid X, \mathcal{D})$

1 Construct a model with architecture $\mathcal{A}$ and parameters sampled
   from $P_\Theta$
2 Update the prior distribution using Bayes' rule (3.1)
3 Infer the predictive distribution $f(Y \mid X, \mathcal{D})$ using Eq. (3.2)

4 **return** $f(Y \mid X, \mathcal{D})$

---

score ($\ell^1$-loss). For Bayesian methods, the construction of a prediction interval suffers from similar ambiguities since it reduces to the classic, but not always straightforward, problem of finding prediction intervals for a given distribution. One of the main advantages of Bayesian methods, compared to other approaches, however, is that we can incorporate domain knowledge into the prior distribution whilst still retaining probabilistic properties. By leveraging this knowledge to more accurately characterize the class of distributions that the data-generating process (hopefully) belongs to, we can obtain improved uncertainty estimates, especially in settings with very limited data.

➢  **Gaussian processes**

One of the most popular probabilistic models for regression problems is the Gaussian process (Williams & Rasmussen, 1996). The main reason for its popularity is that it is one of the only Bayesian methods where the inference step in Eq. (3.1) can be performed exactly, since the marginalization of multivariate normal distributions can be written in closed form and, hence, no approximations are needed (in principle). Formally, a Gaussian process (GP) is defined as a stochastic process (Section A.2.9) for which the joint distribution of any finite number of random variables is (multivariate) Gaussian.[1] Gaussian processes are, therefore, characterized by only two functions: the mean and covariance functions $m : \mathcal{X} \rightarrow \mathbb{R}$ and $k : \mathcal{X} \times \mathcal{X} \rightarrow \mathbb{R}$, where the latter has the structure of a kernel function (Definition A.62).

---

[1]  For Gaussian processes, the index set of the process is not $\mathbb{N}$, but $\mathcal{X}$. Although this would lead us too far astray, it should be noted that this makes the general construction more involved. (For $\mathcal{X} = \mathbb{R}^n$, everything is fine.)

If we assume that the population is distributed according to a GP with given mean and covariance functions, by definition, the responses for the training and test instances $X, x'$ are distributed according to a multivariate Gaussian as follows (functions are calculated elementwise on tuples and matrices):

$$\begin{pmatrix} \boldsymbol{y} \\ y' \end{pmatrix} \sim \mathcal{N}\left(\begin{pmatrix} m(\boldsymbol{X}) \\ m(x') \end{pmatrix}, \begin{pmatrix} k(\boldsymbol{X}, \boldsymbol{X}) & k(\boldsymbol{X}, x') \\ k(x', \boldsymbol{X}) & k(x', x') \end{pmatrix}\right) \equiv \mathcal{N}\left(\begin{pmatrix} \boldsymbol{\mu} \\ \mu' \end{pmatrix}, \begin{pmatrix} \Sigma & \Sigma' \\ (\Sigma')^\mathsf{T} & \Sigma'' \end{pmatrix}\right),$$
(3.3)

where the notations $\boldsymbol{\mu} := m(\boldsymbol{X})$, $\mu' := m(x')$, $\Sigma := k(\boldsymbol{X}, \boldsymbol{X})$, etc. are introduced for brevity. With this convention, $\boldsymbol{\mu}$ and $\Sigma$ are, for example, a tuple and a matrix, respectively. Following Algorithm 5, the posterior distribution is obtained by conditioning on the data set $\mathcal{D} \equiv (\boldsymbol{X}, \boldsymbol{y})$:

$$Y' \mid X', \mathcal{D} \sim \mathcal{N}\left(\mu' + (\Sigma')^\mathsf{T}\Sigma^{-1}(\boldsymbol{y} - \boldsymbol{\mu}), \Sigma'' - (\Sigma')^\mathsf{T}\Sigma^{-1}\Sigma'\right). \qquad (3.4)$$

From this formula, it is evident that the point estimate, i.e. the conditional mean, is given by

$$\mathsf{E}\big[Y' \mid X', \mathcal{D}\big] = \mu' + (\Sigma')^\mathsf{T}\Sigma^{-1}(\boldsymbol{y} - \boldsymbol{\mu}). \qquad (3.5)$$

Although they are conceptually extremely simple, it is also immediately clear that GPs suffer from a major drawback. Due to the matrix inversion in the calculation of both the mean and covariance of the posterior distribution, they are computationally less efficient than neural network-based methods. The memory complexity of matrix inversion scales as $\mathcal{O}(|\mathcal{D}|^2)$, while the computational complexity scales as $\mathcal{O}(|\mathcal{D}|^3)$. For large data sets, this gives rise to considerable overhead. However, many approximations exist in the literature. Some of these try to reduce the complexity of the matrix operations by introducing stochastic approximations, while others belong to the class of variational models that will be introduced further on (Hensman, Matthews, & Ghahramani, 2015; Wilson & Nickisch, 2015).

There is also an important remark to be made with respect to the validity condition (2.10). When using GPs, two important assumptions are made. By definition it is assumed that the data is (conditionally) normally distributed and, moreover, a choice of mean and covariance functions has to be made by the user. When these assumptions do not properly characterize the data, it is to be expected that the resulting distribution will also not correctly model the data. In general, misspecification will lead to invalid models.

➢ **Bayesian neural networks**

Bayesian modelling can also be implemented using neural networks. In this approach, the likelihood function in Eq. (3.1) is given by a parametric function for which the parameters are estimated by a neural network (Goan & Fookes, 2020; MacKay, 1992; Neal, 1996). A common example is that of a homoskedastic Gaussian likelihood (Hinton & van Camp, 1993) where the mean is estimated by a neural network with weights $\theta \in \mathbb{R}^d$:

$$Y \mid X, \theta \sim \mathcal{N}\big(\hat{y}_\theta(X), \sigma^2\big). \tag{3.6}$$

Although ordinary neural networks have the benefit that even for a large number of features and weights they can be implemented very efficiently, their Bayesian incarnation suffers from a problem. The nonlinearities in the activation functions and the sheer number of parameters, although they are the features that make traditional NNs so powerful, lead to the inference steps (3.1) and (3.2) becoming intractable. This is similar to the situation of *graphical models* and *Bayesian networks* (not to be confused with Bayesian neural networks), where exact inference is *NP-hard* (Cooper, 1990).

➢ **Approximate Bayesian inference**

Both the integral in the inference step (3.1) and in the prediction step (3.2) can in general not be computed exactly. (The situation of conjugate priors (Fink, 1997), such as with normal distributions, forms an important exception.) At inference time, two general classes of approximations are available:

1. **Variational inference** (VI): Instead of computing the posterior distribution with Eq. (3.1), the problem is reformulated as a variational problem, i.e. the posterior density function $f(\theta \mid \mathcal{D})$ is replaced by a parametric family of density functions $q_\lambda(\theta)$ and a divergence measure (Definition A.60) is optimized within this family (Blei, Kucukelbir, & McAuliffe, 2017). For the Kullback–Leibler divergence $D_{\mathrm{KL}}$, the resulting loss function (up to a factor -1)

$$\mathcal{L}_{\mathrm{VI}}(\mathcal{D}; \lambda) := D_{\mathrm{KL}}\big(q_\lambda(\theta) \parallel f(\theta)\big) - \mathsf{E}\big[\ln f(\mathcal{D} \mid \theta)\big], \tag{3.7}$$

where the expectation is taken with respect to the variational density $q_\lambda(\theta)$, is called the **variational free energy** in the statistical physics literature or **evidence lower bound** (ELBO) in the machine learning literature. For Bayesian neural networks, this function can be optimized using a backpropagation scheme similar to that for traditional neural networks (Blundell, Cornebise, Kavukcuoglu, & Wierstra, 2015).

2. **Monte Carlo (MC) integration**: The general idea (Gentle, 2009) of Monte Carlo integration is that for a general distribution $P$, an integral of the form [2]

$$\mathsf{E}[f] = \int_{\mathcal{X}} f \, \mathrm{d}P \tag{3.8}$$

can be approximated by a finite sum

$$\mathsf{E}[f] \approx \frac{1}{n} \sum_{i=1}^{n} f(x_i), \tag{3.9}$$

where the points $x_i$ are sampled from $P$. For this sampling step, we usually make use of a suitable Monte Carlo sampling scheme such as *Metropolis–Hastings* or *Hamiltonian* MC. The approximation converges to the true value for large $n$ by the law of large numbers A.46 and, in particular, Corollary A.47.

For making predictions, i.e. calculating the conditional mean, MC integration gives the following expression:

$$\mathsf{E}[Y \mid X, \mathcal{D}] \approx \frac{1}{n} \sum_{i=1}^{n} y_i, \tag{3.10}$$

where the samples $y_i$ are drawn from the posterior distribution $P(Y \mid X, \mathcal{D})$, i.e. the expectation value is approximated by the empirical mean of a sample.

### 3.2.2 Mean-variance estimators

Although Bayesian models are in general superior in terms of probabilistic properties, they are computationally very expensive and require a lot of data. However, when restricting to the set of probability distributions that are determined by their first two moments or, equivalently, their expectation value and variance, a more efficient approach exists: mean-variance estimation (Khosravi, Nahavandi, Creighton, & Atiya, 2011a; Nix & Weigend, 1994). Given a suitable loss function, to be introduced below, a model that predicts both the conditional mean and conditional variance can be trained and, given a functional form for the data-generating process, the parameters of this distribution can be recovered in terms of the predicted moments (not necessarily in closed form).

---

[2] Making use of the law of the unconscious statistician A.33.

This is also a good point to introduce some machinery relevant to training probabilistic or statistical models (Gneiting & Raftery, 2007).

> **Definition** 3.1 (**Scoring rule**). Consider a class of probability distributions $\mathcal{P} \subset \mathbb{P}(\mathcal{X})$. A scoring rule is a function $S : \mathcal{P} \times \mathcal{X} \to \overline{\mathbb{R}}$ such that $S(P, \cdot)$ is $\mathcal{P}$-quasiintegrable for all $P \in \mathcal{P}$ (Definition A.23). By integration, two distributions can be compared:
>
> $$S(P, Q) := \int_{\mathcal{X}} S(P, x)\, \mathrm{d}Q(x).  \tag{3.11}$$
>
> A scoring rule $S$ is said to be **proper** with respect to $\mathcal{P}$ if $S(P, P) \geq S(Q, P)$ for all $P, Q \in \mathcal{P}$, i.e. the maximal score is assigned to the 'true' distribution $P$. When the inequality becomes a strict inequality, the scoring rule is said to be **strictly proper**.

Using strictly proper scoring rules as loss (or reward) functions for training a probabilistic model will guarantee that the predicted distribution converges to the true probability distribution. For probabilistic models, we might expect the log-likelihood, as used in maximum likelihood estimation, to give a proper scoring rule. For example, for heteroskedastic Gaussian distributions, the following loss function is often used:

$$\begin{aligned}
\mathcal{L}_{\mathrm{MLL}}(\mathcal{D}) &:= \ln\left( \prod_{(x,y) \in \mathcal{D}} f_{\mathrm{Gauss}}(y \mid x) \right) \\
&= \sum_{(x,y) \in \mathcal{D}} \ln\left( \frac{1}{\sqrt{2\pi\widehat{\sigma}^2(x)}} e^{-\frac{1}{2}\left(\frac{y - \widehat{\mu}(x)}{\widehat{\sigma}(x)}\right)^2} \right) \\
&= -\frac{|\mathcal{D}|}{2}\ln(2\pi) - \sum_{(x,y) \in \mathcal{D}} \frac{|y - \widehat{\mu}(x)|^2}{2\widehat{\sigma}^2(x)} + \frac{1}{2}\ln\widehat{\sigma}^2(x).
\end{aligned} \tag{3.12}$$

Dropping all constants gives:

$$S_{\mathrm{MLL}}(\mathcal{D}) = -\sum_{(x,y) \in \mathcal{D}} \frac{|y - \widehat{\mu}(x)|^2}{\widehat{\sigma}^2(x)} + \ln\widehat{\sigma}^2(x).  \tag{3.13}$$

It might come as a surprise that this score function is not only a good choice for Gaussian distributions, but is in fact applicable to a much more general class of distributions (Gneiting & Raftery, 2007).

> **Property** 3.2. Equation (3.13) gives a proper scoring rule with respect to the class $\mathbb{P}_2(\mathcal{X})$ of probability distributions that have a finite second moment. Moreover, it is strictly proper with respect to all convex subsets of $\mathbb{P}_2(\mathcal{X}) \cap \mathcal{P}_2(\mathcal{X})$, where $\mathcal{P}_2(\mathcal{X})$ denotes the set of all probability distributions determined by their first two moments. It should be noted that dropping the logarithmic term results in a scoring rule, the so-called *energy score*, which is also proper (but not strictly proper) with respect to $\mathbb{P}_2(\mathcal{X})$.

To extract a prediction interval from these models we, in general, will have to resort to a parametric family and, as before, a Gaussian setting in particular. Given an estimate $(\hat{\mu}, \hat{\sigma}) : \mathcal{X} \to \mathbb{R} \times \mathbb{R}^+$, an interval is constructed as follows:

$$\Gamma^\alpha_{\text{Gauss}}(x) := \left[ \hat{\mu}(x) - z^\alpha \hat{\sigma}(x), \, \hat{\mu}(x) + z^\alpha \hat{\sigma}(x) \right], \qquad (3.14)$$

where $z^\alpha$ is the two-tailed $z$-score at significance level $\alpha$ for a standard normal distribution, e.g. $z^\alpha \approx 1.65$ for $\alpha = 0.1$. For a (conditionally) normal distribution, the correct prediction interval with estimated mean $\hat{\mu}$ and estimated standard deviation $\hat{\sigma}$ is given by

$$\Gamma^\alpha_{\text{Gauss}}(x) := \left[ \hat{\mu}(x) - t^\alpha_{n-1} \hat{\sigma}(x) \sqrt{1 + 1/n}, \, \hat{\mu}(x) + t^\alpha_{n-1} \hat{\sigma}(x) \sqrt{1 + 1/n} \right], \quad (3.15)$$

where $t^\alpha_{n-1}$ is the two-tailed $t$-score with $n-1$ degrees of freedom at significance level $\alpha$. If the training set is assumed to be large enough, i.e. $n \gg 1$, then $1/n \approx 0$ and $t^\alpha_{n-1} \approx z^\alpha$, so Eq. (3.14) is a good approximation.

### 3.2.3 Ensemble methods

Ensemble learning is a popular approach to improve predictions by training multiple machine learning models and aggregating the individual predictions, for example by taking the mean (Sollich & Krogh, 1995). In general, we can consider ensemble methods as an intermediate step between Bayesian methods and pure point predictors. They yield an approximation to Bayesian inference, where the trained models represent a sample from the parameter space and the aggregation corresponds to an MC approximation of the inference step. For instance, mean aggregation corresponds to ordinary MC integration (3.9). This also implies that ensemble methods have a natural notion of uncertainty. However, this uncertainty estimate does not always easily admit a probabilistic interpretation, because the ensembles are

---

**Algorithm 6:** Ensemble learning

**Input** : Number of models $k \in \mathbb{N}$, model architectures $\mathcal{A}_i$, sampling
strategies $\mathcal{S}_i$, data set $\mathcal{D}$ and aggregation strategy $\mathcal{E}$

**Output:** Ensemble predictor $\hat{y}$

**1 for** ( $i = 1$; $i \leq k$; $i \leftarrow i + 1$ ) {

**2**      Use the strategy $\mathcal{S}_i$ to obtain a data set $\mathcal{D}_i$ from $\mathcal{D}$

**3**      Construct the model $\hat{y}_i$ with architecture $\mathcal{A}_i$

**4**      Train $\hat{y}_i$ on $\mathcal{D}_i$

**5** }

**6** Aggregate the models $\{\hat{y}_i\}_{i=1}^k$ into a new model $\hat{y}$ using strategy $\mathcal{E}$

**7 return** $\hat{y}$

---

in practice often of an ad hoc nature and only roughly represent approximate Bayesian inference. The general structure of an ensemble learning scheme is summarized in Algorithm 6.

Every ensemble allows for a naive construction of a prediction interval when the aggregation strategy in Algorithm 6 is given by the arithmetic mean as in Heskes (1996). In this situation, we can treat the predictions of the individual models in the ensemble as elements of a data sample and obtain an empirical measure as in Section A.4.2. As explained there, the empirical mean and variance can be used as moment estimators for a normal distribution:

$$\hat{\mu}_{\text{ens}}(x) := \frac{1}{k} \sum_{i=1}^k \hat{y}_i(x) \,, \tag{3.16}$$

$$\hat{\sigma}^2_{\text{ens}}(x) := \frac{1}{k} \sum_{i=1}^k \left( \hat{y}_i(x) - \hat{\mu}_{\text{ens}}(x) \right)^2 \,, \tag{3.17}$$

where $k \in \mathbb{N}_0$ denotes the ensemble size. These are the expressions obtained by interpreting the ensemble as a mixture of Dirac measures as in Eq. (A.98). Since this puts us in the setting of the previous section, we can construct a Gaussian prediction interval as in Eq. (3.14):

$$\Gamma^{\alpha}_{\text{ensemble}}(x) := \left[ \hat{\mu}_{\text{ens}}(x) - z^{\alpha} \hat{\sigma}_{\text{ens}}(x) \,, \ \hat{\mu}_{\text{ens}}(x) + z^{\alpha} \hat{\sigma}_{\text{ens}}(x) \right] . \tag{3.18}$$

Moreover, recalling the content of Section 2.2.2, ensemble models can also be seen to give a straightforward way to quantify the epistemic uncertainty.

When the data is not sufficient to accurately model the data-generating process, the models in the ensemble will strongly differ from each other. For this reason, the ensemble variance gives a measure of the epistemic uncertainty and not so much the aleatoric uncertainty. The extension to ensembles of probabilistic models will be treated further on in the section on dropout networks.

➢ **Random forests** (**and other** *bagged* **ensembles**)

Random forests, as introduced in Breiman (2001), are an extension of the idea of (regression) **decision trees**, which work as follows:

1. Recursively choose a feature and perform a binary split.

2. Keep doing this until a certain criterion has been reached (entropy, *Gini impurity*, leaf size, ...).

3. Determine for a new point in which leaf node it ends up.

4. Predict the mean of the responses of data points in the given leaf node.

A simple example with three features is shown in Fig. 3.1. Assume that a new data point $x = (15, -5, 7)$ is given. Applying the rules of the decision tree, this point will be assigned to the third leaf node and, accordingly, the prediction will be

$$\hat{y}(x) = \frac{0.2 - 1 - 0.3 + 0}{4} = -0.275 \,. \tag{3.19}$$

To obtain random forests, we do not simply take an ensemble of such decision trees. Their main feature is the use of bagging (Definition A.70), where every decision tree in the forest is constructed from a subset of the original training set (possibly sampled with replacement). Random forests have been shown to be robust against overfitting and, due to their nonparametric nature, they can be reliably applied in a wide variety of situations.

The main disadvantage of random forests for uncertainty quantification or other probabilistic purposes is the fact that they do not produce an uncertainty estimate by themselves aside from the crude construction introduced above in Eq. (3.18). There are two general methods for obtaining a more refined prediction region from random forests. Either we can use the 'out-of-bag samples' — the samples for each tree that have not been used to train

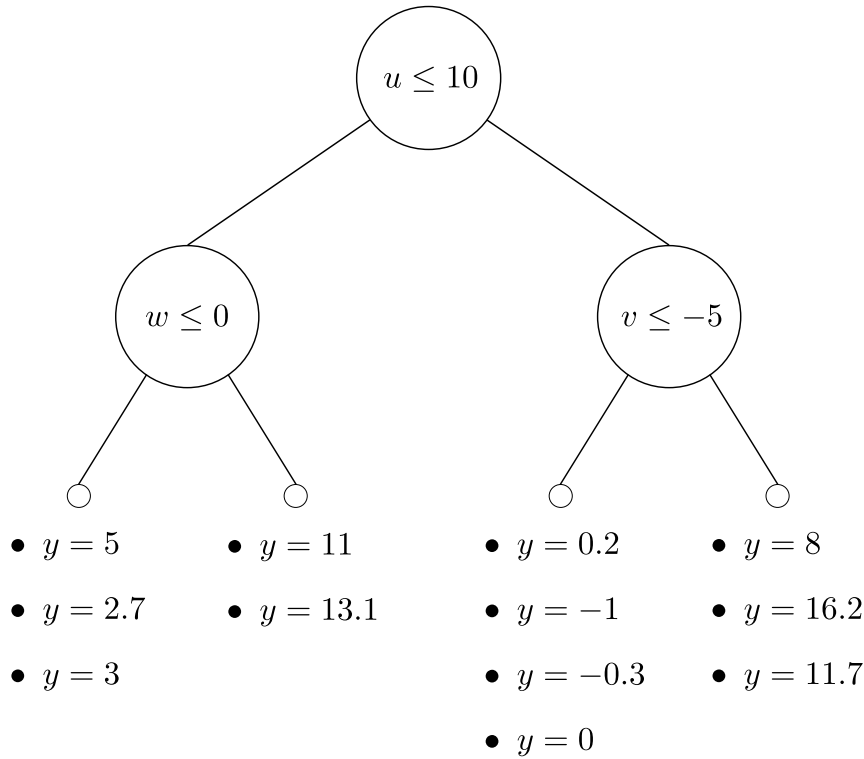

Figure 3.1: A simple decision tree for a univariate regression problem.

it — to calculate an uncertainty estimate or we can use a stand-alone procedure to extract uncertainty estimates (the former could be considered as a specific case of the latter).

Since conformal prediction, the prime example of the second approach, was discussed in Section 2.3, the first one will be considered here. For a random forest, and in fact for any ensemble that implements the bagging procedure (Definition A.70), it is possible to use OOB estimates to construct prediction regions. Because these ensembles do not use the full training set to train the individual models, we can for every point $z_i \equiv (x_i, y_i)$ in the training sample use the subensemble $\hat{y}_{(i)}$ of models that were not trained on $z_i$ to get an estimate of the prediction accuracy. A well-established first step in this direction is an extension of the (infinitesimal) *jackknife* method for variance estimation (Miller, 1974; Quenouille, 1956; Tukey, 1958) to bagged ensembles (Efron, 1992; Wager, Hastie, & Efron, 2014). This approach applies the general construction of Eq. (3.18) to the ensemble consisting of the OOB estimates for all training samples as in Section 2.5.1. A related (conformal) approach, first introduced in Johansson et al. (2014) and later generalized in H. Zhang, Zimmerman, Nettleton, and Nordman (2020), uses the OOB estimates to construct an empirical error distribution using

$$\mathcal{E} := \left\{ y_i - \hat{y}_{(i)}(x_i) \mid (x_i, y_i) \in \mathcal{D} \right\}. \tag{3.20}$$

The prediction interval for a new instance is constructed as follows:

$$\Gamma_{\mathrm{OOB}}^{\alpha}(x) := \left[ \hat{y}(x) + q_{\alpha/2}(\mathcal{E}), \, \hat{y}(x) + q_{1-\alpha/2}(\mathcal{E}) \right]. \tag{3.21}$$

By comparison with Eq. (2.57) in Section 2.3, this method can be seen to reduce to a conformal predictor in the case of symmetric error distributions. A similar idea was used in Barber et al. (2021b) for the *jackknife+* (to be introduced further on in Section 3.3.3). There, a prediction interval is obtained by using $k$-fold or leave-one-out cross-validation to estimate the generalization error for arbitrary regression models. The above interval estimator for bagged ensembles is a specific instance of this approach. It should be noted that both the jackknife+ method and the predictor in Eq. (3.21), also called the jackknife method in Barber et al. (2021b), are nonparametric in contrast to the genuine jackknife estimator from Wager et al. (2014).

➤ **Random forest quantile regression**

Meinshausen (2006) modified the random forest approach from the previous section to be able to directly estimate quantiles. (See Section 3.2.4 further on for a general introduction to quantile estimation.) Ordinary regression forests estimate the conditional mean of the response variable by taking a (weighted) average over the training instances in the leaves where the new instance belongs to as in Eq. (3.19). Quantile regression forests generalize this idea by using the property that the conditional CDF associated to a probability distribution $P \in \mathbb{P}(\mathcal{X} \times \mathbb{R})$ can be expressed as a conditional mean:

$$F(y \mid x) = \mathsf{E}_P \left[ \mathbb{1}_{]-\infty, y]}(Y) \,\middle|\, X = x \right]. \tag{3.22}$$

The algorithm starts by building an ordinary random forest and then extracts the relevant weights to estimate the conditional distribution $\hat{P}_{Y|X}$. The (conditional) quantiles are inferred through their defining equation (A.63):

$$\hat{q}_{\alpha}(x) := \inf\left\{ y \in \overline{\mathbb{R}} \,\middle|\, \alpha \leq \hat{F}_{Y|X}(y \mid x) \right\}. \tag{3.23}$$

The main benefit of this approach over ordinary quantile regressors (see Section 3.2.4 below) is that we immediately obtain an approximation of the full conditional distribution and, accordingly, it is not necessary to train a separate model for each quantile. However, the downside of this method is its computational inefficiency. Because the full distribution needs to be modelled, the information about the leaf nodes of all trees needs to be stored and, for every new data point, both the conditional distribution (3.22) and the infimum in Eq. (3.23) need to be recalculated, whereas for ordinary regression forests, only the mean of every leaf node has to be stored.

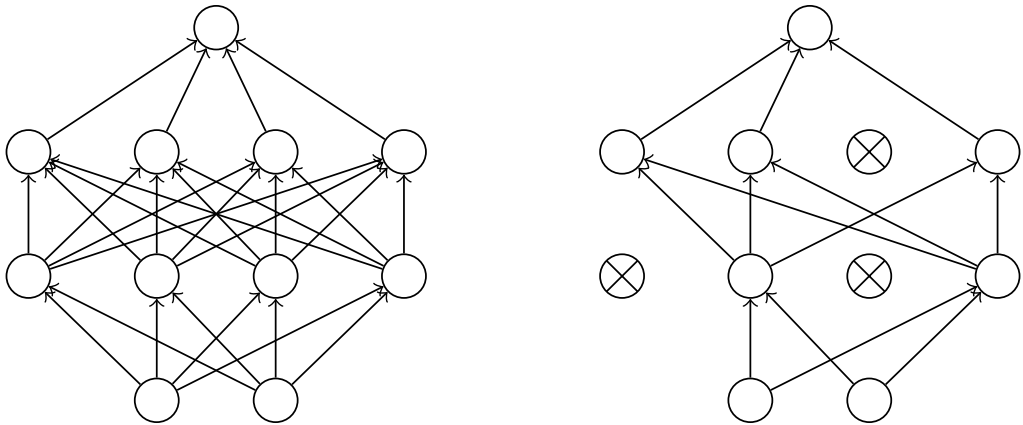

Figure 3.2: Two-layer dropout network. In the left panel, an ordinary dense neural network is shown. The right panel depicts an instance of a neural network where a subset of nodes have been zeroed out.

➢ **Dropout networks**

Dropout layers were initially introduced in Srivastava, Hinton, Krizhevsky, Sutskever, and Salakhutdinov (2014) as a stochastic regularization technique. During each forward pass at training time, a random subset of the network nodes is set to zero with probability $p \in [0, 1]$. More precisely, a subset of the network nodes is (temporarily) removed with a given probability. In essence, this means that a Bernoulli prior with parameter $p$ is placed over every node.[3] The idea behind this method was to reduce the 'co-adaptation' behaviour of fully-connected networks, i.e. the pruning procedure makes it harder for the weights to work together to overfit the data. Further research showed that dropout networks can be obtained as a variational approximation to deep Gaussian processes (Gal, 2016). Therefore, it can be expected that, at least when normality assumptions are satisfied, this method can outperform other ad hoc approaches.

---

[3] A similar idea is given by *flipout*, where the individual weights are assigned a Bernoulli prior (Wen, Vicol, Ba, Tran, & Grosse, 2018).

A second step in the development of dropout networks was the extension of the stochastic behaviour of the dropout layers to test time. By also randomly setting weights to zero during test time, an ensemble of different models can be obtained without having to retrain the model itself. Furthermore, because the Bernoulli priors induce a very sparse structure, MC integration can be expected to give a good approximation. It could even be performed exactly when the number of weights is not too high. A more in-depth study of stochastic regularization and Bayesian inference in dropout models is given in Gal (2016).

Shortly thereafter, a generalization was introduced that also allows to incorporate aleatoric uncertainty in dropout networks (Kendall & Gal, 2017). Instead of having a neural network that produces a single point prediction, the network also estimates the predictive variance, thereby turning it into a mean-variance estimator as introduced in Section 3.2.2. Conform Eq. (3.13), the loss function for this neural network consists of a mean squared error term and a penalty term for the variance:

$$\mathcal{L}_{\text{Gauss}}(\mathcal{D}) := \sum_{(x,y) \in \mathcal{D}} \frac{|y - \hat{y}(x)|^2}{\hat{\sigma}^2(x)} + \ln \hat{\sigma}^2(x) \,. \tag{3.24}$$

As explained in Section 3.2.2, this loss function can be derived from the maximum likelihood principle when assuming a Gaussian probability distribution $\mathcal{N}\big(\hat{y}(x), \hat{\sigma}^2(x)\big)$ or, equivalently, by assuming that the error $y - \hat{y}(x)$ is distributed according to $\mathcal{N}\big(0, \hat{\sigma}^2(x)\big)$. However, as was also noted in that section, this loss function can be applied more generally for all distributions in $\mathbb{P}_2(\mathbb{R})$. Moreover, turning to the Bayesian side of things, instead of maximum likelihood estimation, we could also use maximum a posteriori estimation with a Gaussian or Laplace prior, so that the common $\ell^1$- or $\ell^2$-penalties can also be captured.

When making predictions, the conditional mean is again approximated by MC integration (3.9), i.e. we take the average of multiple forward passes. However, contrary to the case of ordinary ensemble methods, the mean-variance ensemble consists of fully fledged distributional forecasters and can, in particular, be interpreted as a uniformly weighted Gaussian mixture model (Reynolds, 2009). As such, all terms in Eq. (A.98) have to be taken into account. The total predictive variance is given by the sum of the empirical variance of the ensemble and the average variance predicted by the

models:

$$\widehat{\sigma}_{\text{ens}}^2(x) = \underbrace{\frac{1}{k}\sum_{i=1}^{k}\widehat{y}_i(x)^2 - \left(\frac{1}{k}\sum_{i=1}^{k}\widehat{y}_i(x)\right)^2}_{\text{ensemble spread}} + \underbrace{\frac{1}{k}\sum_{i=1}^{k}\widehat{\sigma}_i^2(x)}_{\text{average variance}} . \qquad (3.25)$$

Prediction intervals are constructed using Eq. (3.18) as before. We can, how-
ever, immediately expect that, analogous to general mean-variance estima-
tors with a Gaussian prediction interval, this procedure does not give op-
timal intervals for data sets that do not follow a normal distribution. One
of the consequences is that this model might suffer from the validity prob-
lems discussed in Section 3.2.1. In fact, this issue was already highlighted
in the appendix to Gal (2016). Equation (3.25) also lends itself to study the
decomposition of uncertainty from Section 2.2.2. Whereas the variation in
ordinary ensembles is rather a measure for the epistemic uncertainty, prob-
abilistic ensembles such as the dropout model can be used to characterize
both contributions. The first two terms in Eq. (3.25) are again the variance
coming from the ensemble when only the point predictions are taken into
account and, hence, model the epistemic uncertainty. The last term, how-
ever, is given by the variance as predicted by the probabilistic estimators
themselves and is an estimate of the aleatoric uncertainty.

➢ **Deep ensembles**

The idea behind deep ensembles (Lakshminarayanan, Pritzel, & Blundell,
2017) is the same as for any ensemble technique: training multiple models to
obtain a better and more robust prediction. The loss functions of most (deep)
models have multiple local minima and by aggregating multiple models
we might hope to take into account all these minima. From this point of
view, the approach in Lakshminarayanan et al. (2017) is very similar to that
of Kendall and Gal (2017). However, the underlying philosophy is slightly
different. Instead of obtaining the loss function from a Bayesian perspec-
tive, it is explicitly chosen to be a proper scoring rule (Definition 3.1). As
mentioned before, by training the model with respect to a (strictly) proper
scoring rule, the model is encouraged to approximate the true probability
distribution. A disadvantage of these scoring rules, however, is that they are
only (strictly) proper relative to a certain class of distributions and, hence,
they still introduce distributional assumptions in the model. For example,
as noted in Section 3.2.2, the scoring rule (3.24) is only proper with respect
to probability distributions with a finite second moment and strictly proper with respect to distributions that are determined by their first two moments. Another difference with the approach in Kendall and Gal (2017) is that, instead of constructing an ensemble through Monte Carlo sampling from the prior distribution, multiple models are independently trained and diversity is induced by using different initial parameters. This is motivated by a prior observation that random initialization often leads to superior performance when compared to other ensemble techniques (S. Lee, Purushwalkam, Cogswell, Crandall, & Batra, 2015).

A further modification introduced by Lakshminarayanan et al. (2017) is the use of adversarial perturbations. Instead of using an ordinary gradient descent step

$$\theta \longrightarrow \theta - \varepsilon \nabla_\theta \mathcal{L}(\boldsymbol{X}, \boldsymbol{y}; \theta) \,, \tag{3.26}$$

an additional contribution is added where the training instances are replaced by *adversarial perturbations* obtained using the 'Fast Gradient Sign Method' (the specific choice of perturbation method can be seen as a hyperparameter of the model). For every instance $(x, y) \in \mathbb{R}^n \times \mathbb{R}$, a perturbed version is obtained as follows (Goodfellow, Shlens, & Szegedy, 2015):

$$x' := x + \eta \odot \mathsf{sgn}\big(\nabla_x \mathcal{L}(x, y; \theta)\big) \,, \tag{3.27}$$

where $\odot$ denotes elementwise multiplication, for some constant $\eta \in \mathbb{R}^n$. Allowing $\eta$ to be a vector allows to accommodate features with different ranges. The modified update rule is obtained by replacing the loss function in the gradient descent step with the total loss

$$\mathcal{L}_{\mathsf{tot}}(x, y; \theta) := \mathcal{L}(x, y; \theta) + \mathcal{L}(x', y; \theta) \,, \tag{3.28}$$

where $x'$ denotes the perturbed feature tuple as defined in Eq. (3.27).

Prediction intervals are constructed as in the previous section, i.e. a (conditionally) normal distribution is assumed and the intervals are given by Eq. (3.18). It was observed that this architecture shows improved modelling capabilities and robustness for uncertainty estimation. In Fort, Hu, and Lakshminarayanan (2019), the improved performance is attributed to the multimodal behaviour in the parameter space. The authors argued that most networks tend to have the property that they only 'live' in one specific subspace of the model space, while deep ensembles (due to random initialization) can model different modes.

---

**Algorithm 7:** Direct interval estimation

   **Input**   : Model architecture $\mathcal{A}$, loss function $\mathcal{L}$ and data set $\mathcal{D}$

   **Output:** Interval estimator $\Gamma$

**1** Construct a model $\Gamma$ with architecture $\mathcal{A}$

**2** Train $\Gamma$ on $\mathcal{D}$ using the loss function $\mathcal{L}$

**3** (Optional) Post-process $\Gamma$ to obtain an interval estimator

**4** **return** $\Gamma$

---

Without the adversarial training contribution in the loss function, this model is similar to the one introduced in Khosravi, Nahavandi, Srinivasan, and Khosravi (2015). However, there, instead of training an ensemble of mean-variance estimators, an ensemble of point predictors is trained to first predict the conditional mean and, in a second step, a separate model $\widehat{\sigma} : \mathcal{X} \rightarrow \mathbb{R}^+$ for the data noise is trained using the loss function in Eq. (3.24), while the ensemble model is kept fixed.

### 3.2.4    Direct interval estimation methods

The class of direct interval estimators consists of all methods that are trained to directly output a prediction interval. Instead of modelling a distribution or extracting uncertainty from an ensemble, they are trained using a loss function that is specifically tailored to the construction of prediction intervals. The general structure of this approach is summarized in Algorithm 7. Because these methods are specifically made for estimating uncertainty, they could be expected to perform better than modified point predictors. However, this is also immediately their main disadvantage: they generally do not produce a point estimate. Another disadvantage is that they are in general also specifically constructed for a predetermined confidence level $\alpha \in [0, 1]$. Choosing a different value for $\alpha$ would also require to train a new model.

➢   **Quantile regression**

Instead of estimating the conditional mean of a distribution, quantile regression models estimate the conditional quantiles. This has the benefit that we immediately obtain measures of the spread of the underlying distribution and, therefore, of the uncertainty regarding future predictions. A neural network quantile regressor $\hat{q}_\alpha : \mathcal{X} \to \mathbb{R}$ for the $\alpha$-quantile can easily be optimized by replacing the common MSE loss by the following **pinball loss** or **quantile loss** (Koenker & Hallock, 2001):

$$\mathcal{L}_{\text{pinball}}(\mathcal{D}) := \sum_{(x,y) \in \mathcal{D}} \max\Big((1-\alpha)\big(\hat{q}_\alpha(x) - y\big), \alpha\big(y - \hat{q}_\alpha(x)\big)\Big). \qquad (3.29)$$

This loss function tries to balance the number of data points below and above the (estimated) quantile. It can be obtained by reformulating the definition of quantiles as an optimization problem:

$$Q(\alpha) = \underset{q \in \mathbb{R}}{\arg\min}\left((1-\alpha)\int_{-\infty}^{q}(q-y)\,\mathrm{d}P(y) + \alpha\int_{q}^{+\infty}(y-q)\,\mathrm{d}P(y)\right), \qquad (3.30)$$

where $Q : [0,1] \to \overline{\mathbb{R}}$ is the quantile function corresponding to the probability distribution $P$. If $P$ admits a density function with respect to the Lebesgue measure, differentiating the argument on the right-hand side with respect to $q$ and equating it to zero gives Eq. (3.23), which is the definition of the $\alpha$-quantile. For a given data set $\mathcal{D} \in \mathbb{R}^*$, the minimizer of the pinball loss (3.29) is simply the sample $\alpha$-quantile $q_\alpha(\mathcal{D})$, i.e. the $\alpha$-quantile of the empirical distribution function of $\mathcal{D}$.

An intuition for this loss function can be gained from considering the example of the sample median ($\alpha = 0.5$). The median $\xi$ is defined as the point for which there are as many positive as negative residuals $y_i - \xi$ (Example A.43). From an optimization point of view, this definition is equivalent to that of the minimizer of the average of all absolute residuals (the MAE loss). The pinball loss estimates other quantiles by replacing this average with a weighted average. Moreover, it can be shown that this loss function gives a proper scoring rule (Definition 3.1). It is a specific example of a more general class of quantile scoring rules (where $s(x) = x$ and $h(x) = -\alpha x$).

**Property** 3.3. Consider two $\mathcal{P}$-quasiintegrable functions $s, h : \mathcal{X} \to \overline{\mathbb{R}}$. If $s$ is increasing, the scoring rule

$$S(r, x) := \alpha s(r) + \big(s(x) - s(r)\big)\mathbb{1}_{]-\infty, r]}(x) + h(x) \qquad (3.31)$$

is proper for predicting the $\alpha$-quantile of distributions with strictly increasing CDF and finite moments of any order.

> *Proof.*  See Gneiting and Raftery (2007) for a proof.                    □

To estimate multiple quantiles at the same time, we simply add multiple output nodes to the network and add terms of the form of Eq. (3.29) for every quantile:

$$\mathcal{L}_{\text{pinball}}(\mathcal{D}) := \sum_{\alpha \in \mathfrak{Q}} \lambda_\alpha \sum_{(x,y) \in \mathcal{D}} \max\Big((1-\alpha)\big(\hat{q}_\alpha(x) - y\big), \alpha\big(y - \hat{q}_\alpha(x)\big)\Big), \quad (3.32)$$

where $\mathfrak{Q} \subset [0, 1]$ denotes the set of quantiles to be estimated and $\lambda_\alpha \in \mathbb{R}^+$ are *Lagrange multipliers* that assign importance to the different quantiles. Although quantile regression models do not produce a point estimate in the form of a conditional mean, they can give a prediction of the conditional median. These models do, however, suffer from a problem of underrepresented tails of the distribution. As for every neural network, the performance becomes worse when less data is available. Because the quantiles that are relevant to interval estimation are often located in the tails of the distribution, especially for very low significance levels $\alpha$, the quantile regressors might not give optimal results.

However, when combining such models with conformal prediction, validity is not an issue. Therefore, in Y. Romano et al. (2019), the authors introduced a 'softening' factor $w \in \mathbb{R}^+$ such that the underlying quantile regression model only has to estimate the $w\alpha/2$- and $(1 - w\alpha/2)$-quantiles, while conformal prediction takes care of the validity. This approach (to be introduced further below) circumvents the problems of underrepresented tails while preserving the validity of the model, at the expense of potentially (excessively) broadening the prediction intervals if the smoothening parameter is not correctly tuned.

### ➢  High-Quality principle

As stated before, there are two quantities that are mainly used to evaluate the performance of confidence predictors: the degree of coverage (2.6) and the average size of the prediction regions (2.11). The principle that optimal prediction intervals should saturate inequality (2.10) while minimizing the average size was dubbed the 'High-Quality (HQ) principle' in Pearce (2020); Pearce, Brintrup, Zaki, and Neely (2018). The idea to construct a loss function based on the HQ principle was first proposed in Khosravi, Nahavandi, Creighton, and Atiya (2011b). There, the 'Lower-Upper Bound Estimation' (LUBE) network, predicting the lower and upper bounds of the prediction interval as with quantile regressors, was trained using the following loss function:

$$\mathcal{L}_{\text{LUBE}}(\Gamma, \mathcal{D}) := \frac{\text{MPIW}(\Gamma, \mathcal{D})}{r} \Big( 1 + \exp\Big(\lambda \max\big(0, (1-\alpha) - \mathcal{C}(\Gamma, \mathcal{D})\big)\Big)\Big), \quad (3.33)$$

where $\mathcal{C}$ denotes the coverage degree as in Eq. (2.6) and, similarly, the **mean prediction interval width** (MPIW) denotes the average width of the intervals as in Eq. (2.11):

$$\text{MPIW}(\Gamma, \mathcal{D}) := \frac{1}{|\mathcal{D}|} \sum_{(x,y)\in\mathcal{D}} |\hat{y}_+(x) - \hat{y}_-(x)| = \mathcal{W}(\Gamma, \mathcal{D}), \quad (3.34)$$

where, as before, $\hat{y}_{\pm} : \mathcal{X} \to \mathbb{R}$ denote the upper and lower bounds of the prediction intervals:

$$\Gamma(x) \equiv \big[\hat{y}_-(x), \hat{y}_+(x)\big]. \quad (3.35)$$

The numbers $r \in \mathbb{R}$ and $\lambda \in \mathbb{R}$ are, respectively, the range of the response variable and a constant determining how much a deviation from optimal coverage should be penalized. The main idea is that the penalty should be proportional to the size of the intervals and that it should be greater if the desired coverage is not achieved. Note that the presence of the range implies that this method is harder to apply for data-generating distributions with unbounded support.

Although intuitively sensible, this method admits no true theoretical derivation. It is derived solely from heuristic arguments. Later, in Pearce et al. (2018), the authors introduced an alternative to the LUBE loss where some of the ad hoc choices were formalized. The most important modifications are the replacement of MPIW by 'captured' MPIW, i.e. only the lengths of the intervals that contain the true value are taken into account to avoid artificially shrinking already invalid intervals, and replacing the exponential function by a term derived from a likelihood principle, since the number of captured points can be modelled by a binomial distribution. The resulting loss function is

$$\mathcal{L}_{\text{QD}}(\Gamma, \mathcal{D}) := \text{MPIW}_{\text{capt}}(\Gamma, \mathcal{D}) + \lambda \frac{n}{\alpha(1-\alpha)} \max\big(0, (1-\alpha) - \mathcal{C}(\Gamma, \mathcal{D})\big)^2, \quad (3.36)$$

where $n \in \mathbb{N}$ is the total number of data points and $\lambda \in \mathbb{R}$ is again a *Lagrange multiplier* determining the relative importance of the two contributions. Although this loss function is slightly more substantiated, it remains rather heuristic in nature and, hence, it does not admit any theoretical guarantees regarding its performance.

## 3.3    Conformal regression

In Section 2.3, various conformal prediction algorithms and their theoretical guarantees were introduced. Now, it is time to apply them to regression problems. Although the ICP algorithm already strongly improved upon the computational efficiency of the transductive algorithm, the most important obstruction to the application of conformal prediction methods still remains. The nonconformity scores have to be calculated for all possible response values, which for infinite spaces is practically unfeasible. For this reason we should make an intelligent choice of nonconformity measure.

### 3.3.1    Point predictors

In the case of metric target spaces $(\mathcal{Y}, d)$, as defined in Section A.3, the most common nonconformity measure is simply the distance between the prediction and the ground truth (Shafer & Vovk, 2008):

$$A_d(x, y) := d\big(y, \hat{y}(x)\big). \tag{3.37}$$

By Property A.58, if $\mathcal{Y}$ is a vector space and $d : \mathcal{Y} \times \mathcal{Y} \to \mathbb{R}$ is induced by a norm on $\mathcal{Y}$, the level sets of $A_d$ for fixed $x \in \mathcal{X}$ are the extreme points of a convex subset and, by the *Krein–Milman theorem* (Grothendieck & Chaljub, 1973), we can reconstruct the full prediction region from these extreme points.[4] However, it is still very well possible that the set of extreme points is itself infinite (this occurs whenever the set is not a *polytope*). To resolve this issue, we should either find a coordinate transformation such that the resulting sets have a simple expression or we must further restrict the choice of metric. For example, when replacing the standard Euclidean $\ell^2$-metric with the $\ell^\infty$-metric, the ($n$-)ball is replaced by its bounding hypercube and, accordingly, we only need to calculate the corners. In one dimension, however, all $\ell^p$-metrics coincide and the induced nonconformity measure is the absolute residual.

> **Example** 3.4 (**Residual measure**). Consider a univariate regression regression problem, i.e. $\mathcal{Y} = \mathbb{R}$. The simplest error measure in this case is

---

[4] This can be generalized to nonconformity measures that are *quasiconvex* in $y \in \mathcal{Y}$. This means that preimages of sets of the form $]-\infty, a]$ are convex.

the (absolute) residual:

$$A_{\text{res}}(x, y) := \|\hat{y}(x) - y\|_p = |\hat{y}(x) - y| \, . \tag{3.38}$$

Since the prediction regions are necessarily convex in this case, they are given by (closed) intervals. Moreover, the uncertainty is homoskedastic since there is no feature-dependent scaling of the nonconformity measure.

Denote the lower and upper bounds of the prediction intervals by $\hat{y}_{\pm} : \mathcal{X} \to \mathbb{R}$ as in Eq. (2.11). An explicit expression is easily derived from the form of the nonconformity measure:

$$\begin{aligned} a^* &= |\hat{y}(x) - \hat{y}_{\pm}(x)| \\ \Longleftrightarrow a^* &= \pm\big(\hat{y}_{\pm}(x) - \hat{y}(x)\big) \\ \Longleftrightarrow \hat{y}_{\pm}(x) &= \hat{y}(x) \pm a^* \, , \end{aligned} \tag{3.39}$$

where $a^* \in \mathbb{R}$ is the critical nonconformity score. Note that this formula allows us to construct new prediction intervals with $\mathcal{O}(1)$-time complexity:

$$\Gamma^{\alpha}_{\text{point}}(x) := \big[\hat{y}(x) - a^*, \hat{y}(x) + a^*\big] \, . \tag{3.40}$$

The class of metric nonconformity measures is very useful to work with due to the reduced complexity, but it has an important downside. As mentioned above, these functions are completely independent of the feature space $\mathcal{X}$. So, even if we have more knowledge about a certain subspace and, accordingly, would be able to take into account additional information, the metric does not know anything about this. For this reason, Johansson, Boström, and Löfström (2021); Papadopoulos, Gammerman, and Vovk (2008) considered a modification with a feature-dependent scaling.

**Example** 3.5 (**Normalized measure**). Consider a function $\rho : \mathcal{X} \to \mathbb{R}^+$, called the **difficulty function**, and a metric $d : \mathcal{Y} \times \mathcal{Y} \to \mathbb{R}^+$. The $\rho$-normalized nonconformity measure associated with $d$ is defined as follows:

$$A^{\rho}_d(x, y) := \frac{A_d(x, y)}{\rho(x)} = \frac{d\big(\hat{y}(x), y\big)}{\rho(x)} \, . \tag{3.41}$$

Nonconformity measures of this form are often said to be **normalized** (or **locally adaptive**) and the resulting algorithm is called **normalized con-** **formal prediction** (NCP) (or **locally adaptive conformal prediction**).

This function can be interpreted as a conditional or feature-dependent Mahalanobis metric (see further below). In the case of mean-variance estimators (see Section 3.2.2 and Example 4.8 further on), the difficulty function could be an estimate of the predictive variance (or standard deviation), resulting in the **standardized nonconformity measure**

$$A_{\text{st}}(x, y) := A_{t1}^{\widehat{\sigma}} = \frac{|\hat{\mu}(x) - y|}{\widehat{\sigma}(x)} .$$

(3.42)

Another possible choice is the empirical uncertainty of a data subset found using, for example, a $k$-nearest neighbour algorithm (Papadopoulos, Vovk, & Gammerman, 2011). It should be clear that this choice, although interesting because it takes into account the number of similar data points and, hence, the data uncertainty, does introduce extra computational overhead. For the standard metric on $\mathbb{R}^n$, we can further generalize (3.41) to a proper **Mahalanobis metric**:

$$A_{\text{Mah}}(x, y) := \sqrt{\big(\hat{y}(x) - y\big)^{\mathsf{T}} \rho(x) \big(\hat{y}(x) - y\big)} ,$$

(3.43)

where $\rho : \mathcal{X} \to \mathbb{R}^{n \times n}$ is now a positive-definite matrix-valued function of the features. On $\mathbb{R}$, the functions $A_{\text{st}}$ and $A_{\text{Mah}}$ are clearly equivalent (up to inverting $\rho$).

As with Eq. (3.39), an explicit expression for the resulting prediction interval can be derived, where $\hat{y}_{\pm} : \mathcal{X} \to \mathbb{R}$ denote the lower and upper bounds as before:

$$a^* = \frac{|\hat{y}(x) - \hat{y}_{\pm}(x)|}{\widehat{\sigma}(x)}$$

$$\Longleftrightarrow a^* \widehat{\sigma}(x) = \pm \big(\hat{y}_{\pm}(x) - \hat{y}(x)\big)$$

$$\Longleftrightarrow \hat{y}_{\pm}(x) = \hat{y}(x) \pm a^* \widehat{\sigma}(x) ,$$

(3.44)

where $a^*$ is again the critical nonconformity score.

**Extra** 3.6. By further generalizing the above construction of normalized conformal predictors, we could consider a family $d_- : \mathcal{X} \to \text{Met}(\mathcal{Y})$ of metric structures modulated by the feature space $\mathcal{X}$. The procedure would virtually remain the same with the nonconformity measure now

given by

$$A(x, y) := d_x\big(y, \hat{y}(x)\big).\tag{3.45}$$

### 3.3.2 Interval predictors

In this section, the models that directly predict the lower and upper bounds of prediction intervals are considered. A classic example would be the $\alpha/2$- and $(1 - \alpha/2)$-quantile estimates for a given significance level $\alpha \in [0, 1]$. For this class of estimators, a reasonable choice of nonconformity measure is

$$A_{\text{int}}(x, y) := \max\big(\hat{y}_-(x) - y, y - \hat{y}_+(x)\big),\tag{3.46}$$

where, as before, $\hat{y}_\pm : \mathcal{X} \to \mathbb{R}$ denote the lower and upper interval bounds. This function was originally introduced for quantile regression in Y. Romano et al. (2019). As was the case for point predictors with Eq. (3.37), it only considers the distance between points in the target space and, therefore, the critical value of the ICP algorithm will be independent of the underlying features. This again allows us to compute the critical value and construct a prediction interval for every new instance with $\mathcal{O}(1)$-time complexity as follows:

$$\Gamma_{\text{int}}^\alpha(x) := \big[\hat{y}_-(x) - a^*, \hat{y}_+(x) + a^*\big].\tag{3.47}$$

The idea behind this construction is also rather straightforward. The conformal predictor estimates the amount by which the constructed intervals are too small (or too wide) on average on the calibration set and corrects for these errors in the future. Two related normalized nonconformity measures were considered in Kivaranovic, Johnson, and Leeb (2020); Sesia and Candès (2020):

$$A_{\text{int},m}(x, y) := \max\left(\frac{\hat{y}_{1/2}(x) - y}{\hat{y}_{1/2}(x) - \hat{y}_-(x)}, \frac{y - \hat{y}_{1/2}(x)}{\hat{y}_+(x) - \hat{y}_{1/2}(x)}\right),\tag{3.48}$$

$$A_{\text{int},w}(x, y) := \max\left(\frac{\hat{y}_-(x) - y}{\hat{y}_+(x) - \hat{y}_-(x)}, \frac{y - \hat{y}_+(x)}{\hat{y}_+(x) - \hat{y}_-(x)}\right),\tag{3.49}$$

where $\hat{y}_{1/2} : \mathcal{X} \to \mathbb{R}$ is an estimate of the conditional median. However, it was empirically observed that the inherent heteroskedasticity of quantile regression models is already sufficient and that the added normalization only increases the size of the prediction intervals without gaining much in terms of validity.

**Property** 3.7. All nonconformity measures introduced in the preceding paragraphs are induced by a metric structure.

*Proof*. For the metric measure and normalized metric measure this was already covered. To include, and vastly generalize, the interval non-conformity measure, we should not just consider the distance between points, but also the distance between sets (A.83). In this way, Eq. (3.46) becomes

$$A_{\text{int}}(x, y) = \underline{d}\big(y, \Gamma^\alpha_{\text{int}}(x)\big),  \tag{3.50}$$

where $\underline{d}$ denotes the signed distance (A.84).                    □

To unify all these approaches, consider the following construction. Given the critical nonconformity score $a^* \in \mathbb{R}^+$ and the confidence predictor $\Gamma^\alpha : \mathcal{X} \to 2^{\mathcal{Y}}$, which can now just be a point estimate $\Gamma^\alpha(x) = \{\hat{y}(x)\}$, the conformalized predictor is given by

$$\widetilde{\Gamma}^\alpha(x) := \sup\big\{S \in 2^{\mathcal{Y}} \,\big|\, \forall y \in S : \underline{d}\big(y, \Gamma^\alpha(x)\big) \le a^*\big\},  \tag{3.51}$$

where the supremum is taken with respect to the partial order on $2^{\mathcal{Y}}$ given by inclusion of sets.

To finish this section, it is also interesting to recall Section 2.3.5. There, we saw how conformal prediction in and of itself does not present us with the tools to handle the distinction between aleatoric and epistemic uncertainty. However, in certain cases, such as the one treated in this section, we can say slightly more about this decomposition. When the baseline model is already a confidence predictor, the critical value of nonconformity measures such as Eq. (3.46) gives an idea of how much the model in its current epistemic state misjudges the uncertainty present in the problem. Since the same aleatoric uncertainty is essentially present during the training process of the model and the determination of the nonconformity scores, this quantity represents a contribution to the epistemic uncertainty.

### 3.3.3 Jackknife extension

Some other algorithms strongly related to the cross-conformal predictors from Section 2.5.1 were further developed in Barber et al. (2021b). The starting point is again the concept of jackknife estimates (Miller, 1974; Quenouille, 1956; Tukey, 1958). Consider all **jackknife estimates** $\hat{y}_{(i)}$, sometimes called **leave-one-out estimates**, for a data set $\mathcal{D} \equiv \{(x_i, y_i)\}_{i \leq n}$. These are the estimates obtained by training models on the reduced data sets $\mathcal{D} \backslash \{(x_i, y_i)\}$. Akin to the conformal prediction interval (3.39), the jackknife interval is constructed as follows:

$$\Gamma^\alpha_{\text{jk}}(x) := \left[ \hat{y}(x) - A^*_{\text{jk}}, \hat{y}(x) + A^*_{\text{jk}} \right],\tag{3.52}$$

where

$$A^*_{\text{jk}} := q_{(1 + 1/n)(1 - \alpha)}\big(\big\{ |y_i - \hat{y}_{(i)}(x_i)| \mid (x_i, y_i) \in \mathcal{D} \big\}\big)\tag{3.53}$$

with $q_\alpha : \mathbb{R}^* \to \mathbb{R}$ the quantile function (A.65) as before. Note that in this definition, the interval is obtained by subtracting (resp. adding) a certain quantile of the jackknife residuals from (resp. to) the model trained on the full data set $\mathcal{D}$. The jackknife+ method modifies this approach by interchanging the quantile and subtraction/addition operations:

$$\Gamma^\alpha_{\text{jk+}}(x) := \left[ q^-_{(1 + 1/n)\alpha}\big(\mathcal{J}^-(x)\big), q_{(1 + 1/n)(1 - \alpha)}\big(\mathcal{J}^+(x)\big) \right],\tag{3.54}$$

where

$$\mathcal{J}^\pm(x) := \big\{ \hat{y}_{(i)}(x) \pm |y_i - \hat{y}_{(i)}(x_i)| \,\big|\, (x_i, y_i) \in \mathcal{D} \big\}\tag{3.55}$$

and

$$q^-_\alpha(\{x_1, \ldots, x_n\}) := -q_{1 - \alpha}\big(\{-x_1, \ldots, -x_n\}\big) = x_{\lfloor n\alpha \rfloor}.\tag{3.56}$$

Note that the ordinary jackknife intervals have virtually the same form after replacing $\mathcal{J}^\pm$ with

$$\mathcal{I}^\pm(x) := \big\{ \hat{y}(x) \pm |y_i - \hat{y}_{(i)}(x_i)| \,\big|\, (x_i, y_i) \in \mathcal{D} \big\}.\tag{3.57}$$

As for cross-conformal predictors, a validity guarantee at the $2\alpha$-level can be obtained.

**Property 3.8.** Jackknife+ estimators at significance level $\alpha \in [0, 1]$ are conservatively valid at the $1 - 2\alpha$ level.

*Proof.* See Section 6 of Barber et al. (2021b). The authors also prove that the factor 2 is required and that explicit counterexamples to the validity of the naive jackknife method can be constructed. A modification, attaining the $1 - \alpha$ coverage goal, is provided by replacing the global es-timate in Eq. (3.52) by the minimum and maximum of the jackknife estimates. (However, this modification is, in practice, too conservative.) □

Instead of having to obtain a jackknife estimate for every data point, we can also use the cross-validation approach from Vovk (2015). In this case, the jackknife estimates in Eq. (3.54) are replaced by cross-validation estimates. It is important to note that the output of a CV+ estimator is not necessarily the same as that of the associated cross-conformal predictor.

**Property** 3.9. CV+ estimates are more conservative than the associated cross-conformal estimates:

$$\Gamma^\alpha_{\text{CCP}}(x) \subseteq \Gamma^\alpha_{\text{CV+}}(x). \tag{3.58}$$

Since CV+ prediction regions are necessarily intervals, we also have the following stronger result for the convex hull:

$$\text{Conv}\big(\Gamma^\alpha_{\text{CCP}}(x)\big) \subseteq \Gamma^\alpha_{\text{CV+}}(x). \tag{3.59}$$

*Proof.* See Barber et al. (2021b).                                        □

### 3.3.4 Implementation

The code excerpt in Algorithm 8 shows how easy it is to implement (inductive) conformal prediction in a basic Python setup. Note that the `NumPy` package is used for convenience. The numerical simplicity of (inductive) conformal prediction clearly does not do any justice to the beauty and usefulness of the framework.

The most popular conformal prediction library (again in Python) at the time of writing is the `MAPIE`[5] library (Taquet, Blot, Morzadec, Lacombe, & Brunel, 2022). This library is based on the `scikit-learn` package (Pedregosa et al., 2011) and uses the same conventions. It supports all main conformal prediction approaches, such as ICP (Algorithm 2), CCP (Algorithm 4), Jackknife+ (Section 3.3.3), etc. Moreover, it can also be used for classification and calibration. In Algorithm 9, a small example is shown where the same task as in Algorithm 8 is performed using `MAPIE`.

Another library that is gaining traction is the `crepes` library (Boström, 2022).

---

[5] This stands for 'Model Agnostic Prediction Interval Estimator'.

---

**Algorithm 8:** (Inductive) Conformal Regression with `NumPy`

---

```python
1  import math
2  import numpy as np

3  # The function calculates the critical level (empirical quantile) of the
     nonconformity scores
4  def critical(scores, alpha):
5      scores = np.sort(scores)
6      level = min((1 + (1 / scores.shape[0])) * (1 - alpha), 1)
7      index = math.ceil(level * scores.shape[0]) - 1
8      return scores[index]

9  # Assume that a model has been trained before
10 # and that the nonconformity measure is given by the absolute
     residuals
11 preds = model.predict(X_val)
12 scores = np.abs(preds - y_val)
13 crit = critical(scores, 0.9) # The significance level was chosen to be
     α = 0.1 as an example

14 pi = np.zeros((X_test.shape[0], 2))
15 pi[:, 0] = preds - crit
16 pi[:, 1] = preds + crit
```

---

**Algorithm 9:** (Inductive) Conformal Regression with `MAPIE`

---

```python
1  import numpy as np
2  from mapie.regression import MapieRegressor

3  # Assume that a model algorithm has been defined using the
     scikit-learn conventions
4  mapie_regressor = MapieRegressor(estimator = model, cv = 'split')
5  mapie_regressor = mapie_regressor.fit(X_train, y_train)

6  preds, pi = mapie_regressor.predict(X_test, alpha = [0.1])
7  # As before, the significance level is fixed at α = 0.1
```

---

**Algorithm 10:** (Inductive) Conformal Regression with `crepes`

---

1  import numpy as np
2  from crepes import WrapRegressor

3  # Assume that a model algorithm has been defined using the
     `scikit-learn` conventions
4  crepes_regressor = WrapRegressor(model)
5  crepes_regressor.fit(X_train, y_train)
6  crepes_regressor.calibrate(X_val, y_val)

7  pi = crepes_regressor.predict_int(X_test, confidence = 0.9)
8  # As before, the significance level is fixed at $\alpha = 0.1$

---

This package also allows to wrap existing models. It only supports inductive conformal prediction, but includes an extension for conformal predictive systems (Section 2.5.2). The small code snippet in Algorithm 10 performs the same task as Algorithms 8 and 9 in `crepes`.

## 3.4    Experimental comparison

### 3.4.1    Data

Most of the data sets were obtained from the UCI repository (Dua & Graff, 2017). Table 3.1 shows the number of data points and (selected[6]) features and the skewness and (Pearson) kurtosis of the response variable. Specific references are given in Table 3.2.

All data sets were standardized before training (both features and response variables). The data sets `blog` and `fb1` were also analysed after first taking the logarithm of the response variable because these data sets are extremely skewed, which is reflected in the high skewness and kurtosis as shown in the fourth column of Table 3.1, and are believed to follow a power law distribution. This strongly improved the $R^2$-coefficient (Definition A.68) of the various models, but did not improve the prediction intervals, and therefore, these results are not included. The `crime` data set comes in two versions: the original data set consists of integer-valued data (count data), whereas the version used here was preprocessed using an unsupervised standardization algorithm (Redmond & Baveja, 2002). Although standardized, the data set retains (some of) its count data properties. The `traffic` data set, aside from being very small, is also extremely sparse (on average 14 features are zero). It should be noted that all of the data sets used in this study were considered as ordinary (static) data sets. Even though some of them could be considered in a time series context, no autoregressive features were included. The main reason to exclude autoregressive features is that most, if not all, methods considered in this study assume the data to be i.i.d. (or exchangeable), a property that is generically not valid for time series data.

---

[6]  Some data sets provided additional features that were either not useful due to their values or due to additional information provided by the source.

Table 3.1: Details of the data sets.

| Name | Samples | Features | Skewness / Kurtosis |
|---|---|---|---|
| concrete | 1030 | 8 | 0.42 / 2.68 |
| naval | 11934 | 14 | 0.0 / 1.8 |
| turbine | 9568 | 4 | 0.31 / 1.95 |
| puma32H | 8192 | 32 | 0.02 / 3.04 |
| residential | 372 | 105 | 1.26 / 5.15 |
| crime2 | 1994 | 123 | 1.52 / 4.83 |
| fb1 | 40949 | 54 | 14.29 / 301.44 |
| blog | 52397 | 280 | 12.69 / 235.3 |
| traffic | 135 | 18 | 1.07 / 3.74 |
| star | 2161 | 39 | 0.29 / 2.63 |

### 3.4.2 Models and training

All neural networks were constructed using the `PyTorch` (Paszke et al., 2019) Python package. The general architecture for all neural network-based models was fixed. For all differentiable models, the *Adam-optimizer* was used for weight optimization with a fixed learning rate of $5 \times 10^{-4}$, as in Y. Romano et al. (2019). The number of epochs was limited to 100, unless stated otherwise. All neural networks contained only a single hidden layer with 64 neurons. The activation functions after the first layer and the hidden layer were of the ReLU type, while the activation function at the output node was simply a linear function.

Table 3.2: Sources of the data sets.

| Name | Source |
| --- | --- |
| concrete | Yeh (1998) |
| naval | Coraddu et al. (2016) |
| turbine | Kaya, Tüfekci, and Gürgen (2012) |
| puma32H | Corke (1996) |
| residential | Rafiei and Adeli (2016) |
| crime2 | Redmond and Baveja (2002) |
| fb1 | Singh, Sandhu, and Kumar (2015) |
| blog | Buza (2014) |
| traffic | Ferreira, Affonso, and Sassi (2011) |
| star | Achilles et al. (2008) |

For each of the four classes of interval predictors in Section 3.2, at least one example was chosen for a general comparison. Furthermore, to handle calibration issues, conformal prediction was chosen as a post-hoc method due to its nonparametric and versatile nature. Every model that produces a prediction interval (or an uncertainty estimate) is also re-examined after 'conformalization'. For this step, we chose the residual score (3.38) for point predictors and interval score (3.46) for interval predictors. Hyperparameter optimization, if applicable, was performed on a validation set containing 5% of the training samples, unless stated otherwise. The specific architectures are as follows:[7]

1. Bayesian methods:

   (1) Gaussian Process: As kernel, a *radial basis function* (RBF) kernel was used. This automatically incorporates out-of-distribution uncertainty. The default implementation from the GPyTorch library was used (Gardner, Pleiss, Bindel, Weinberger, & Wilson, 2018).
This library provides a multitude of different approximations and
deep learning adaptions for Gaussian processes. A heteroskedas-
tic likelihood function with trainable noise parameter was chosen
and the model was trained for 50 epochs (following the library's
heuristics). Optimization of the noise parameter was handled in-
ternally, so no additional validation set was required.

---

[7] All code was made publicly available at https://github.com/nmdwolf/ValidPredictionIntervals.

(2) Approximate GP: Because in most present-day cases the data sets
are too large to efficiently apply GPs, a variational approxima-
tion was included. The stochastic variational GP approximation
(SVGP) was chosen (Hensman et al., 2015), which is also imple-
mented in `GPyTorch`. The same likelihood function and number
of epochs were used as for the exact model.

2. Ensemble methods:

(1) Dropout ensemble: An ensemble average of 50 MC samples was
used. An early stopping criterion, based on the minimum of the
loss function, was employed to avoid overfitting. The dropout
probability was optimized over the interval $[0.05, 0.5]$ with steps
of 0.05.

(2) Mean-variance ensemble: The ensemble mean-variance estimator
from Kendall and Gal (2017) was chosen because this model ex-
tends the dropout-based ensemble by explicitly incorporating un-
certainty. An ensemble average of 50 MC samples was used. The
dropout probability, which was the same for all nodes, was op-
timized over the interval $[0.05, 0.5]$ with steps of 0.05 and early
stopping was employed to counteract overfitting.

(3) Deep ensembles: The ensemble consisted of five models and the
adversarial step size was equal to 0.01 times the range of the cor-
responding dimension as in Lakshminarayanan et al. (2017). To
prevent overfitting, $\ell^2$-regularization with $\lambda = 10^{-6}$ was applied,
i.e. the loss function $\mathcal{L}$ was replaced by

$$\mathcal{L}_{\text{tot}}(\mathcal{D}) := \mathcal{L}(\mathcal{D}) + \lambda \|\boldsymbol{w}\|_2^2, \tag{3.60}$$

where $\boldsymbol{w}$ is the vectorized weight matrix of the neural network.

3. Direct interval estimation methods:

Table 3.3: Abbreviations of model names.

| Model | Abbr. | Model | Abbr. |
|-------|-------|-------|-------|
| Neural network | NN | Random forest | RF |
| Quantile regressor | QR | Deep ensemble | DE |
| Dropout ensemble | Drop | Gaussian dropout-MVE | MVE |
| Exact Gaussian Process | GP | Stochastic Variational GP | SVGP |

(1) Neural network quantile regression: The same softening factor $w = 2$ as in Y. Romano et al. (2019); Sesia and Candès (2020) was used. The early stopping criterion for neural networks was also modified to work with the average length and coverage degree of the prediction intervals instead of the loss function of the network. Dropout with $p = 0.1$ and $\ell^2$-regularization with $\lambda = 10^{-6}$ were applied at training time.

4. Conformal prediction:

(1) Neural network: A standard neural network point predictor was chosen as a baseline model. Early stopping, dropout with $p = 0.1$ and $\ell^2$-regularization with $\lambda = 10^{-6}$ were applied at training time as a regularization method.

(2) Random forest: The default implementation from `scikit-learn` was used (Pedregosa et al., 2011). The number of trees was fixed at 100, while all other hyperparameters were left at their default value.

These models will be abbreviated as shown in Table 3.3 in the remainder of this section.

All experimental results were obtained by evaluating the models on 50 different train/test splits of the data sets in Table 3.1. The test set always contained 20% of the data. If a calibration set was needed for post-hoc calibration, the training set was further divided into two equal-sized sets. Because the models were tested both with and without post-hoc calibration, this approach also allows to see how the performance changes when halving the size of the training set. The confidence level was fixed at 0.9 (equivalently, the significance level was fixed at $\alpha = 0.1$).

### 3.4.3 Results

➢ **Overview**

The experimental results, aggregated over 50 random splits, are shown in two ways: graphs and tables. For the graphical representation, Figs. 3.3 and 3.4 show the empirical coverage degree (2.6), average width (2.11) and $R^2$-coefficient (Definition A.68) for a selection of data sets. For an overview of the results of all models and data sets, see Figs. 3.8 to 3.13. The models NN and RF are shown separately, since these do not provide a baseline uncertainty estimate and, consequently, only a conformalized variant exists. Except for these two models, Fig. 3.3 reports three variants for every model. The first one is trained on half of the training set, the second one is trained on all of the training set and the third one is conformalized using a 50/50 split of the training set. The results for the $R^2$-coefficient of the conformalized models are the same as for the models trained on half of the data set, since conformal prediction is a post-hoc method. Therefore, only the first two variants are shown in Fig. 3.4.

For a tabular representation, see Tables B.1 to B.6. The results for every metric (coverage, average width and $R^2$-coefficient) are each split into two tables for clarity, with five data sets each. The rows represent the various models and the columns correspond to the different data sets. For every combination, the average over 50 random train-test splits is reported. The standard deviation is shown between parentheses.

For the exact Gaussian processes (GP), no results are reported on the data sets `fb1` and `blog` due to the excessively large memory and time consumption. While these sets are not extraordinary compared to modern, real-life data sets, they already require a considerable amount of memory due to the bad scaling behaviour as noted in Section 3.2.1. Approximate models, such as the variational approximation used in this study, are imperative in these cases. When the runtime limit for training was exceeded, the value is replaced by the label 'OoT' (out of time) in the tables. Moreover, certain values were deemed to be out of range, e.g. $R^2$-coefficients below 0 or cases where the standard deviation was unreasonably large. This is indicated by the label

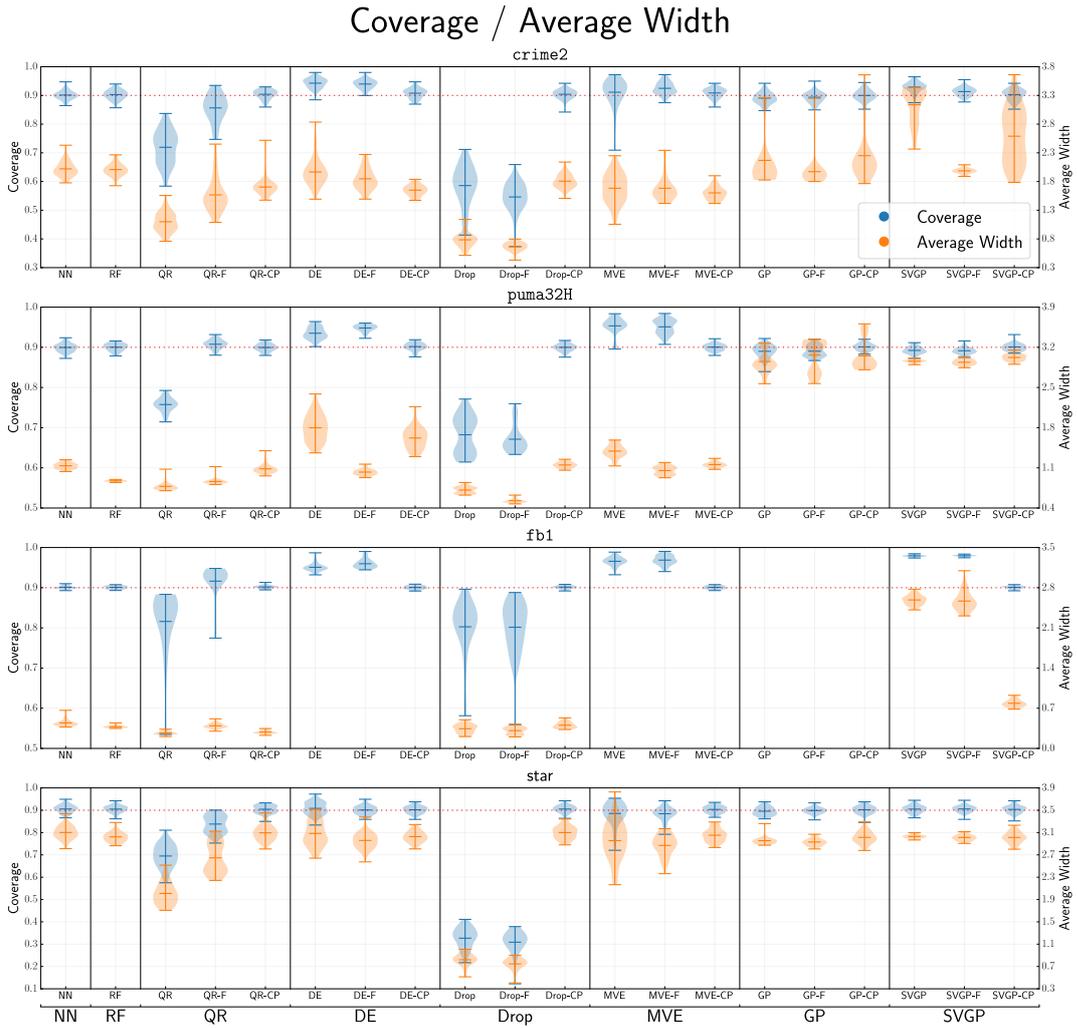

Figure 3.3:  Prediction interval properties for a selection of data sets.  Blue and orange regions show the coverage degree and the average interval width, respectively. For every model, three variants are shown (except for NN and RF). Fully trained and conformalized models are indicated by the tags 'F' and 'CP', respectively.  The red dashed line indicates the nominal coverage level.

'OoR' (out of range) in the tables.

➢   **General impression**

When comparing the results, it should be immediately clear that the coverage is much more concentrated around the nominal confidence level of 0.9 for the conformalized models than for the baseline models (as guaranteed by the marginal validity theorems 2.27 and 2.29).  Comparing Figs. 3.3 and 3.4, we can also see that higher variability in the predictive power often corresponds to higher variability in the interval quality (both coverage and efficiency). This is not surprising since a model in general performs worse when the uncertainty is higher. As most of the models explicitly use the predictions to build prediction intervals, this relation can be expected to be even stronger. More recently, this has also been studied in more detail in Cherubin (2023).

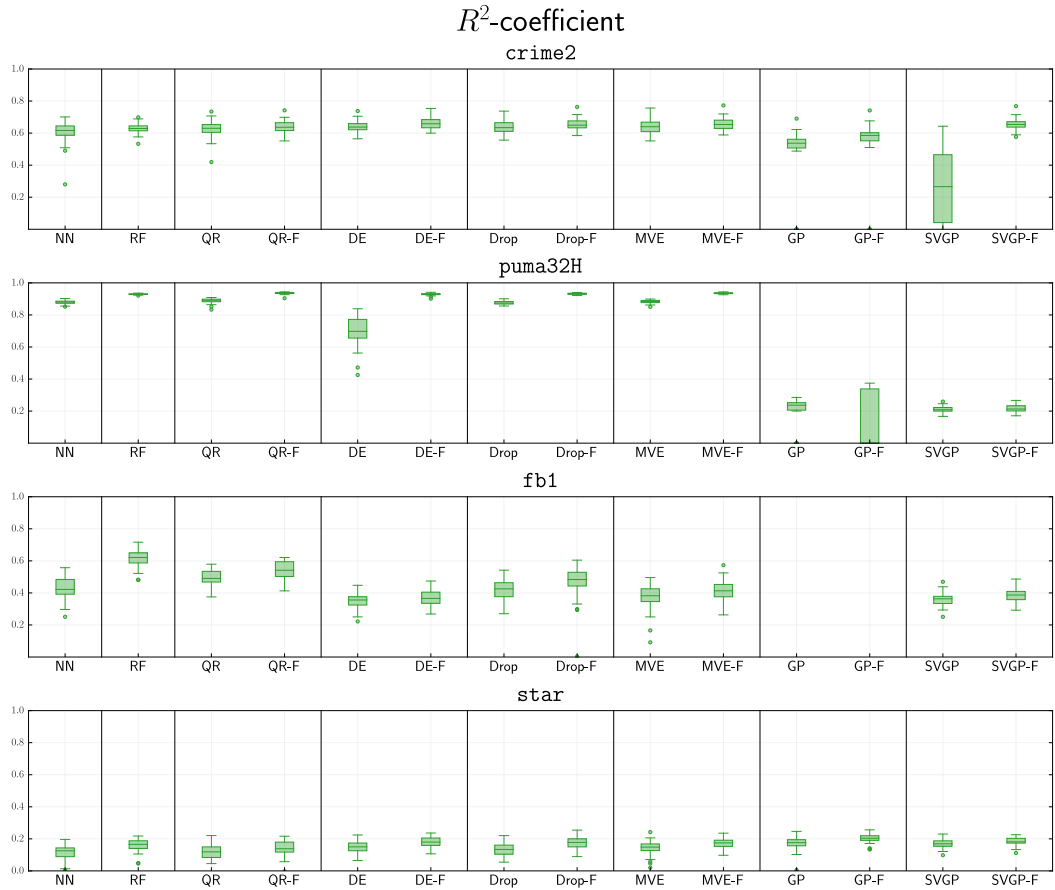

Figure 3.4: $R^2$-coefficients for a selection of data sets. For every model, two variants are shown (except for NN and RF): trained on the full data set and conformalized. Fully trained models are indicated by the tag 'F'.

The $R^2$-coefficient for the NN and Drop-CP models on the `blog` data set show a strong variability, while the average width remains small and almost constant. This can be explained by the strong skewness present in the data set. The $R^2$-coefficient is sensitive to extreme outliers, while the prediction intervals are not, as long as the outlier proportion is less than the significance level $\alpha \in [0, 1]$. This also explains why almost all models give reasonably good intervals for both the `fb1` and `blog` data sets. However, for both the DE and MVE models, the results for the `fb1` and `blog` data sets are missing because the average widths differed by about two orders of magnitude compared to the other data sets and models and were, therefore, deemed to be nonsensical. These methods inherently take into account the data uncertainty and, therefore, cannot discard outliers.

➢ **Calibration**

As mentioned above, Fig. 3.3 shows, except for the models NN and RF, three variants for every model: one trained on half of the training set, one trained on all of the training set and one conformalized using a 50/50 split. This allows to see both the influence of the data set size (variants 1 vs. 2) and of the post-hoc conformalization step (variants 1 vs. 3). As before, the average width for the DE and MVE models is not shown for the data set `fb1` because these were too large to be meaningful.

From this figure, it is again clear that the uncalibrated models do not (not even approximately) satisfy the validity constraint. They either underestimate the uncertainty or produce overly conservative prediction intervals. When comparing the models trained on half of the data set and the ones trained on the full data set, it is generally the case that the fully trained model achieves better results in terms of both coverage and efficiency. The models incorporating a probabilistic prior (DE, MVE and GP) come out as the most stable ones with respect to data size change.

For each of the selected data sets, Fig. 3.5 shows the best five models in terms of average width, excluding those that do not (approximately) satisfy the coverage constraint (2.10). This figure also shows that there is quite some variation in the models. There is not a clear best choice. Because on most data sets the models produce uncalibrated prediction intervals, post-hoc calibrated models generally produce superior intervals. However, when the models already produce (approximately) calibrated intervals from the start, they tend to give tighter intervals than their conformalized counterparts, because there is more data available for training.

In Figs. 3.3 and 3.5, we can also see that the models incorporating a Gaussian assumption (GP, SVGP, MVE and DE) fail to deliver optimal results in the case of non-Gaussian data (e.g. the `fb1` data). The desired confidence level can be attained, but there is a large trade-off with the average width. As noted before, the models (MVE and DE) using scoring rule (3.24) some-times produce intervals where the average width is some orders of magnitude larger than for the other models. Leaving out the inherent uncertainty contribution to the variance, the third term in Eq. (3.25), resolves this prob-lem, but then the model loses its ability to approximately reach the required coverage level, similar to the ordinary dropout networks. (Removing this term turns the models into ordinary ensemble models that are not forced to learn the aleatoric uncertainty.) However, as stated in the previous section, the data sets `fb1` and `blog` are strongly skewed. After reducing the skewness by discarding data points greater than a certain threshold, the average width also strongly decreases. This observation, combined with the fact that it are exactly the models explicitly incorporating data uncertainty that suffer from excessively wide prediction intervals, might also point to a different interpretation. Namely, in the case of outliers, the model has almost no information of the conditional distribution and, hence, this corresponds to a situation of almost full uncertainty. From this point of view, an extremely wide prediction interval might be more sensible than a mild extrapolation of the lower-uncertainty regions containing the vast majority of the data points.

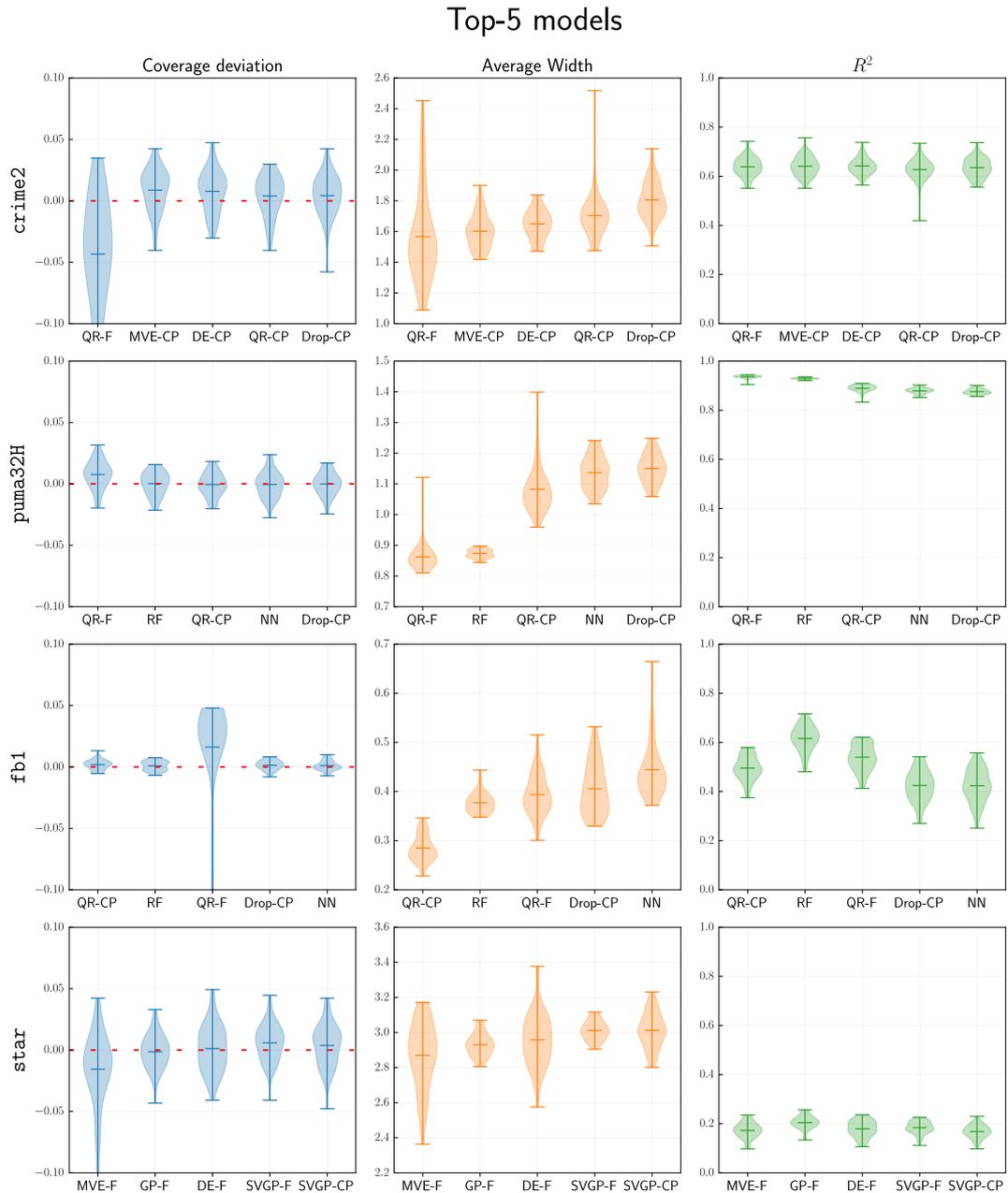

Figure 3.5: Overview of the best models based on average interval width. The left column shows the deviation of the coverage from the desired level ($\alpha = 0.9$), which is indicated by the red dashed line. The middle and right columns show the average width of the prediction intervals and the $R^2$-coefficient of the models, respectively. Error bars represent the standard deviation over train-test splits.

Table 3.4: Quantile gap of selected data sets.

| crime2 | puma32H | fb1 | star |
|--------|---------|------|------|
| 3.17   | 3.49    | 0.86 | 3.24 |

➢ **Efficiency**

Because all data sets were standardized, the average widths of the prediction intervals can also be compared across data sets. From this perspective, the models seem to give significantly better results on the `fb1` data set. However, when making such a comparison, we should also look at the distribution of the data itself. The difference between the $\alpha/2$- and $(1-\alpha/2)$-quantiles gives a rough estimate of the interval widths. The values of this quantity for the four selected data sets, averaged over the train-test splits, are given in Table 3.4. Similar to the $R^2$-coefficient or the *relative absolute error*, we could define a 'relative width' as the ratio of the average width as estimated by the given model and the quantile gap obtained from the unconditional data distribution, i.e. the interval width we would obtain when ignoring the underlying features in the data set. If this quantity is smaller than 1, the model is able to improve on a simple featureless model. From this point of view, it is clear that the models do not perform as well as could be thought on first sight for the `fb1` data set. At the same time, it is also clear that for the `star` data set, the models only barely improve on the naive estimate.

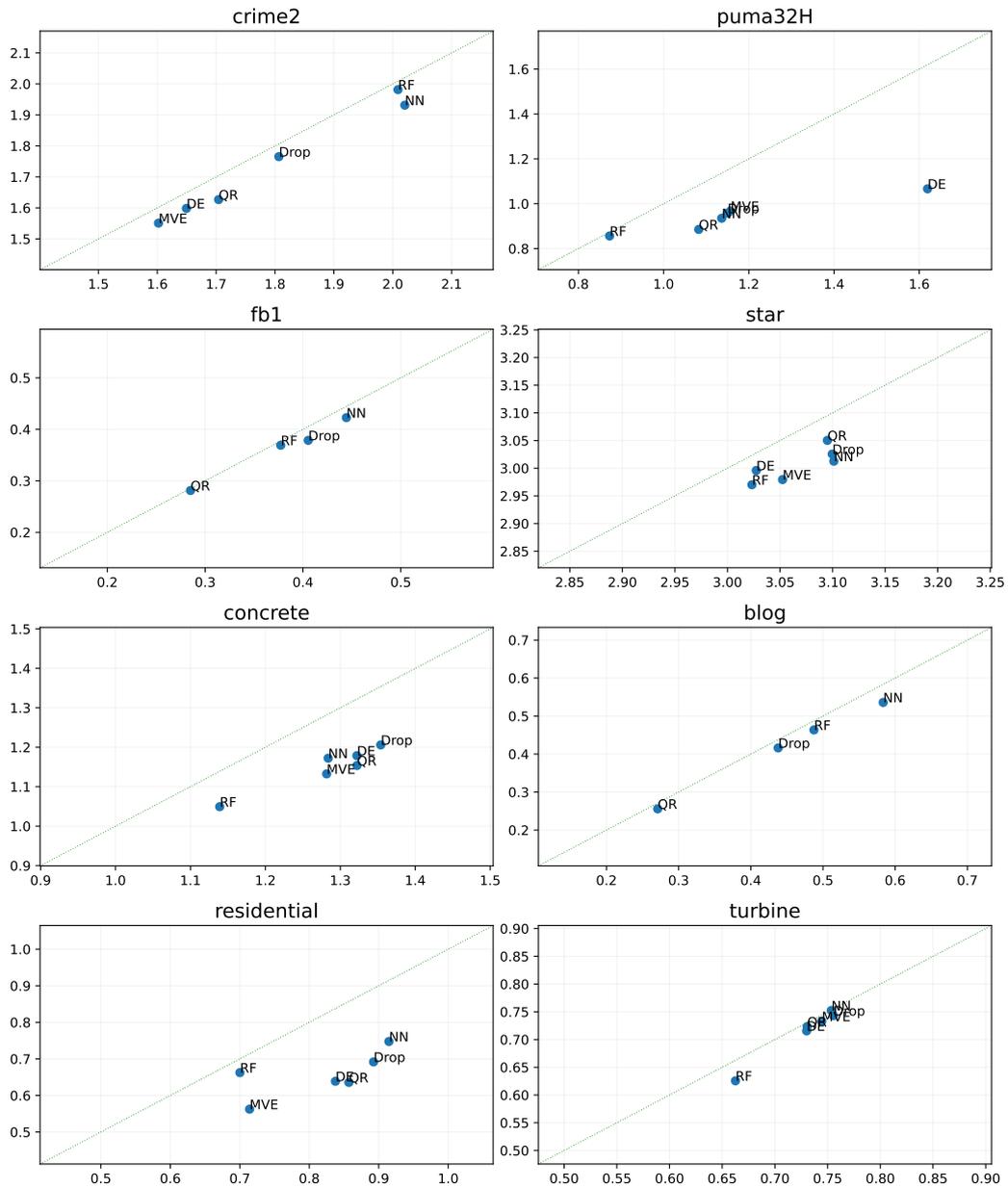

Figure 3.6: Comparison of average interval lengths between conformalized models trained on 50% of the data and models trained on 75% of the data.

➢ **Training size**

To see the influence of the training-calibration split on the resulting prediction intervals, two smaller experiments were performed where the training-calibration ratio was modified. In the first experiment, shown in Fig. 3.6, the split ratio was changed from 50/50 to 75/25, i.e. more data was used in the training step. The average coverage was not significantly changed. However, the average width of the intervals decreased on average by 7%. This corresponds to all points lying below the diagonal for all data sets.  Using more data to train the underlying models, thereby obtaining better predictions, will lead to tighter prediction intervals as long as the calibration set is not too small.  This conclusion is in line with the observations from Figs. 3.3 and 3.4.

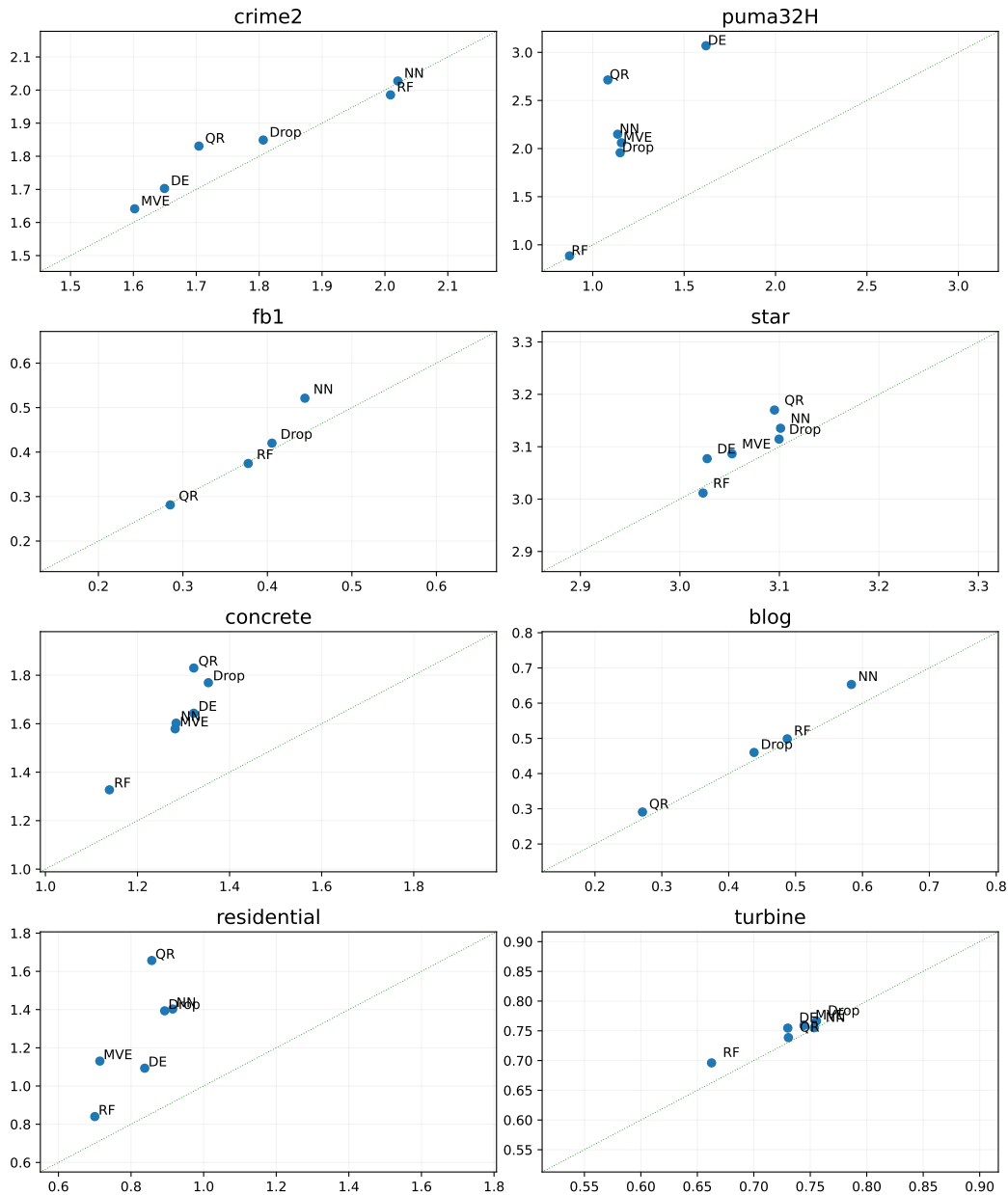

Figure 3.7: Comparison of average interval lengths between conformalized models trained on 50% of the data and models trained (and calibrated) on 100% of the data (thereby suffering from overfitting).

This experiment was then repeated in an extreme fashion, where the mod-els were trained on the full data set. This is shown in Fig. 3.7. Due to the lack of an independent calibration set, the post-hoc calibration was also performed on the same training set. This way, the influence of violating the assumptions of Algorithm 2 and the associated validity theorem could also be investigated. It was found that for some models the coverage decreased while, at the same time, the average widths increased sharply. Both of these observations should not come as a surprise. The amount by which the intervals are scaled can be interpreted as a parameter that the model has to fit. In general, it is better to use more data to train a model than to calibrate, as long as the calibration data set is representative of the true population. Moreover, optimizing hyperparameters on the training set is known to lead to overfitting, which in this case corresponds to overly optimistic prediction intervals. Conformal prediction looks at how large the errors are on the calibration set, so as to be able to correct for them in the future. However, by using the training set to calibrate, future errors are underestimated and, therefore, the ICP algorithm cannot fully correct for them.

## 3.5  Discussion

In this chapter, the general problem of (univariate) regression was treated from the point of view of uncertainty estimation. Several types of interval predictors were reviewed and compared. (An overview of the main properties of these methods is shown in Table 3.5.) For this study, two important properties were taken into account: the coverage degree and the average width (or efficiency) of the prediction intervals. We saw that without post-hoc calibration, the methods derived from a probabilistic model attained the best coverage degree. However, after calibration with conformal prediction, all methods attained the desired coverage (as expected) and, in certain cases, the calibrated models even produced more efficient intervals, i.e. intervals with a smaller average width. We also observed that the predictive power of the model and the quality of the prediction intervals are correlated. For the post-hoc calibration step, the framework of conformal prediction was used. To obtain the desired results in its inductive form, this method requires the data set to be split in a training and a calibration set. A small experiment confirmed that violating this condition leads to strongly uncalibrated intervals. Although this can be expected from an overfitting point of view, this also shows that the assumption does not merely serve a theoretical purpose.

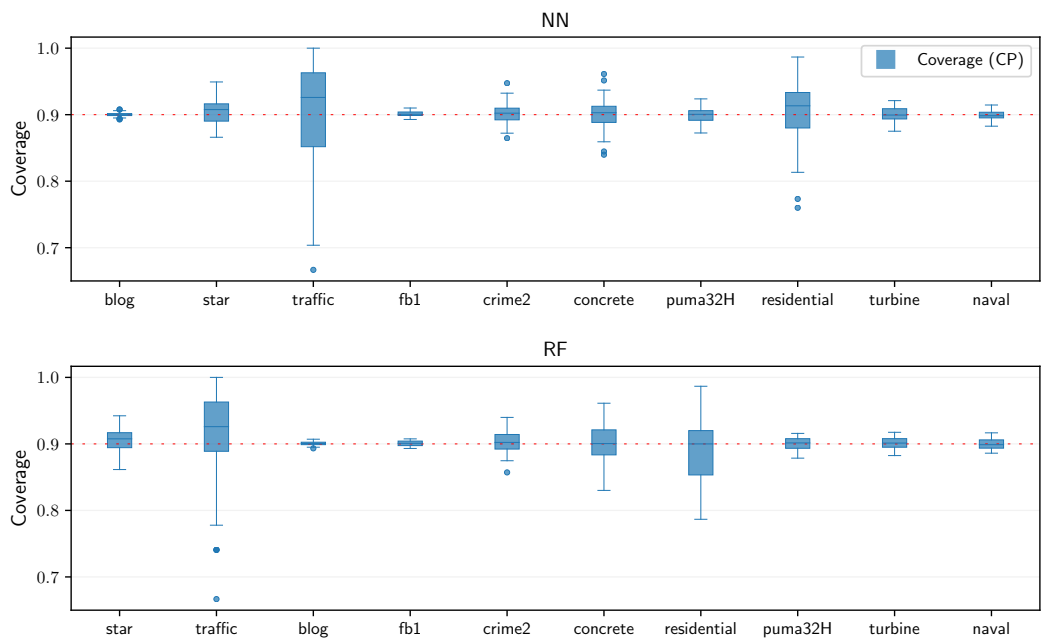

Figure 3.8: The coverage degree of the neural network and random forest models for all data sets from Section 3.4.3. The data sets are sorted along the $x$-axis in order of increasing $R^2$-coefficient.

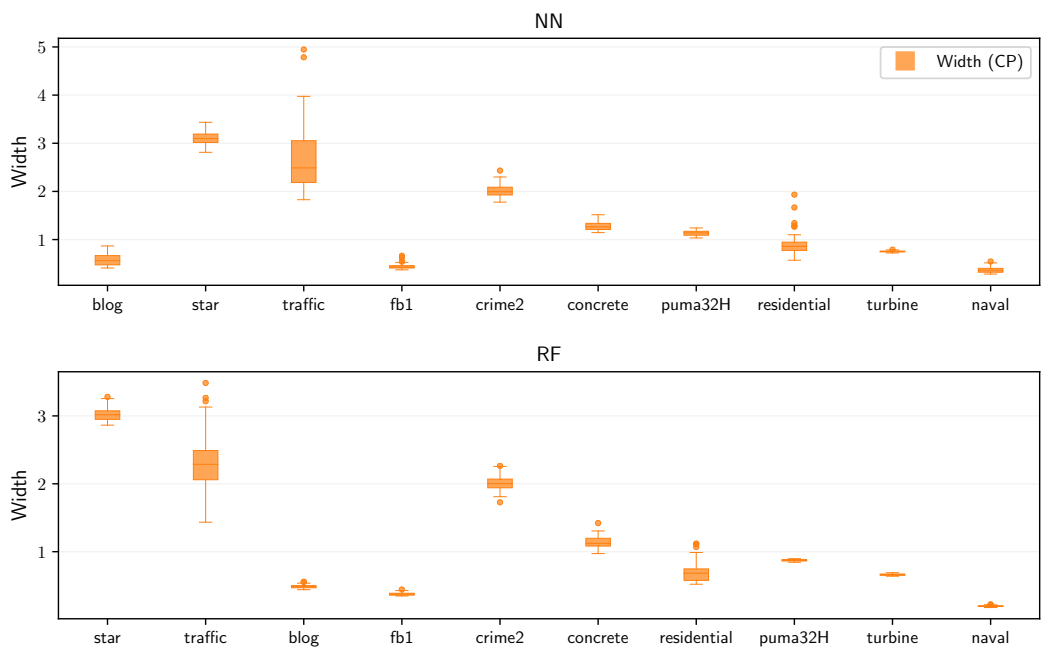

Figure 3.9: The average interval widths of the neural network and random forest models for all data sets from Section 3.4.3. The data sets are sorted along the $x$-axis in order of increasing $R^2$-coefficient.

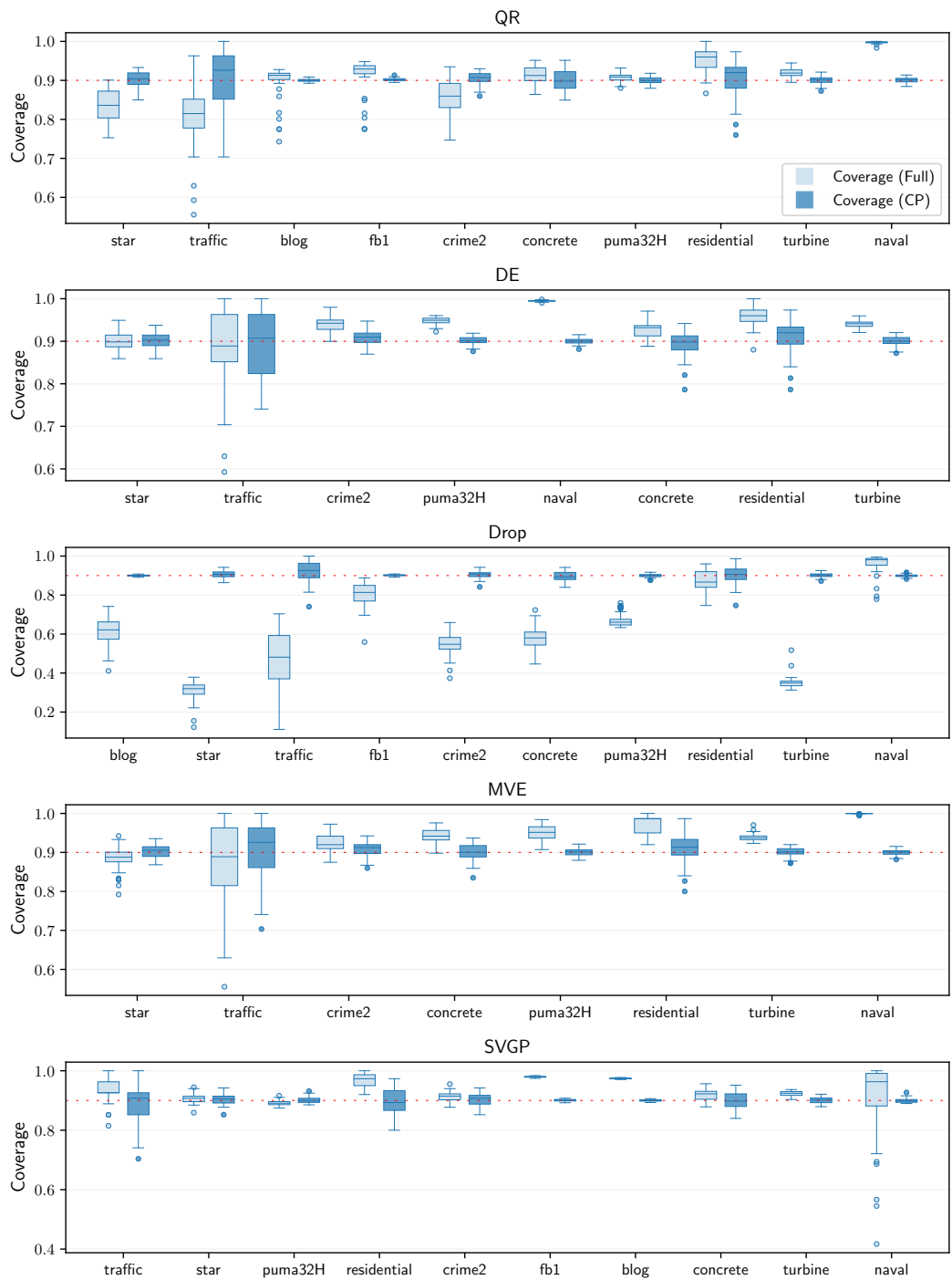

Figure 3.10: The coverage degree of the confidence predictors and data sets from Section 3.4.3. For each model, two versions are shown: one trained on the full training set, indicated by 'Full', and one trained on half of it (corresponding to the conformalized models), indicated by 'CP'. The data sets are sorted along the $x$-axis in order of increasing $R^2$-coefficient of the conformalized model.

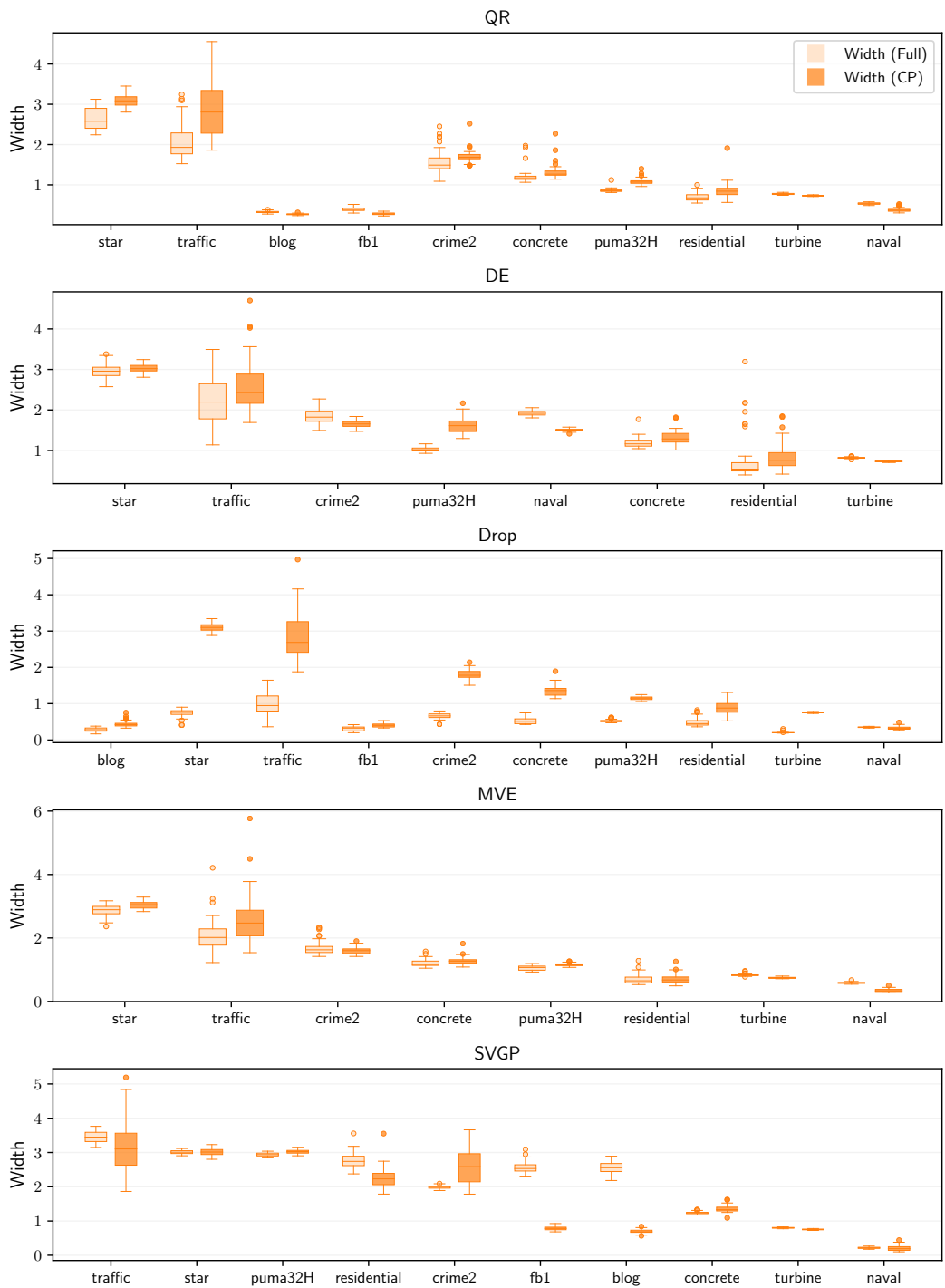

Figure 3.11: The average interval widths of the confidence predictors and data sets from Section 3.4.3. For each model, two versions are shown: one trained on the full training set, indicated by 'Full', and one trained on half of it (corresponding to the conformalized models), indicated by 'CP'. The data sets are sorted along the $x$-axis in order of increasing $R^2$-coefficient of the conformalized model.

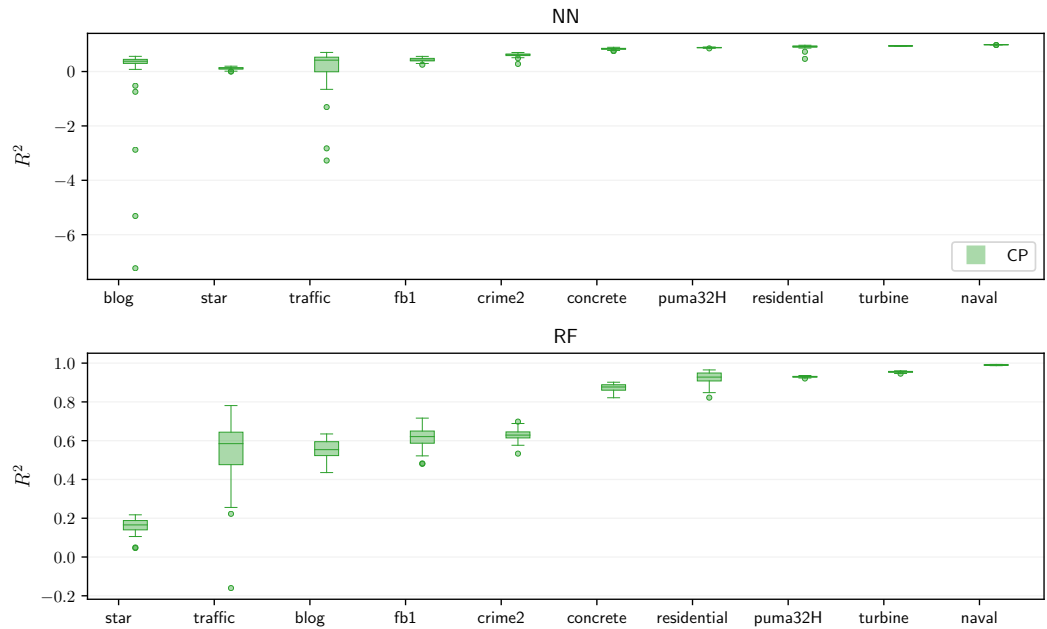

Figure 3.12: The $R^2$-coefficient of the neural network and random forest models for all data sets from Section 3.4.3.

On the other hand, increasing the relative size of the training set can result in smaller prediction intervals without having a negative influence on the calibration.

Although a variety of methods was considered, it is not feasible to include all of them. The most important omission is a more detailed overview of Bayesian neural networks (although one can argue, as was done in the section on dropout networks, that some common neural networks are, at least partially, Bayesian by nature). The main reason for this omission is the large number of choices in terms of priors and approximations, both of which strongly depend on the problem at hand. At the level of calibration, there are also some methods that were not included in this study, mostly because they were either too specific or too complex for simple regression problems. For general regression models, the literature on calibration methods is not as extensive as it is for classification models. Recently, some advances were made in which *β-calibration* (Kull, Filho, & Flach, 2017) was generalized to regression problems using a Gaussian process approach (Song, Diethe, Kull, & Flach, 2019). However, as mentioned before, GPs do not exhibit a favorable scaling behaviour and some approximations would be necessary. Another technique that was recently introduced calibrates the cumulative distribution function produced by a distribution predictor using *isotonic regres-*

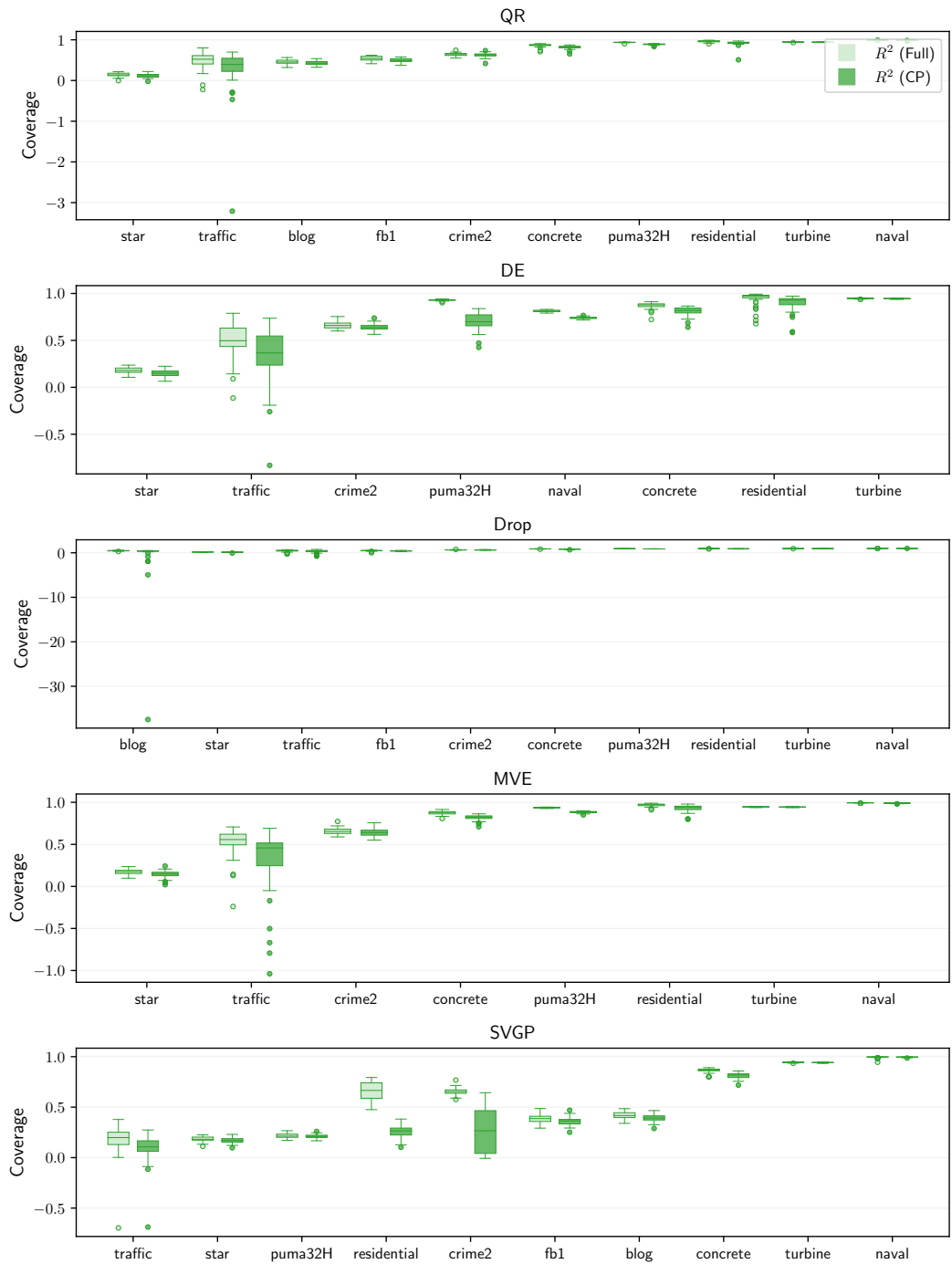

Figure 3.13: The $R^2$-coefficient of the confidence predictors and data sets from Section 3.4.3. For each model, two versions are shown: one trained on the full training set, indicated by 'Full', and one trained on half of it (corresponding to the conformalized models), indicated by 'CP'.

*sion* (Kuleshov, Fenner, & Ermon, 2018). Even though this technique is simple in spirit, it is only applicable to predictors that construct the full output distribution. By dividing the target space in a finite number of bins, Keren, Cummins, and Schuller (2018) introduced an approach where the regression problem is approximated by a classification problem such that the usual tools for classifier calibration can be applied. The main downside of this approach is that one loses the continuous nature of the initial problem. Another concept that was not covered is that of *predictive distributions* (Schweder & Hjort, 2016; Shen et al., 2018), where not only a single interval is considered, but a full distribution is estimated. This approach was combined with conformal prediction in Vovk et al. (2017) giving rise to the conformal predictive systems of Section 2.5.2.

The choice of data sets in this comparative study was very general and no specific properties were taken into account a priori. After comparing the results of the different models, it did become apparent that certain assumptions or properties can have a major influence on the performance of the models. The main examples were the normality assumption for mean-variance estimators or the proper scoring rule (3.24). Models using this loss function appear to behave very badly when used for data sets with strongly skewed variables. Since such data sets are gaining importance in the digital age, it would be interesting to both study methods tailored to these properties and study how existing models behave on outliers. Another type of data that was not specifically considered in this study, but that is also becoming increasingly important, is time series data. The main issue with this kind of data, as was also mentioned in Section 3.4.1, is the autocorrelation, which generally invalidates methods or theorems that make use of i.i.d. or exchangeability properties.

A last aspect that was not considered in this study is the conditional behaviour of the models. When constructing a model that optimizes the coverage probability (2.6), only the marginal coverage is controlled, i.e. the specific properties of an instance are not taken into account. In certain cases, it might be relevant to not only attain global validity, but also guarantee the validity on a certain subset of the instance space. A general notion of conditional validity was considered in J. Lei and Wasserman (2014); Vovk (2012). There, it was also shown how to incorporate this notion in the conformal prediction framework. This will be the content of the next chapter.

Table 3.5: Summary of method characteristics. Extensions that are not covered in the experimental section are indicated in grey.

| Method | Validity | Scalability / Complexity | Domain knowledge | Calibration data |
|---|---|---|---|---|
| Bayesian methods | No (except for exact inference with correct priors) | Only scalable with approximate inference | Yes | No |
| Ensemble methods | No | Yes (when scalable models are used) | No | No |
| Direct interval estimation | No | Only model complexity | No | Yes |
| (Inductive) conformal prediction[a] | Yes | Only model complexity | No | No |
| Cross-conformal prediction | At the $2\alpha$ level | Number of folds on top of model complexity | No | No |
| Jackknife+ | At the $2\alpha$ level | Number of data points on top of model complexity | No | No |

[a] The sorting of scores for the calculation of quantiles is ignored here.

# Conditional Validity 4

$$\partial_\mu A_\nu - \partial_\nu A_\mu + [A_\mu, A_\nu] = 0$$

Yang–Mills (1954)

The content of this chapter is based on the preprint *Conditional validity of heteroskedastic conformal regression* (Dewolf et al., 2023a), jointly written with my promotors W. Waegeman and B. De Baets.

## 4.1 Introduction

In the previous chapters, we introduced a lot of machinery, both in terms of theory and computational tools, to model the uncertainty in statistical learning theory. Some examples (Section 3.2) are Gaussian processes, quantile regression, Monte Carlo dropout or, in a model-independent and distribution-free way, conformal prediction (Sections 2.3 and 3.3). Each of these methods allows us to construct prediction regions with statistical guarantees and the validity of inductive conformal prediction has been verified numerous times, see e.g. Bosc et al. (2019); Toccaceli, Nouretdinov, and Gammerman (2017) and M. Zhang et al. (2020). A comparison of the aforementioned uncertainty quantification methods and further improvements resulting from applying conformal prediction as a (post-hoc) calibration method was investigated in the previous chapter. However, as mentioned in the conclusion, there is a very important limitation present. These methods do not take the structure inherent to the data and problem setting into account. All definitions, properties and models are stated in 'global' terms and, as a consequence, the results only hold marginally. Nonetheless, an important problem in the field of uncertainty quantification is exactly the conditional behaviour. Data points with large uncertainty, be it aleatoric or epistemic, are often underrepresented and this allows the models to 'cheat' and ignore these points in favour of the 'easier' points, even though the more difficult points are often more important for the modeler and, more critically, the end user.

Let us illustrate this with a very simple example.

> **Example** 4.1 (**Cheating models**).  Consider the following process:
>
> $$f : [0, 10] \to \mathbb{R} : x \mapsto x + \varepsilon(x), \tag{4.1}$$
>
> where
>
> $$\varepsilon \sim \begin{cases} \mathcal{U}(0, 1) & \text{if } x \geq 0.1, \\ \mathcal{U}(0, 10) & \text{otherwise.} \end{cases} \tag{4.2}$$
>
> If $X$ is uniformly sampled, a simple model will probably approximate the generating process $f(X) \mid X$ by a uniform distribution $\mathcal{U}(X, 1)$. This model will still attain a very high coverage.  It will be valid at the 1% significance level even when completely ignoring points with features in the interval $S := [0, 0.1]$.  However, if the points in $S$ are of particular importance, this would be very bad, since, conditional on $S$, the coverage will only be 10%.

This is exactly what tends to happen when applying 'naive' conformal prediction methods as can be seen in Fig. 4.1.  If the two samples in this figure would make up a data set, the prediction intervals would be valid at the significance level $\alpha = 0.2$, because 80% of the points is covered.  Since these intervals were generated with a standard conformal predictor based on the absolute residuals for the significance level $\alpha = 0.2$, this figure illustrates its marginal guarantees.  However, all of the data points in the blue subgroup are covered, while only 60% of the red subgroup is covered.  The conformal predictor might work marginally as promised, but this is definitely not sufficient when working with data sets in which more structure is present.

With the rise of conformal prediction, the interest in distribution-free conditional uncertainty quantification has also increased.  Although Venn and Mondrian conformal predictors (to be introduced in the next section) are actually almost as old as the field itself (Vovk et al., 2022; Vovk, Lindsay, Nouretdinov, & Gammerman, 2003), adoption by machine learning practitioners has remained even more limited than in the case of their nonconditional counterparts.  Just like these counterparts, the conditional variants provide strict statistical validity guarantees.

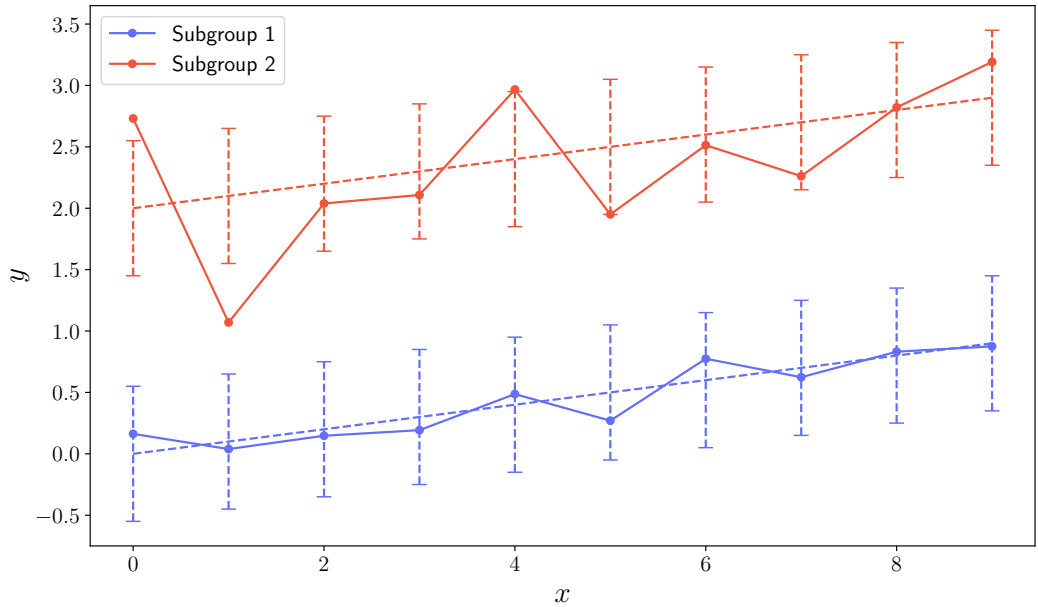

Figure 4.1: Two data samples with the same trend but with different noise levels: $y \mid x, s \sim 0.1x + 2s + \varepsilon(s)$, where $s \in \{0, 1\}$ is a dummy variable labelling the subgroups. The blue subgroup ($s = 0$) has standard deviation 0.1, while the red subgroup ($s = 1$) has standard deviation 0.5. Although the prediction intervals are valid at the $\alpha = 0.2$ significance level, both marginally and for the blue subgroup, this is not the case for the red subgroup.

The focus in this chapter lies on modelling heteroskedastic noise with guarantees conditional on the level of heteroskedasticity, i.e. where the data set is divided based on an estimate of the residual variance as in Fig. 4.1, and the validity of different models with respect to such a division. In this respect, it can be seen as a continuation of Boström and Johansson (2020). Aside from comparing the conditional validity of various standard nonconformity measures, with and without Mondrian taxonomies, theoretical conditions are derived for attaining conditional validity.

This chapter is structured as follows. The general notion of conditional validity is treated in Section 4.2. Section 4.3 introduces a straightforward approach to achieve conditional validity and Section 4.4 shortly recalls normalized conformal prediction, since the focus in this chapter lies on the study of heteroskedastic data. The main contribution of this chapter is presented in Section 4.5, where the conditions for achieving conditional validity with a general conformal predictor are analysed and related to the notion of pivotal quantities. Before moving to an experimental analysis, the results are first evaluated on synthetic data in Section 4.6.  This is also the perfect time
to study possible deviations from conditional validity and present some di-
agnostic tools.  Although Chapter 3 already introduced four classes of uncer-
tainty quantification methods, a fifth class was lacking.  *Generative models* or,
more specifically, normalizing flows are introduced in Section 4.7.  Last but
not least, the insights gained in this chapter are tested against some real data
sets in a short experimental analysis in Section 4.8.

## 4.2   Conditional validity

For the reasons outlined in the introduction, Eqs. (2.6) and (2.10) should be
replaced by their conditional counterparts (Barber, Candès, Ramdas, & Tib-
shirani, 2021a; J. Lei, G'Sell, Rinaldo, Tibshirani, & Wasserman, 2018; Vovk,
2012).

**Definition 4.2 (Conditional coverage).**  Let $\Gamma : \mathcal{X} \to 2^{\mathcal{Y}}$ be a confidence
predictor.  The conditional coverage with respect to a distribution $P \in$
$\mathbb{P}(\mathcal{X} \times \mathcal{Y})$ is defined as follows:

$$\mathcal{C}(\Gamma, P \mid A) := \mathsf{E}_P\big[\mathbb{1}_{\Gamma(X)}(Y)\,\big|\,A\big] = P\big(Y \in \Gamma(X) \mid A\big), \qquad (4.3)$$

where $A \subseteq \mathcal{X} \times \mathcal{Y}$ is any measurable subset satisfying $P(A) > 0$.

The confidence predictor is said to be **conditionally valid** at significance
level $\alpha \in [0, 1]$ if

$$\mathcal{C}(\Gamma, P \mid A) \geq 1 - \alpha \qquad (4.4)$$

for all measurable subsets $A \subseteq \mathcal{X} \times \mathcal{Y}$ satisfying $P(A) > 0$.  As in the
previous chapter, a confidence predictor at significance level $\alpha \in [0, 1]$
will be denoted by $\Gamma^\alpha$.

Although the condition $P(A) > 0$ in the above definition seems rather in-
nocuous, its consequences are far-reaching.  Naively, we would want our
estimator to be **object-conditionally valid** (Vovk, 2012):

$$P\big(Y \in \Gamma(X) \mid X = x\big) \geq 1 - \alpha \qquad (4.5)$$

for $P_X$-almost all $x \in \mathcal{X}$, where $P_X$ is the marginal distribution on $\mathcal{X}$ induced
by the joint distribution $P$.  However, unless $P_X$ has atoms, this expression is not even well-defined. Moreover, when this condition does make sense, the following no-go result holds.

**Theorem** 4.3. If the confidence predictor $\Gamma : \mathcal{X} \times (\mathcal{X} \times \mathcal{Y})^* \to 2^{\mathcal{Y}}$, where $X$ is a *separable* metric space, is object-conditionally valid at the $1 - \alpha$ level, then for $P_X$-almost all $x \in \mathcal{X}$ the following results hold:

- Regression ($\mathcal{Y} = \mathbb{R}$):

$$\mathrm{Prob}\big(\Lambda(\Gamma(x \mid V)) = +\infty\big) \geq 1 - \alpha \,, \qquad (4.6)$$

 where $\Lambda$ is the Lebesgue measure (A.20) on $\mathbb{R}$. In fact, the convex hull of $\Gamma(x)$ will be $\mathbb{R}$ with probability greater than $1 - \alpha$.

- Classification ($\mathcal{Y}$ finite):

$$\mathrm{Prob}\big(y \in \Gamma(x \mid V)\big) \geq 1 - \alpha \qquad (4.7)$$

 for all $y \in \mathcal{Y}$.

*Proof*. See the proofs in J. Lei et al. (2018); Vovk (2012). □

## 4.3 Mondrian conformal prediction

While the previous chapter introduced plenty of methods to construct prediction regions and prediction intervals in particular, none of those had any conditional guarantees in the sense of Definition 4.2. However, any ordinary conformal prediction algorithm can be extended to the conditional setting without major modifications in the following two situations:

1. The subregions of interest are given as a finite collection of disjoint measurable sets $\{S_i\}_{i \in [k]}$ covering $\mathcal{X} \times \mathcal{Y}$. These are equivalently characterized by a **taxonomy function** $\kappa : \mathcal{X} \times \mathcal{Y} \to [k]$.[1]

2. The true heteroskedastic nature of the data-generating process can be approximated by such a taxonomy.

For convenience, the codomain $[k]$ of the taxonomy function and the distribution $P_C$ of the taxonomy class $C := \kappa(X, Y)$, given by the pushforward

---

[1] For the full transductive approach, these taxonomies should also take a multiset of training points as argument: $\kappa : (\mathcal{X} \times \mathcal{Y})^* \times \mathcal{X} \times \mathcal{Y} \to [k]$.

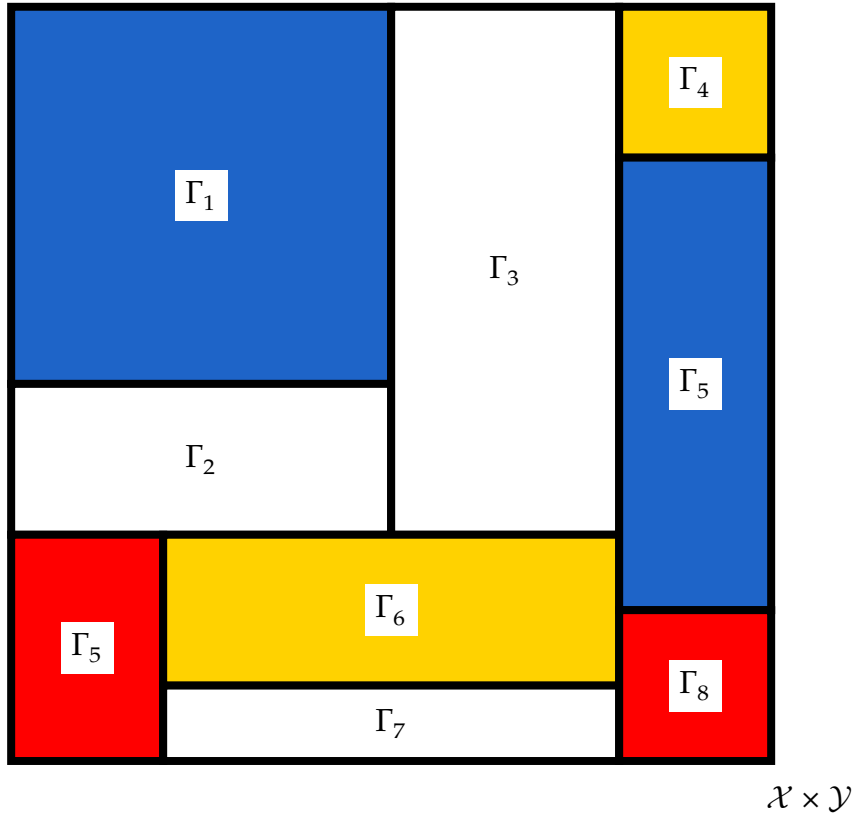

Figure 4.2: Artistic impression of a Mondrian conformal predictor

rule (Definition A.19)

$$P_C(c) := \kappa_* P_{X,Y}(c) = P_{X,Y}\big(\kappa^{-1}(c)\big),\qquad(4.8)$$

are assumed to be finite[2], with only the empty set having probability zero. This latter restriction is to ensure that conditioning on a taxonomy class does not lead to measure-zero issues. Note that the taxonomy function $\kappa$ can, in general, be any measurable function.

In this situation, we can use the Mondrian extension (Vovk et al., 2022, 2003) as shown in Algorithm 11. Comparing it to the ICP Algorithm 2, it is not hard to see that **Mondrian conformal prediction**[3] (MCP) methods work in two steps. First, the data set is, just like the instance space, divided into multiple classes using the taxonomy function $\kappa : \mathcal{X} \times \mathcal{Y} \to [k]$. This partitioning also

---

[2]  In theory, $[k]$ could be generalized to any countable or discrete measurable space. The restriction to finite sets is such that the algorithm is numerically feasible.

[3]  Conformal predictors that are not constructed in a Mondrian fashion will be called **non-Mondrian** in the remainder of this dissertation.

induces a decomposition of the calibration set:

$$\mathcal{V} = \bigcup_{c \in [k]} \mathcal{V}_c \tag{4.9}$$

with

$$\mathcal{V}_c := \mathcal{V} \cap \kappa^{-1}(c) = \left\{ (x, y) \in \mathcal{V} \mid \kappa(x, y) = c \right\}. \tag{4.10}$$

Then, the algorithm proceeds by constructing a conformal predictor for every class $c \in [k]$, where the calibration set is given by $\mathcal{V}_c$. This is schematically shown in Fig. 4.2. This figure should also explain the name of the framework: 'Mondrian conformal prediction'. If we would make the assumption that the taxonomy class only depends on the projection of an instance onto the feature space, i.e. if the taxonomy function is of the form $\kappa : \mathcal{X} \to [k]$, then for every new instance $x \in \mathcal{X}$ with $\kappa(x) = c$, the critical nonconformity score $a_c^*$ could be used to construct a prediction region. A Mondrian taxonomy will be called **feature-dependent** if it only depends on the feature space $\mathcal{X}$. However, not all taxonomy functions are of such a simple nature. For general taxonomies, the same approach as for cross-conformal predictors (Algorithm 4) has to be followed, i.e. a separate comparison has to be made for every possible response $y \in \mathcal{Y}$.

**Remark** 4.4 (**Terminology**). The method introduced in this section was initially called 'Venn conformal prediction' (Vovk et al., 2022), while Mondrian conformal prediction had a different definition. Due to the specific structure (Fig. 4.2), however, the more recent literature adopted the Mondrian terminology.

**Remark** 4.5 (**Distinct measures**). Although a separate ICP instance is constructed for every taxum in the definition of Mondrian conformal predictors, some could object that the same nonconformity measure is used every time. It is not hard to see that it is indeed a perfectly valid choice to use a distinct nonconformity measure (or even a different conformal prediction algorithm) for every taxonomy class. Note that this distinction is essentially artificial and that, without loss of generality, only a single nonconformity measure has to be mentioned, since for given a collection $\{A_c\}_{c \in [k]}$ of nonconformity measures, we can just define

$$A(x, y) := A_{\kappa(x,y)}(x, y). \tag{4.11}$$

---

**Algorithm 11:** Mondrian Conformal Prediction

---

**Input** : Significance level $\alpha \in [0, 1]$, nonconformity measure
$A : \mathcal{X} \times \mathcal{Y} \to \mathbb{R}$, training set $\mathcal{T} \in (\mathcal{X} \times \mathcal{Y})^*$, calibration set
$\mathcal{V} \in (\mathcal{X} \times \mathcal{Y})^*$ and taxonomy function $\kappa : \mathcal{X} \times \mathcal{Y} \to [k]$

**Output:** Mondrian conformal predictor $\Gamma^\alpha$

1 (Optional) Train the underlying model of $A$ on $\mathcal{T}$

2 **foreach** $c \in [k]$ **do**

3      Initialize an empty list $\mathcal{A}_c$

4      Construct the $c$-stratum $\mathcal{V}_c \leftarrow \{(x, y) \in \mathcal{V} \mid \kappa(x, y) = c\}$

5      **foreach** $(x_i, y_i) \in \mathcal{V}_c$ **do**

6          Calculate the nonconformity score $A_{c,i} \leftarrow A(x_i, y_i)$

7          Add the score $A_{c,i}$ to $\mathcal{A}_c$

8      **end**

9      Determine the critical score $a_c^* \leftarrow q_{(1-\alpha)(1+1/|\mathcal{A}_c|)}(\mathcal{A}_c)$

10 **end**

11 **procedure** $\Gamma^\alpha(x : \mathcal{X})$

12      **return** $\{y \in \mathcal{Y} \mid A(x, y) \leq a_{\kappa(x,y)}^*\}$

13 **return** $\Gamma^\alpha$

---

In the case of (Mondrian) taxonomies $\kappa : \mathcal{X} \times \mathcal{Y} \to [k]$, Definition 4.2 becomes:

$$\mathcal{C}(\Gamma^\alpha, P_{X,Y} \mid c) := \mathsf{Prob}\big(Y \in \Gamma^\alpha(X) \mid \kappa(X, Y) = c\big). \qquad (4.12)$$

Accordingly, **conditional validity with respect to** $\kappa$ means that

$$\mathcal{C}(\Gamma^\alpha, P_{X,Y} \mid c) \geq 1 - \alpha \qquad (4.13)$$

for all $c \in [k]$. The Mondrian approach benefits from the theoretical guarantees of Theorem 2.27 of the (I)CP framework in that validity will be guaranteed for every class in $[k]$ individually, as long as the data is exchangeable in every class (Vovk et al., 2022).

> **Corollary 4.6 (Conditional validity).** Let $\Gamma^\alpha : \mathcal{X} \times (\mathcal{X} \times \mathcal{Y})^* \to 2^{\mathcal{Y}}$ be a Mondrian conformal predictor at significance level $\alpha \in [0, 1]$ with nonconformity measure $A : \mathcal{X} \times \mathcal{Y} \to \mathbb{R}$ and taxonomy function $\kappa : \mathcal{X} \times \mathcal{Y} \to [k]$. If the nonconformity scores are exchangeable within each taxum of any calibration set and any new observation, then $\Gamma^\alpha$ is conditionally conservatively valid:
>
> $$\mathrm{Prob}\big(Y \in \Gamma^\alpha(X \mid V) \mid \kappa(X, Y) = c\big) \geq 1 - \alpha \qquad (4.14)$$
>
> for all $c \in [k]$, where the probability is taken over both $(X, Y)$ and $V$.

As before, when the assumption on the taxonomy function that it only depends on the feature space is relaxed, the algorithm is less numerically efficient. In this situation, for every $x \in \mathcal{X}$, there can be multiple taxonomy classes possible, depending on the response $y \in \mathcal{Y}$. Accordingly, instead of determining the taxonomy class of $x$ and using the associated critical nonconformity score to construct a prediction set, the prediction sets for the different possible taxonomy classes have to be aggregated. In the case of univariate regression, $\mathcal{Y} = \mathbb{R}$, this would correspond to taking the union of multiple prediction intervals (or the convex hull of the union if a new interval is required).

To avoid this issue, we will only work with feature-dependent taxonomy function. In fact, we will actually consider an even more specific type of taxonomy function. The taxonomy functions of interest divide the instance space based on an estimate of the uncertainty (Boström & Johansson, 2020). These taxonomy functions, which will be said to be **uncertainty-dependent**, will be derived from a proxy of the uncertainty or, more concretely, the heteroskedastic noise such as the (conditional) standard deviation. More formally, given such an uncertainty estimate $\rho : \mathcal{X} \to \mathbb{R}^+$ and a clustering function $\mathcal{B} : \mathbb{R}^+ \to [k]$, the induced taxonomy function is given by

$$\kappa = \mathcal{B} \circ \rho : \mathcal{X} \to [k]. \qquad (4.15)$$

A straightforward choice would be the case where $\rho = \widehat{\sigma}$ is an estimate of the (conditional) standard deviation and $\mathcal{B}$ corresponds to equal-frequency binning for some predetermined number of classes (see Fig. 4.8a further below for the case of three classes).

For clarity's sake, a simple example is in order.

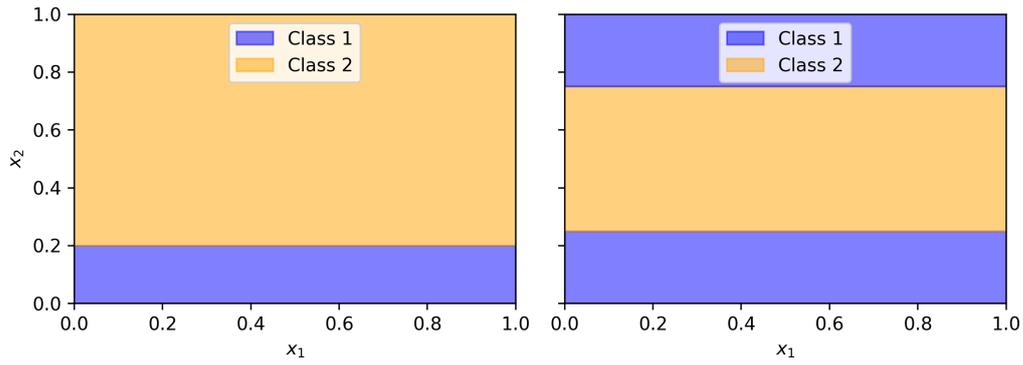

Figure 4.3: Division of the feature space $\mathcal{X} = [0,1]^2$ based on two differ­ent taxonomy functions. The first one simply thresholds the second dimen­sion (at $\xi = 0.2$), whereas the second one performs equal-frequency bin­ning on the (conditional) standard deviation, which has the form shown in Eq. (4.16).

**Example** 4.7. Consider the two-dimensional feature space $\mathcal{X} = [0,1]^2$, equipped with the uniform distribution $\mathcal{U}^2$. As data-generating process, take[a]

$$y(x) \sim \mathcal{N}\big(x^1 + x^2, 1 + |x^2 - 0.5|\big). \tag{4.16}$$

One possible choice of taxonomy function is one that divides the instance space based on some parameters such as

$$\kappa_{\xi}(x, y) := \mathbb{1}_{[0,\xi]}(x^2), \tag{4.17}$$

where $\xi \in [0,1]$ and $\mathbb{1}$ denotes the indicator function defined in Eq. (1.6). The resulting division of the feature space is shown in the left panel of Fig. 4.3. Another possibility, giving an example of the general class of tax­onomies defined by Eq. (4.15), would be to divide the feature space by binning the (conditional) standard deviation. For the case of two classes, this is shown in the right panel of Fig. 4.3. Note that the division of the instance space by the former, feature-dependent taxonomy is entirely in­dependent of the conditional distribution $P_{Y|X}$, whereas the division by the latter, uncertainty-dependent taxonomy is strongly influenced by the form of $P_{Y|X}$.

---

[a] The reason for this choice will be explained in Section 4.6, where this example will be studied in more detail.

## 4.4 Normalized conformal prediction

The standard residual measure (3.38) does not take into account any information about subregions of $\mathcal{X}$, such as where the model might perform subpar. This has as a consequence that the resulting prediction intervals are all of the same size, given by twice the critical nonconformity score and, as such, this method implicitly assumes domain knowledge about the homoskedasticity of the problem. To resolve this issue, knowledge about the data noise can be explicitly incorporated to obtain more realistic and more efficient intervals (meaning that the intervals will be smaller when possible cf. Section 2.2.3). In Section 3.3, Eq. (3.41) to be precise, we therefore introduced a class of nonconformity measures that allowed to scale the scores and, accordingly, the prediction intervals:

$$A_{\ell 1}^{\rho}(x, y) := \frac{|\hat{y}(x) - y|}{\rho(x)} \, , \tag{4.18}$$

where $\rho : \mathcal{X} \to \mathbb{R}^+$ was called the difficulty function. Note that the uncertainty measure inducing uncertainty-dependent taxonomies as introduced in the previous section was suggestively given the same notation.

For general taxonomies, this normalized approach does not have conditional guarantees, but when the taxonomy can be expressed in terms of the difficulty function as considered before, it begs the question of whether this algorithm might have good conditional behaviour even without considering a Mondrian approach. Moreover, another advantage of normalized conformal prediction is that the calibration set does not have to be split in as many folds as there are taxonomy labels. When the data set is relatively small, the number of labels is relatively high, or the taxonomy is imbalanced, such a data split might lead to inefficient or even invalid confidence predictors.

**Example** 4.8 (**Mean-variance estimation**). Consider the mean-variance estimators introduced in Section 3.2.2. Instead of simply estimating a point predictor, this approach assumes a parametric model[a] for the data-generating process, usually a normal distribution, characterized by a conditional mean $\hat{\mu} : \mathcal{X} \to \mathbb{R}$ and conditional standard deviation $\hat{\sigma} : \mathcal{X} \to \mathbb{R}^+$. The canonical choice of nonconformity measure for these models

is Eq. (3.42):

$$A_{\text{st}}(x, y) = \frac{|\hat{\mu}(x) - y|}{\hat{\sigma}(x)}.  \tag{4.19}$$

For stability issues, $\hat{\sigma}$ can be replaced by $\hat{\sigma} + \varepsilon$ for some (small) $\varepsilon \in \mathbb{R}^+$ as in Johansson et al. (2021). Note that this transformation induces a strong bias when $\varepsilon$ is not carefully tuned, especially when the estimate of the standard deviation is small compared to $\varepsilon$, since $\varepsilon \gg \hat{\sigma}$ forces the resulting nonconformity measure to essentially be that of a homoskedastic model with standard deviation $\varepsilon$.

As was also explained in Section 3.2.2, more specifically in Eq. (3.14), given a mean-variance estimator $(\hat{\mu}, \hat{\sigma}) : \mathcal{X} \to \mathbb{R} \times \mathbb{R}^+$, a prediction interval at significance level $\alpha \in [0, 1]$ can be obtained for normal distributions as follows:

$$\Gamma^\alpha_{\text{Gauss}}(x) := \left[ \hat{\mu}(x) - z^\alpha \hat{\sigma}(x), \hat{\mu}(x) + z^\alpha \hat{\sigma}(x) \right],  \tag{4.20}$$

where $z^\alpha$ is the $(1 - \alpha/2)$-quantile of the standard normal distribution.

---

[a]  Note that for conformal prediction to work, the data does not have to be generated by some parametric model. The estimates $\hat{\mu}$ and $\hat{\sigma}$ can be obtained by any means. However, assigning a (valid) probabilistic interpretation to these functions rests on the choice of parametric model. Moreover, the importance of parametric assumptions will become apparent throughout the remainder of this chapter.

## 4.5  Theoretical analysis

In this section, the conditional validity of non-Mondrian conformal predictors, i.e. conformal predictors that are not constructed using the Mondrian approach with a separate instance for every taxonomy class, is studied from a theoretical perspective. First, the general case is considered, where no parametric assumptions are made about the nonconformity measures. In a second step, by making some stricter assertions, an explicit expression in the form of Eq. (4.28) will be derived for the data-generating process as a sufficient condition for conditional validity to hold with respect to any uncertainty-dependent taxonomy function.

### 4.5.1 Pivotal quantities

In the situation at hand, the hope for non-Mondrian conformal prediction is that the heteroskedasticity in a data set can be treated by choosing a suitable nonconformity measure, such as the normalized one (4.18), without having to resort to conditional methods such as MCP, since these are more data intensive (and data sparsity is often higher in regions with high heteroskedastic noise). So, we should ask ourselves when general conformal prediction methods will satisfy the same guarantees as MCP with its conditional validity result (Corollary 4.6). Every ICP instance in an MCP ensemble predicts the elements of $\mathcal{Y}$ for which the nonconformity score is smaller than a certain (empirical) quantile of the nonconformity distribution conditioned on the selected taxonomy class. Now, if these quantiles or, more generally, the conditional nonconformity distributions, coincide for all taxa, they will also be equal to the ones calculated on the whole calibration set, since a mixture of identical distributions is again equal to the given distributions.

The following theorem formalizes this idea and gives a sufficient condition for conditionally valid conformal predictors (Definition 4.2). As before, the joint distribution over calibration sets and test instances will be denoted by $P$ and, as always, it will be assumed to be exchangeable.

**Theorem 4.9 (Independence).** Consider a taxonomy function $\kappa : \mathcal{X} \times \mathcal{Y} \to [k]$ and nonconformity measure $A : \mathcal{X} \times \mathcal{Y} \to \mathbb{R}$. Denote the $\sigma$-algebra on $\mathcal{Y}$ by $\Sigma$ and denote the random variable representing the taxonomy class of the test instance by $C := \kappa(X, Y)$. If both the nonconformity distribution and the distribution of the calibration sets are independent of the taxonomy class of the test instance, i.e.

$$P_{A|C}(B \mid c) = P_A(B) \qquad \text{and} \qquad P(\mathcal{V} \mid c) = P(\mathcal{V}) \qquad (4.21)$$

for all $c \in [k]$, $B \in \Sigma$ and $\mathcal{V} \in (\mathcal{X} \times \mathcal{Y})^n$, the conformal predictor associated with $A$ is conditionally valid with respect to $\kappa$.

*Proof.* For any taxonomy class $c \in [k]$, the following relation holds:

$$\mathsf{Prob}\big(Y \in \Gamma^\alpha(X, V) \mid C = c\big)$$
$$= \int_{(\mathcal{X} \times \mathcal{Y})^n} \mathsf{Prob}\big(Y \in \Gamma^\alpha(X, V) \mid C = c, V = \mathcal{V}\big) \, dP(\mathcal{V} \mid C = c).$$

By the definition of ICPs, the first factor can be expressed in terms of the distribution of nonconformity scores:

$$\mathsf{Prob}\big(Y \in \Gamma^\alpha(X, V) \mid C = c\big)$$
$$= \int_{(\mathcal{X} \times \mathcal{Y})^n} F_{A|C}\big(a_\mathcal{V}^* \mid C = c, V = \mathcal{V}\big) \, dP(\mathcal{V} \mid C = c),$$

where $a_\mathcal{V}^* = q_{(1-\alpha)(1+\frac{1}{n})}\big(A(\mathcal{V})\big)$ as usual. By assumption, neither the distribution of nonconformity scores nor the distribution of calibration sets depends on the taxonomy class $C$ and, hence,

$$\mathsf{Prob}\big(Y \in \Gamma^\alpha(X, V) \mid C = c\big) = \int_{(\mathcal{X} \times \mathcal{Y})^n} F_A\big(a_\mathcal{V}^* \mid V = \mathcal{V}\big) \, dP(\mathcal{V})$$
$$= \mathsf{Prob}\big(Y \in \Gamma^\alpha(X, V)\big)$$
$$\geq 1 - \alpha,$$

which proves the statement. $\qquad\square$

Note that this theorem implies that mere independence of the nonconformity score and taxonomy class of any data point is not sufficient in general. The property of exchangeability is too weak for the conclusion to hold. In the case of i.i.d. data, the second condition is, however, trivially satisfied. The conditions can also be slightly weakened and unified by noting that the only thing that matters about the calibration data are the nonconformity scores. Hence, the condition can be jointly relaxed to

$$\mathsf{Prob}\big(A\big(V \cup \{(X, Y)\}\big) \mid C = c\big) = \mathsf{Prob}\big(A\big(V \cup \{(X, Y)\}\big)\big) \quad (4.22)$$

for all $c \in [k]$.

More generally, assume that there exists a permutation $(Z^{1:k}, \widetilde{Z}^{1:l})$ of the random variables $X$ and $Y$, such that $A$ only depends on the variables $Z^{1:k}$ and $\kappa$ only depends on the variables $\widetilde{Z}^{1:l}$. The theorem will hold as soon as the joint distribution $P$ can be represented as a graphical model shown in Fig. 4.4, where both the upper and lower layer are exchangeable sequences. To see why this holds, consider the probability $P\big(A(V) \mid C = c\big)$. By assumption, none of the variables $(Z_i^{1:k}, \widetilde{Z}_i^{1:l})$ depend on $\widetilde{Z}^{1:l}$. Consequently, since $\kappa(X, Y)$ is only a function of the random variables $\widetilde{Z}^{1:l}$, this also implies that

$$P\big(A(V) \mid C = c\big) = P\big(A(V)\big) \tag{4.23}$$

for all $c \in [k]$.

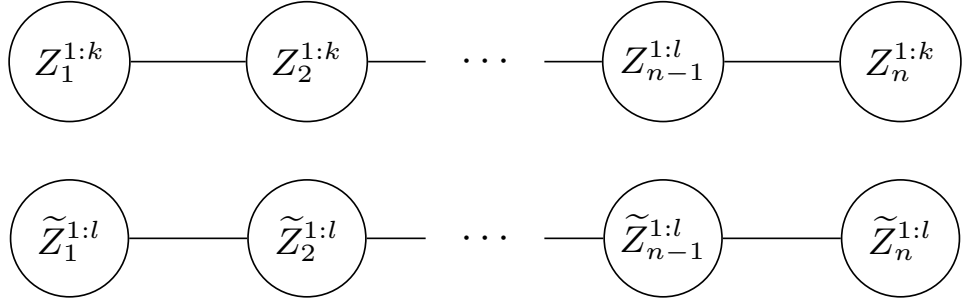

Figure 4.4: (Sufficient) dependency structure for the joint distribution over calibration sets (of size $n \in \mathbb{N}$) and test instances in Theorem 4.9.

**Remark** 4.10. Assuming that the second condition holds, the first condition is also a very reasonable condition to require. By exchangeability, the dependence between any two data points is exactly the same. So, if only the second condition holds, the data-generating process would be such that, for every index $i \leq n$, the nonconformity scores of all data points $(X_j, Y_j)$ depend on the taxonomy class of $(X_i, Y_i)$ except for $i = j$.

In practice, it is often more interesting to not only attain conditional validity with respect to a fixed taxonomy function, but with respect to an entire family of taxonomy functions. A straightforward example would be where an adaptive or hierarchical conditioning strategy is used. To this end, a concept from the classical statistical literature becomes relevant (Arnold, 1984; Toulis, 2017).

**Definition** 4.11 (**Pivotal quantity**). Consider a family of probability distributions

$$\{P_\theta \mid \theta \in \Theta\} \subseteq \mathbb{P}(\mathcal{X}) \tag{4.24}$$

parametrized by a set $\Theta$ (the parametrization does not have to be smooth or continuous). A function $g : \mathcal{X} \to \mathcal{Y}$ of observations is called a **pivotal** **quantity** (or simply **pivot**) with respect to this family if its distribution does not depend on the particular choice of parameter:

$$\forall \theta, \theta' \in \Theta : \left( X \sim P_\theta \wedge X' \sim P_{\theta'} \right) \implies \left( g(X) \overset{d}{=} g(X') \right), \qquad (4.25)$$

where $\overset{d}{=}$ denotes **equality in distribution**, i.e. $g(X)$ and $g(X')$ have the same distribution. The distribution of such a pivotal quantity will be called the **pivotal distribution**.

Combined with the Independence Theorem 4.9, this gives rise to the following result, which states how assuming the conditional validity of a conformal predictor is tied to making assumptions about the data-generating distribution.

**Corollary** 4.12. The conformal predictor associated with a nonconformity measure $A : \mathcal{X} \times \mathcal{Y} \to \mathbb{R}$ is conditionally valid for any feature-dependent taxonomy if $A$ is a pivotal quantity for the family of conditional distributions $\left\{ P_{Y|X}(\cdot \mid x) \mid x \in \mathcal{X} \right\}$ and if the data is such that the calibration set does not depend on the taxonomy class of any test point, i.e.

$$P\left( V \mid \kappa(X, Y) = c \right) = P(V) \qquad (4.26)$$

for all $c \in [k]$.

## 4.5.2   Normalization

Although the results in the preceding section do not make any assumptions about the form of the conditional distribution $P_{Y|X}$, beyond the nonconformity measure $A : \mathcal{X} \times \mathcal{Y} \to \mathbb{R}$ being pivotal, it is interesting to look at some more specific examples. In this section, we will see that a large class of common distributions gives rise to often-used nonconformity measures.

Recall the mean-variance estimators from Section 3.2.2 and Example 4.8 with the induced (normalized) nonconformity measure (4.19). If the absolute value was not present in that equation, the nonconformity scores would actually be the (estimated) $z$-scores of the data. Now, assuming the idealized situation where the conditional distribution $P_{Y|X}$ is a normal distribution and we have access to an **oracle**, i.e. both the (conditional) mean and variance can be obtained with arbitrary precision, such a transformation would lead to the random variable $A \equiv A(X, Y)$, the nonconformity score, following a normal distribution with

$$\mathsf{E}[A] = 0 \qquad \text{and} \qquad \mathsf{Var}[A] = 1 \tag{4.27}$$

independent of whether we consider the marginal distribution or condition on a specific taxonomy class. Using the terminology of Definition 4.11, this can be rephrased by saying that the nonconformity measure $A : \mathcal{X} \times \mathcal{Y} \to \mathbb{R}$ is a pivotal quantity for the random variable $(X, Y)$ representing a new instance. The next theorem shows that such a situation does not only present itself for normal distributions, but actually holds more generally. In the remainder of the text, $f_A$ and $f_{A|C}$ denote the probability density functions of the marginal and conditional nonconformity distributions $P_A$ and $P_{A|C}$, respectively.

**Theorem 4.13 (Standardization).** If the (conditional) data-generating distribution $P_{Y|X}$ admits a density function of the form

$$f_{Y|X}(y \mid x) = \frac{1}{\sigma(x)} g\left(\frac{y - \mu(x)}{\sigma(x)}\right) \tag{4.28}$$

for some smooth function $g : \mathbb{R} \to \mathbb{R}^+$, the probability distribution of the standardized nonconformity measure[a]

$$A_{\text{st}}(x, y) := \frac{y - \mu(x)}{\sigma(x)} \tag{4.29}$$

is independent of the classes of any feature-dependent Mondrian taxonomy $\kappa : \mathcal{X} \times \mathbb{R} \to [k]$.

*Proof.* The joint density of nonconformity scores and taxonomy classes can be rewritten as

$$\begin{aligned}
f_{A_{\text{st}}, C}(a, c) &= \frac{\partial}{\partial a} F_{A_{\text{st}}, C}(a, c) \\
&= \frac{\partial}{\partial a} P_{X, Y}\left(\left\{(x, y) \mid A_{\text{st}}(x, y) \in \,]-\infty, a] \land \kappa(x) = c\right\}\right) \\
&= \frac{\partial}{\partial a} \int_{\kappa^{-1}(c)} \int_{-\infty}^{\mu(x) + a\sigma(x)} f_{X, Y}(x, y) \, \mathrm{d}y \, \mathrm{d}x \\
&= \int_{\kappa^{-1}(c)} f_{X, Y}\left(x, \mu(x) + a\sigma(x)\right) \sigma(x) \, \mathrm{d}x \,,
\end{aligned}$$

where, in the second line, we wrote $\kappa(x)$ instead of $\kappa(x, y)$ to emphasize that $\kappa$ does not depend on the response $Y$ and where in the last step,

Leibniz's integral rule was applied:

$$\frac{\mathrm{d}}{\mathrm{d}x}\left(\int_{a(x)}^{b(x)} f(x,y)\,\mathrm{d}y\right) = \int_{a(x)}^{b(x)} \frac{\partial f}{\partial x}(x,y)\,\mathrm{d}y$$
$$+ f\big(b(x),y\big)b'(x) - f\big(a(x),y\big)a'(x)\,. \qquad (4.30)$$

Finally, the joint density function $f_{X,Y}$ is factorized as

$$f_{X,Y}(x,y) = f_{Y|X}(y \mid x)f_X(x)$$

to obtain:

$$f_{A_{\mathrm{st}},C}(a,c) = \int_{\kappa^{-1}(c)} \underbrace{\sigma(x)f_{Y|X}\big(\mu(x)+a\sigma(x) \mid x\big)}_{=f_{A_{\mathrm{st}}|X}(a|x)} f_X(x)\,\mathrm{d}x\,.$$

If the first factor is (functionally) independent of $x$ and, hence, of $\mu$ and $\sigma$, it can be moved out of the integral:

$$f_{A_{\mathrm{st}},C}(a,c) = f_{A_{\mathrm{st}}}(a)\int_{\kappa^{-1}(c)} f_X(x)\,\mathrm{d}x = f_{A_{\mathrm{st}}}(a)P_C(c)\,.$$

To see when this holds, define a three-parameter function $\tilde{f}$ as a generalization of $f_{Y|X}$ as follows:

$$\tilde{f}\big(a,\mu(x),\sigma(x)\big) := f_{Y|X}\big(\mu(x)+a\sigma(x) \mid x\big)\,.$$

If after standardization the density should only depend on $y$, the following system of partial differential equations should hold:

$$\begin{cases} \partial_\mu\big(\sigma\tilde{f}(a,\mu,\sigma)\big) = 0\,, \\ \partial_\sigma\big(\sigma\tilde{f}(a,\mu,\sigma)\big) = 0\,. \end{cases}$$

The first partial differential equation immediately yields that $\tilde{f}$ is (functionally) independent of $\mu$. Analogously, the second equation says that $\sigma\tilde{f}$ is independent of $\sigma$ or, equivalently, that

$$\tilde{f}(a,\mu,\sigma) = \frac{g(a)}{\sigma}$$

for an arbitrary smooth function $g : \mathbb{R} \to \mathbb{R}^+$ (requiring that $\tilde{f}$ gives a density function imposes further conditions on $g$ such as integrability). Transforming back to the original function gives

$$f_{Y|X}(y \mid x) = \frac{1}{\sigma(x)} g\left( \frac{y - \mu(x)}{\sigma(x)} \right),$$

which concludes the proof.					□

<sup>a</sup> Note that what is written in Eq. (4.19) is technically not the standardization of $y \mid x$, but its absolute value. From here on, we will make this distinction more explicitly, where Eq. (4.19) will be called the normalized measure (even though normalization often means *min-max scaling* in the field of data science).

In view of Corollary 4.12, the standardized variable $A_{\mathrm{st}}(X, Y)$ is a pivotal quantity whenever the conditional data-generating distributions form a $\mathcal{X}$-parametrized family of the form of Eq. (4.28). Moreover, the function $g : \mathbb{R} \to \mathbb{R}^+$ is exactly its pivotal distribution.

Although the standardized measure (4.29) is not exactly the same as the normalized measure (4.19), it is also of interest on its own. It is for example used in the construction of conformal predictive systems (Section 2.5.2), where, instead of calibrating at a single quantile, the whole predictive distribution $P_{Y|X}$ is modelled. The results in this chapter carry over to that setting accordingly. Although requiring invariance under standardization seems weaker than what is required to handle the normalized nonconformity measure, the following result shows that this difference is actually irrelevant.

**Property** 4.14 (**Composition**). Distributions with a density function of the form of Eq. (4.28) also lead to a pivotal distribution for the normalized measure (4.19). More generally, if a nonconformity measure $A$ is pivotal, any nonconformity measure obtained by (post)composing it with a feature-independent function $g : \mathbb{R} \to \mathbb{R}$ will also be pivotal.

*Proof.* The change-of-variables formula (Corollary A.31) says that:

$$f_{g(A)|X}(a \mid x) = \sum_{a' \in g^{-1}(a)} \frac{f(a' \mid x)}{|g'(a')|}. \tag{4.31}$$

By assumption, $f$ is probabilistically independent of $x \in \mathcal{X}$, i.e. $f(a' \mid x) = f(a')$ for all $x \in \mathcal{X}$ and $a' \in \mathbb{R}$, and $g$ is functionally independent of $x \in \mathcal{X}$. It follows that the left-hand side is also independent of $x \in \mathcal{X}$.□

**Remark 4.15.** Note that, instead of making a detour via the preceding theorem, the proof of Theorem 4.13 could have been generalized to work directly with the normalized residual measure (4.19). In the step before applying Leibniz's integral rule, the lower integration bound for $y$, here given by $-\infty$, would then become $\mu(x) - a\sigma(x)$ for the normalized nonconformity measure. The integral rule would then lead to an additional term $\sigma(x)f\big(x, \mu(x) - a\sigma(x)\big)$. However, Property 4.14 is more generally applicable for any nonconformity measure.

**Example 4.16.** Some common examples of distributions, where $\mu$ and $\sigma$ represent the conditional mean and standard deviation, satisfy the above theorem. Their pivotal distributions are:

- Normal distribution:

$$g(x) = \frac{1}{\sqrt{2\pi}}\exp\big(-x^2/2\big). \tag{4.32}$$

- Laplace distribution:

$$g(x) = \frac{1}{\sqrt{2}}\exp\big(-\sqrt{2}|x|\big). \tag{4.33}$$

- Uniform distribution:

$$g(x) = \frac{1}{2\sqrt{3}}\mathbb{1}_{\left[-\sqrt{3},\sqrt{3}\right]}(x), \tag{4.34}$$

where $\mathbb{1}$ denotes the indicator function defined in Eq. (1.6).

To give an example of how general the allowed distributions in this theorem are, consider the following asymmetric one (see Fig. 4.5):

$$f_{Y|X}(y \mid x) = \frac{2y}{\lambda(x)^2}\mathbb{1}_{[0,\lambda(x)]}(y) \tag{4.35}$$

for some positive function $\lambda : \mathcal{X} \to \mathbb{R}^+$. Although it is seemingly not of the form of Eq. (4.28), it can, with some work, be rewritten as such:

$$\frac{2y}{\lambda(x)^2}\mathbb{1}_{[0,\lambda(x)]}(y) = \frac{1}{\sigma(x)}\left(\frac{1}{9}\left(\frac{y - \mu(x)}{\sigma(x)}\right) + \frac{2\sqrt{2}}{9}\right)\mathbb{1}_{\left[-2\sqrt{2},\sqrt{2}\right]}\left(\frac{y - \mu(x)}{\sigma(x)}\right), \tag{4.36}$$

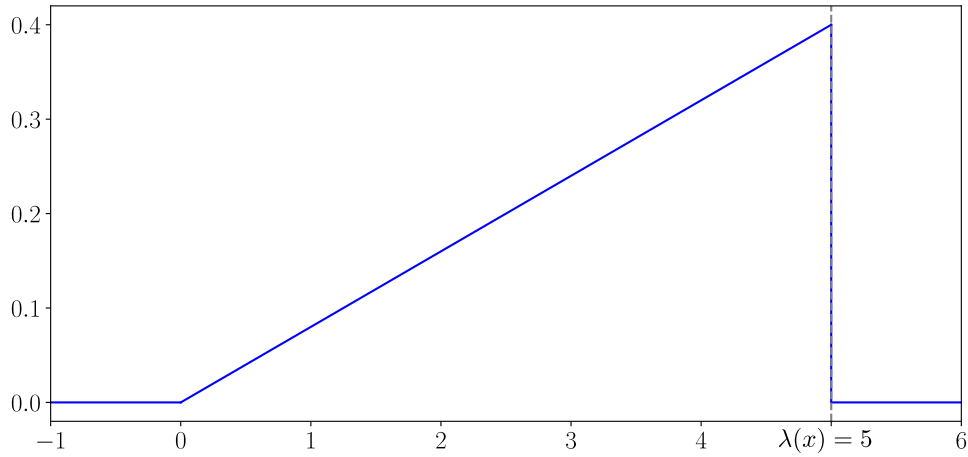

Figure 4.5: Probability density function of the triangular distribution in Eq. (4.35) with (conditional) width parameter $\lambda(x) = 5$.

where

$$\mu(x) := \frac{2\lambda(x)}{3} \qquad \text{and} \qquad \sigma(x) := \frac{\lambda(x)}{3\sqrt{2}}. \qquad (4.37)$$

To obtain a pivotal distribution for the standardized (or normalized) nonconformity measure, the conditional distribution $P_{Y|X}$ should be obtained as a member of the location-scale family with parameters $\mu(x)$ and $\sigma(x)$ induced by the distribution $g$. As a consequence, requiring invariance under standardization also allows for conditional distributions such as exponential distributions. Although exponential distributions in general only form a scale family and not a location family, because the mean and standard deviation coincide for exponential distributions ($\mu = \sigma$), they do form a location-scale family generated by $\mu : \mathcal{X} \to \mathbb{R}^+$. Accordingly, they give rise to the following pivotal distribution for standardized variables:

$$g(x) = \exp(-x - 1)\theta(x + 1), \qquad (4.38)$$

where $\theta : \mathbb{R} \to \{0, 1\}$ denotes the Heaviside step function:

$$\theta(x) := \mathbb{1}_{[0, +\infty[}(x). \qquad (4.39)$$

Note that the functions $\mu : \mathcal{X} \to \mathbb{R}$ and $\sigma : \mathcal{X} \to \mathbb{R}^+$ in the preceding theorems in general do not have to be the conditional mean and standard deviation. Moreover, it is not even strictly necessary that the parameters $\mu$ and $\sigma$ are estimated perfectly. For example, from the form of Eq. (4.28) it can immediately be seen that the estimates $\hat{\mu}$ and $\hat{\sigma}$ only need to satisfy

$$\hat{\sigma} = \lambda \sigma \,, \tag{4.40}$$

$$\hat{\mu} - \mu = \lambda' \hat{\sigma} \,, \tag{4.41}$$

for some $\lambda \in \mathbb{R}^+$ and $\lambda' \in \mathbb{R}$. (These relaxations also follow from Property 4.14.)

**Example** 4.17 (**Additive noise**). Theorem 4.13 also shows that any data set having a conditional data-generating process of the form (James et al., 2013; Johansson et al., 2014; J. Lei et al., 2018)

$$y = \mu(x) + \sigma(x)\varepsilon \,, \tag{4.42}$$

where $\varepsilon$ is sampled from some fixed distribution, will lead to a conditionally valid NCP algorithm. In other words, NCP models will be conditionally valid for any data set obtained by adding noise to a fixed (i.e. deterministic) trend. It is important to remark that the distribution of $\varepsilon$ is entirely unconstrained and does not have to be of the functional form of Eq. (4.28).

To finish this section, it can be useful for future work to note that the proof of Theorem 4.13 can be generalized to other nonconformity measures and distributions with a differentiable density function.

**Theorem** 4.18. Assume that $P_{Y|X}$ admits a density function $f_{X|Y}$ that is smoothly parameterized by some parameters $\boldsymbol{\theta} \equiv \boldsymbol{\theta}(x) \in \mathbb{R}^n$. Furthermore, assume that the nonconformity measure $A : \mathcal{X} \times \mathbb{R} \to \mathbb{R}$ only depends on $\mathcal{X}$ through $\boldsymbol{\theta}$. If, for a fixed $x \in \mathcal{X}$, the transformation $y \mapsto A(x, y)$ is increasing, then $A$ is pivotal if the following condition is satisfied:

$$f_{Y|X}\big(g(a, \boldsymbol{\theta}) \mid x; \boldsymbol{\theta}\big)\nabla_{\boldsymbol{\theta}}\frac{\partial g}{\partial a}(a, \boldsymbol{\theta}) + \frac{\partial g}{\partial a}(a, \boldsymbol{\theta})\nabla_{\theta}f_{Y|X}\big(g(a, \boldsymbol{\theta}) \mid x; \boldsymbol{\theta}\big) = \mathbf{0}, \tag{4.43}$$

where $g(\cdot, \boldsymbol{\theta}) : \mathbb{R} \to \mathbb{R}$ is defined by the equation

$$A\big(g(a, \boldsymbol{\theta}), x\big) = a \,, \tag{4.44}$$

i.e. it is the (right) inverse of $A$ for fixed $x \in \mathcal{X}$.[a]

---
[a] Since $A$ is assumed to only depend on $x$ through $\boldsymbol{\theta}$, the dependency of $g$ on $\boldsymbol{\theta}$ suffices.

---

This differential equation could be solved, for example, in the case of the interval nonconformity measure in Eq. (3.46). However, although technically viable, it would be less sensible than in the case of the standardized or normalized nonconformity measures. The latter do not take into account the significance level at which the prediction intervals are going to be constructed. They simply gives a statistic of the predictive distribution. However, any reasonable interval predictor does assume a predetermined significance level in some way and, hence, it should only become a 'pivotal quantity' at that given significance level.

**Remark 4.19 (Box–Cox transformation).** The theorems in this chapter were strongly motivated by studying **variance-stabilizing transformations**, i.e. those transformations that turn a heteroskedastic distribution into a homoskedastic one. For distributions with

$$\mathsf{Var}[X] = \mu[X]^{1-\lambda},\tag{4.45}$$

a candidate is given by the Box–Cox transformation[a] (Box & Cox, 1964)

$$f(x) := \begin{cases} \dfrac{x^\lambda - 1}{\lambda} & \text{if } \lambda \neq 0, \\ \ln(x) & \text{if } \lambda = 0. \end{cases}\tag{4.46}$$

Depending on the situation, it might be helpful to first apply a transformation to stabilize the variance and then apply, for example, conformal prediction with the ordinary residual measure from Eq. (3.38).

---
[a] This follows from Method A.27.

---

## 4.5.3  Intervals for Gaussian models

Recall the interval measure from Eq. (3.46):

$$A_{\text{int}}(x, y) = \max\big(\hat{y}_-(x) - y, y - \hat{y}_+(x)\big)\tag{4.47}$$

In the case of Gaussian prediction intervals as defined in Eq. (3.14), this expression can be rewritten as follows:[4]

$$
\begin{aligned}
\max\big(\hat{y}_-(x) - y, y - \hat{y}_+(x)\big) &= \max\big(\hat{\mu}(x) - z\hat{\sigma}(x) - y, y - \hat{\mu}(x) - z\hat{\sigma}(x)\big) \\
&= \max\big(\hat{\mu}(x) - y, y - \hat{\mu}(x)\big) - z\hat{\sigma}(x) \\
&= |\hat{\mu}(x) - y| - z\hat{\sigma}(x)\,.
\end{aligned}
\tag{4.48}
$$

The resulting conformalized interval of a new data point is then given by

$$
\begin{aligned}
\hat{y}_\pm^{\text{int}}(x) &= \hat{\mu}(x) \pm \big(|\hat{\mu}(x^*) - y^*| - z\hat{\sigma}(x^*) + z\hat{\sigma}(x)\big) \\
&= \hat{\mu}(x) \pm \big[|\hat{\mu}(x^*) - y^*| + z\big(\hat{\sigma}(x) - \hat{\sigma}(x^*)\big)\big]\,,
\end{aligned}
\tag{4.49}
$$

where $(x^*, y^*)$ is the data point corresponding to the critical nonconformity score, i.e. $A(x^*, y^*) = a^*$.

To relate this expression to the NCP method, Eq. (3.44) can be rewritten as follows:

$$
\begin{aligned}
\hat{y}_\pm^{\text{NCP}}(x) &= \hat{\mu}(x) \pm |\hat{\mu}(x^*) - y^*| \frac{\hat{\sigma}(x)}{\hat{\sigma}(x^*)} \\
&= \hat{\mu}(x) \pm |\hat{\mu}(x^*) - y^*| \left(1 + \frac{\hat{\sigma}(x) - \hat{\sigma}(x^*)}{\hat{\sigma}(x^*)}\right) \\
&= \hat{\mu}(x) \pm \left[|\hat{\mu}(x^*) - y^*| + \frac{|\hat{\mu}(x^*) - y^*|}{\hat{\sigma}(x^*)}\big(\hat{\sigma}(x) - \hat{\sigma}(x^*)\big)\right] \\
&= \hat{\mu}(x) \pm \big[|\hat{\mu}(x^*) - y^*| + z^*\big(\hat{\sigma}(x) - \hat{\sigma}(x^*)\big)\big]\,,
\end{aligned}
\tag{4.50}
$$

where the (absolute) critical $z$-score is given by

$$
z^* := \frac{|\hat{\mu}(x^*) - y^*|}{\hat{\sigma}(x^*)}\,.
\tag{4.51}
$$

So, for Gaussian prediction intervals, the difference between interval conformal prediction and $\hat{\sigma}$-normalized conformal prediction is that the standard $z$-score at significance level $\alpha$, the $(1-\alpha)$-quantile of the (standard) *folded* normal distribution, is replaced by the $z$-score of the inflated $(1-\alpha)$-quantile of the nonconformity scores. When both the conditional mean and the conditional variance can be consistently estimated, the algorithms should asymptotically converge to the same result.

---

[4]  A similar reasoning holds for all central prediction intervals, i.e. intervals of the form $[\hat{y}(x) - \xi(x), \hat{y}(x) + \xi(x)]$.

> **Property 4.20.** If the estimates $\hat{\mu}, \hat{\sigma} : \mathcal{X} \to \mathbb{R}$ are consistent, normalized conformal prediction converges to interval conformal prediction:
>
> $$\lim_{|\mathcal{V}| \to \infty} \hat{y}^{\mathrm{NCP}}_{\pm} = \hat{y}^{\mathrm{int}}_{\pm} . \qquad (4.52)$$

A similar analysis could be performed for the Student $t$-interval (3.15). The resulting interval coming from the interval CP method is given by:

$$\hat{y}^{\mathrm{int}}_{\pm}(x) = \hat{\mu}(x) \pm \left[ |\hat{\mu}(x^*) - y^*| + t_{n-1}\sqrt{1 + 1/n}\big(\hat{\sigma}(x) - \hat{\sigma}(x^*)\big) \right] . \qquad (4.53)$$

For NCP, a similar approach can be followed. If instead of normalizing by $\hat{\sigma}$, we normalize by $\sqrt{1 + 1/n}\,\hat{\sigma}$, the intervals are given by:

$$\hat{y}^{\mathrm{NCP}}_{\pm}(x) = \hat{\mu}(x) \pm \left[ |\hat{\mu}(x^*) - y^*| + t^*_{n-1}\sqrt{1 + 1/n}\big(\hat{\sigma}(x) - \hat{\sigma}(x^*)\big) \right] . \qquad (4.54)$$

More generally, whenever the intervals are of a central form

$$\Gamma^{\alpha}(x) = \left[ \hat{y}(x) - \xi^{\alpha}(x), \hat{y}(x) + \xi^{\alpha}(x) \right] , \qquad (4.55)$$

for a location estimate $\hat{y} : \mathcal{X} \to \mathbb{R}$ and spread estimate $\xi^{\alpha} : \mathcal{X} \to \mathbb{R}^+$, the induced interval CP and NCP predictions can be related and, if these estimates are consistent, an analogue of Property 4.20 will hold.

## 4.6   Experiments on synthetic data

In this section, the results from Section 4.5 will be validated by performing and analysing some synthetic experiments. Moreover, a diagnostic tool will be developed to help assess whether a non-Mondrian conformal predictor could be conditionally valid (with respect to a given taxonomy function) and the general influence of estimates that deviate from the ground truth will be investigated.

### 4.6.1   Data types

To compare the different methods in a controlled manner, some synthetic data sets are considered. Four different types are of importance and representative of real-world situations:

Type 1.   Constant mean:

$$\mathsf{E}\big[Y \,|\, X\big] = \mathsf{E}\big[Y\big] . \qquad (4.56)$$

Even though the (conditional) mean is a constant function on the feature space $\mathcal{X}$ and, hence, fluctuations in the data are purely of a stochastic nature, most models will still try to attribute the behaviour to a nonconstant mean and only to a lesser extent to a nonzero (and nonconstant) variance. (Mainly because most training algorithms heavily focus on the mean and only to a lesser extent on the variance.)

Type 2. Functional dependence:

$$\mathsf{Var}\big[Y \,|\, X\big] = \varphi\big(\mathsf{E}\big[Y \,|\, X\big]\big),  \tag{4.57}$$

for some function $\varphi : \mathbb{R} \to \mathbb{R}^+$. In its most basic form, where $\varphi$ is a monomial, this captures examples such as Poisson and exponential distributions (also recall Remark 4.19).

Type 3. Low-dimensional representation:

$$\mathsf{Var}\big[Y \,|\, X\big] = f(X^\downarrow),  \tag{4.58}$$

where $X^\downarrow$ denotes the projection of $X$ onto a subspace of $\mathcal{X}$, such as the projection of a vector onto its first components. (A generalization could be considered in light of the *manifold hypothesis* (Fefferman, Mitter, & Narayanan, 2016).)

Type 4. Mixture models:

$$P_{Y|X} = \sum_{i \in \mathfrak{I}} \lambda_i Q_i(Y \,|\, X),  \tag{4.59}$$

for some finite[5] set $\mathfrak{I}$, where the supports of the $Q_i$ are often disjoint or have only partial overlap.

Note that, similar to data of Type 1, we could also consider data with a constant variance function. However, this implies homoskedasticity (at least aleatorically) and is not considered here, since the taxonomy functions of interest are exactly those that are functions of the uncertainty.

To generate data sets of each of the above types synthetically, a general sampling procedure is followed. The main steps are as follows:

---

[5] In theory, the index set does not have to be finite, but in practice it always is (at least for sampling purposes).

1. $n \in \mathbb{N}$ feature tuples $x_i \in \mathcal{X}$ are sampled from a fixed distribution $P_X$, e.g. a uniform distribution over a $k$-dimensional (unit) hypercube.

2. A (parametric) family of distributions $P_{Y|X} \in \mathcal{P}_2(\mathcal{Y})$ is chosen, e.g. normal distributions $\mathcal{N}(\mu, \sigma^2)$.

3. A choice of (conditional) mean and variance functions $\mu : \mathcal{X} \to \mathbb{R}$ and $\sigma^2 : \mathcal{X} \to \mathbb{R}^+$, parametrizing the family of conditional distributions $P_{Y|X}$, is made.

4. For every feature tuple $x_i \in \mathcal{X}$, a response $y \in \mathbb{R}$ is sampled from the distribution $P_{Y|X}\big(\cdot \mid x_i ; \mu(x_i), \sigma^2(x_i)\big)$.

To evaluate the quality of interval predictors $\Gamma^\alpha : \mathcal{X} \to 2^\mathbb{R}$, two performance metrics are used as in the previous chapter: the coverage (2.6) and the average size of the prediction intervals (2.11). We explicitly give the conditional counterparts for simplicity as in Eq. (4.12):

$$\begin{aligned}
\mathcal{C}\big(\Gamma^\alpha, P_{X,Y} \mid \kappa(X,Y)\big) &= \mathsf{Prob}\big(Y \in \Gamma^\alpha(X) \mid \kappa(X,Y)\big) \\
&= \mathsf{E}\big[\mathbb{1}_{\Gamma^\alpha(X)}(Y) \mid \kappa(X,Y)\big]
\end{aligned} \qquad (4.60)$$

and

$$\mathcal{W}\big(\Gamma^\alpha, P_{X,Y} \mid \kappa(X,Y)\big) = \mathsf{E}\big[|\hat{y}_+(X) - \hat{y}_-(X)| \mid \kappa(X,Y)\big], \qquad (4.61)$$

where, as before, the functions $\hat{y}_\pm : \mathcal{X} \to \mathbb{R}$ denote the upper and lower bounds of the prediction intervals produced by $\Gamma^\alpha$. Note that whereas the (marginal) distribution over $\mathbb{R}$ is irrelevant for the marginal measures, it plays a crucial role in the conditional definitions, since the taxonomy function can, in general, also depend on the response variable.

## 4.6.2 Deviations

Up to some very specific relaxations in the form of Eqs. (4.40) and (4.41) and, more generally, Property 4.14, the theorems and methods from the previous sections require the parameters to be estimated exactly (or at least consistently in the large data setting). However, this assumption is not a very realistic one in practice. Estimating higher conditional moments, such as the variance, to high precision usually requires state-of-the-art methods and, even more so, a large amount of data, since without strong parametric assumptions multiple samples with nearly identical features are necessary.

It should be clear that this requirement is hard to satisfy, especially in high-dimensional settings.

Because obtaining perfect estimates is often near impossible, it is interesting to see what happens when the estimates deviate from the values obtained from an oracle, i.e. the ground truth. In general, these deviations can be attributed to two sources:

- **Misspecification**: General nonconformity measures usually depend on the parameters that have been estimated. As a consequence, if the estimators were misspecified, the transformation $(X, Y) \mapsto A(X, Y)$ might not remove all dependency on the feature space $\mathcal{X}$.

- **Contamination**: Similar to the case of misspecification, the taxonomy function will, in general, depend on the estimated parameters and, if the estimators were misspecified, the taxonomy classes can get mixed up. Even though the distribution of the nonconformity scores might not depend on the true taxonomy, it might depend on the estimated taxonomy.

**Example** 4.21. Recall Example 4.7 and consider the residual nonconformity measure in Eq. (3.38):

$$A(x, y) = |y - \hat{\mu}(x)|.  \tag{4.62}$$

The taxonomy function is still the indicator function:

$$\kappa_{\hat{\xi}}(x, y) = \mathbb{1}_{[0, \hat{\xi}]}(x^2),  \tag{4.63}$$

where $\hat{\xi} \in \mathbb{R}$. For $\hat{\xi} = 0.5$, it is not hard to see that the conformal predictor associated with $A$ will be conditionally valid, even though $A$ itself is not pivotal for the given data-generating process.

Analyzing the effect of misspecification and contamination is now also quite straightforward. If the location $\hat{\mu}$ is misspecified (in such a way that the residuals have a mean that depends on $x^2$), the distribution of the nonconformity scores will also depend on the taxonomy, no matter how perfect the parameter $\hat{\xi}$ is fine-tuned. On the other hand, as soon as $\hat{\xi}$ deviates from 0.5, e.g. when it would be estimated based on a data sample, the distribution of nonconformity scores would also no longer be independent of the classes, even when $\hat{\mu}$ is modelled perfectly.

Although misspecification and contamination are, generally, two distinct sources of heterogeneity in the nonconformity distributions, in the setting of this chapter, where both the taxonomy function and the nonconformity measure depend explicitly on (an estimate of) the data noise as in Eq. (4.15), misspecification and contamination go hand in hand.

In Figs. 4.6 and 4.7, the coverage results for the four types of synthetic data are shown (Types $1-4$) for different kinds of misspecification and for the three types of nonconformity measure introduced in Section 2.3: the residual (3.38), interval (3.46) and $\sigma$-normalized measures (4.19). For each of these figures, the conditional distribution $P_{Y|X}$ has the general form of Theorem 4.13. More precisely, for these synthetic experiments, the conditional distributions $P_{Y|X}$ are chosen to be normal distributions. The misspecification is simulated by adding random noise to the values of the mean and variance functions and six different situations are shown in columns $2-7$. The first (green) column, indicated by the label 'Oracle', uses the true mean and standard deviation.

The columns (orange, blue and pink) indicated by the label '$\sigma$-shifted $(\lambda)$' show the empirical coverage for increasing values of noise in the standard deviation estimates:

$$\widehat{\sigma}(x) = \sigma(x) + \varepsilon \qquad \text{with} \qquad \varepsilon \sim \mathcal{N}\left(0, \lambda^2\right). \qquad (4.64)$$

Note that these values are clipped to the positive real line $\mathbb{R}^+$ to guarantee the positivity of the standard deviation. This simulates the behaviour of models that are not able to estimate the variance accurately or even consistently. It is clear that for larger values of the noise parameter $\lambda$, the (non-Mondrian) conformal predictor using the normalized conformal measure stops being conditionally valid as expected. The light green column, indicated by the label '$\sigma$-scaled' shows the coverage when the variance is scaled by a fixed value ($\lambda = 5$ in this case). As expected from Property 4.14 and, in particular, Eq. (4.40), this does not change anything in terms of the conditional validity of the normalized conformal predictors. However, it does break the conditional validity of the conformal predictor using the interval nonconformity measure. This observation was also to be expected. For the mean-variance estimators from Example 4.8 with a normality assumption, the intervals are of the form of Eq. (3.14). As shown in Eq. (4.48), it follows that the interval score can be rewritten as follows:

$$\max\left(\widehat{y}_-(x) - y, y - \widehat{y}_+(x)\right) = |\widehat{\mu}(x) - y| - z^\alpha \widehat{\sigma}(x). \qquad (4.65)$$

From this expression, it is immediately clear that scaling the standard deviation is not equivalent to applying a feature-independent transformation and, accordingly, Property 4.14 is not applicable.

The last two columns (yellow and brown), indicated by the label '$\mu$-shifted', show the coverage when the mean is shifted by, respectively, a constant and a value proportional to the standard deviation:

$$\hat{\mu}(x) = \mu(x) + \varepsilon \qquad \text{with} \qquad \varepsilon \sim \mathcal{N}\left(0, \lambda^2 \widehat{\sigma}(x)^2\right). \qquad (4.66)$$

It is immediately clear that whereas the constant shift breaks the conditional validity of all three conformal predictors, a shift proportional to the standard deviation does preserve the conditional validity of the normalized model. This is entirely in line with the above results, in particular Eq. (4.41).

A similar plot for a data set of Type 4 is shown in the lower half of Fig. 4.7. In this case, the data is sampled from the following bimodal mixture:

$$y \sim \begin{cases} \mathcal{N}\left(\mu(x) - 1, 0.01\mu(x)^2\right) & \text{if } \mu(x) \leq 2, \\ \mathcal{N}\left(\mu(x) + 1, 0.01\mu(x)^2\right) & \text{if } \mu(x) > 2. \end{cases} \qquad (4.67)$$

Even in the 'Oracle'-column, where the variance of the mixture is given by Example A.69, it is already clear that the non-Mondrian conformal predictors are not conditionally valid in general for such a distribution. This observation is expected to hold for all mixture distributions $P_{Y|X}$. Unless a consistent model can be obtained for every component in the mixture and unless each of these components admits the same pivotal distribution, conditional validity will not be attained.

### 4.6.3 Diagnostics

The results in the preceding sections actually provide us with a means to get an idea of the conditional behaviour of conformal predictors without having to consider a test set. If the nonconformity distributions in every taxonomy class are the same, the conformal predictor will attain conditional validity. Consequently, comparing the distributions of the nonconformity scores over the different taxonomy strata of the calibration set can give insight into how well the methods will perform and what impact misspecification and contamination might have.

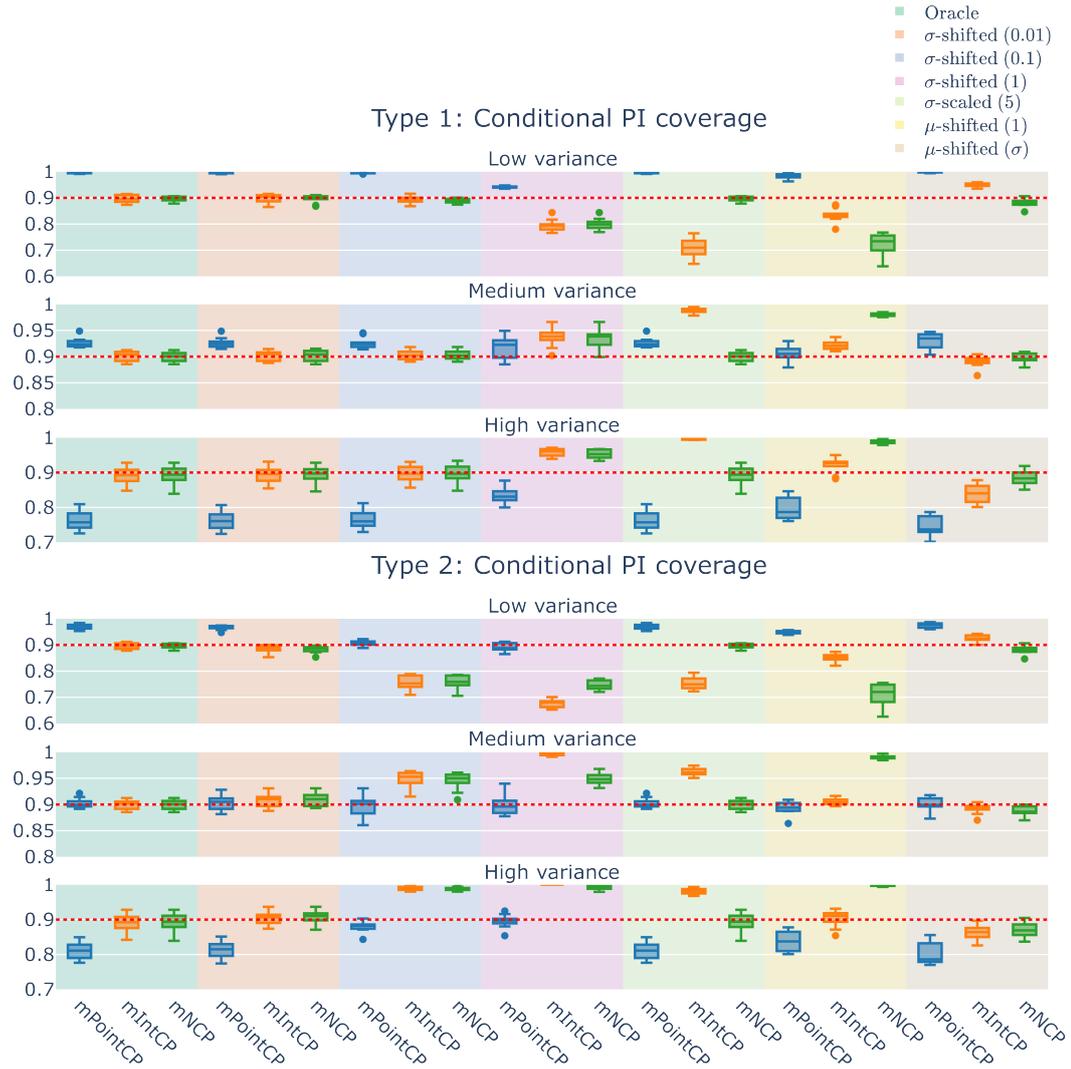

Figure 4.6: Conditional coverage at significance level $\alpha = 0.1$ for synthetic data sets of Types 1 and 2. For every type, the data is divided in three folds based on equal-frequency binning of the estimated variance. The coloured columns indicate the type of misspecification (from left to right): oracle, additive noise on the standard deviation (means of 0.01, 0.1 and 1), scaling by a factor 5 of the standard deviation and additive noise on the mean (means of 1 and $\hat{\sigma}$). For every model, three nonconformity measures are shown (from left to right): residual, interval and $\hat{\sigma}$-normalized nonconformity measure.

As a toy example, a family of normal distributions with constant *coefficient of variation*, fixed at $c_v = 0.1$, is chosen as data-generating process:

$$Y \mid X \sim \mathcal{N}\big(\mu(X), 0.01\,\mu(X)^2\big). \tag{4.68}$$

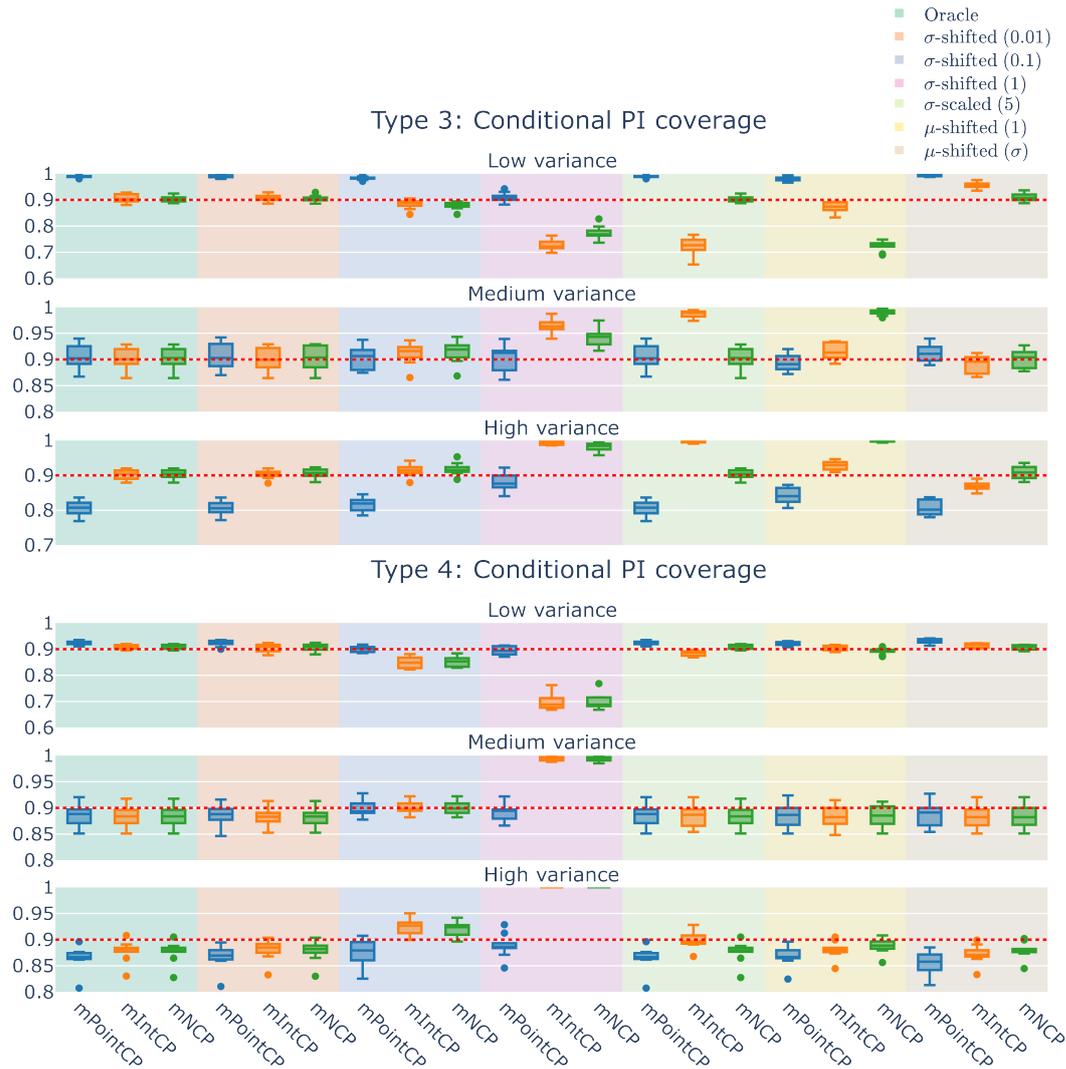

Figure 4.7: Conditional coverage at significance level $\alpha = 0.1$ for synthetic data sets of Types 3 and 4. For every type, the data is divided in three folds based on equal-frequency binning of the estimated variance. The coloured columns indicate the type of misspecification (from left to right): oracle, additive noise on the standard deviation (means of 0.01, 0.1 and 1), scaling by a factor 5 of the standard deviation and additive noise on the mean (means of 1 and $\widehat{\sigma}$). For every model, three nonconformity measures are shown (from left to right): residual, interval and $\widehat{\sigma}$-normalized nonconformity measure.

Figure 4.8a shows a CDF plot of the variance of these distributions with

$$\mu(x) = \text{mean}(x) \tag{4.69}$$

and $X \sim \mathcal{U}^n([0, 100])$. The colors indicate the taxonomy classes corresponding to equal-frequency binning of the variances as in Eq. (4.15). The CDF

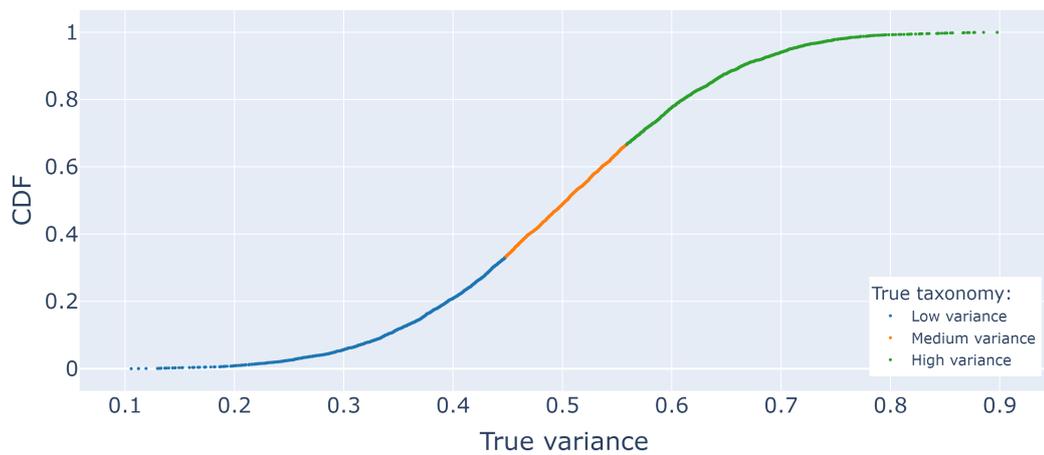

(a) CDF plot of the (true) variance. The colors indicate the taxonomy classes corresponding to equal-frequency binning (with $n = 3$ classes).

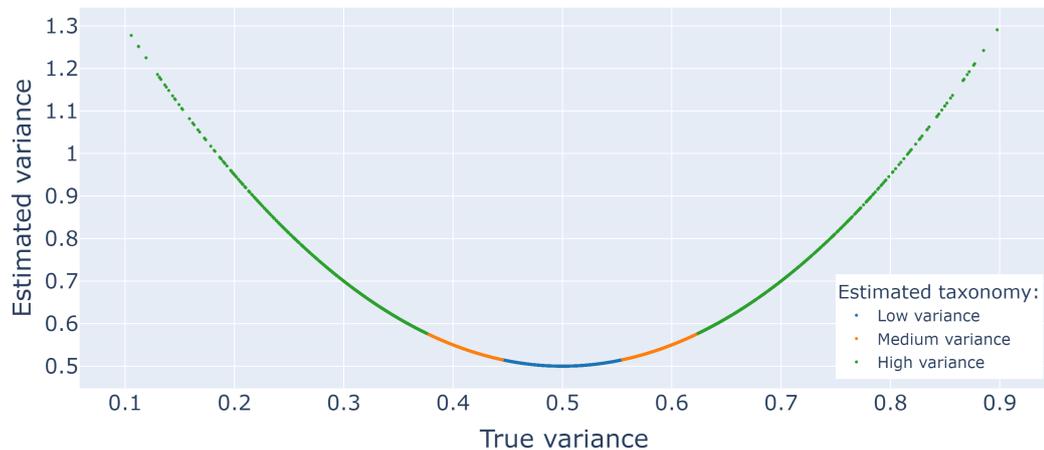

(b) Comparison of the true and estimated variance. The colors indicate the taxonomy classes corresponding to equal-frequency binning (with $n = 3$ classes) of the estimated variance.

Figure 4.8: Characterization of the true (a) and estimated (b) variance in Eq. (4.68). Comparing the subfigures shows that misspecification strongly mixes the taxonomy classes.

plots of the residual and ($\sigma$-)normalized nonconformity scores are shown in Figs. 4.9a and 4.9c, respectively. When applying a non-Mondrian conformal predictor with the residual nonconformity measure to 20 random test sets, each containing 1000 instances, sampled from the distribution in Eq. (4.68), the results shown in the first row of Table 4.1 are obtained.

While performing experiments on synthetic data, it was observed that most methods showed quadratic deviations when comparing the true variances to the predicted variances. In the region with high levels of noise, the estimates were approximately correct, but in the regions with low noise levels, the estimates were often much higher than expected (of the same magnitude as in region with high noise levels). This effect can be modelled by using the following explicitly misspecified model:

$$\hat{\mu}(x) = \mu(x) \qquad \text{and} \qquad \widehat{\sigma}^2(x) = 5\big(\sigma^2(x) - 0.5\big)^2 + 0.5 \,. \qquad (4.70)$$

Table 4.1: Empirical coverage for different methods across variance classes (at significance level $\alpha = 0.1$). Mean and standard deviation over 20 samples are reported.

|            | Marginal      | Low variance  | Medium variance | High variance |
|------------|---------------|---------------|-----------------|---------------|
| $A_{\text{res}}$ | $0.905 \pm 0.008$ | $0.951 \pm 0.010$ | $0.904 \pm 0.017$ | $0.856 \pm 0.018$ |
| $A_{\sigma}$ | $0.902 \pm 0.008$ | $0.905 \pm 0.018$ | $0.902 \pm 0.015$ | $0.898 \pm 0.017$ |

The consequences can be seen in Fig. 4.8b for the variance and Fig. 4.9 for the nonconformity scores. In the first figure, the estimated variance is shown in function of the true variance. The colors again indicate the taxonomy classes, but this time those determined by the estimated variance. As is clearly visible by comparing Fig. 4.8a to Fig. 4.8b, the taxonomy classes are completely different from what the true classes would be. The true class with medium variance corresponds to the class with low estimated variance and the classes with low and high true variance have been reshuffled to become 50/50 mixtures of low and high estimated variance. When comparing the CDF plots in Figs. 4.9a and 4.9c to those in Figs. 4.9b and 4.9d, an interesting effect can be seen. Whereas the residual score for the oracle did not give rise to conditional coverage, it does do so for the misspecified model. On the other hand, for the $\widehat{\sigma}$-normalized nonconformity score, the effect acts in the opposite direction. For the oracle, conditional validity is attained (for all significance levels), but for the misspecified model, only marginal validity is attained. These results are also reflected in Table 4.2.

The analysis of the toy model in Eq. (4.68) leads to the following diagnostic method.

**Method** 4.22 (**CDF plots**).  Assuming that the conformal predictors are not overly conservative, meaning that ties do not arise by Theorem 2.28, and that the calibration set is representative, creating a CDF plot of both the marginal nonconformity scores and the conditional nonconformity scores for all taxonomy classes can give an idea of how well the method will perform conditionally. If the plots coincide at some significance level, the conformal predictor can be expected to be conditionally valid at that level.

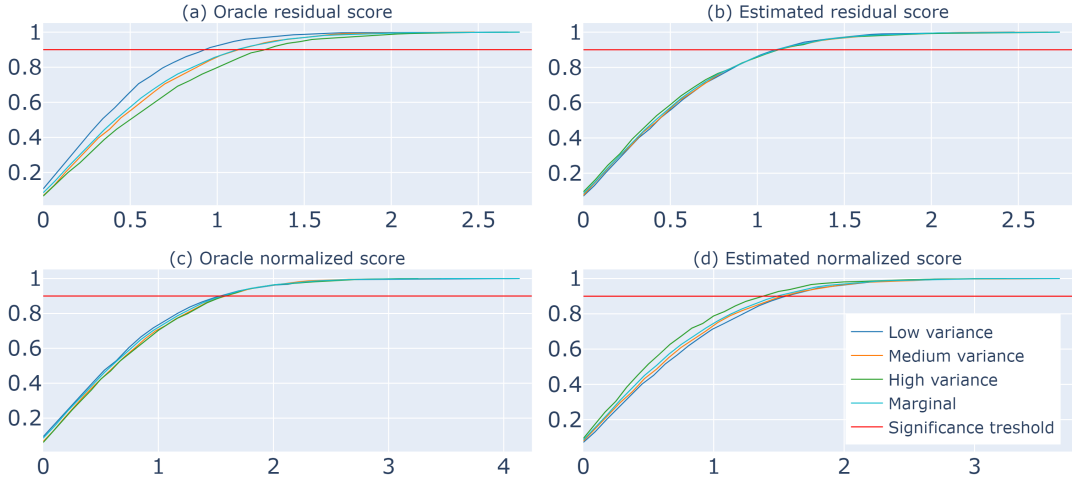

Figure 4.9: (a) and (c): CDF plots of the residual and $\sigma$-normalized nonconformity scores for the oracle. (b) and (d): CDF plots of the residual and $\sigma$-normalized nonconformity scores for the misspecified model in Eq. (4.68). The (empirical) distributions are shown marginally (all data) and for the taxonomy classes corresponding to equal-frequency binning of the estimated variance (with $n = 3$ classes).

Table 4.2: Empirical coverage for different methods across variance classes (at significance level $\alpha = 0.1$) for the misspecified model in Eq. (4.70). Mean and standard deviation over 20 samples are provided.

| | Marginal | Low variance | Medium variance | High variance |
|---|---|---|---|---|
| $A_{\mathrm{res}}$ | $0.905 \pm 0.008$ | $0.905 \pm 0.017$ | $0.905 \pm 0.014$ | $0.905 \pm 0.012$ |
| $A_{\sigma}$ | $0.905 \pm 0.008$ | $0.886 \pm 0.017$ | $0.897 \pm 0.014$ | $0.931 \pm 0.008$ |

As is immediately clear from Fig. 4.9a, the quantiles for low and high variance are, respectively, smaller and greater than the marginal quantile at significance level $\alpha = 0.1$. This directly translates to over- and undercoverage in these regions, although the model is valid marginally (as expected from Theorem 2.27). By happenstance, the CDF for the taxon-omy class with medium variance intersects the marginal CDF at around this significance level and this is also visible in the coverage values. For medium variance, the coverage is close to the nominal level.

For the normalized conformal predictor with the scores from Fig. 4.9c, the coverage degrees are shown in the second row of Table 4.1 (again for 20 test sets with 1000 instances). The fact that all CDFs approximately coincide at all levels in the figure is reflected in the conditional coverage values in the table. The model can be seen to be valid for all classes.

Instead of inspecting nonconformity distributions purely visually, a more analytical approach can be followed. In the situations where a single calibration set is used and, accordingly, consistency of the $(1 - \alpha)$-sample quantile is assumed, the framework of hypothesis testing can be applied.

**Method** 4.23 (**Bootstrap analysis**). A statistical test can be used to determine whether the required quantiles coincide for the different taxonomy classes. To compare the $(1 - \alpha)$-quantiles between two different taxonomy classes, the method from Wilcox, Erceg-Hurn, Clark, and Carlson (2014) can be used. Choose two taxonomy classes $c_1, c_2 \in [k]$ and an integer $B \in \mathbb{N}_0$. Then, consider a fixed number of bootstrap samples $\{\mathcal{V}_i^1\}_{i=1,\dots,B}$ and $\{\mathcal{V}_i^2\}_{i=1,\dots,B}$, sampled from the data sets $\mathcal{V}_{c_1}$ and $\mathcal{V}_{c_2}$, respectively. For every $i \in \{1, \dots, B\}$, the quantile difference

$$d_i := \hat{q}_{(1-\alpha)(1+1/n)}(\mathcal{V}_i^1) - \hat{q}_{(1-\alpha)(1+1/n)}(\mathcal{V}_i^2) \tag{4.71}$$

can be calculated using the Harrell–Davis quantile estimator (Harrell & Davis, 1982). A bootstrap confidence interval at significance level $\beta \in [0, 1]$, or at least an approximate one, is given by $[d_{(B\beta/2)}, d_{(B-B\beta/2)}]$. If 0 is not contained in any of the pairwise confidence intervals for a suitable value of $\beta$, e.g. $\beta = 0.01$, evidence has been found against the use of the conformal predictor for attaining conditional validity.

**Remark** 4.24. The above diagnostic tools, especially the bootstrap analysis, are primarily useful in case only a fixed significance level (or a finite number of them) is of interest. To test whether validity will hold at a range of significance levels (or even at all levels), another test for comparing distributions, such as the *Kolmogorov–Smirnov test*, might be preferred.

## 4.7 Generative models

In the previous chapter, we considered four classes of confidence predictors: Bayesian methods, mean-variance estimators, ensemble methods and methods that directly construct intervals. In this section, a fifth class belonging to the family of *generative models* will be introduced. The idea behind generative models is that they represent the conditional distribution $P(Y \mid X)$, but not necessarily in a Bayesian way. For a general introduction, see e.g. Goodfellow et al. (2016). Here, we will focus on the subclass of flow models.

### 4.7.1 Flows

**Definition 4.25 (Flow).** A continuous[a] function $\phi : \mathbb{R} \times \mathcal{X} \to \mathcal{X} : (t, x) \mapsto \phi_t(x)$ is called a flow if it satisfies:

1. **Unity**: $\phi_0 = \mathbb{1}_{\mathcal{X}}$,

2. **Multiplication**: $\phi_t \circ \phi_s = \phi_{t+s}$ for all $s, t \in \mathbb{R}$, and

3. **Invertibility**: $\phi_t^{-1} = \phi_{-t}$ for all $t \in \mathbb{R}$.

[a] Depending on the context, differentiability, smoothness, or any other number of specializations can be considered.

**Extra 4.26 (Group theory).** A flow on a set $\mathcal{X}$ can be defined more concisely as a group action of $\mathbb{R}$ on $\mathcal{X}$ (Definition A.6).

For most practical purposes, continuous flows are approximated by discrete flows $\phi : \mathbb{Z} \times \mathcal{X} \to \mathcal{X}$ (which would be equivalent to a group action of the integers). The most straightforward approximation is given by the 'power flow' generated by a bijection $\Phi : \mathcal{X} \to \mathcal{X}$:

$$\forall k \in \mathbb{Z} : \phi_k := \Phi^k . \tag{4.72}$$

This could be for example the *time-1 map* of a continuous flow. However, for our purpose, the more general form[6]

$$\phi_k := \Phi_k \circ \cdots \circ \Phi_1 \tag{4.73}$$

for a sequence $(\Phi_n)_{n \in \mathbb{N}}$ of invertible transformations is considered.

---

[6] Note that this is technically no longer a discrete flow since it fails to satisfy the multiplication axiom in Definition 4.25.

**Example** 4.27 (**Euclidean flows**). Some common flows (or their generators) on $\mathbb{R}^d$ are listed here:

- Affine flow:

$$\Phi(x) = Ax + v \qquad (4.74)$$

  for some $v \in \mathbb{R}^d$ and $A \in \mathsf{GL}(d)$, where $\mathsf{GL}(d)$ denotes the general linear group in $d$ dimensions, i.e. the group of invertible $d \times d$-matrices.

- Planar flow:

$$\Phi(x) = x + h(\boldsymbol{w} \cdot x + \lambda)\boldsymbol{v}\,, \qquad (4.75)$$

  where $\lambda \in \mathbb{R}$, $\boldsymbol{v}, \boldsymbol{w} \in \mathbb{R}^d$ and $h : \mathbb{R} \to \mathbb{R}$ is a *diffeomorphism*. In general, the inverse of a planar flow cannot be expressed in closed form.

- Coupling flow:

$$\Phi(x) = \left(x^\alpha, h(x^\beta, x^\alpha)\right)\,, \qquad (4.76)$$

  where $(\alpha, \beta)$ is a partition of $[d]$ and $h(-, x^\alpha) : \mathbb{R}^{|\beta|} \to \mathbb{R}^{|\beta|}$ is a *diffeomorphism* for every $x^\alpha \in \mathbb{R}^{|\alpha|}$.

Given such a sequence $(\Phi_n)_{n \in \mathbb{N}}$ and a probability density function $f_X : \mathcal{X} \to \mathbb{R}^+$, the change-of-variables formula (Property A.30) implies the following expression:

$$f_X(x) = f_{\phi_k(X)}\big(\phi_k(x)\big) \prod_{i=1}^{k} \left| \frac{\partial \Phi_i}{\partial \Phi_{i-1}}\big(\Phi_{i-1}(x)\big) \right| \qquad (4.77)$$

for all $x \in \mathcal{X}$, where $f_{\phi_k} := \phi_{k,*} f$ is simply the density function of the transformed random variable $\phi_k(X)$. The idea of normalizing flows (or of using flows for generative modelling) is that by assuming a certain form for the final density function $f_{\phi_k(X)}$, the above formula allows to model the target density function $f_X$.

### 4.7.2 Normalizing flows

On $\mathbb{R}^d$, a specific type of flow is useful in this respect (Kobyzev, Prince, & Brubaker, 2020; Rezende & Mohamed, 2015).

**Definition** 4.28 (**Normalizing flow**). Let $P \in \mathbb{P}(\mathbb{R}^d)$ be a probability distribution. A flow $\phi : \mathbb{Z} \times \mathbb{R}^d \rightarrow \mathbb{R}^d$ is called a normalizing flow for $P$ if

$$\lim_{n \to \infty} \phi_{n,*} P = \mathcal{N}(\mu, \Sigma) \tag{4.78}$$

for some mean vector $\mu \in \mathbb{R}^d$ and covariance matrix $\Sigma \in \mathbb{R}^{d \times d}$. More generally, a sequence $(\phi_n)_{n \in \mathbb{N}}$ of transformations constitutes a normalizing flow if

$$(\phi_n \circ \cdots \circ \phi_1)_* P = \mathcal{N}(\mu, \Sigma)$$

for some mean vector $\mu \in \mathbb{R}^d$ and covariance matrix $\Sigma \in \mathbb{R}^{d \times d}$.

The following theorem supports the use of normalizing flows for the generative modelling of any (continuous) distribution on $\mathbb{R}^d$. Moreover, it also points to a specific type of transformation to achieve this goal.

**Theorem** 4.29. *Every two absolutely continuous distributions on $\mathbb{R}^d$ are related by an increasing triangular Borel function.*

*Proof*. See Bogachev, Kolesnikov, and Medvedev (2005).                    □

Before moving on, it might be useful to explain the adjectives in this theorem. 'Borel' indicates the fact that we are working with the Borel $\sigma$-algebra on $\mathbb{R}^n$ (Definition A.11) and for the purpose of this dissertation simply means the standard notion of measurable function between Euclidean spaces. A 'triangular' function is a function $\Phi : \mathbb{R}^d \rightarrow \mathbb{R}^d$ for which there exist bases of $\mathbb{R}^d$ with respect to which $\Phi$ admits the following coordinate expression:

$$\Phi(x)^i = f_i(x^i, \dots, x^1) \tag{4.79}$$

for some suitable function $f_i : \mathbb{R}^i \rightarrow \mathbb{R}$ for all $i \leq d$. In vector form, this becomes (making the terminology more evident):

$$\Phi(x) = \begin{pmatrix} f_1(x^1) \\ f_2(x^2, x^1) \\ \vdots \\ f_d(x^d, \dots, x^1) \end{pmatrix}. \tag{4.80}$$

A schematic representation of a triangular function is shown in Fig. 4.10. An 'increasing triangular' function is a triangular function $\Phi$ for which every component $f_i$ is increasing in its first argument, i.e.

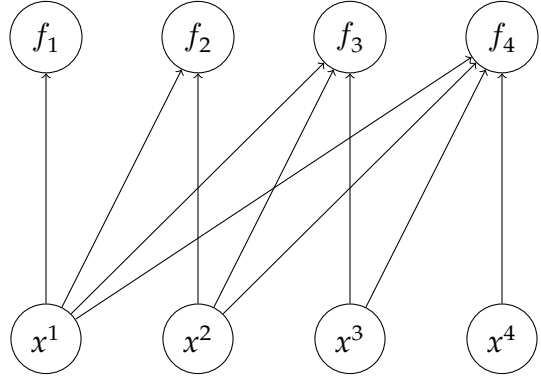

Figure 4.10: Schematic view of triangular function on $\mathbb{R}^4$.

$$x \leq y \implies f_i\big(x, x^{i-1} \dots, x^1\big) \leq f_i\big(y, x^{i-1}, \dots, x^1\big) \tag{4.81}$$

for all $i \leq d$ and $\{x^1, \dots, x^{i-1}, x, y\} \subset \mathbb{R}$. Flows generated by transformations of this form are also called **autoregressive flows** (Kingma et al., 2016; Kobyzev et al., 2020).

At this point, implementing such a model seems easy. The components of the transformation $\Phi$ can be taken to be the heads of a neural network and optimizing the parameters can be achieved through maximum likelihood optimization with respect to the density in Eq. (4.77). However, the situation is slightly more complicated. In regression problems, we do not simply have a distribution $P \in \mathbb{P}(\mathbb{R}^d)$. We have a family of conditional distributions $P_{Y|X} \in \mathbb{P}(\mathbb{R}^d \mid \mathcal{X})$. This translates to the parameters $\theta_i \in \mathbb{R}^k$ of the model not simply being constants, but in fact becoming functions $\theta_i : \mathcal{X} \to \mathbb{R}^k$.

For $d = 1$, i.e. univariate regression, conditional autoregressive networks can be obtained through a simple procedure (at least on paper). First, a nonnegative function $g_\omega : \mathbb{R} \times \mathcal{X} \to \mathbb{R}^+$ is constructed.[7] Note that this function does not have to be increasing. Then, to obtain an increasing function, it is integrated (Wehenkel & Louppe, 2019):

$$\Phi_\omega(y \mid x) = \int_0^y g_\omega(z, x) \, \mathrm{d}z . \tag{4.82}$$

Since the integral of a nonnegative function is always an increasing function with respect to the upper integration bound, this construction circum-vents the need to incorporate the monotonicity constraint in the learning process. With neural networks, for example, obtaining such a function is simple enough. Just take any neural network and add an ELU (exponential linear unit) activation at the end (shifted by 1).[8] An alternative is to construct a neural network with only positive weights and monotonically increasing activation functions (Huang, Krueger, Lacoste, & Courville, 2018). Moreover, backpropagation of the monotonic transformation in Eq. (4.82) is straightforward through the fundamental theorem of calculus or, more specifically, Leibniz's integral rule (4.30). For the network parameters $\omega$, this leads to the following expression:

$$\frac{\partial}{\partial \omega} \Phi_\omega(y \mid x) = \int_0^y \frac{\partial}{\partial \omega} g_\omega(z, x) \, \mathrm{d}z \,. \tag{4.83}$$

---

[7] The parameters characterizing this function are explicitly shown as a subscript for further convenience.

If the gradient with respect to the input $y$ is required, the expression is even simpler:

$$\frac{\partial}{\partial y} \Phi_\omega(y \mid x) = g_\omega(y, x) \,. \tag{4.84}$$

Note that these equations imply that backpropagation can be performed efficiently. Calculating the gradient of the integral would have required to propagate the gradient through the whole numerical integration scheme, while integrating the gradient has the same complexity as calculating the integral in Eq. (4.82).

## 4.8 Experiments on real data

In the synthetic experiments of Section 4.6, it was possible to investigate the impact of misspecification. In practice, however, there is no way to know the exact model parameters and misspecification almost always occurs. For this reason, it is also useful to see how the models perform on realistic data sets.

### 4.8.1 Data

The selected data sets are a subset of those considered in the previous chapter (see Section 3.4). More specifically, the `fb1` and `blog` data sets are not included due to their size and the fact that, as noted in Section 3.4.3, the uncertainty estimates from the models differed by some orders of magnitude. Since we want to condition on these estimates, it was decided to leave them aside in further analyses. More information about the data sets can be found in Tables 3.1 and 3.2. The former shows the number of data points and (selected) features, together with the skewness and (Pearson) kurtosis of the response variable, while the latter gives specific references.

---

[8] An ELU activation is preferred over a ReLU activation to obtain strictly increasing functions.

All data sets were standardized (both features and target variables) before training. For the taxonomy function $\kappa : \mathcal{X} \times \mathbb{R} \to [k]$, equal-frequency binning (4.15) of the variance estimates with $n = 3$ bins was chosen, as in the synthetic case. Although these data sets all have different characteristics, e.g. dimensionality, sparsity, count data vs. continuous data, etc., they are treated equally as in Chapter 3. The experimental results could be interpreted in more detail in light of these differences, but this would require more insight in the underlying data-generating mechanisms.

## 4.8.2    Models and training

Since the focus lies on uncertainty-dependent conditioning, only models that actually provide estimates of the conditional variance will be considered. This excludes point predictors such as standard neural networks or random forests. The methods considered in this benchmarking effort are listed below (abbreviations that will be used in the remainder of the text are indicated between parentheses):

- quantile regressors (QR),

- quantile regression forests (QRF),

- mean-variance estimators (MV),

- mean-variance ensembles (MVE), and

- normalizing flows (NF).

The first four methods form a subset of those listed in Table 3.3 and are explained in more detail in Section 3.2, whereas the fifth method was introduced in Section 4.7 above. Each of these methods comes in different flavours when augmented with conformal prediction. All of them have a baseline version where no conformal prediction is applied, but a paramet-ric interval formula like Eq. (3.14) is assumed, a marginal CP incarnation giving rise to three models (residual score (3.38), interval score (3.46) and normalized score (4.18)) and a Mondrian CP incarnation, again giving rise to three models. In the remainder of this section, the three variants are, respectively, denoted by PointCP, IntCP and NCP. The marginal variants are further indicated by the prefix 'm-'. For the conditional performance, this results in a total of seven different options. The aim of this experimental part is to analyse whether the results from the previous section hold, i.e. to check when the marginal models, the normalized variant in particular, have the same conditional performance as the Mondrian ones.

All neural networks were constructed using `PyTorch` (Paszke et al., 2019). Just as in the previous chapter (Section 3.4), the *Adam-optimizer* was used for weight optimization with a fixed learning rate of $5 \times 10^{-4}$. The number of epochs was fixed at 300 and all networks contained three hidden layers, each with 64 neurons. (The number of epochs and hidden layers is greater than in the previous chapter, since accurate predictions are more important to reduce the effect of misspecification and contamination.) The activation functions after all layers, except the final one, were of the ReLU type, while the activation function at the output node was simply a linear function. For regularization purposes, all models, except the normalizing flows which used the implementation from Wehenkel and Louppe (2019), also had a dropout layer (Srivastava et al., 2014) before each hidden layer, with the dropout probability fixed at $p = 0.2$.

For completeness, each of the models from the list above, is explained here in a little more detail:

1. Quantile regression: Since, except for the baseline version, the models are augmented by conformal prediction, the models do not have to be trained at extreme significance levels, where data might be scarce. Therefore, a 'softening factor' as in Y. Romano et al. (2019); Sesia and Candès (2020) was adopted. As in previous chapter, the value was fixed at $w = 2$.

   (1) Neural networks: A neural network with two (or three if the median is used as a point estimate) outputs is trained using the pinball loss (Y. Romano et al., 2019).

   (2) Random forests: Instead of taking the mean of samples in the

leaf nodes of the trees, all the samples are retained to construct a full distribution function (Meinshausen, 2006). The default implementation from Nelson (2023) was used.

2. Mean-variance estimation:  As for quantile regression, a neural network with two outputs is used.  Up to a logarithmic transformation for positivity and numerical stability, these yield the mean and the variance.  The model is trained by optimizing the (log-)likelihood of a normal distribution with the given mean and variance.

3. (Dropout) Ensemble estimation:  Using dropout at test time allows to create an ensemble without having to actually (re)train a network multiple times.  This considerably speeds up the training process.  In this study, the mean-variance approach from Kendall and Gal (2017) with an ensemble of 50 Monte Carlo samples was used, i.e. 50 mean-variance models were obtained.  The dropout ensemble constitutes a mixture of Gaussian distributions with parameters $\{(\hat{\mu}_i, \hat{\sigma}_i)\}_{i \leq 50}$.  Aggregating these estimates gives a total mean and total variance as in Eqs. (A.96) and (A.98).  The problem of finding the quantiles is simplified by approximating the mixture model by a Gaussian distribution with the same mean and variance.  This allows to find the prediction intervals in the standard way.

4. Normalizing flows: As explained in Section 4.7, these model a bijective transformation between the data-generating distribution and a normal distribution by optimizing the (log-)likelihood.  Since inference for a normal distribution is straightforward and the transformation is bijective, inference for the data-generating distribution also becomes manageable.  For the experiments, the unconstrained monotonic neural networks from Wehenkel and Louppe (2019) are used.

All experimental results were obtained by evaluating the models on 10 different train/test-splits, where the test set contained 20% of the data. To obtain a calibration set, the training set was further split in half.  The significance level was fixed at $\alpha = 0.1$ as in the previous chapter.

### 4.8.3   Results

The coverage of the prediction intervals of different models on the data sets from Table 3.1 are shown in Figs. 4.11 and 4.12. (For figures of the conditional interval widths and marginal performance, both in terms of coverage and efficiency, for all data sets, see Figs. 4.13 to 4.18.) It should be immediately clear that the Mondrian conformal prediction methods, indicated by the labels 'PointCP', 'IntCP' and 'NCP', have the most stable and desirable behaviour. For these methods, the coverage is centered around the target confidence level of 90% (corresponding to the predetermined significance level of $\alpha = 0.1$), whereas for the baseline and marginal conformal prediction methods, indicated by the label 'Base' and the prefixes 'm-', the coverage can fluctuate heavily, lying either far above or far below 90%. Of course, this is entirely to be expected, since only the Mondrian approach satisfies the strict conditional coverage guarantees of Corollary 4.6.

Another feature of these figures is that the deviations of the marginal models are not always in the same direction. Comparing the different subplots, it can be seen that whereas some models show undercoverage on one data set, they exhibit overcoverage on another. The same occurs for the different methods on a single data set. If the true variances were known, as in Fig. 4.8 in the previous section, an in-depth analysis could be made. However, in contrast to estimates of the true response, where we can use metrics such as the MSE or $R^2$ to quantify deviations from the truth, no analogues exist for the residual variance. For higher conditional moments, multiple measurements for the same feature tuple are required to get a good estimate. Any uncertainty-dependent taxonomy function will, therefore, also lead to a wrong decomposition of the instance space. Moreover, there is, in general, no way to determine where the lack of validity stems from bad estimates of the prediction intervals or incorrect taxonomy classes (cf. the distinction between misspecification and contamination in Section 4.6).

## 4.9   Discussion

The work in this chapter was motivated by the surge in interest in conformal prediction and accurate uncertainty quantification over the past few years. As mentioned in the discussion of the previous chapter, many confidence predictors, conformal prediction in particular, are constructed for marginal usage. Although the benefits and validity of these methods have been illustrated on numerous data sets and in various domains, an important aspect remains less understood: the conditional performance. This can be especially important for heteroskedastic problems, where we can condition on (estimates of) the residual variance, so as to make sure the models do not neglect the often underrepresented regions of high uncertainty.

In this work, uncertainty-independence of the resulting nonconformity distribution was studied in both a general setting, leading to the use of pivotal quantities, and in the case of explicit, parametric nonconformity measures. The latter allowed us to derive families of probability distributions for which conditional validity will hold whenever the instance space is subdivided based on the data noise, provided it can be expressed in terms of the chosen parameterization.

One issue that was not covered in this chapter or, at least, ignored throughout it, was the data availability. All of the considerations in this chapter (except those in Section 4.6.2) are contingent on having sufficient data to get accurate parameter estimates or, in light of Theorem 4.9, correctly specified nonconformity measures. So, when we know that only a limited amount of data is available and that we can expect, in advance, that the models might perform subpar, we might want to try something else. This will be the content of the next chapter, where we will start from a Mondrian conformal predictor and try to combine taxonomy classes in smart way to mitigate the lack of data.

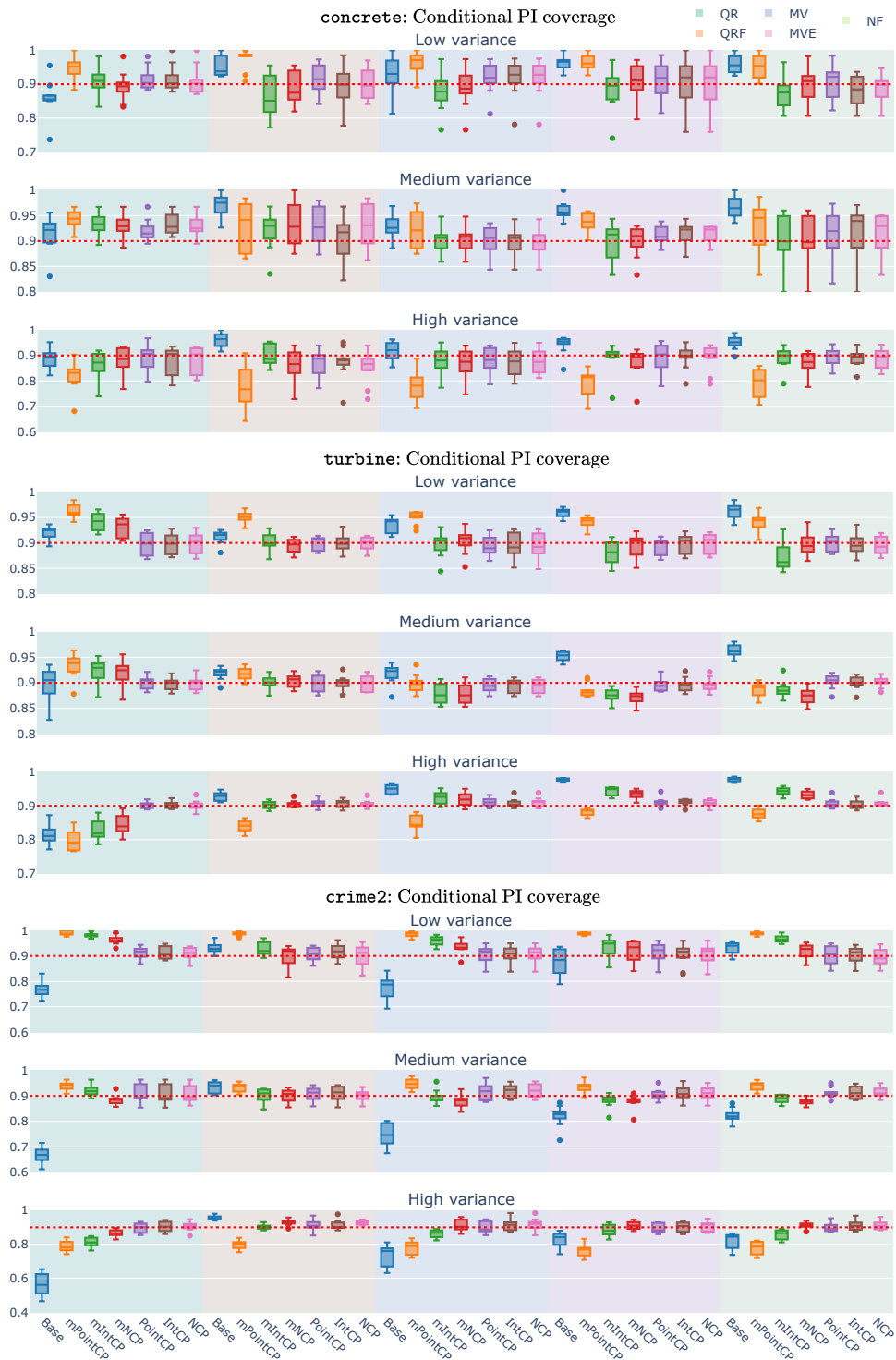

Figure 4.11: Conditional coverage at significance level $\alpha = 0.1$ for the `concrete`, `turbine` and `crime2` data sets. The data is divided in three folds based on equal-frequency binning of the estimated variances. The coloured columns indicate the different models. For every model, a baseline result and six nonconformity measures are shown.

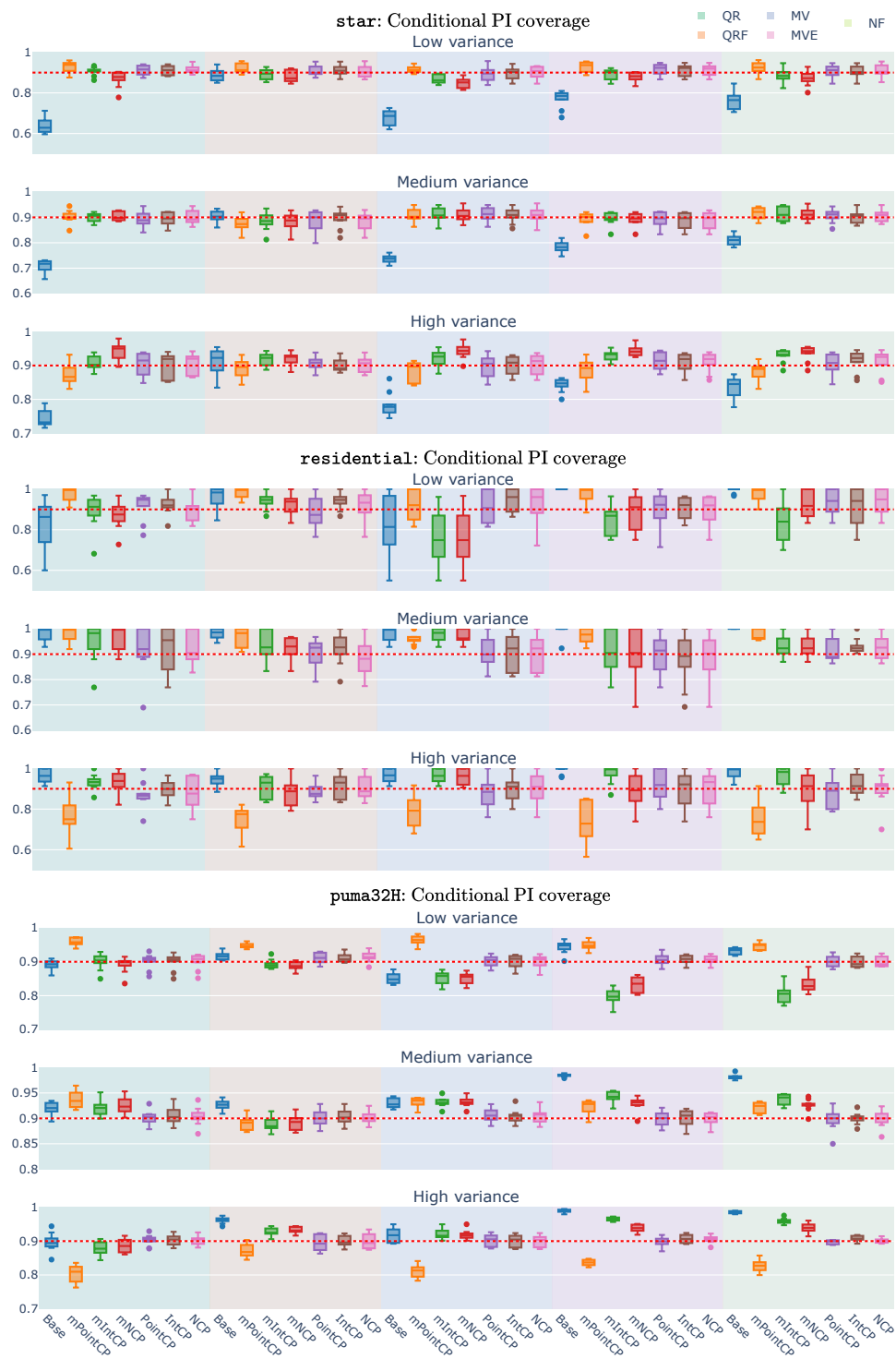

Figure 4.12: Conditional coverage at significance level $\alpha = 0.1$ for the `star`, `residential` and `puma32H` data sets. The data is divided in three folds based on equal-frequency binning of the estimated variances. The coloured columns indicate the different models. For every model, a baseline result and six nonconformity measures are shown.

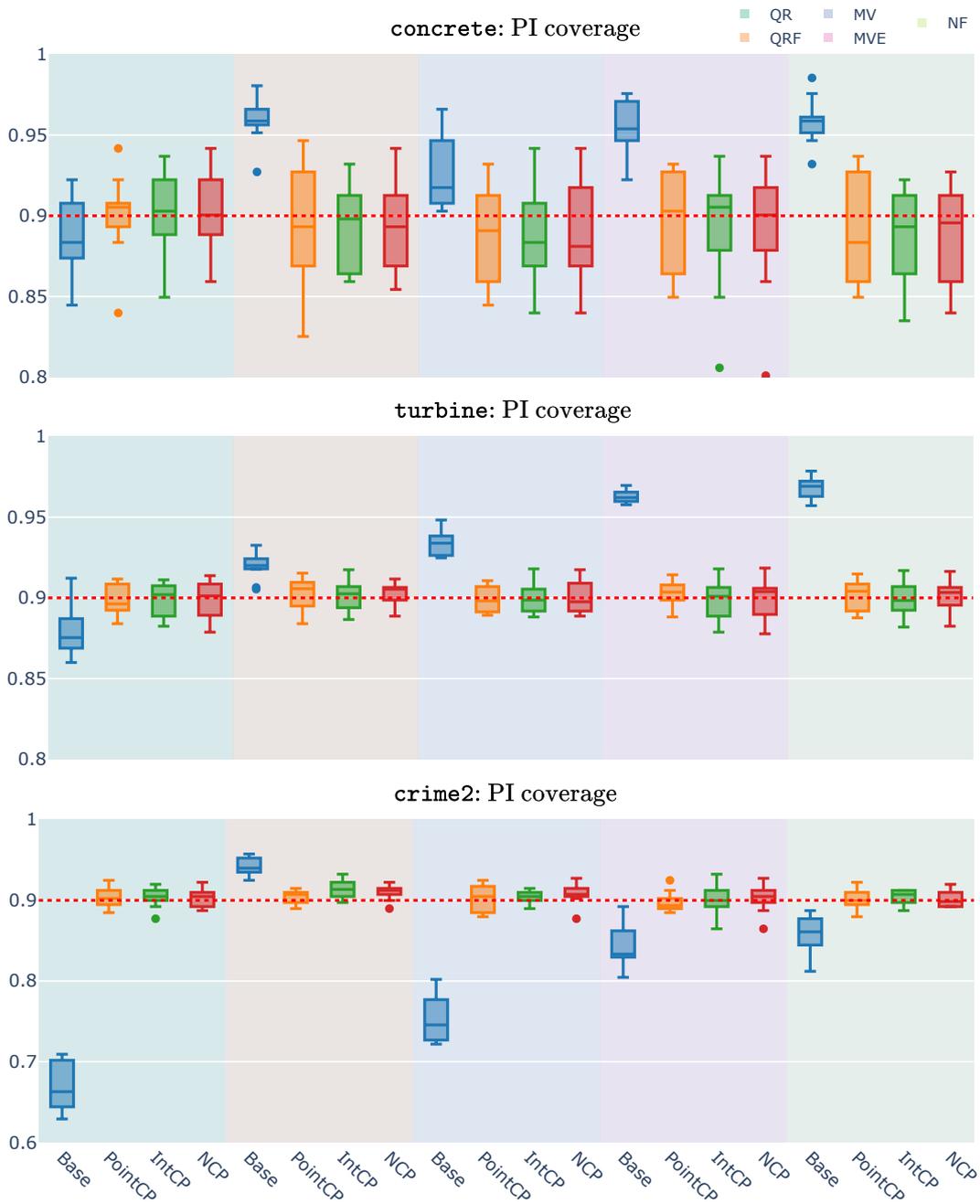

Figure 4.13: Marginal coverage at significance level $\alpha = 0.1$ for the `concrete`, `turbine` and `crime2` data sets. The coloured columns indicate the different models. For every model, a baseline result and three nonconformity measures are shown.

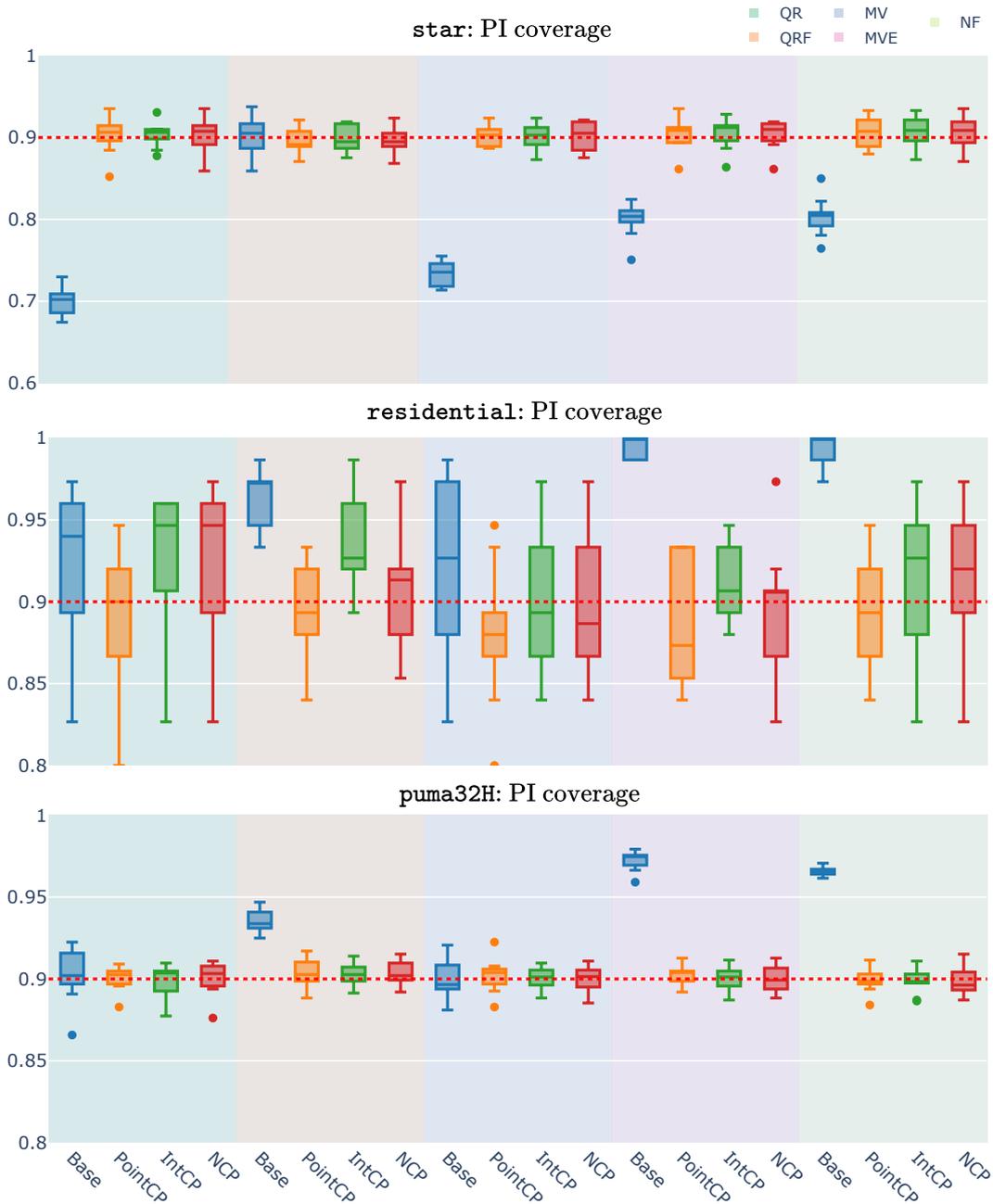

Figure 4.14:  Marginal coverage at significance level $\alpha = 0.1$ for the `star`, `residential` and `puma32H` data sets.  The coloured columns indicate the different models.  For every model, a baseline result and three nonconformity measures are shown.

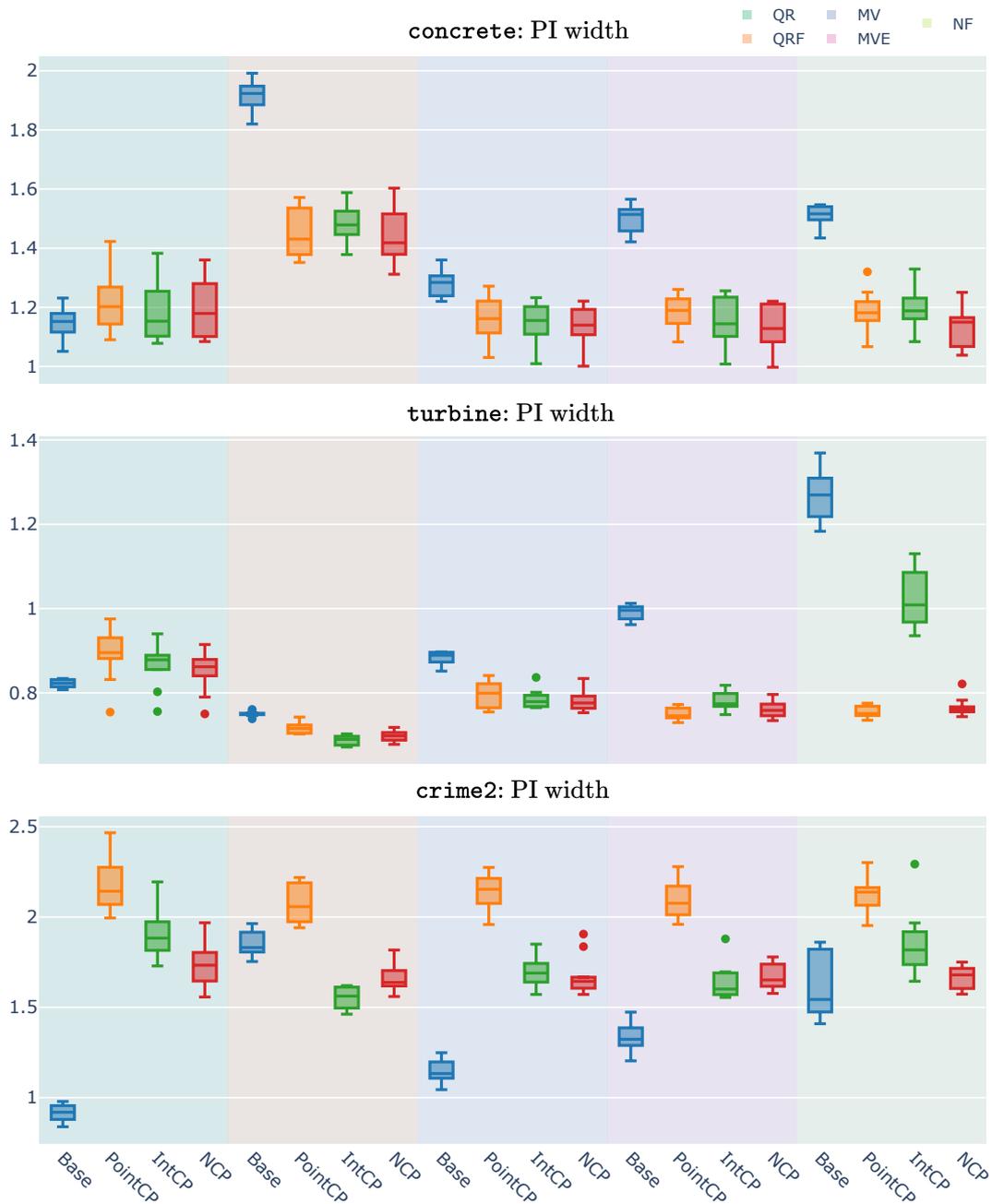

Figure 4.15: Marginal interval widths at significance level $\alpha = 0.1$ for the concrete, turbine and crime2 data sets. The coloured columns indicate the different models. For every model, a baseline result and three nonconformity measures are shown.

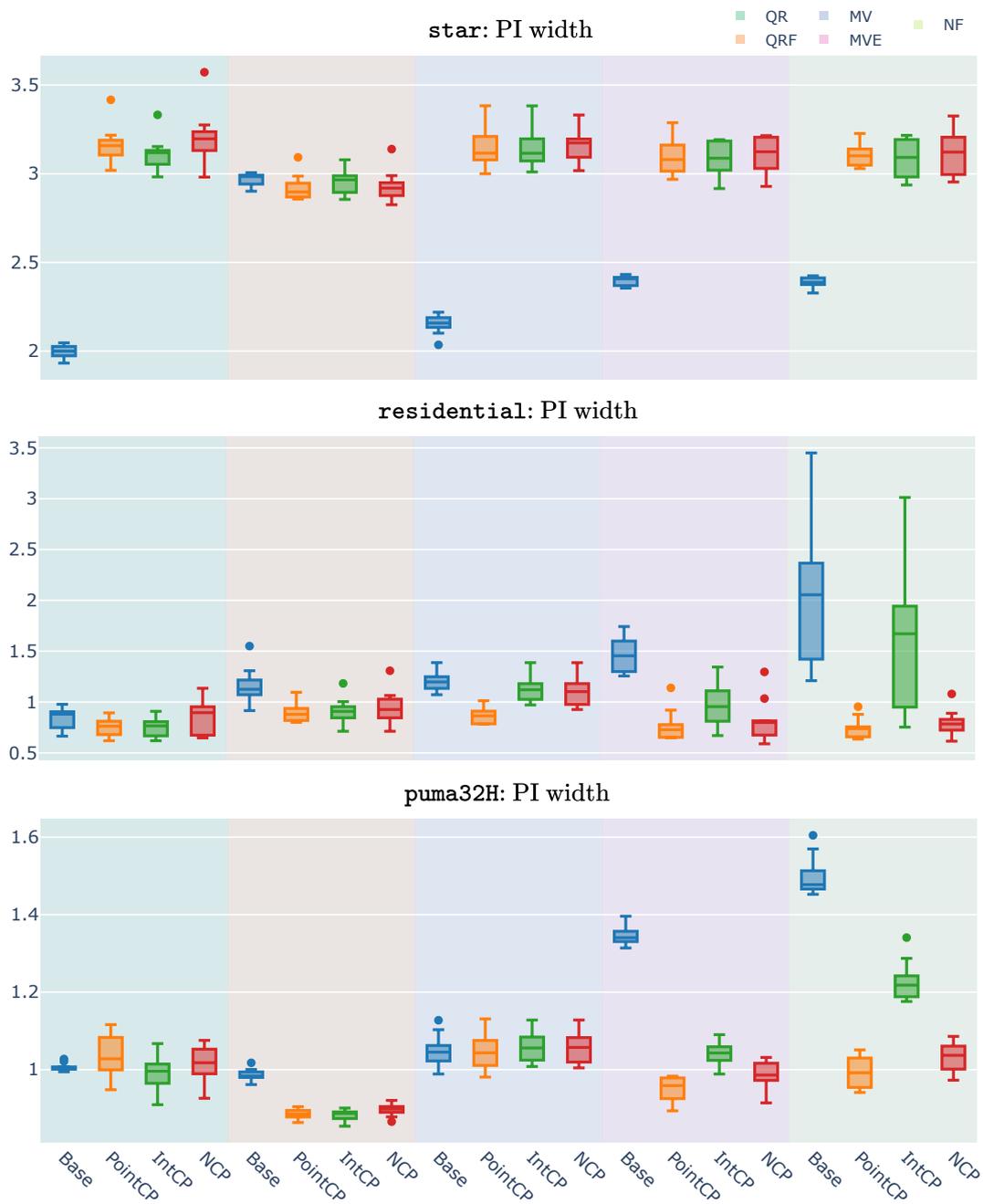

Figure 4.16:  Marginal interval widths at significance level $\alpha = 0.1$ for the star, residential and puma32H data sets.  The coloured columns indicate the different models.  For every model, a baseline result and three nonconformity measures are shown.

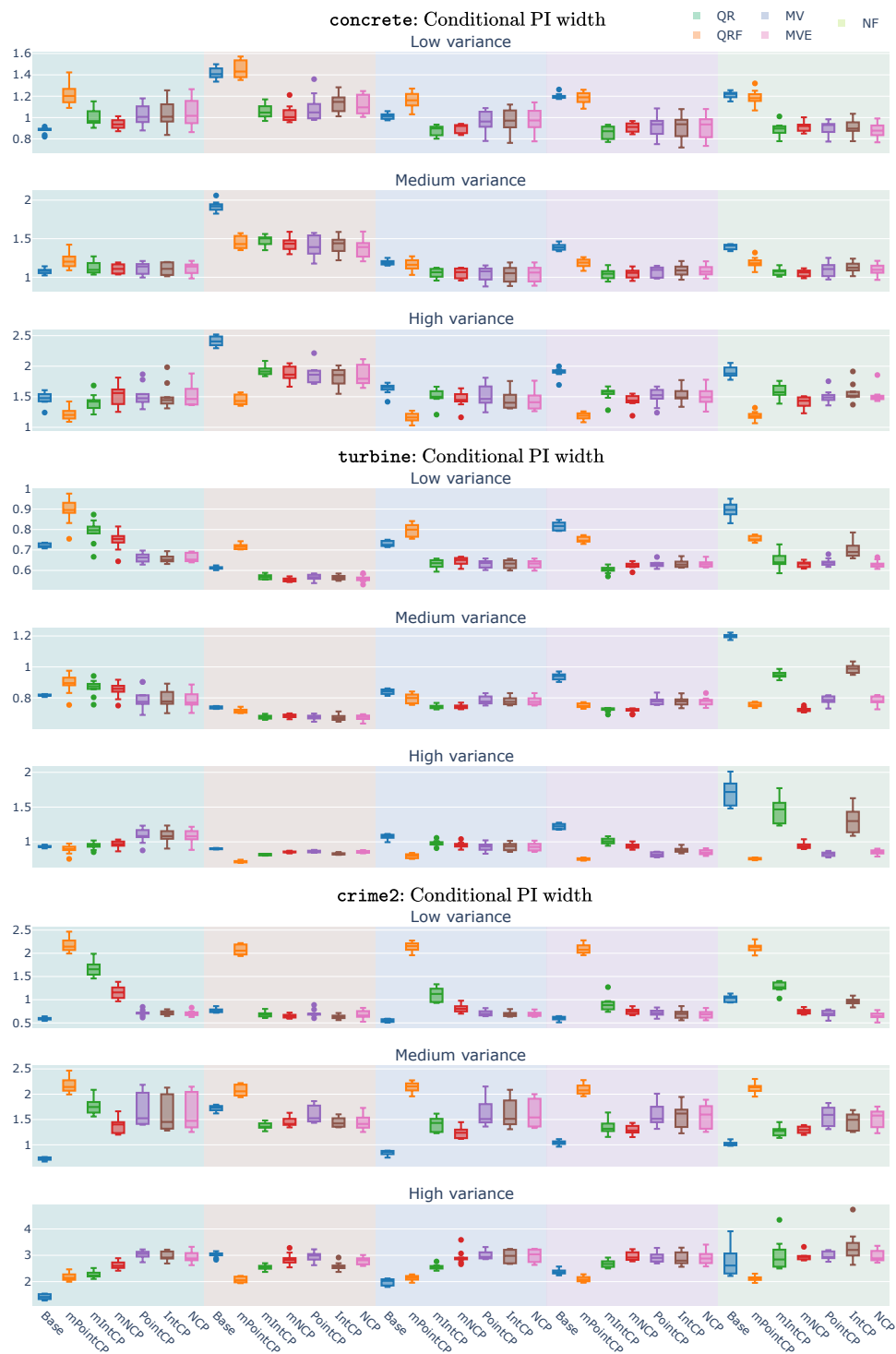

Figure 4.17: Conditional interval widths at significance level $\alpha = 0.1$ for the `concrete`, `turbine` and `crime2` data sets. The data is divided in three folds based on equal-frequency binning of the estimated variance. The coloured columns indicate the different models. For every model, a baseline result and six nonconformity measures are shown.

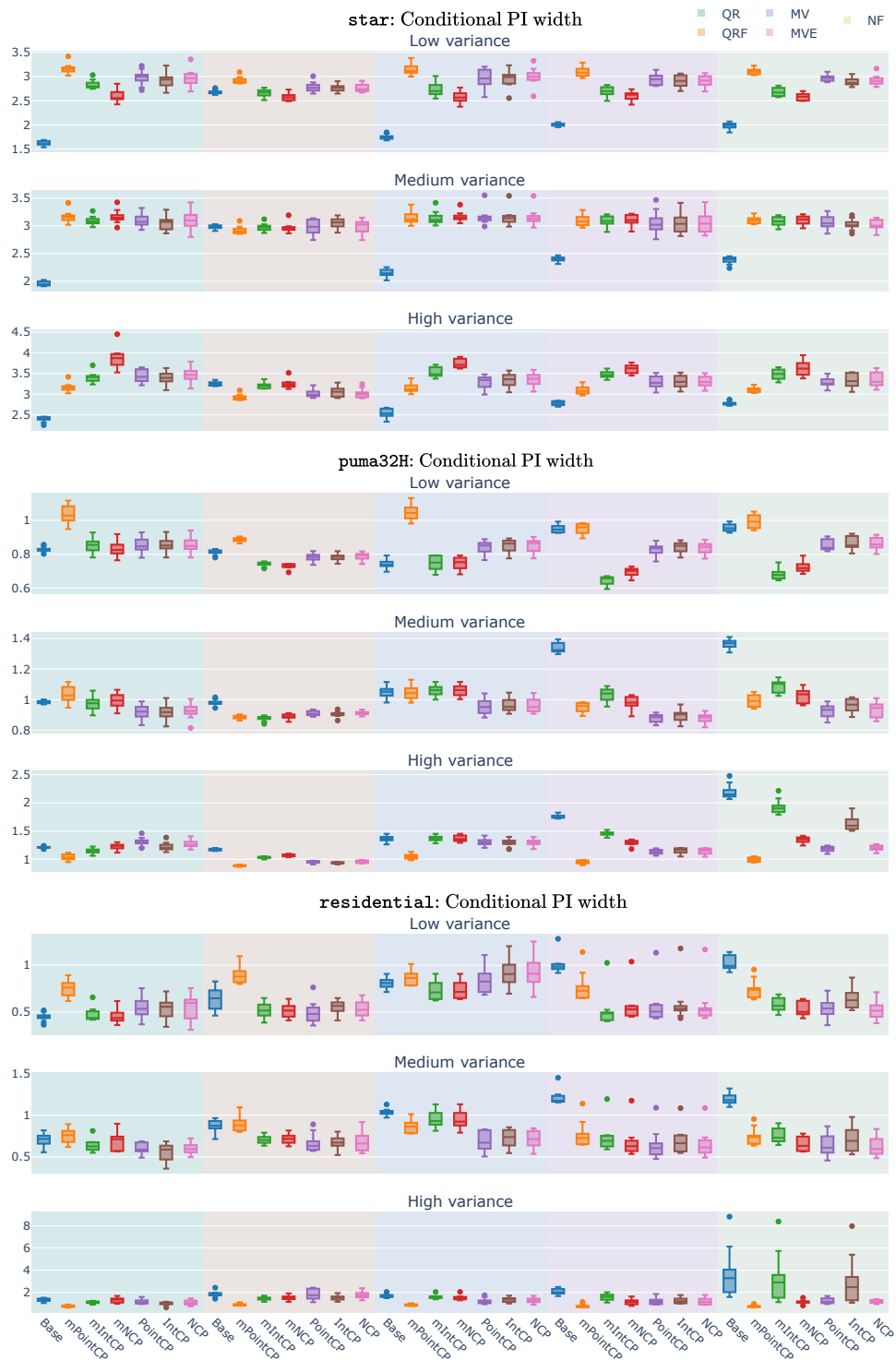

Figure 4.18: Conditional interval widths at significance level $\alpha = 0.1$ for the `star`, `residential` and `puma32H` data sets. The data is divided in three folds based on equal-frequency binning of the estimated variances. The coloured columns indicate the different models. For every model, a baseline result and six nonconformity measures are shown.

# Clusterwise Validity

<div style="text-align: right">5</div>

$$A(X) \cong B(\widehat{X})$$

$$\mathsf{Fuk}(X) \cong D_{\infty}^{b}(\widehat{X})$$

Kontsevich (1994)

## 5.1 Introduction

In Chapter 3, we treated conformal prediction and uncertainty quantification from a marginal point of view. Then, in Chapter 4, we treated the same topics from a conditional point of view, where all taxonomy classes were considered separately. This chapter can be seen as covering the middle ground.

Consider the general case of a conditionally valid conformal predictor $\Gamma^{\alpha} : \mathcal{X} \to 2^{\mathcal{Y}}$, possibly a non-Mondrian one, with respect to a taxonomy function $\kappa : \mathcal{X} \times \mathcal{Y} \to [k]$. By Definition 4.2, this means that

$$\mathsf{Prob}\big(Y \in \Gamma^{\alpha}(X) \mid \kappa(X, Y) = c\big) \geq 1 - \alpha \tag{5.1}$$

for all $c \in [k]$. Recall that this condition of validity holds only when also taking the probability over nuisance parameters (like the calibration data) as explained in Remark 2.5. It does not hold for a fixed calibration set due to Property 2.29. When the calibration set is small, which is often the case in realistic settings (especially with imbalanced data), the empirical coverage of the model might deviate substantially from the nominal confidence level $1 - \alpha$. This problem might even persist when resampling calibration sets in the extreme classification setting (see Section 5.4 below), where the distribution over the taxonomy classes is so skewed that, in 'usual' samples, some classes occur only a few times. Although validity will hold in probability, this will not be the case when taking empirical frequencies, even over many samples. A possible solution, the one followed in this chapter, is given by sacrificing strict conditional validity and clustering 'similar' taxonomy classes.

The clusterwise approach to conformal prediction covers only one part of this chapter. Two other aspects are also examined in this exploratory chapter. On the one hand, we will see how hierarchies over the taxonomy space can be used to facilitate the clustering and how it gives us an additional tool to quantify the quality of prediction sets. These hierarchies structure the taxonomy space in such a way that related classes are grouped together and, by incorporating this information, we can hope to obtain more informative results. On the other hand, we also extend the clusterwise approach, which was originally introduced for classification problems, to multitarget prediction. This latter framework is gaining traction in recent years due to its unifying power and the widespread use of related models.

This chapter is structured as follows. In Section 5.2, the general idea of clusterwise conformal prediction is introduced and some clustering approaches are studied. A first contribution here is the study of the validity of feature-based clustering. Another approach to clustering, making use of hierarchies, is considered in Section 5.3. Since class-conditional validity and clustering is very natural in the setting of classification problems, conformal classification is treated in Section 5.4. The content of this chapter is, however, applicable in a much more general setting and Section 5.5 explains exactly how the preceding sections can be combined. At last, in Section 5.6, an application to extreme classification is presented.

## 5.2   Clustering

### 5.2.1   Clusterwise validity

The idea behind clusterwise conformal prediction is closely related to the results of the previous chapter. Independence Theorem 4.9 roughly stated that when the nonconformity distributions of two taxonomy classes coincide, the classes can be combined and a single conformal predictor can be used. This result is based on the fact that when two data sets are sampled from the same distribution, they can be aggregated and the union of the data sets is again a sample from that same distribution. However, we end up with a larger data set and, hence, the empirical distribution, the one that matters for the actual conformal prediction algorithm, will be a better approximation to the true distribution as exemplified by the following property.

**Property 5.1.** Consider two data sets $A, B$ sampled from the same distribution $P \in \mathbb{P}(\mathcal{X})$. The empirical distribution $P_{\mathrm{emp}}^{A \cup B} \in \mathbb{P}(\mathcal{X})$ is a better approximation to $P$, with respect to the Kolmogorov–Smirnov distance (Example A.53), than the worst of $P_{\mathrm{emp}}^A$ and $P_{\mathrm{emp}}^B$.

*Proof.* This is a direct application of Corollary A.59. (This result actually implies that the statement holds for any distance measure induced by a norm on $\mathsf{Meas}^{\pm}(\mathcal{X})$, the space of *signed measures* [a].)  $\square$

───────

[a] The reason for working with signed measures instead of probability distributions is that the latter are not closed under taking differences, the operation that turns norms into distances.

The same approach will be followed here. When the available calibration data for a given taxonomy class is not large enough, we can try to cluster different classes in a smart manner. The following extension of Corollary 4.6 quantifies the worst case loss in validity due to clustering classes with different nonconformity distributions.

**Theorem 5.2 (Clusterwise validity).** Let $\kappa : \mathcal{X} \times \mathcal{Y} \to [k]$ be a taxonomy function and consider a cluster of taxonomy classes $C \subseteq [k]$. If $\Gamma^\alpha : \mathcal{X} \to 2^{\mathcal{Y}}$ is a conformal predictor at significance level $\alpha \in [0, 1]$, then

$$\mathsf{Prob}\big(Y \in \Gamma^\alpha(X \mid V) \mid \kappa(X, Y) = c, V \in \kappa^{-1}(C)^*\big)$$
$$\geq 1 - \alpha - \max_{c' \in C} d_{\mathrm{KS}}(P_{A|c}, P_{A|c'}), \tag{5.2}$$

where $P_{A|c} := A_* P_{X,Y}\big(\cdot \mid \kappa(X, Y) = c\big)$ and $d_{\mathrm{KS}} : \mathbb{P}(\mathbb{R}) \times \mathbb{P}(\mathbb{R}) \to \mathbb{R}$ is the Kolmogorov–Smirnov distance (Example A.53).

*Proof.* See Proposition 3 in Ding, Angelopoulos, Bates, Jordan, and Tibshirani (2023). This is also an application of Corollary A.59 on convex combinations and distances.  $\square$

This theorem says that if we condition on a specific taxonomy class, but use calibration sets that also contain data from other classes in the cluster, the coverage is at most decreased by the maximal (KS) distance between the members of the cluster. In other words, the more distinct the cluster members are, the worse the coverage can be.

**Remark 5.3 (Total variation distance).** The original preprint of Ding et al. (2023) stated the preceding theorem in terms of the total variation (TV) distance $d_{\mathrm{TV}} : \mathbb{P}(\mathbb{R}) \times \mathbb{P}(\mathbb{R}) \to \mathbb{R}$ (Example A.54). It is not hard to see that the KS distance gives a lower bound on the TV distance and, hence, the former leads to a crisper bound on the conditional coverage.

However, for analytical purposes, the total variation distance is actually slightly easier to use due to Property 5.4 shown below. In the remainder of this section, if the exact choice of distance is irrelevant, the term 'statistical distance' will be used.

The presence of the supremum in the statistical distances might lead us to think that it can be hard to manipulate these expressions, especially for uncountable sets. However, the following property says that the TV distance admits a much simpler formulation.

**Property 5.4 (Scheffé's lemma[1]).** Consider a measurable space $(\mathcal{X}, \Sigma)$ together with two probability measures $P, Q \in \mathbb{P}(\mathcal{X})$ that admit density functions $f, g \in L^1(\mu)$ with respect to some ($\sigma$-finite[a]) base measure $\mu \in \mathbb{P}(\mathcal{X})$. The total variation distance can be expressed as an $L^1$-distance (A.77):

$$d_{\mathrm{TV}}(P, Q) = \frac{1}{2} \int_{\mathcal{X}} |f - g| \, \mathrm{d}\mu. \qquad (5.3)$$

*Proof.* See Lemma 2.1 in Tsybakov (2008). □

[a] Note that this condition is nonrestrictive since $P + Q$ is already a valid choice.

Together with the methods from the previous chapters, we end up with three approaches, each with their own validity guarantee:

1. Marginal conformal prediction: This only gives a marginal validity guarantee, but it will not suffer from a lack of data[2].

2. Classwise/Mondrian conformal prediction: This has the strongest conditional validity guarantees, but it will suffer severely from any lack of data.

[1] The name stems from a convergence result by Scheffé (1947). However, it is not used in most proofs of this result.

[2] We ignore the extreme case where there is in general barely any data, since this would also prevent the use of ordinary learning methods.

3. Clusterwise conformal prediction: This is the middle ground between vanilla (marginal) and Mondrian conformal prediction. With well-chosen clusters, it can achieve near-optimal conditional validity.

## 5.2.2 Score-based clustering

In Ding et al. (2023), the clustering is performed directly in the space of probability measures by comparing quantiles of the empirical nonconformity distributions (see Algorithm 12). Given a calibration set $\mathcal{V} \in (\mathcal{X} \times \mathcal{Y})^*$ and a taxonomy function $\kappa : \mathcal{X} \times \mathcal{Y} \to [k]$, we first construct the conditional nonconformity sets $\mathcal{A}_c := \{(x, y) \in \mathcal{V} \mid \kappa(x, y) = c\}$ for all $c \in [k]$ as for Mondrian conformal prediction. Then, we choose $l \in \mathbb{N}_0$ quantile levels $\{\alpha_i\}_{i \leq l} \subset ]0, 1[$ and, for every $c \in [k]$, we compute the sample $\alpha_i$-quantiles of $\mathcal{A}_c$ to obtain an $l$-dimensional embedding $\xi_c \in \mathbb{R}^l$. This gives a set of vectors $\{\xi_c\}_{c \in [k]} \subset \mathbb{R}^l$ on which we can apply a clustering algorithm such as *k-means clustering* (Goodfellow et al., 2016) or *hierarchical clustering* (Nielsen, 2016).

The argument for choosing this clustering approach is straightforward. By clustering the quantile embeddings, classes with similar distributions are matched together and, hence, the TV distance is minimized. (This is very similar in spirit to the content of Section 4.6.3.) However, although this approach might be superior from the point of view of conformal prediction, other approaches can also be useful in the bigger picture of data analysis. These will be treated in the next section.

## 5.2.3 Feature-based clustering

In the previous section, we (metrically) clustered the conditional nonconformity distributions as to obtain an optimal result in Theorem 5.2, since it is exactly the distance between these distributions that modifies the standard validity result. However, two vastly different phenomena might have similar distributions, examples abound, and, hence, the resulting clusters do not necessarily tell us much about the underlying process. They do not necessarily carry any semantic information.

Moreover, as also noted in Ding et al. (2023), the score-based clustering approach only leads to the theoretically optimal result (in terms of conformal prediction) when used in conjunction with an oracle model, i.e. when we

**Algorithm 12:** $A$-Clusterwise Conformal Prediction

**Input**   : Significance level $\alpha \in [0, 1]$, nonconformity measure
$A : \mathcal{X} \times \mathcal{Y} \to \mathbb{R}$, taxonomy function $\kappa : \mathcal{X} \times \mathcal{Y} \to [k]$, training
set $\mathcal{T} \in (\mathcal{X} \times \mathcal{Y})^*$, calibration set $\mathcal{V} \in (\mathcal{X} \times \mathcal{Y})^*$, quantile
levels $\{\rho_i\}_{i \leq l} \subset ]0, 1[$ and clustering algorithm $\mathcal{S} : \mathbb{R}^l \to [m]$

**Output:** Conformal predictor $\Gamma^\alpha$

1 (Optional) Train the underlying model of $A$ on $\mathcal{T}$

2 **foreach** $c \in [k]$ **do**
3   Initialize an empty list $\mathcal{A}_c$
4   Select the data set $\mathcal{D}_c \leftarrow \{(x, y) \in \mathcal{D} \mid \kappa(x, y) = c\}$
5   **foreach** $(x_i, y_i) \in \mathcal{D}_c$ **do**
6     Calculate the nonconformity score $A_i \leftarrow A(x_i, y_i)$
7     Add the score $A_i$ to $\mathcal{A}_c$
8   **end**
9   Calculate the embedding $\xi_c$ as the vector of $\{\rho_i\}_{i \leq l}$-quantiles of $\mathcal{A}_c$
10 **end**

11 Cluster the embeddings $\{\xi_c\}_{c \in [k]}$ using $\mathcal{S}$
12 Construct the clusterwise calibration sets
   $\mathcal{A}_\omega \leftarrow \bigcup_{c \in [k]} \{\mathcal{A}_c \mid \mathcal{S}(\xi_c) = \omega\}$

13 **procedure** $\Gamma^\alpha(x : \mathcal{X})$
14   Initialize an empty list $\mathcal{R}$
15   **foreach** $y \in \mathcal{Y}$ **do**
16     Determine the cluster $\omega \leftarrow \mathcal{S}(\xi_{\kappa(x,y)})$
17     Determine the critical score $a^* \leftarrow q_{(1-\alpha)(1+1/|\mathcal{A}_\omega|)}(\mathcal{A}_\omega)$
18     **if** $A(x, y) \leq a^*$ **then**
19       Add $y$ to $\mathcal{R}$
20     **end**
21   **end**
22   **return** $\mathcal{R}$

23 **return** $\Gamma^\alpha$

have access to the ground truth. When we only have access to finite data samples, we have to work with the empirical distributions, which might behave completely differently. In the asymptotic limit (and given some general assumptions), we can resort to consistency results such as Property A.45, but asymptotic results are not always very useful in practice as reflected in this quote by Le Cam (1986):[3]

> From time to time results are stated as limit theorems obtainable as something called $n$ "tends to infinity." [...] Indeed, limit theorems "as n tends to infinity" are logically devoid of content about what happens at any particular $n$. All they can do is suggest certain approaches whose performance must then be checked on the case at hand. Unfortunately the approximation bounds we could get were too often too crude and cumbersome to be of any practical use. Thus we have let n tend to infinity, but we would urge the reader to think of the material in approximation terms [...]

In many situations, however, the natural taxonomy classes are induced by some underlying structure or process for which side information is provided (an example will be studied in Section 5.5). When possible, it can be useful to leverage this information to construct an embedding, similar in fashion to the quantile embeddings from the previous section, and cluster the classes based on these embeddings (again using algorithms such as *k-means clustering* or *hierarchical clustering*). Assuming this gives reasonable results, this approach would be preferable over clustering in the nonconformity space, since the side information does carry semantic information.

One situation where using the functional, feature-based approach should also lead to (near) optimal results is where the clustering of feature embeddings is equivalent to the clustering of nonconformity distributions. This amounts to turning Theorem 5.2 into a Lipschitz condition (Definition A.63). As long as the assignment $\mathcal{X} \to \mathbb{P}(\mathbb{R}) : x \mapsto P_{A|x}$ is Lipschitz-continuous with respect to the statistical distance, clusterwise validity for a feature-based clustering algorithm will be near optimal. Lipschitz continuity of the former can be ensured as follows.

**Property 5.5.** Let $(\Theta, d)$ be a metric space and consider a parametric family of distributions $\left\{ P_\theta \in \mathbb{P}(\mathcal{X}) \mid \theta \in \Theta \right\}$ on a measurable space $(\mathcal{X}, \Sigma)$. The family is Lipschitz-continuous with respect to the total variation distance if there exists a family of functions $\left\{ f_\theta : \mathcal{X} \to \mathbb{R} \mid \theta \in \Theta \right\}$ satisfying the following conditions:

---

[3] This applies to all consistency statements in this work.

1. **Density**: There exists a measure $\mu$ on $(\mathcal{X}, \Sigma)$ such that

$$\frac{\mathrm{d}P_\theta}{\mathrm{d}\mu} = f_\theta \tag{5.4}$$

for all $\theta \in \Theta$, where the derivative is the Radon–Nikodym derivative (Theorem A.29).

2. **Lipschitz continuity**: The functions $f_\theta$ are Lipschitz-continuous with respect to the $L^1$-distance induced by $\mu$:

$$\int_{\mathcal{X}} \left| f_\theta(x) - f_{\theta'}(x) \right| \mathrm{d}\mu \le \zeta \|\theta - \theta'\| \tag{5.5}$$

for some $\zeta \in \mathbb{R}^+$ and for all $\theta, \theta' \in \Theta$.

*Proof*. This is simply an application of Scheffé's lemma 5.4. $\qquad\square$

For the setting of this chapter, the parametric family is given by the conditional probability distributions $P_{A|c}$ for $\mathcal{X} = [k]$, modelled as Markov kernels (Definition A.38).

**Corollary** 5.6. Since the TV distance is bounded from below by the KS distance by Property A.55, the theorem also holds for the KS distance.

At this point, we have Lipschitz results for the data-generating process. However, one final piece in the puzzle is still missing, since Theorem 5.2 is stated in terms of the nonconformity distribution and not the data-generating distribution. The missing link is given by the following property.

**Property** 5.7.  Consider a parametric family of probability distributions $\{P_\theta \in \mathbb{P}(\mathcal{X}) \mid \theta \in \Theta\}$. If $(\Theta, d)$ is a metric space such that the assignment $\theta \mapsto P_\theta$ is Lipschitz continuous with respect to the TV distance on $P(\mathcal{X})$, then, for any measurable function $f : \mathcal{X} \to \mathcal{X}'$, the assignment $\theta \mapsto f_* P_\theta$ of pushforward measures is also Lipschitz continuous, with the same Lipschitz constant, with respect to the statistical distance.

*Proof*. The statement is a simple consequence of the data processing inequality of the TV distance:

$$d_{\text{TV}}(f_* P_\theta, f_* P_{\theta'}) = \sup_{B \in \Sigma_{\mathcal{X}'}} |f_* P_\theta(B) - f_* P_{\theta'}(B)|$$

$$= \sup_{B \in \Sigma_{\mathcal{X}'}} |P_\theta(f^{-1}(B)) - P_{\theta'}(f^{-1}(B))|$$

$$\leq \sup_{B \in \Sigma_{\mathcal{X}}} |P_\theta(B) - P_{\theta'}(B)|$$

$$= d_{\text{TV}}(P_\theta, P_{\theta'}),$$

where $\Sigma_{\mathcal{X}}, \Sigma_{\mathcal{X}'}$ are the $\sigma$-algebras of $\mathcal{X}$ and $\mathcal{X}'$, respectively. Since $d_{\text{KS}} \leq d_{\text{TV}}$, the result also holds for the KS distance. $\qquad \square$

If we take the function $f$ in the preceding theorem to be the nonconformity measure, we can extend the clusterwise validity theorem to (metric) clustering in the instance space.

**Corollary** 5.8. Let the feature space be a metric space $(\mathcal{X}, d)$ and consider an associated distance-based clustering method $\kappa : \mathcal{X} \rightarrow [k]$. If $P_{A|\cdot} : \mathcal{X} \rightarrow \mathbb{P}(\mathbb{R})$, where $A : \mathcal{X} \times \mathcal{Y} \rightarrow \mathbb{R}$ is a nonconformity measure, is Lipschitz-continuous as a map $(\mathcal{X}, d') \rightarrow (\mathbb{P}(\mathbb{R}), d_{\text{TV}})$ for an equivalent metric $d'$ (Definition A.56), then Theorem 5.2 holds with respect to clustering by $\kappa$:

$$\text{Prob}(Y \in \Gamma^\alpha(X \mid V) \mid X = x, V \in \kappa^{-1}(c)^*)$$
$$\geq 1 - \alpha - \zeta \max_{x' \in \kappa^{-1}(c)} d'(x, x') \qquad (5.6)$$

for all $x \in \mathcal{X}$, where $c := \kappa(x)$ and $\zeta \in \mathbb{R}^+$ is the Lipschitz constant of $P_{A|\cdot} : \mathcal{X} \rightarrow \mathbb{P}(\mathbb{R})$.

Although these results might seem interesting from a theoretical point of view, this does, however, not mean that they are necessarily useful in practice. The following examples show that the assumptions are at least satisfied for some simple situations. [4]

**Example** 5.9 (**Uniform distributions**). Consider the following family of

---

[4] It can be shown that for the second example, that of shifted normal distributions, the TV and KS distance coincide.

uniform distributions

$$f(y \mid x) = \frac{\mathbb{1}_{[a(x),a(x)+w(x)]}(y)}{w(x)} \tag{5.7}$$

for some location function $a : \mathcal{X} \to \mathbb{R}$ and width function $w : \mathcal{X} \to \mathbb{R}^+$. After some tedious calculations, it can be shown that the total variation distance is given by the following expression (without loss of generality, we can assume that $a(x) \leq a(x')$):

$$d_{\mathrm{TV}}\big(P(Y \mid x), P(Y \mid x')\big) \tag{5.8}$$

$$= \begin{cases} 1 & \text{if } a(x') - a(x) \geq w(x), \\ \dfrac{w(x) - w(x')}{w(x)} & \text{if } a(x) + w(x) \geq a(x') + w(x'), \\ \dfrac{\big(a(x') - a(x)\big) + \big(w(x') - w(x)\big)}{w(x')} & \text{if } w(x') \geq w(x), \\ \dfrac{a(x') - a(x)}{w(x)} & \text{otherwise.} \end{cases}$$

Assuming that the width function $w : \mathcal{X} \to \mathbb{R}^+$ is bounded from below by some constant $W \in \mathbb{R}^+$, the total variation distance is bounded as follows:

$$d_{\mathrm{TV}}\big(P(\cdot \mid x), P(\cdot \mid x')\big) \leq \frac{|a(x') - a(x)| + |w(x') - w(x)|}{W}. \tag{5.9}$$

It follows that if the map $x \mapsto \big(a(x), w(x)\big)$ is Lipschitz-continuous with respect to the $\ell^1$-distance on $\mathbb{R}^2$, the map $x \mapsto \mathcal{U}\big([a(x), a(x) + w(x)]\big)$ is also Lipschitz-continuous.

**Example** 5.10 (**Shifted normal distributions**). Consider the parametric family of homoskedastic normal distributions

$$f(y \mid x) = \frac{1}{\sqrt{2\pi}\sigma} \exp\left(-\frac{(y - \mu(x))^2}{2\sigma^2}\right). \tag{5.10}$$

As a function $g : \mathbb{R} \to \mathbb{R}^+$ of the mean $\mu(x)$, the $L^1$-distance

$$\int_{\mathbb{R}} \frac{1}{\sqrt{2\pi}\sigma}\left|\exp\left(-\frac{y^2}{2\sigma^2}\right) - \exp\left(-\frac{(y - \mu(x))^2}{2\sigma^2}\right)\right| \mathrm{d}y \tag{5.11}$$

is differentiable and, by Property A.65, if it has a bounded derivative, it is also Lipschitz-continuous with the least upper bound of the derivative as Lipschitz constant. The derivative can be shown to be equal to

$$\frac{\mathrm{d}g}{\mathrm{d}\mu} = \int_{\mathbb{R}} \frac{1}{\sqrt{2\pi}\sigma} \operatorname{sgn}\left[\exp\left(-\frac{y^2}{2\sigma^2}\right) - \exp\left(-\frac{(y-\mu)^2}{2\sigma^2}\right)\right] \tag{5.12}$$
$$\frac{\mu - y}{\sigma^2} \exp\left(-\frac{(y-\mu)^2}{2\sigma^2}\right) \mathrm{d}y\,.$$

To resolve the sign function, we have to solve the equation

$$0 = \exp\left(-\frac{y^2}{2\sigma^2}\right) - \exp\left(-\frac{(y-\mu)^2}{2\sigma^2}\right)$$
$$\iff 0 = -\frac{y^2}{2\sigma^2} + \frac{(y-\mu)^2}{2\sigma^2}$$
$$\iff 0 = y\mu - \mu^2/2 \tag{5.13}$$

and, hence, the sign flips at $y = \mu/2$. This allows to express the derivative as

$$\frac{\mathrm{d}g}{\mathrm{d}\mu} = \int_{-\infty}^{\mu/2} \frac{1}{\sqrt{2\pi}\sigma} \frac{\mu - y}{\sigma^2} \exp\left(-\frac{(y-\mu)^2}{2\sigma^2}\right) \mathrm{d}y$$
$$- \int_{\mu/2}^{+\infty} \frac{1}{\sqrt{2\pi}\sigma} \frac{\mu - y}{\sigma^2} \exp\left(-\frac{(y-\mu)^2}{2\sigma^2}\right) \mathrm{d}y$$
$$= \frac{2}{\sqrt{2\pi}\sigma} \exp\left(-\frac{\mu^2}{8\sigma^2}\right)$$
$$\leq \frac{2}{\sqrt{2\pi}\sigma}\,. \tag{5.14}$$

If the mean function $\mu : \mathcal{X} \to \mathbb{R}$ is Lipschitz-continuous with constant $\kappa \in \mathbb{R}^+$, Property A.64 implies that the map $x \mapsto \mathcal{N}(\mu(x), \sigma^2)$ is Lipschitz-continuous with constant $2\kappa/\sqrt{2\pi}\sigma$.

**Example** 5.11 (**Centered normal distributions**). If the standard deviation (or variance) is varied instead of the mean, a similar result can be obtained. To this end, consider the following family of centered normal distributions:

$$f(y \mid x) = \frac{1}{\sqrt{2\pi}(\sigma + \theta(x))} \exp\left(-\frac{y^2}{2(\sigma + \theta(x))^2}\right). \tag{5.15}$$

In this case, Eq. (5.12) becomes (assuming $\theta > 0$ without loss of gener-

ality):

$$\frac{\mathrm{d}g}{\mathrm{d}\theta} = \int_{\mathbb{R}} \mathrm{sgn} \left[ \frac{1}{\sqrt{2\pi}\sigma} \exp\left(-\frac{y^2}{2\sigma^2}\right) - \frac{1}{\sqrt{2\pi}(\sigma+\theta)} \exp\left(-\frac{y^2}{2(\sigma+\theta)^2}\right) \right]$$
$$\left[ \frac{1}{\sqrt{2\pi}(\sigma+\theta)^2} \exp\left(-\frac{y^2}{2(\sigma+\theta)^2}\right) \right.$$
$$\left. - \frac{y^2}{\sqrt{2\pi}(\sigma+\theta)^4} \exp\left(-\frac{y^2}{2(\sigma+\theta)^2}\right) \right] \mathrm{d}y. \quad (5.16)$$

The main issue now is resolving the sign function. In stark contrast to the previous example, the sign does not change at a single point, i.e. the zeros are not given by a linear equation. It can be shown that the zeros are of the form

$$y_{\pm} = \pm \frac{\sqrt{2}\sigma(\sigma+\theta)}{\sqrt{\theta(2\sigma+\theta)}} \sqrt{\ln\frac{\sigma+\theta}{\sigma}}. \quad (5.17)$$

From the shape of normal distributions, we also know that the higher the standard deviation, the lower the density at $y = 0$, so the sign will be positive in between $y_-$ and $y_+$. Using this fact, the derivative can be rewritten as follows:

$$\frac{\mathrm{d}g}{\mathrm{d}\theta} = \int_{\mathbb{R}} \frac{1}{\sqrt{2\pi}(\sigma+\theta)^2} \exp\left(-\frac{y^2}{2(\sigma+\theta)^2}\right) \mathrm{d}y$$
$$- 2 \int_{-x_+}^{x_+} \frac{1}{\sqrt{2\pi}(\sigma+\theta)^2} \exp\left(-\frac{y^2}{2(\sigma+\theta)^2}\right) \mathrm{d}y$$
$$- \int_{\mathbb{R}} \frac{y^2}{\sqrt{2\pi}(\sigma+\theta)^4} \exp\left(-\frac{y^2}{2(\sigma+\theta)^2}\right) \mathrm{d}y$$
$$+ 2 \int_{-x_+}^{x_+} \frac{y^2}{\sqrt{2\pi}(\sigma+\theta)^4} \exp\left(-\frac{y^2}{2(\sigma+\theta)^2}\right) \mathrm{d}y. \quad (5.18)$$

The first term is, up to a factor $(\sigma + \theta)$, simply equal to the probability of the real line (which is 1) and, similarly, the third term is, up to a factor $(\sigma + \theta)^3$, equal to the variance of the normal distribution. Since the variance is $(\sigma + \theta)^2$ and these terms have opposite signs, they cancel out. In a similar way, the second term is, again up to a factor $(\sigma + \theta)$, equal to $\mathrm{erf}\left(\frac{x_+}{\sqrt{2}(\sigma+\theta)}\right)$, where the error function erf was defined in Eq. (2.48). By using integration by parts twice on the last term in Eq. (5.18), two contributions are obtained:

$$2 \int_{-x_+}^{x_+} \frac{y^2}{\sqrt{2\pi}(\sigma + \theta)^4} \exp\left(-\frac{y^2}{2(\sigma + \theta)^2}\right) \mathrm{d}y \tag{5.19}$$

$$= 2\sqrt{\frac{2}{\pi}} \frac{x_+}{(\sigma + \theta)^2} \exp\left(-\frac{x_+^2}{2(\sigma + \theta)^2}\right)$$

$$+ 2 \int_{-x_+}^{x_+} \frac{1}{\sqrt{2\pi}(\sigma + \theta)^2} \exp\left(-\frac{y^2}{2(\sigma + \theta)^2}\right) \mathrm{d}y \,.$$

The second term is again of the form of the error function and cancels out and, hence, the only remaining contribution is:

$$\frac{\mathrm{d}g}{\mathrm{d}\theta} = 2\sqrt{\frac{2}{\pi}} \frac{x_+}{(\sigma + \theta)^2} \exp\left(-\frac{x_+^2}{2(\sigma + \theta)^2}\right) = \frac{4\sigma\left(\frac{\sigma+\theta}{\sigma}\right)^{-\frac{\sigma^2}{\theta(2\sigma+\theta)}} \sqrt{\ln\frac{\sigma+\theta}{\sigma}}}{\sqrt{\pi}(\sigma + \theta)\sqrt{\theta(2\sigma + \theta)}} \,. \tag{5.20}$$

This expression is a bit harder to manipulate than the one in the previous example. Luckily, it can be shown that it is strictly decreasing and, therefore, we simply have to calculate the limit for $\theta \longrightarrow 0$. Using a Taylor approximation of the (natural) logarithm and the limit definition of Euler's number, we are quickly led to the following bound:

$$\frac{\mathrm{d}g}{\mathrm{d}\mu} \leq \sqrt{\frac{2}{\pi e}} \frac{1}{\sigma} \,. \tag{5.21}$$

Consequently, the map $x \mapsto \mathcal{N}\left(0, \sigma(x)^2\right)$ will be Lipschitz-continuous as soon as the function $\sigma : \mathcal{X} \to \mathbb{R}^+$ is bounded from below as in Example 5.9.

The last example is also shown in Fig. 5.1. For normal distributions centered at 0, the figure shows the total variation distance when varying the standard deviation. This figure, the green line in particular, also shows that when the standard deviations of both normal distributions increase linearly, the TV distance is a constant.[5] (This can easily be seen by writing out the $L^1$-distance and noting that the dependence on the parameter $x$ can be removed through a rescaling.)

---

[5] This actually happens whenever they are proportional.

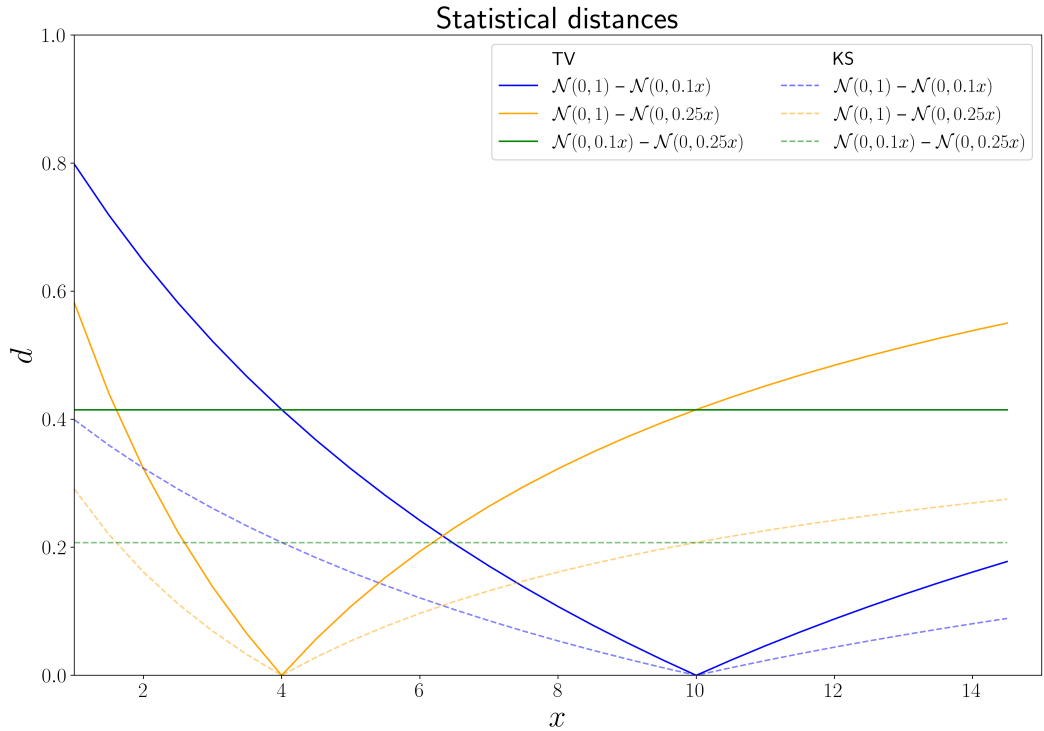

Figure 5.1: Statistical distances between normal distributions with mean 0, but where the standard deviation scales linearly: $\sigma(x) = x$. The solid lines indicate the total variation distance and the dashed lines indicate the Kolmogorov–Smirnov distance.

**Remark** 5.12. Because the statistical distances are themselves bounded from above by 1, the Lipschitz condition is usually only relevant for small parameter differences (small relative to the Lipschitz constant). This might also allow for considering cases where strict Lipschitz continuity is replaced by a weaker notion such as *Hölder continuity* or *moduli of continuity* that do not increase too rapidly (e.g. concavely [a]).

---

[a] Concavity implies subadditivity and, hence, these moduli give rise to equivalent metrics, which are already allowed by Corollary 5.8.

### 5.2.4    Strength of the bounds

To get an idea of how sharp the statistical distance bound is, we can again look at Fig. 5.1. As can be seen there, the distances are only relatively small when the standard deviations are approximately equal. Once the difference in $\sigma$ becomes appreciably large, the distance term will quickly turn the bound in Theorem 5.2 essentially meaningless.

In Figs. 5.2 and 5.3, a cluster (i.e. a mixture) of three normal distributions is considered. In the former, the standard deviation is fixed at $\sigma = 1$ and the means are shifted:

$$\mu_1(x) = 0 \qquad \mu_2(x) = x \qquad \mu_3(x) = 2x \,, \tag{5.22}$$

whereas in the latter, the means are fixed at $\mu = 0$ and the standard deviations are scaled as in Fig. 5.1:[6]

$$\sigma_1(x) = 1 \qquad \sigma_2(x) = 0.1x \qquad \sigma_3(x) = 0.25x \,. \tag{5.23}$$

On these figures, we can see four types of data points:

1. Empirical coverage (blue markers): These are obtained by generating 10000 samples (consisting of 100 calibration points and one test point) to get a good MC approximation of the probability over calibration and test data.

2. Total variation bound (pink markers): These indicate the worst-case coverage as given by the total variation term in Theorem 5.2. These can be calculated in the same way as in Examples 5.10 and 5.11.

3. Mixture bound (green line): This lower bound is simply a consequence of the mixture nature of the population distribution. (It is derived in Property 5.13 below.)

4. Asymptotic coverage (orange markers): These points give the asymptotic coverage levels, i.e. the classwise coverage when the size of the calibration sets would go to infinity. (This corresponds to Property 5.14 below.)

As can be seen in both figures, the total variation distance gives a lower bound that is too weak in general. Moreover, the empirical coverage clearly exhibits asymptotic behaviour for large deviations in the parameters. Although the upper bound is a simple consequence of the fact that the total variation distance takes values in $[0, 1]$, the lower (asymptotic) bound at 0.55 also has a simple explanation given by the following result.

---

[6] In fact, half-normal distributions were used here to mimic the behaviour of the absolute residual measure (3.38).

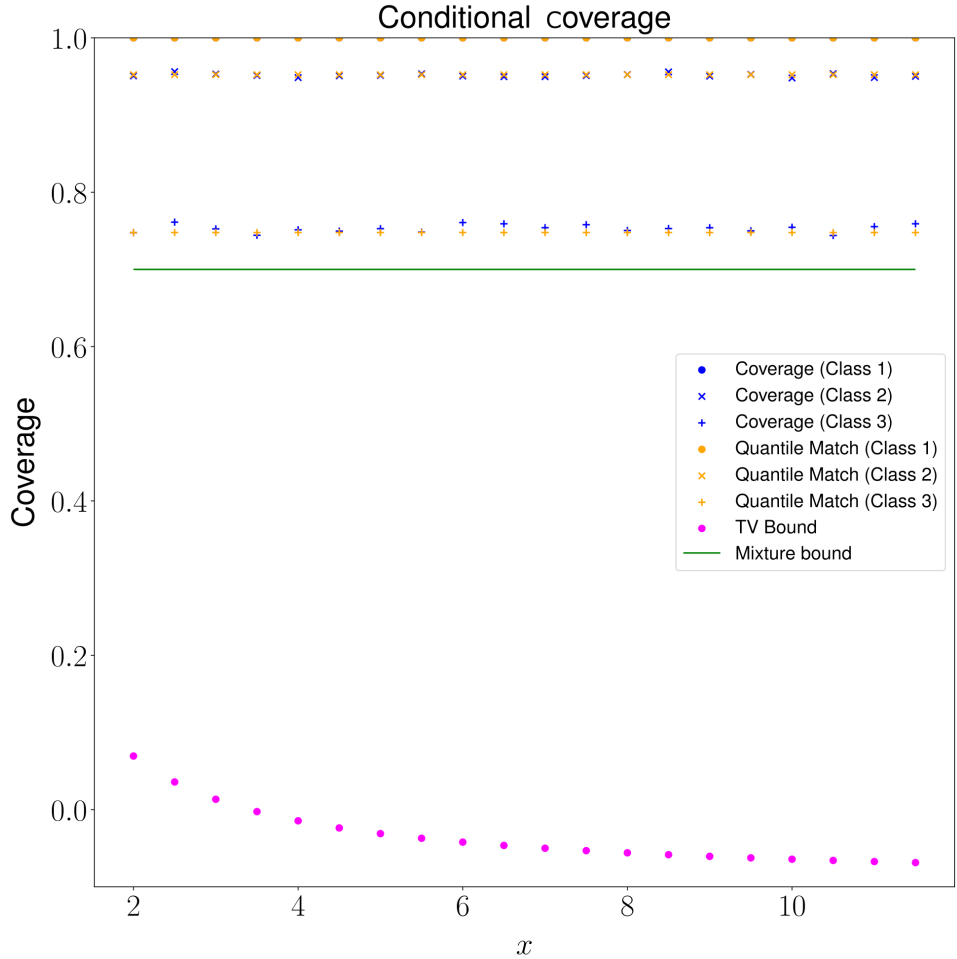

Figure 5.2: Conditional coverage for clusterwise conformal prediction with three (centered) normal distributions. For every value of the mean parameter $\mu(x) = x$, three points are indicated: the empirical coverage in blue, the value predicted by quantile matching (see Property 5.14) in orange and the lower bound given by the total variation distance in pink. The green line indicates the lower bound obtained from the mixture coefficients (see Property 5.13).

**Property 5.13 (Mixture bound).** Assume that the calibration sets are sampled from the mixture distribution (without loss of generality we assume the cluster to be all of $[k]$)

$$P_{X,Y} = \sum_{c \in [k]} \lambda_c P_{X,Y}\big( \cdot \mid \kappa(X,Y) = c \big), \tag{5.24}$$

with $\sum_{c \in [k]} \lambda_c = 1$. The conditional coverage at significance level $\alpha \in$

[0, 1] is bounded from below by

$$\frac{1-\alpha}{\lambda_c} - \sum_{c' \in [k] \setminus \{c\}} \frac{\lambda_{c'}}{\lambda_c}. \tag{5.25}$$

*Proof.* At the marginal level, Validity Theorem 2.27 implies:

$$1 - \alpha \leq \mathsf{Prob}\big(Y \in \Gamma^\alpha(X \mid V)\big)$$
$$= \sum_{c \in [k]} \lambda_c \, \mathsf{Prob}\big(Y \in \Gamma^\alpha(X \mid V) \mid \kappa(X, Y) = c\big).$$

The conditional coverage of a given class $c \in [k]$ attains its minimal value when the terms belonging to other classes are maximal, i.e. when the other terms are equal to 1, and exact validity holds. This gives:

$$1 - \alpha = \lambda_c \, \mathsf{Prob}\big(Y \in \Gamma^\alpha(X \mid V) \mid \kappa(X, Y) = c\big) + \sum_{c' \in [k] \setminus \{c\}} \lambda_{c'}.$$

A simple rearrangement proves the result. □

Although this bound is applicable in any situation, it is still rather weak and clearly does not capture the more complex behaviour of the statistical distances, since it only depends on the mixture composition and not on the probability distributions themselves. When the calibration sets are large enough, a stronger result holds and the precise clusterwise coverage probability can be obtained.

**Property 5.14 (Quantile matching).** Choose a significance level $\alpha \in [0, 1]$ and consider the *Lebesgue decomposition* $\nu_0 + \nu_\perp$ of the mixture distribution $P_A := A_* P_{X,Y}$ with respect to the Lebesgue measure (Capiński & Kopp, 2004). If the calibration sets are sampled from the mixture distribution

$$P_{X,Y} = \sum_{c \in [k]} \lambda_c P_{X,Y}\big(\,\cdot\,\mid \kappa(X, Y) = c\big) \tag{5.26}$$

and the critical quantile $Q_A(1 - \alpha)$ (Definition A.42) of the nonconformity distribution satisfies:

1. it is unique, i.e. the density function of $\nu_0$ does not vanish at $Q_A(1 - \alpha)$, and

2. it does not belong to the support of $\nu_\perp$,

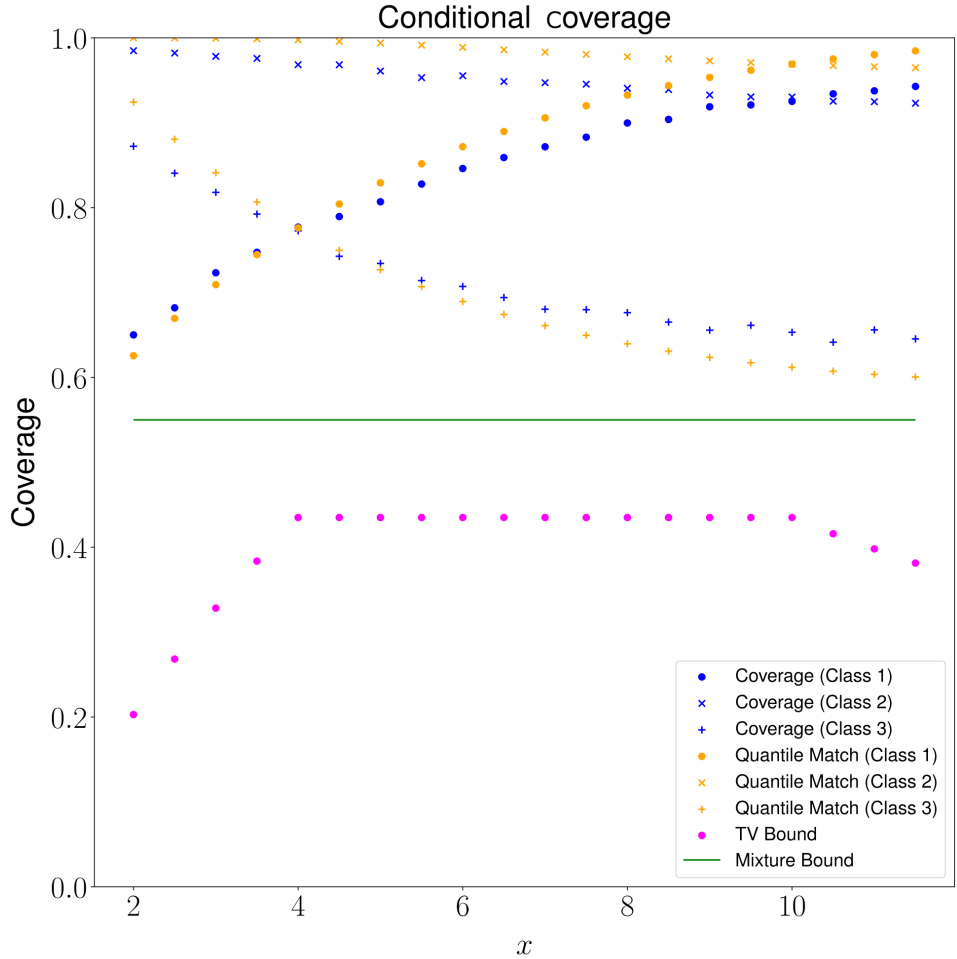

Figure 5.3: Conditional coverage for clusterwise conformal prediction with three (centered) normal distributions. For every value of the standard deviation parameter $\sigma(x) = x$, three points are indicated: the empirical coverage in blue, the value predicted by quantile matching (see Property 5.14) in orange and the lower bound given by the total variation distance in pink. The green line indicates the lower bound obtained from the mixture coefficients (see Property 5.13).

then

$$\lim_{n \to \infty} \mathrm{Prob}\big(Y \in \Gamma^{\alpha}(X \mid V) \mid \kappa(X,Y) = c, |V| = n\big) \qquad (5.27)$$
$$= P_{A|c}\big(A \leq Q_A(1-\alpha)\big),$$

where $P_{A|c} := A_* P_{X,Y}\big(\cdot \mid \kappa(X,Y) = c\big)$.

*Proof*. The probability on the left-hand side can be rewritten as

$$\text{Prob}\big(Y \in \Gamma^\alpha(X \mid V) \mid \kappa(X, Y) = c, |V| = n\big)$$
$$= \int_{(\mathcal{X} \times \mathcal{Y})^n} \int_{-\infty}^{q_{(1-\alpha)(1+1/n)}(A(V))} \mathrm{d}P_{A|c} \, \mathrm{d}P_V \, .$$

By the above assumptions [a], Property A.45 implies that the sample quantile $q_{(1-\alpha)(1+1/n)}(A(V))$ is a consistent estimate of $Q_A(1-\alpha)$. Moreover, under the same assumptions [b], the inner integral is continuous at $Q_A(1-\alpha)$ and the *continuous mapping theorem* (Mann & Wald, 1943) implies that it is consistent as well. This further entails that

$$\left| \int_{(\mathcal{X} \times \mathcal{Y})^n} \left( \int_{-\infty}^{q_{(1-\alpha)(1+1/n)}(A(V))} \mathrm{d}P_{A|c} - \int_{-\infty}^{Q_A(1-\alpha)} \mathrm{d}P_{A|c} \right) \mathrm{d}P_V \right|$$

$$\leq \int_{(\mathcal{X} \times \mathcal{Y})^n} \left| \int_{-\infty}^{q_{(1-\alpha)(1+1/n)}(A(V))} \mathrm{d}P_{A|c} - \int_{-\infty}^{Q_A(1-\alpha)} \mathrm{d}P_{A|c} \right| \mathrm{d}P_V$$

$$\leq \int_{(\mathcal{X} \times \mathcal{Y})^n} \varepsilon \, \mathrm{d}P_V$$

$$= \varepsilon$$

for any $\varepsilon > 0$ when $n \longrightarrow +\infty$. The result follows.     □

---

[a]  Existence of a density to the right of $Q_A(1-\alpha)$.

[b]  Existence of a density to the left of $Q_A(1-\alpha)$ together with the quantile not being an atom.

**Remark** 5.15. The proof is simply an instantiation of the general fact that *convergence in probability* implies *convergence in distribution*, where the latter is equivalent to

$$\lim_{n \to \infty} \mathsf{E}\big[f(X_n)\big] = \mathsf{E}\big[f(X)\big] \tag{5.28}$$

for all bounded, continuous functions $f : \mathcal{X} \to \mathbb{R}$.

## 5.3   Hierarchies

Another tool that will be of interest in the study of clusterwise conformal prediction is the notion of a hierarchy. When working with a multitude of categories, we can often group these together based on some structure or semantic information. A simple example is that of biological taxonomies, where biological species are grouped together based on properties such as their morphology, biochemical character, behaviour or ecological properties (Judd, Campbell, Kellogg, Stevens, & Donoghue, 2007). As is often the case, this grouping is controlled by a tree or **dendrogram** (Definition A.3).[7] For example, the dendrogram in Fig. 5.4 represents a hierarchy on the set $\{1,2,3\}$ with three layers: a root node equal to the total set, an intermediate layer and then a final layer with only singleton leaves.[8] Given a depth $d \in \mathbb{N}_0$, a hierarchy of the given depth on a set $\mathcal{X}$ is a function $\mathcal{H} : [d] \to P(\mathcal{X})$ such that $\mathcal{H}(i)$ is a partition of $\mathcal{X}$ for all $i \in [d]$ consistent with the tree structure, i.e. such that $\mathcal{H}(i + 1)$ is the union of partitions of the nodes in $\mathcal{H}(i)$ for all $i \in [d - 1]$. For simplicity, we will also denote the image of this function by $\mathcal{H}$.

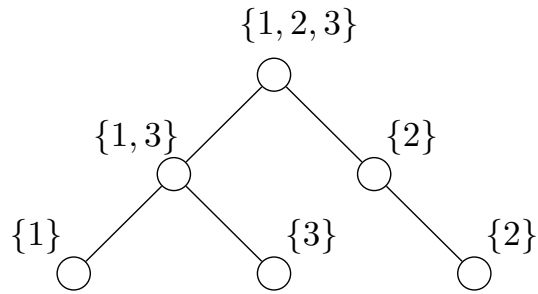

Figure 5.4: Dendrogram representing a hierarchy with three layers on the set $\{1,2,3\}$.

Not only can such hierarchies be used to improve predictive power over standard flat classifiers (Chzhen, Denis, Hebiri, & Lorieul, 2021; Mortier, 2023), they will also allow us to perform conformal prediction in the extreme classification setting, where we do not have sufficient data points for every class. If the hierarchy contains sufficient information about the data-generating process, combining the data in clusters corresponding to nodes in the hierarchy might provide enough information for conformal prediction to achieve results that are better than mere marginal validity and hopefully approximate conditional validity.

---

[7]  More general taxonomy structures exist, such as *directed acyclic graphs*, but these will not be considered here.

[8]  Note that the final layer in a general hierarchy could consist of more complex sets than singletons. However, we will use the convention where the leaf nodes are always singletons.

### 5.3.1 Induced taxonomies

Consider a hierarchy $\mathcal{H}$ on the set $[k]$. Not only can this hierarchy be used to provide side information for predictive methods, it also induces a natural class of taxonomy functions. Consider any (finite) partition $\mathfrak{P} \equiv \{B_i\}_{i \in \mathfrak{I}}$ of $[k]$ by nodes of $\mathcal{H}$ with index set $\mathfrak{I}$:

$$\bigcup_{i \in \mathfrak{I}} B_i = [k].\tag{5.29}$$

The induced taxonomy function $\kappa_{\mathfrak{P}} : [k] \to \mathfrak{I}$ is simply the projection on the index:

$$\kappa_{\mathfrak{P}}(c) := i \qquad \text{such that} \qquad c \in B_i.\tag{5.30}$$

Given a general hierarchy, the above definition of induced taxonomies clearly allows for a vast amount of different choices of partition. On one end of the spectrum lies the top node. This represents the entire target space and, as such, clusterwise validity with respect to this partition amounts to standard Marginal Validity Theorem 2.27. On the other hand, we can choose the bottom layer of the hierarchy, where we decompose the target space into singletons. The validity guarantee associated to this partition is exactly the Conditional Validity Corollary 4.6, since the corresponding model is a classwise Mondrian conformal predictor. In this sense, the induced taxonomies interpolate between marginal and Mondrian conformal predictors in a very natural way.

### 5.3.2 Size thresholding

The question becomes what the right or at least optimal choice of taxonomy is. This in turn depends on the intended purpose and the characteristics of the data-generating process. When the distribution over classes is balanced, i.e. all of them are sufficiently frequently observed, the optimal taxonomies would be those that are induced by one of the final layers of the hierarchy (the deeper the layer, the closer to conditional validity we get). However, when the distribution is skewed, as in the case of extreme classification, the issue is often that for some classes we observe little to no data points. To avoid this problem, a size-adaptive approach can be used.

---

**Algorithm 13:** Size thresholding

---

**Input**   : Significance level $\alpha \in [0, 1]$, nonconformity measure
$A : \mathcal{X} \times \mathcal{Y} \to \mathbb{R}$, taxonomy function $\kappa : \mathcal{X} \times \mathcal{Y} \to [k]$,
hierarchy $\mathcal{H}$ on $[k]$, training set $\mathcal{T} \in (\mathcal{X} \times \mathcal{Y})^*$, calibration set
$\mathcal{V} \in (\mathcal{X} \times \mathcal{Y})^*$ and size threshold $\lambda \in \mathbb{N}_0$

**Output**: Conformal predictor $\Gamma^\alpha$

---

1  (Optional) Train the underlying model of $A$ on $\mathcal{T}$

2  Initialize an empty list $\mathcal{C}$

3  **foreach** $c \in [k]$ **do**

4      **if** $c$ *not in* $\cup_{\omega' \in \mathcal{C}} \omega'$ **then**

5          Initialize the cluster $\omega \leftarrow \{c\}$

6          Select the data set $\mathcal{A}_\omega \leftarrow \{A(x, y) \in \mathcal{D} \mid \kappa(x, y) \in \omega\}$

7          **while** $|\mathcal{A}_\omega| < \lambda$ **do**

8              With respect to $\mathcal{H}$, replace the cluster $\omega \leftarrow \text{Parent}(\omega)$

9              Recalculate the data set $\mathcal{A}_\omega \leftarrow \{A(x, y) \in \mathcal{D} \mid \kappa(x, y) \in \omega\}$

10         **end**

11         Add $\omega$ to $\mathcal{C}$

12 **end**

13 **procedure** $\Gamma^\alpha(x : \mathcal{X})$

14     Initialize an empty list $\mathcal{R}$

15     **foreach** $y \in \mathcal{Y}$ **do**

16         Select the cluster $\omega$ from $\mathcal{C}$ such that $\kappa(x, y) \in \omega$

17         Determine the critical score $a^* \leftarrow q_{(1-\alpha)(1+1/|\mathcal{A}_\omega|)}(\mathcal{A}_\omega)$

18         **if** $A(x, y) \leq a^*$ **then**

19             Add $y$ to $\mathcal{R}$

20         **end**

21     **end**

22     **return** $\mathcal{R}$

23 **return** $\Gamma^\alpha$

---

**Remark** 5.16 (**Multiple taxonomies**).  For the sake of completeness and clarity, it is important to emphasize that, in general, we will be working with two taxonomy functions. The initial one, $\kappa : \mathcal{X} \times \mathcal{Y} \to [k]$, assigns classes to data points $(x, y) \in \mathcal{X} \times \mathcal{Y}$, while the induced taxonomy $\kappa_{\mathfrak{P}}$ is used to cluster in the taxonomy space $[k]$ itself.

To this end, fix a hierarchy $\mathcal{H}$ on the taxonomy space $[k]$, a size threshold $\lambda \in \mathbb{N}_0$ (corresponding to the minimum number of data points that any calibration set should contain) and a data set $\mathcal{V} \in (\mathcal{X} \times \mathcal{Y})^*$. First, we initialize a list of clusters $\mathfrak{P}$. Then, for every class $c \in [k]$, we check whether $\mathfrak{P}$ contains a node $B$ such that $c \in B$ and if this is not the case, we traverse the (unique) path from the leaf node corresponding to $c$ to the root node until we find a node $B \in \mathcal{H}$ such that the calibration set

$$\mathcal{V}_B := \big\{ (x, y) \in \mathcal{V} \mid \kappa(x, y) \in B \big\} \tag{5.31}$$

satisfies the size constraint $|\mathcal{V}_B| \geq \lambda$ and we add this node to $\mathfrak{P}$. By construction, $\mathfrak{P}$ gives a partition of $[k]$ by nodes of $\mathcal{H}$ and the Mondrian conformal predictor associated to the induced taxonomy $\kappa_{\mathfrak{P}}$ will not suffer from any lack of data. This procedure is shown in Algorithm 13.

### 5.3.3 Representation complexity

Consider a prediction problem $f : \mathcal{X} \to \mathcal{Y}$ with target hierarchy $\mathcal{H}$ and an induced taxonomy $\kappa_{\mathfrak{P}} : \mathcal{X} \times \mathcal{Y} \to \big[ |\mathfrak{P}| \big]$. The clusterwise conformal prediction methods from the previous sections will lead to a conformal predictor $\Gamma^{\alpha} : \mathcal{X} \to 2^{\mathcal{Y}}$ such that

$$\mathsf{Prob}\big( Y \in \Gamma^{\alpha}(X \mid V) \mid \kappa(X, Y) = c, V \in \kappa^{-1}(C)^* \big)$$
$$\geq 1 - \alpha - \max_{c' \in C} d_{\mathrm{TV}}(P_{A|c}, P_{A|c'}), \tag{5.32}$$

holds for all $C \in \mathfrak{P}$ and $c \in C$. Of course, as before, if we consider the clusterwise validity, Corollary 4.6 implies that validity will hold at this level.

Although coverage is not really an issue, the size of the prediction sets remains an important factor. In the previous chapters, we used an efficiency measure given by the length of prediction intervals (2.11). For finite sets, as in the case of multiclass classification (see the next section), the most straightforward choice is the cardinality. However, in light of the hierarchy, cardinality is not the only property that needs to be taken into account. As explained in Mortier (2023), hierarchies induce a notion of structural complexity.

**Definition** 5.17 (**Representation complexity**). Consider a hierarchy $\mathcal{H}$ on a finite set $[k]$ and consider a subset $B \subseteq [k]$. Denote by $\mathcal{P}_{\mathcal{H}}(B)$ the set of all disjoint covers of $B$ by nodes in the hierarchy:

$$\mathcal{P}_{\mathcal{H}}(B) := \left\{ \mathfrak{P} \subset \mathcal{H} \;\middle|\; \bigcup_{S \in \mathfrak{P}} S = B \land \forall S, S' \in \mathfrak{P} : S \cap S' = \emptyset \right\}. \quad (5.33)$$

The representation complexity of $B$ (with respect to $\mathcal{H}$) is defined as the minimal size of a disjoint covers of $B$ by nodes in the hierarchy:

$$\mathcal{R}_{\mathcal{H}}(B) := \min_{\mathfrak{P} \in \mathcal{P}_{\mathcal{H}}(B)} |\mathfrak{P}|. \quad (5.34)$$

In the case of, for example, flat classification, any two prediction sets with the same cardinality can be considered equivalent (assuming that they were constructed by valid confidence predictors). However, when considering hierarchies, representation complexity allows us to further disambiguate them. As an example, recall Fig. 5.4 and consider the prediction sets $\Gamma_1 = \{1, 2\}$ and $\Gamma_2 = \{1, 3\}$. According to the foregoing definition, the first set has representation complexity 2, while the second one has representation complexity 1, thereby preferring the latter as a more suitable prediction set. Especially when the hierarchies carry semantical information, i.e. when they are related to the dynamics of the data-generating process, will this be a relevant approach, since a lower representation complexity implies that elements contained in the prediction sets adhere more to the natural structure of the data. A *greedy branch-and-bound* implementation is shown in Algorithm 14.

## 5.4 Classification

Whereas Chapters 3 and 4 solely considered the case of (univariate) regression, classification problems will lend themselves more easily to studying conditional methods (in Section 5.5 we will generalize the setting again). For the purpose of this chapter, multiclass classification and, more specifically, extreme classification will be of major importance.

---

**Algorithm 14:** Representation complexity (greedy search)

**Input** : Hierarchy $\mathcal{H}$, label set $\Gamma$

**Output** : Representation complexity $\mathcal{R}_{\mathcal{H}}(\Gamma)$

1 Initialize a sorted list $\mathcal{A}$ (e.g. a *priority queue*)

2 Add the root node of $\mathcal{H}$ to $\mathcal{A}$

3 Initialize a set $\mathcal{R}$

4 **while** $\cup_{C \in \mathcal{R}} C \neq \Gamma$ **do**

5     Let $N$ be the first element of $\mathcal{A}$ and remove it from $\mathcal{A}$

6     **if** $N \subseteq \Gamma$ **then**

7        Add $N$ to $\mathcal{R}$

8     **else if** $N \cap \Gamma \neq \emptyset$ **then**

9        Add the children of $N$ to $\mathcal{A}$

10 **end**

11 **return** $|\mathcal{R}|$

---

## 5.4.1   Binary classification

In classification problems, we are concerned with deciding whether a given instance $(x, y) \in \mathcal{X} \times \mathcal{Y}$ belongs to a certain 'class' $c \in [k]$ (in which case a natural initial taxonomy function is simply also the identity function). We will first start with the relatively simple case of binary classification ($k = 2$). For the general modelling methodology of (binary) classification problems, see James et al. (2013). We will restrict our focus to probabilistic models, since deterministic classification models $\rho : \mathcal{X} \to [k]$ do not easily lend themselves to refined conformal prediction methods. For example, in the binary case, the only nonconformity measure would be the **zero-one loss**

$$A(x, y) = \begin{cases} 0 & \text{if } y \neq \rho(x), \\ 1 & \text{if otherwise}. \end{cases} \tag{5.35}$$

Now, given a probabilistic classifier[9] $\hat{\rho} : \mathcal{X} \to [0, 1]$, we can wonder what the most relevant choices of nonconformity measure are. In the case of regression, the default choice was the (absolute) residual measure (3.38). A sim-ilar default choice exists for classification problems (Angelopoulos, Bates, Jordan, & Malik, 2021; Shafer & Vovk, 2008):

$$A_{\mathrm{softmax}}(x, y) := \begin{cases} \hat{\rho}(x) & \text{if } y = 0\,, \\ 1 - \hat{\rho}(x) & \text{if } y = 1\,. \end{cases} \tag{5.36}$$

---

[9] Since we consider binary classification, we can simply consider a model that predicts the probability of the positive class.

The idea behind this choice is simple. If the true label is 1, the prediction should be close to 1 and, hence, the more it deviates from it, the more nonconform the value is (and the same for the negative class).

> **Remark** 5.18 (**Metric score**).  The softmax score $A_{\mathrm{softmax}}$ can also be seen as an instance of the more general class of metric nonconformity measures as defined in Eq. (3.37).  If the ground truth is represented by a Dirac measure, e.g. $(0, 1)$ for the positive class, and the prediction is expressed as $\left(1 - \hat{\rho}(x), \hat{\rho}(x)\right)$, then $A_{\mathrm{softmax}}$ can be seen to be induced by the $\ell^1$-distance (A.75): $A_{\mathrm{softmax}}(x, y) = |y - \hat{\rho}(x)|$.

Before generalizing the classification problem to a higher number of classes, let us quickly recall Section 2.3.5. There we considered the decomposition, or rather the impossibility thereof, of conformal prediction sets in terms of epistemic and aleatoric uncertainty. In (binary) classification, the situation is slightly more manageable (Linusson, Johansson, Boström, & Löfström, 2018; Papadopoulos, Harris, 2008; Shafer & Vovk, 2008).  For both classes, 0 and 1, we obtain a conformal $p$-value. In contrast to the case of conformal regression, we can now make a distinction between predicting all classes and predicting nothing at all. To this end, order the $p$-values: $p_{(0)} \leq p_{(1)}$. If the significance level $\alpha \in [0, 1]$ satisfies $\alpha < p_{(0)}$, both possibilities are predicted and we get a meaningless result. The quantity $1 - p_{(0)}$ is sometimes called the **confidence** of the prediction, it represents the greatest probability with which we can predict a single label. On the other hand, if $\alpha > p_{(1)}$, no labels are predicted. The quantity $1 - p_{(1)}$ is sometimes called the **credibility**, it represents the greatest probability for which the model makes a necessarily wrong prediction.

The credibility and confidence of a prediction can be interpreted in light of the decomposition of uncertainty. Low credibility implies that the data point is very nonconforming and might point towards it being an anomalous point (cf. Section 2.4.2 or *out-of-distribution detection*).  A low confidence, on the other hand, means that the data point is not necessarily unusual, but rather that, given our epistemic state, the best probabilistic approximation to the ground truth is spread out over the class space.

### 5.4.2 Multiclass classification

To extend the modelling methodology from the previous section to situations with more than two classes, two different approaches can be used. Either we consider every class to represent a distinct (binary) classification problem, leading to **multilabel classification**, or we consider a discrete distribution over the class space $[k]$, leading to **multiclass classification**. If someone asks you what colour a given (homogeneous) object is, you are presented with a multiclass classification problem. There are multiple possibilities but only one is correct. However, if someone asks you to identify all objects in a given picture, you are presented with a multilabel classification problem. There are multiple possibilities and more than one can be correct.

By using a flexible multi-output architecture $\hat{y} : \mathcal{X} \to \mathbb{R}^k$, one output node for each class, both approaches can be unified and it will be the choice of final layer that determines which approach is chosen. If, for example, a **softmax layer**

$$\mathsf{softmax}(z)^i := \frac{\exp(z^i)}{\sum_{j=1}^k \exp(z^j)} \tag{5.37}$$

is used, the model represents a probability function $P_{Y|X} \in \mathbb{P}([k]) \cong \Delta^{k-1}$, where $\Delta^{k-1}$ denotes the $(k-1)$-simplex from Property A.17. However, if, for example, every output is interpreted as a 'logit' and the **logistic function**

$$\mathsf{logistic}(z) := \frac{\exp(z)}{1 + \exp(z)} \tag{5.38}$$

is applied, the model represents the (tensor) product $P_{Y|X} \in \mathbb{P}(\{0,1\}^{\otimes k})$ of $k$ binary distributions.

For the purpose of multiclass classification, we will be using the first approach. (The second approach is the one used for multilabel classification.) The simplest nonconformity measure for multiclass models is a straightforward extension of Eq. (5.36):

$$A_{\mathsf{softmax}}(x, y) := 1 - \hat{\rho}(y \mid x). \tag{5.39}$$

As with the binary nonconformity measure, we could replace the definition $1 - p$ with $-p$ and nothing would change. The only point of this transformation is to turn the ground truth probability into a nonconformity score. Including the term 1 does allow us to interpret this value as the probability mass that is assigned to the 'wrong classes'.

In Y. Romano, Sesia, and Candès (2020), a different approach was introduced based on modifying the oracle construction of optimal prediction sets. If such an oracle $\pi : \mathcal{X} \to \Delta^{k-1}$ for the conditional distribution $P_{Y|X}$ were provided, the optimal set can be constructed as follows. Consider the ordered probabilities $\pi_{(1)}(x) \leq \cdots \leq \pi_{(k)}(x)$ and define the function

$$L(x, \tau) := \max\left\{ c \in [k] \,\middle|\, \sum_{i=c}^{k} \pi_{(i)}(x) \geq \tau \right\}. \tag{5.40}$$

If there was a $c \in [k]$ that saturated the inequality, the most efficient set, in the sense of Section 2.2.3, would be

$$\Gamma^{\alpha}(x) = \left\{ y \in [k] \,\middle|\, \pi_y(x) \geq \pi_{(L(x, 1-\alpha))}(x) \right\}. \tag{5.41}$$

However, in general, the sum in Eq. (5.40) will be strictly greater than $\tau$ and the resulting prediction sets will be conservative. To this end, whether $\pi_{(L(x,1-\alpha))}(x)$ should be included ought to be randomly chosen as in the case of tied nonconformity scores (cf. Eq. (2.31)):

$$\Gamma^{\alpha}_{\text{oracle}}(x) = \left\{ y \in [k] \mid \pi_y(x) \geq \pi_{(L(x, 1-\alpha)+\varepsilon)}(x) \right\}, \tag{5.42}$$

where

$$\varepsilon \sim \text{Bern}\left( \frac{1}{\pi_{(L(x,1-\alpha))}(x)} \left[ \sum_{i=L(x,1-\alpha)}^{k} \pi_{(i)}(x) - (1-\alpha) \right] \right). \tag{5.43}$$

In practice, however, we do not have access to an oracle and replacing it by an estimate would lead to prediction sets without any guarantees. The solution will be to replace the true confidence level $\tau = 1 - \alpha$ by a conformalized version $\hat{\tau}$. For a given estimate $\hat{\pi} : \mathcal{X} \to \Delta^{k-1}$, the **adaptive prediction sets** (APS) nonconformity measure is defined as follows:

$$A_{\text{APS}}(x, y) := \min\left\{ \tau \in [0, 1] \,\middle|\, y \in \tilde{\Gamma}^{1-\tau}_{\text{oracle}}(x) \right\}, \tag{5.44}$$

where $\tilde{\Gamma}_{\text{oracle}}$ is obtained by inserting the estimate $\hat{\pi}$ in Eq. (5.42). The resulting prediction sets can also be rewritten as

$$\Gamma^{\alpha}_{\text{APS}}(x) = \left\{ y \in [k] \,\middle|\, \sum_{i=R(x,y)}^{k} \hat{\pi}_{(i)}(x) \leq a^{*}_{\text{APS}} \right\}, \tag{5.45}$$

where $R(x, y)$ is the rank of $\hat{\pi}_y(x)$ among the probabilities and $a^*_{\text{APS}}$ is the critical APS score. Note that this approach is also common in the set-valued classification literature (Chzhen et al., 2021).

The APS measure can also be seen as an analogue of Eq. (3.46) for classification. It determines how much the prediction set has to shrink or expand to reach the required significance level. The interesting result for the APS measure is that it empirically showed good conditional coverage compared to other baselines, even when the Mondrian approach is not used (Y. Romano et al., 2020). However, it does have a drawback. In Angelopoulos et al. (2021), it was noted that, especially when there are many classes, the APS sets can be very inefficient due to what is called the 'permutation problem'. If there are many classes with an approximately uniform distribution, their ordering is essentially random and, accordingly, if the true class is part of this long tail, the predicted set will be large. The solution proposed by Angelopoulos et al. (2021) is simply to include a regularization term, leading to 'regularized adaptive prediction sets' (RAPS):

$$A_{\text{RAPS}}(x, y) := A_{\text{APS}}(x, y) + \lambda \max\left(0, R(x, y) - k_{\text{reg}}\right), \qquad (5.46)$$

where $R(x, y)$ is again the rank of $\hat{\pi}_y(x)$ among the probabilities and $\lambda, k_{\text{reg}} \in \mathbb{R}^+$ are hyperparameters.

One peculiar type of classification problem that is of interest for this chapter and where the considerations behind the APS and RAPS measures are of relevance is that of **extreme** (multilabel or multiclass) **classification** (Bengio, Dembczynski, Joachims, Kloft, & Varma, 2019; W. Zhang, Yan, Wang, & Zha, 2018). In this situation, the number of classes $k \in \mathbb{N}_0$ is very large and, in some cases, even comparable to the number of available data points. Some examples are fingerprint identification (Zabala-Blanco, Mora, Barrientos, Hernández-García, & Naranjo-Torres, 2020) or ad recommendation (Aggarwal, Deshpande, & Narasimhan, 2023). Here, the permutation problem is especially problematic. Moreover, even if this were not the case, the large number of classes will also often prohibit the use of conformal prediction due to some classes not having sufficient data points to construct a calibration set as mentioned in the introduction.

**Remark** 5.19. Some other approaches to conformal classification were introduced in Cauchois et al. (2021). The idea behind *conformal quantile classification* is similar to that of conformal quantile regression (Sec-tion 3.3.2). Given a general score function $\rho : \mathcal{X} \times \mathcal{Y} \to \mathbb{R}$, not necessarily a probability mass or density, these methods first fit a quantile function to the training scores using the pinball loss (3.29) as in quantile regression, and then apply conformal prediction to the estimated quantile function.

## 5.5    Multitarget prediction

In Chapters 3 and 4, we solely focused on (univariate) regression problems, whereas in the previous section, we only considered classification problems. However, these settings only account for a small fraction of the domain of application of conformal prediction and, in this section, we will pass to a much larger domain, that of multitarget prediction.

### 5.5.1    Unification

Driven by the explosion of (supervised) modelling techniques and areas of application, and the need for a unified approach to study these various learning methodologies, multitarget prediction (MTP) emerged as an umbrella term (Iliadis, De Baets, & Waegeman, 2022; Waegeman, Dembczyński, & Hüllermeier, 2019). This umbrella covers all kinds of problem settings in which multiple target variables are jointly modelled. Which type of target is considered, is essentially unconstrained. They can be nominal (for classification problems), ordinal (for ranking problems) or real-valued (for regression problems). Due to its flexibility, this framework also gradually attracts more interest because almost all problems in our daily lives involve multiple variables.

The fields being unified include, for example, multilabel classification, multivariate regression and multitask learning (see further on). Usually, each of these subfields apply specialized techniques and ideas without clear reference to the others. A comprehensive survey by Waegeman et al. (2019) reviewed a wide variety of models from the point of view of multitarget prediction. Moreover, in this review, a formal framework to collectively characterize these subfields as a single entity was introduced.

**Definition** 5.20 (**Multitarget prediction**). The estimation of a function $f : \mathcal{X} \times \mathcal{T} \to \mathcal{Y}$ given a finite data set $\mathcal{D} \in (\mathcal{X} \times \mathcal{T} \times \mathcal{Y})^*$. The sets $\mathcal{X}, \mathcal{T}$

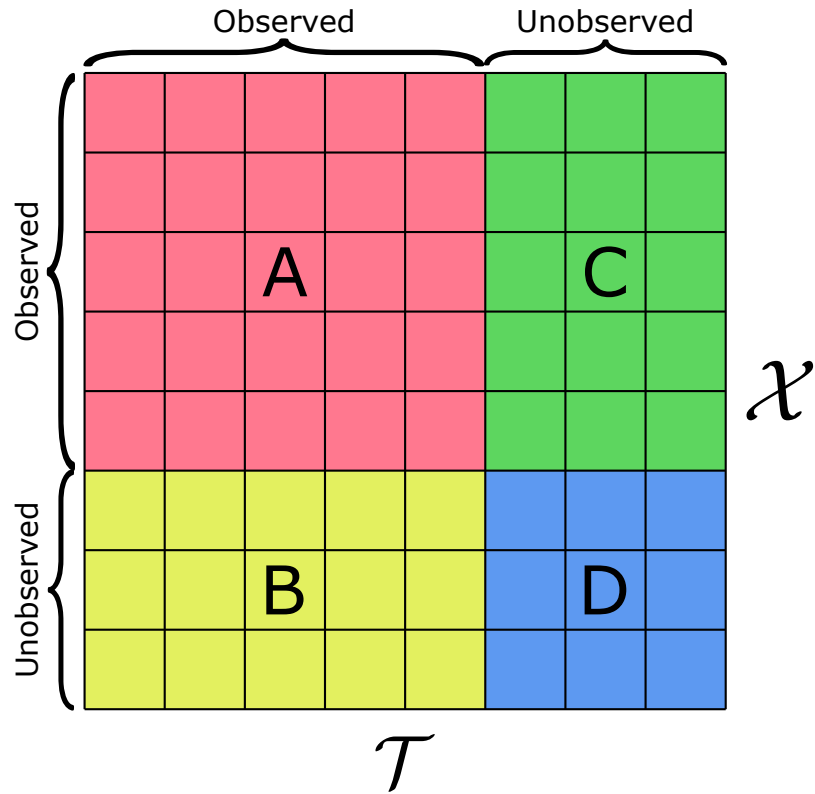

Figure 5.5: Matrix representation of a multitarget prediction problem. The four colours represent the four prediction settings.

and $\mathcal{Y}$ are called the **instance**, **target** and **score spaces**, respectively.

Restricting to finite input spaces $\mathcal{X}$ and $\mathcal{T}$ allows us to represent the training set for the (estimated) function $\widehat{f}$ as a matrix $M_f \in \mathcal{Y}^{|\mathcal{X}| \times |\mathcal{T}|}$. See Fig. 5.5 for such a representation. This matrix consists of four parts, each representing a specific problem setting:

(A) Red area: In this submatrix, all the instances and targets have been observed at least once.

(B) Yellow area: This submatrix consists of instances that have not been observed before.

(C) Green area: This submatrix consists of targets that have not been observed before.

(D) Blue area: This submatrix consists of both instances and targets that have not been observed before.

From a formal point of view, problem settings B and C are equivalent. They are transductive with respect to one factor and inductive with respect to the other. Such settings would, for example, correspond to matching known medicines to unknown diseases (or the other way around). Settings A and D represent the transductive-transductive and inductive-inductive settings, respectively. The former would, for example, correspond to determining the interaction between a known drug and a known disease, while the latter would correspond to the situation where neither the drug nor the disease were observed in the past. This matrix perspective shows that problems such as *matrix completion* (Candès & Recht, 2009; Laurent, 2009) can be incorporated in this paradigm.

Alternatively, the functional perspective easily allows us to see that the following settings can also be unified under the MTP canopy (Breiman & Friedman, 1997; Venkatesan & Er, 2016; Zadrozny & Elkan, 2002):

1. *Multivariate regression*: By *currying*, functions $\mathcal{X} \times \{1, \dots, n\} \to \mathbb{R}$ are in bijection with functions $\mathcal{X} \to \mathbb{R}^n$.

2. *Multilabel classification*: Again, by *currying*, functions $\mathcal{X} \times \{1, \dots, n\} \to \{0, 1\}$ are in bijection with functions $\mathcal{X} \to \{0, 1\}^n$.

By similar reasoning, the more general problem of *multitask learning* (Caruana, 1997; Sener & Koltun, 2018) can also be included under this umbrella.

**Remark** 5.21 (**Multiclass classification**). Note that multiclass classification, as mentioned in Waegeman et al. (2019), is the odd one out in the MTP family when considering it as the modelling of a function $f : \mathcal{X} \to [k]$. For this work, this will not present any issues since we are concerned with probabilistic (multiclass) classification, where we have a prediction for every class.

**Extra** 5.22 (**Currying**). The attentive reader might object to the use of *currying* in the examples of MTP problems, since we are working with measurable functions and not ordinary set functions (Heunen, Kammar, Staton, & Yang, 2017). A general treatment of this issue would lead us too far astray, but suffice it to say that restricting to continuous functions on well-behaved spaces such as the real line $\mathbb{R}$ (with its Borel $\sigma$-algebra) or finite sets $[n]$ avoids this issue (Aumann, 1961).

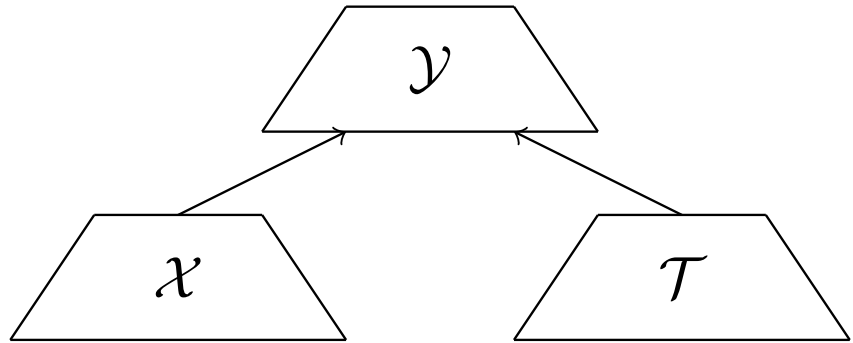

Figure 5.6: Graphical representation of a dual-branch network. For both the feature space $\mathcal{X}$ and target space $\mathcal{T}$, we have a separate embedding layer. These are then aggregated using a third layer.

### 5.5.2 Model architecture

A very natural and useful model architecture for multitarget prediction is that of dual-branch neural networks (Iliadis et al., 2022). In this architecture, we use separate embedding layers for the feature space $\mathcal{X}$ and the target space $\mathcal{T}$. These embeddings are then combined using a third model. See Fig. 5.6 for a graphical representation.

This architecture allows for great flexibility, since the features and targets do not have to be represented in the same way. For example, one can be presented as ordinary numerical vectors (for use with an MLP), while the other could consist of images (for use with a CNN). Two common choices for the third model are either concatenating the feature and target embeddings to feed them to an MLP or choosing the output dimensions of these embeddings to be the same and taking the dot product (Iliadis, De Baets, Pahikkala, & Waegeman, 2024).

Although the previous section might seem to indicate that the purpose of multitarget prediction is to apply machine learning methods to problems on product spaces $\mathcal{X} \times \mathcal{Y}$, there exists another area where the dual-branch structure is of practical use.

An important part of statistical learning and machine learning is constructing methods in such a way that domain knowledge can be incorporated. Instead of taking the risk that the model learns the wrong relations from the data, it is better to bake any existing knowledge directly into the model, e.g. rotationally invariant CNNs when the orientation of an object has no influence on what kind of object it is (Cohen & Welling, 2016).[10]

Not only does the dual-branch structure allow us to incorporate side information, seen as the target variable, the type of model in the target branch allows to encode it in a meaningful way. In the next section, we will see how this allows us to analyse the usefulness of side information-based clustering methods (for which hierarchies, as considered in Section 5.3, form a discrete approximation).

### 5.5.3    Multitarget conformal prediction

Multitarget prediction allows for its own natural class of taxonomy functions, those that condition on the rows (or columns) of the MTP matrix (Fig. 5.5). Unless both $\mathcal{X}$ and $\mathcal{T}$ are countable, conditioning on a single entry will be impossible (J. Lei et al., 2018; Vovk, 2012) and, for simplicity, we will from now on implicitly assume that the space on which we want to condition is finite.[11]

When the taxonomy function is induced by this structure, we can try to exploit the dual-branch architecture to condition clusters in an efficient and meaningful way. For prediction settings $B$ and $C$ (as noted before, these can be considered equivalent), the natural taxonomy associated to it is straightforward. For setting $A$, the choice strongly depends on the situation at hand and, consequently, will not be treated in detail (the general idea is analogous to the other settings). Setting $D$ is, as in the purely predictive setting, the hardest to handle, since neither features nor targets have been observed and, hence, building a Mondrian model is impossible. It would, however, be ideal if we can obtain some kind of guarantee even for those classes that we have not yet observed. Each of these settings will be handled separately below.

Consider the situation where one of the factors of the instance-target pair was previously unobserved (settings B or C). For instance, assume that the instance $x \in \mathcal{X}$ was observed and that the target $t \in \mathcal{T}$ was unobserved. In

---

[10] For more information on the incorporation of inductive biases and, in particular, symmetries, see Section 6.2.3.

[11] Section 5.2 presents a way of reducing an infinite target space to a finite one.

this case, a natural taxonomy function is given by the projection onto $\mathcal{X}$:

$$\kappa_{\mathcal{X}}(x, t, y) := \pi_{\mathcal{X}}(x, t, y) = x. \tag{5.47}$$

Given the taxonomy function $\kappa_{\mathcal{X}}$, the procedure is straightforward. For every new pair $(x, t) \in \mathcal{X} \times \mathcal{T}$, Mondrian conformal prediction (Algorithm 11) constructs a calibration set consisting of points with the same instance label and calculates the critical score of the chosen nonconformity measure. If not enough data is available, the methods from the previous sections can be used to obtain a clusterwise model.

Aside from this straightforward conditioning and clustering procedure, another possibility exists. Assume that a similarity function $S : \mathcal{T} \times \mathcal{T} \to \mathbb{R}$ is given (the real numbers can be replaced by any poset). This function also allows us to condition on the targets as follows. Choose an integer $k \in \mathbb{N}_0$. For every new pair $(x, t) \in \mathcal{X} \times \mathcal{T}$, construct the calibration set $\mathcal{V}_{(x,t)}$ by calculating for every calibration target its similarity with $t$ and retaining the top-$k$ similar targets:

$$\mathcal{V}_{(x,t)} := \left\{ (x', t', y') \in \mathcal{V} \mid S(t, t') \geq s^{(k)} \right\}, \tag{5.48}$$

where $s^{(k)}$ is the $k^{\text{th}}$ greatest element of $\{S(t, t') \mid t' \in \pi_{\mathcal{T}}(\mathcal{V})\}$. By fine-tuning the parameter $k \in \mathbb{N}_0$, we can choose how much data will be included in the calibration set. It should be noted that this algorithm does not lead to disjoint clusters and, as such, is strictly speaking not an example of the methods from the previous sections. However, Theorem 5.2 still applies since it does not depend on any disjointness. If the similarity measure is induced by a metric on $\mathcal{T}$, the Lipschitz results from Section 5.2.3 can be extended through the following property, which simply says that (under reasonable conditions) marginalizing away the feature dependence preserves Lipschitz continuity.

**Property 5.23.** Consider a (parametric) family of probability distributions $P : \mathcal{X} \times \mathcal{T} \to \mathbb{P}(\mathcal{Y}) : (x, t) \mapsto P(\cdot \mid x, t)$, where $(\mathcal{X} \times \mathcal{T}, d)$ is a metric space. Assume that the metric $d$ can be obtained by aggregating metrics on $\mathcal{X}$ and $\mathcal{T}$, i.e.

$$d\big((x, t), (x', t')\big) = f\big(d_{\mathcal{X}}(x, x'), d_{\mathcal{T}}(t, t')\big) \tag{5.49}$$

for some function $f : \mathbb{R}^+ \times \mathbb{R}^+ \to \mathbb{R}^+$ with $f(0, \alpha) = \alpha$.[a] If $P$ is Lipschitz-continuous with respect to the statistical distance, the marginal distri-

bution

$$P(\cdot \mid t) := \int_{\mathcal{X}} P(\cdot \mid x, t)\, \mathrm{d}P_X(x) \tag{5.50}$$

is also Lipschitz-continuous with respect to the statistical distance as a function $\mathcal{T} \to \mathbb{P}(\mathcal{Y})$.

*Proof.*

$$
\begin{aligned}
d_{\mathrm{TV}}\big(P(\cdot \mid t), P(\cdot \mid t')\big) &= \sup_{A \in \Sigma_{\mathcal{Y}}} |P(A \mid t) - P(A \mid t')| \\
&= \sup_{A \in \Sigma_{\mathcal{Y}}} \left| \int_{\mathcal{X}} \big(P(A \mid x, t) - P(A \mid x, t')\big)\, \mathrm{d}P_X(x) \right| \\
&\le \sup_{A \in \Sigma_{\mathcal{Y}}} \int_{\mathcal{X}} |P(A \mid x, t) - P(A \mid x, t')|\, \mathrm{d}P_X(x) \\
&= \int_{\mathcal{X}} \sup_{A \in \Sigma_{\mathcal{Y}}} |P(A \mid x, t) - P(A \mid x, t')|\, \mathrm{d}P_X(x) \\
&\le \int_{\mathcal{X}} d\big((x, t), (x, t')\big)\, \mathrm{d}P_X(x) \\
&= \int_{\mathcal{X}} f\big(d_{\mathcal{X}}(x, x), d_{\mathcal{T}}(t, t')\big)\, \mathrm{d}P_X(x) \\
&= \int_{\mathcal{X}} d_{\mathcal{T}}(t, t')\, \mathrm{d}P_X(x) \\
&= d_{\mathcal{T}}(t, t'),
\end{aligned}
$$

where $\Sigma_{\mathcal{T}}$ denotes the $\sigma$-algebra on $\mathcal{T}$.                    $\square$

---

[a]  Note that this includes many cases such as $p$-product metrics, averaging and even more exotic possibilities such as *warped products* (Petersen, 2006).

Similar to the clustering approach for dimensionwise conditioning, the fully inductive setting (Setting D) can be conquered by using similarities for both instances and targets. To this end, consider two similarity functions $S_{\mathcal{X}} : \mathcal{X} \times \mathcal{X} \to \mathbb{R}$ and $S_{\mathcal{T}} : \mathcal{T} \times \mathcal{T} \to \mathbb{R}$ and two integers $k, l \in \mathbb{N}_0$. The calibration set in Eq. (5.48) then becomes:

$$\mathcal{V}_{(x,t)} := \big\{ (x', t', y') \in \mathcal{V} \mid S_{\mathcal{X}}(x, x') \ge s^{(k)} \wedge S_{\mathcal{T}}(t, t') \ge s^{(l)} \big\}, \tag{5.51}$$

where, as above, $s^{(k)}$ is the $k^{\mathrm{th}}$ greatest element of $\{S(x, x') \mid x' \in \pi_{\mathcal{X}}(\mathcal{V})\}$ and $s^{(l)}$ is the $l^{\mathrm{th}}$ greatest element of $\{S(t, t') \mid t' \in \pi_{\mathcal{T}}(\mathcal{V})\}$. Note that, as before, the choice of $k$ and $l$ lets us interpolate between marginal conformal predic-tion, where $k = |\mathcal{X}|$ and $l = |\mathcal{T}|$, and object-conditional conformal prediction, where $k = l = 1$.[12]

Techniques such as the hierarchy-based clustering method from Section 5.3 can also be included in this story. Given a hierarchy $\mathcal{H}$ of depth $d \in \mathbb{N}_0$, every class can be encoded as a $(d-1)$-tuple (possibly after suitable padding). The space of tuples can then be equipped with the **Fréchet metric**[13] (Choquet-Bruhat, DeWitt-Morette, & Dillard-Bleick, 1991)

$$d_{\text{Fr}}(v, w) := \sum_{i=1}^{d-1} \frac{2^{-i} \rho(v^i, w^i)}{1 + \rho(v^i, w^i)} \tag{5.52}$$

induced by the **discrete metric**

$$\rho(x, y) := \begin{cases} 0 & \text{if } x = y \\ 1 & \text{if } x \neq y \,, \end{cases} \tag{5.53}$$

which, in the sense of Eq. (3.37), is the underlying metric for the zero-one loss (5.35).

Note that this distance makes sense intuitively. The use of the discrete metric at every level of the hierarchy corresponds to the fact that there is no real structure or order amongst the nodes at a given level. On the other hand, it is also clear that deviations from the true path between a (leaf) node and the root node matter less the further along the path we go. Given this construction, the use of hierarchies can be placed under the multitarget prediction umbrella and the Lipschitz results can be applied.

## 5.6  Experiments

### 5.6.1  Data

Although this chapter rather serves an exploratory purpose than to introduce state-of-the-art techniques, we do find it useful to see how the methods perform on real-life data sets. To this end, we illustrate the techniques introduced in this chapter on two multiclass data sets: `Proteins` (Li et al., 2018) and `Bacteria` (RIKEN, 2013). Instead of working with the raw features, the author of Mortier (2023) provided us with precomputed *tf-idf representations*. Moreover, for both data sets, a preconstructed hierarchy was provided. Note that the hierarchies are rather shallow (4 and 6 layers, respectively[14]). This implies that the most parent nodes have a large number of classes/child nodes and, hence, the hierarchy gives a crude partitioning of the class space. (All technical details about the data sets can be found in Table 5.1.)

---

[12] The latter only gives object-conditional conformal prediction when the similarity functions are nondegenerate, i.e. when $S(x, x') = 0 \implies x = x'$.

[13] Note that the factor 2 could be replaced by any other number $\lambda \geq 1$ to obtain an equivalent metric. For $\lambda = 1$; the standard *graph metric* is obtained.

Table 5.1: Details of the data sets.

| Name | Samples | Classes | Features | Hierarchy depth | Source |
|------|---------|---------|----------|-----------------|--------|
| `Proteins` | 22009 | 3485 | 26276 | 5 | Li et al. (2018) |
| `Bacteria` | 12881 | 2659 | 2472 | 7 | RIKEN (2013) |

The provided data sets were split in train and test sets, in such a way that the training set contained all observed classes. From the test set, we constructed 10 random calibration/test splits to perform conformal prediction. Figs. 5.7a and 5.7b show the distribution of the classes in the training set for the data sets `Proteins` and `Bacteria`, respectively. It should be immediately clear that a very strong class imbalance is present in both data sets. Over 50% of the classes have less than five data points and over 90% of the classes lie below the size threshold of 50 data points.

## 5.6.2   Models and training

In this experimental section, we will consider three different classification models:

1. **Probabilistic label trees** (PLT): To obtain a PLT, we first construct a binary tree over the label space [$k$]. Then, we proceed by training a probabilistic (binary) classifier at every node of the tree (Jasinska et al., 2016). These classifiers predict whether the path to this node is possible for the given features. Since this allows for multiple paths to be accessible, PLTs can also be used for multilabel classification. However, in our case, a 'top-1' aggregation strategy is used to obtain a multiclass model.

---

[14] These include the root node and the leaf nodes

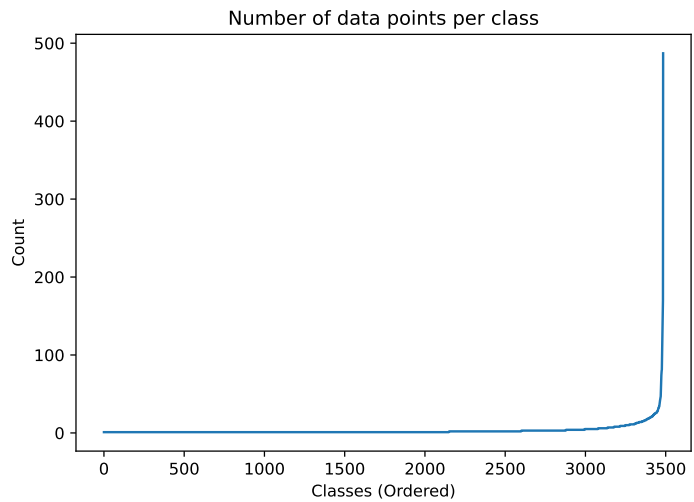

(a) `Proteins` data set.

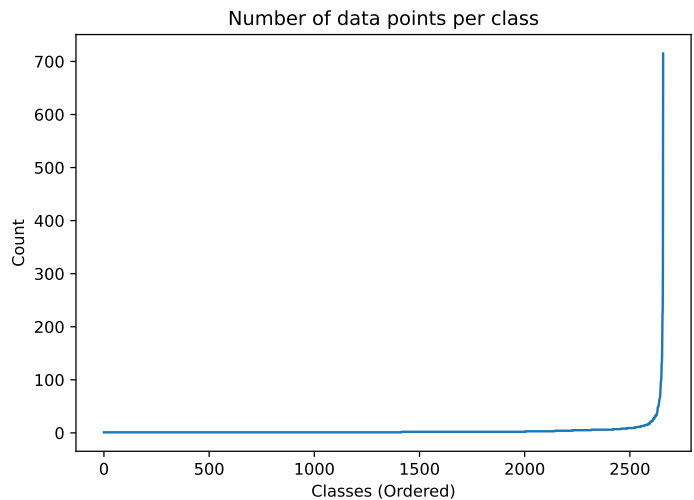

(b) `Bacteria` data set.

Figure 5.7: Number of available data points per class (in the training set) shown in increasing order. Note that most classes have an extremely low number of points.

2. **Hierarchical softmax classifier** (HSM): Similar to PLTs, we start by constructing a tree over the label space, however, it need not be binary. Moreover, we only train a (softmax) classifier at every internal node.

These internal classifiers are used to determine to which child the features correspond (Morin & Bengio, 2005):

$$P(v' \mid v, x) := \frac{\exp\big(\boldsymbol{w}_{vv'} \cdot \boldsymbol{\varphi}(x)\big)}{\sum_{z \in \text{Child}(v)} \exp\big(\boldsymbol{w}_{vz} \cdot \boldsymbol{\varphi}(x)\big)}, \tag{5.54}$$

where Child : $\mathcal{H} \to 2^{\mathcal{H}}$ gives all children of a given node, $\boldsymbol{w}_{vv'} \in \mathbb{R}^d$ denotes the weights corresponding to the path $v \to v'$ and $\boldsymbol{\varphi} : \mathcal{X} \to \mathbb{R}^d$ denotes the feature map, which is shared among all nodes in this case.

3. **One-versus-rest classifier** (OVR): In this approach, we construct a binary classifier for every class, which predicts whether a given data point belongs to that class or not. To obtain a single prediction, the argmax of the predicted probabilities is taken.

All three models use the implementation from the `napkinXC` library for extreme classification in Python (Jasinska-Kobus, Wydmuch, Dembczynski, Kuznetsov, & Busa-Fekete, 2020).

As a prior taxonomy function, we choose the natural one for (multiclass) classification, i.e. we condition on the class:

$$\kappa(x, y) := y. \tag{5.55}$$

Since a (functional) hierarchy is provided, different (clusterwise) conformal prediction approaches can be compared. As in the previous chapters, the two baseline methods will be marginal conformal prediction (Section 2.3), which uses a single calibration set, and Mondrian conformal prediction (Section 4.3), which uses a separate calibration set for every class. As nonconformity measures, both the softmax and APS score from Section 5.4.2 are considered. To interpolate between these extremes, we follow the approach of Section 5.3.2, where classes are clustered according to the size of the calibration sets in such a way that the clusters are consistent with the hierarchy, i.e. such that the clusters are given by nodes in the hierarchy. The size threshold is fixed at $\lambda = 50$. Aside from the predetermined (functional) hierarchies, we also consider the nonconformity-based clustering from Ding et al. (2023) as described in Section 5.2.2. To this end, we choose five quantile levels $\mathfrak{I} = \{0.1, 0.25, 0.5, 0.75, 0.9\}$ and calculate the quantile embeddings

$$\xi_c = \big(q_\beta(A(\mathcal{V}_c))\big)_{\beta \in \mathfrak{I}}. \tag{5.56}$$

Moreover, we combine this score-based approach with the size-adaptive approach. After selecting a size threshold, again chosen to be $\lambda = 50$, we cluster in different ways. All classes with less than $|\mathfrak{I}| = 5$ observations are grouped together in a rest class. On the classes with less than $\lambda$ observations, we run a clustering algorithm such as $k$-means on the quantile embeddings and the classes with more than $\lambda$ observations are kept separately as in the Mondrian approach. We shortly list all four options for clarity with the labels that are used in the figures:

1. Marginal (label: 'Marginal'),

2. Mondrian (label: 'Mondrian'),

3. size-adaptive hierarchical (label: 'Hierarchy'), and

4. size-adaptive score-based (label: 'Score').

To analyse these methods, we will consider different metrics. The main metrics will be the (class-)conditional and clusterwise coverage. While the latter should approximate the nominal value of 90% by the standard validity results of conformal prediction, the hope is that the former also approximates this value as good as possible. The third metric of interest will be the (in)efficiency of the prediction sets, i.e. their size. Especially in the extreme classification setting, where there are many classes and where the permutation problem from Section 5.4.2 plays a role, this is an important factor. Very large prediction sets might say something about the uncertainty, but they are not very useful in practice. At last, we will also consider the representation complexity (Definition 5.17) of the prediction sets. Since we have access to predefined (functional) hierarchies, it is interesting to see whether the confidence predictors model the uncertainty in a way that is consistent with these hierarchies. If this is not the case, this might indicate that the structure behind the data-generating process might be different from the structure determining the hierarchies.

### 5.6.3 Results

The accuracy (Definition A.72), balanced accuracy (Definition A.73) and (weighted) $F_1$-score (Definition A.74) of the different models are shown in Tables 5.2 and 5.3 for the `Proteins` and `Bacteria` data sets, respectively. The mean and standard deviation over the 10 data splits are shown.

In Fig. 5.8, we can see histograms of the classwise and clusterwise coverage for the three models (PLT, HSM and OVR) for a single train-calibration-test split using the softmax nonconformity measure (see Fig. B.1 for the APS measure) as calculated on the `Proteins` data set. The rows indicate the four different ways to construct clusters: marginal (all classes together), Mondrian (one cluster per class), Hierarchy (use the predetermined hierarchy together with a size threshold) and Score (cluster quantile embeddings of the nonconformity distributions together with a size threshold). For each subfigure, two different histograms are shown. For the blue one (with the label 'Clusterwise'), the coverage is calculated on the level of the clusters and, hence, the standard coverage guarantee predicts that this histogram should be centered around the nominal confidence level of $1 - \alpha$ (with $\alpha = 0.1$). For the orange histogram (with the label 'Classwise'), the coverage is calculated on the level of the classes and, hence, corresponds to the conditional coverage.

Table 5.2:  Accuracy measures for the three models (PLT, HSM, OVR) on the `Proteins` data set.  Mean and standard deviation over 10 samples are reported.

| Model | Acc | bAcc | $F_1$ |
|-------|-----|------|-------|
| PLT | $0.793 \pm 0.003$ | $0.594 \pm 0.008$ | $0.761 \pm 0.003$ |
| HSM | $0.792 \pm 0.003$ | $0.587 \pm 0.009$ | $0.759 \pm 0.004$ |
| OVR | $0.806 \pm 0.002$ | $0.614 \pm 0.007$ | $0.775 \pm 0.003$ |

Table 5.3:  Accuracy measures for the three models (PLT, HSM, OVR) on the `Bacteria` data set.  Mean and standard deviation over 10 samples are reported.

| Model | Acc | bAcc | $F_1$ |
|-------|-----|------|-------|
| PLT | $0.793 \pm 0.003$ | $0.594 \pm 0.008$ | $0.761 \pm 0.003$ |
| HSM | $0.792 \pm 0.003$ | $0.587 \pm 0.009$ | $0.759 \pm 0.004$ |
| OVR | $0.806 \pm 0.002$ | $0.614 \pm 0.007$ | $0.775 \pm 0.003$ |

For the classwise approach, we see some interesting behaviour. First of all, it makes sense that the histogram is not necessarily centered around $1-\alpha$, since we are clustering classes and the Clusterwise Validity Theorem 5.2 predicts a possible deviation from validity. However, even in the Mondrian case, there are strong deviations from validity. The main issue here is the low amount of data and this manifests itself in two ways. There will be classes for which not enough calibration data is available and there will be classes for which not enough test data is available. Especially the latter contributes to the peaks at 0% and 100% (this also holds for the other clustering methods).[15]

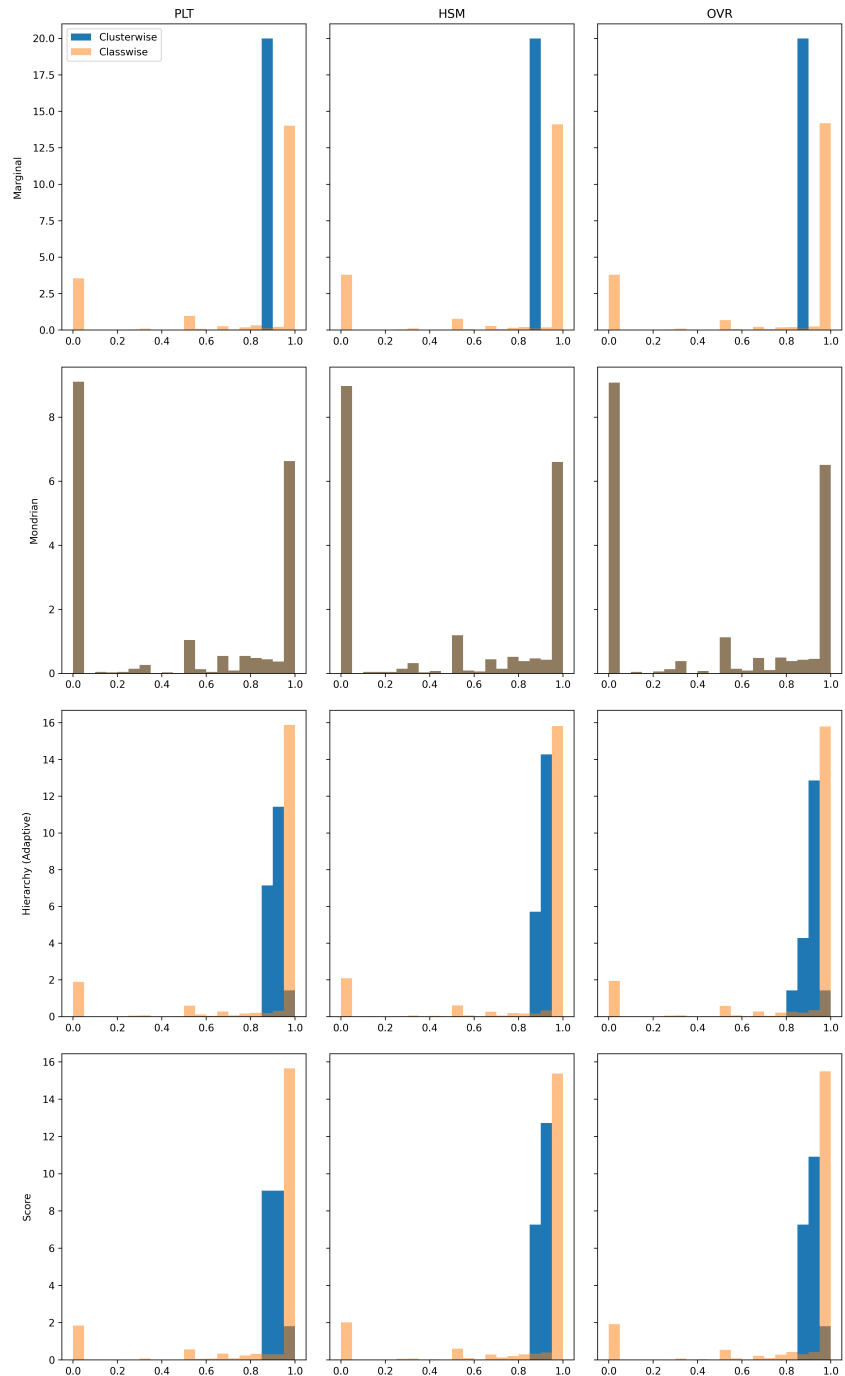

Figure 5.8: Histogram of classwise and clusterwise coverage for the softmax nonconformity measure for three models (PLT, HSM and OVR) and four clustering methods (marginal, Mondrian, hierarchy-based and score-based) on the `Proteins` data set.

Since these histograms depend on the specific calibration-test split (especially in this extreme setting), a boxplot of the clusterwise histogram values is also shown in Figs. 5.9 and B.2, and for the classwise values in Figs. 5.10 and B.3. We can see that for all methods, except the Mondrian one, a valid result is obtained (as expected). Note that the deviation for the Mondrian method was also to be expected as mentioned in the previous paragraph. Moreover, by comparing the first row (the marginal method) to the third and fourth rows (proper clustering methods) in Figs. 5.10 and B.3, we can also see that the latter exhibit slightly better classwise (i.e. conditional) coverage behaviour.

Histogram plots for the average prediction set size (Eq. (2.11)) and the representation complexity (Definition 5.17) are shown in Figs. 5.11 and B.5 for the `Proteins` data set. It can be seen that the proper clustering methods give slightly larger prediction sets than the marginal method. Also note that for the APS score, the average prediction set size is almost equal to the cardinality of the class space. This can be explained by the skewness of the class distribution (recall Fig. 5.7a), together with the permutation problem as explained in Section 5.4.2.

All figures for the `Bacteria` data set are shown in Section B.3 in Figs. B.6 to B.15.

## 5.7  Discussion

Active research in conformal prediction has mainly focused on either further refining marginal methods or obtaining conditionally valid models. Although there is no doubt that these approaches can be applied in many different situations (and that the latter is the holy grail of conformal prediction), a possible lack of data remains an important issue in practice. In this chapter, we explored and investigated a possible solution to counteract data short-age while retaining the conditional guarantees as much as possible. To this end, two possible approaches were considered. One where the classes are clustered using embeddings of the (empirical) conditional nonconformity distributions and one where we use side information in the form of a hierarchy over the classes to group them together. While the former is optimal for conformal prediction, the latter gives more insight into the underlying process. Moreover, a sufficient condition for the optimality of the latter was studied and the theoretical relation between classwise and clusterwise coverage was analysed.

---

[15] Classes for which no test data is available at all are discarded for the classwise calculations.

To assess the different approaches, we also considered two extreme (multiclass) classification problems. The methods were compared based on various characteristics such as the classwise and clusterwise coverage, but also the efficiency and, since hierarchies were provided, the average representation complexity of the prediction sets. Two main conclusions could be drawn from the results. First, that the Mondrian approached is doomed to fail in these limited data settings. Second, that, without additional engineering, the clusterwise approaches are only marginally better than the marginal ones. On the one hand, the classwise coverage is slightly better for the clusterwise methods, but, on the other hand, the efficiency decreases.

Moreover, since the methods used for multiclass classification can be seen as a subset of those relevant to multitarget prediction, we also studied how these architectures form a natural setting for (clusterwise) conformal prediction. Although the experimental results were not overly positive, this work opens up some possible roads for future endeavours, such as studying which nonconformity measures are well-suited for these special settings.

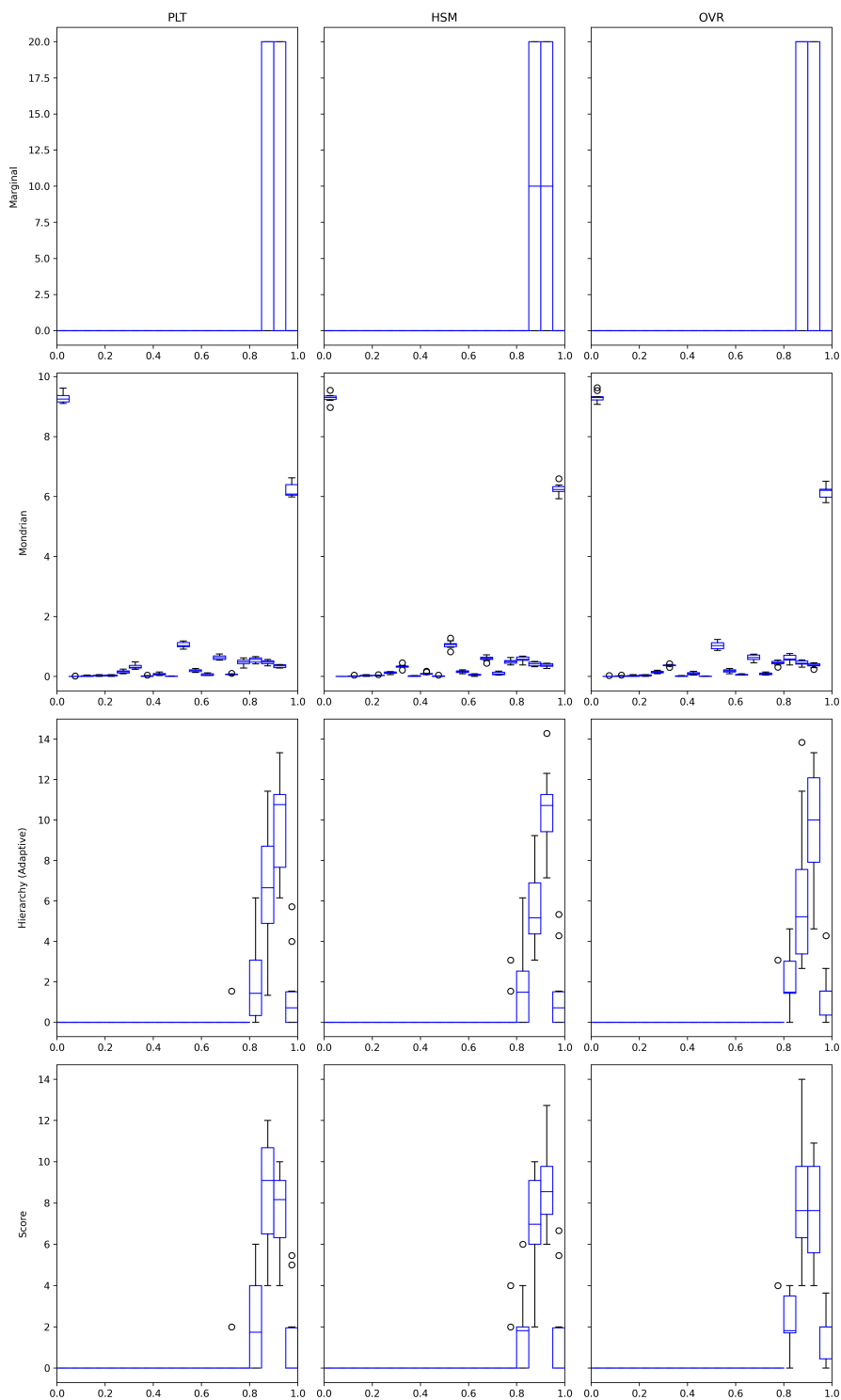

Figure 5.9:  Boxplot of clusterwise histogram densities for the softmax non-conformity measure for three models (PLT, HSM and OVR) and four clustering methods (marginal, Mondrian, hierarchy-based and score-based) over all calibration-test splits on the `Proteins` data set.

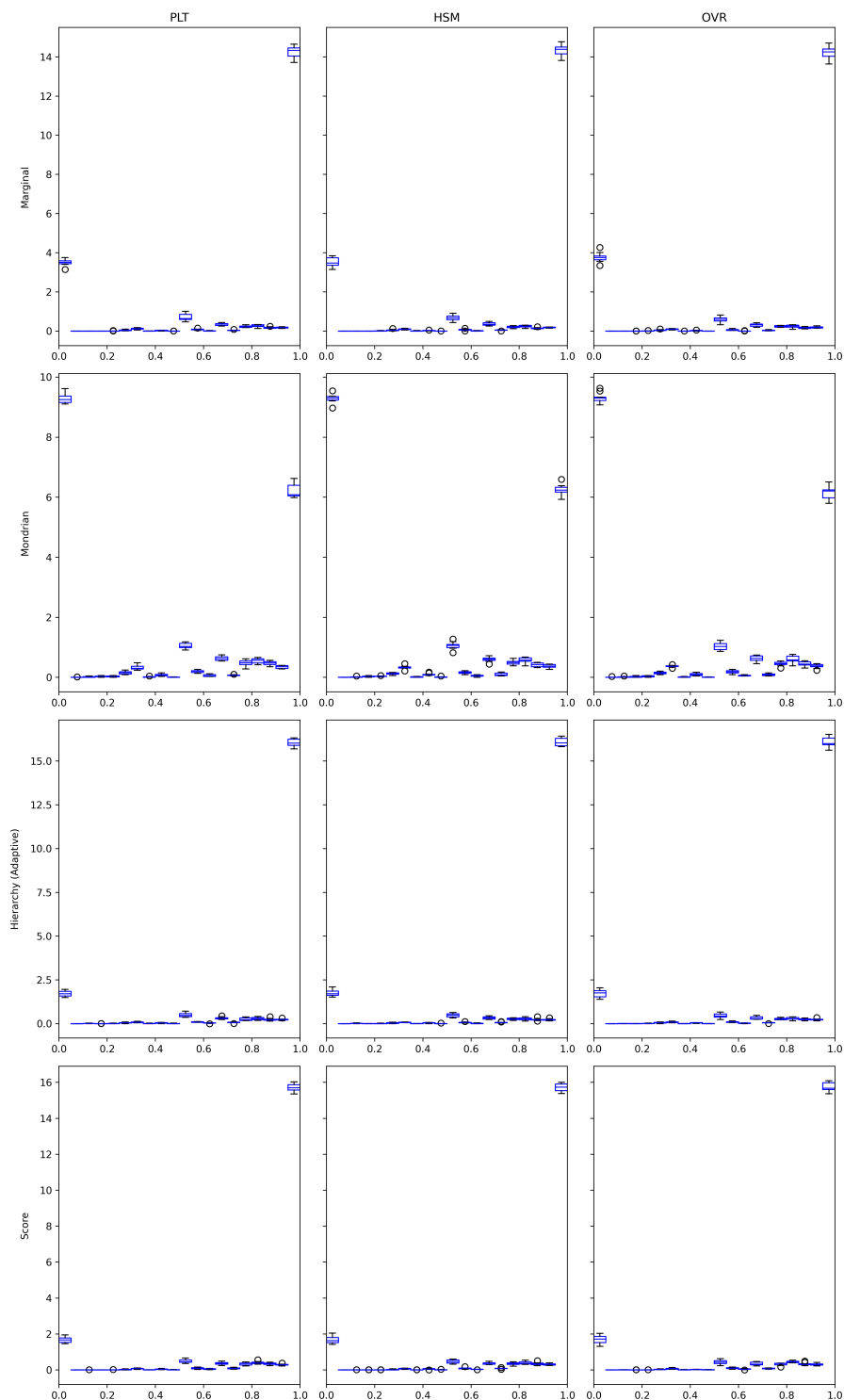

Figure 5.10: Boxplot of classwise histogram densities for the softmax noncon-formity measure for three models (PLT, HSM and OVR) and four clustering methods (marginal, Mondrian, hierarchy-based and score-based) over all calibration-test splits on the `Proteins` data set.

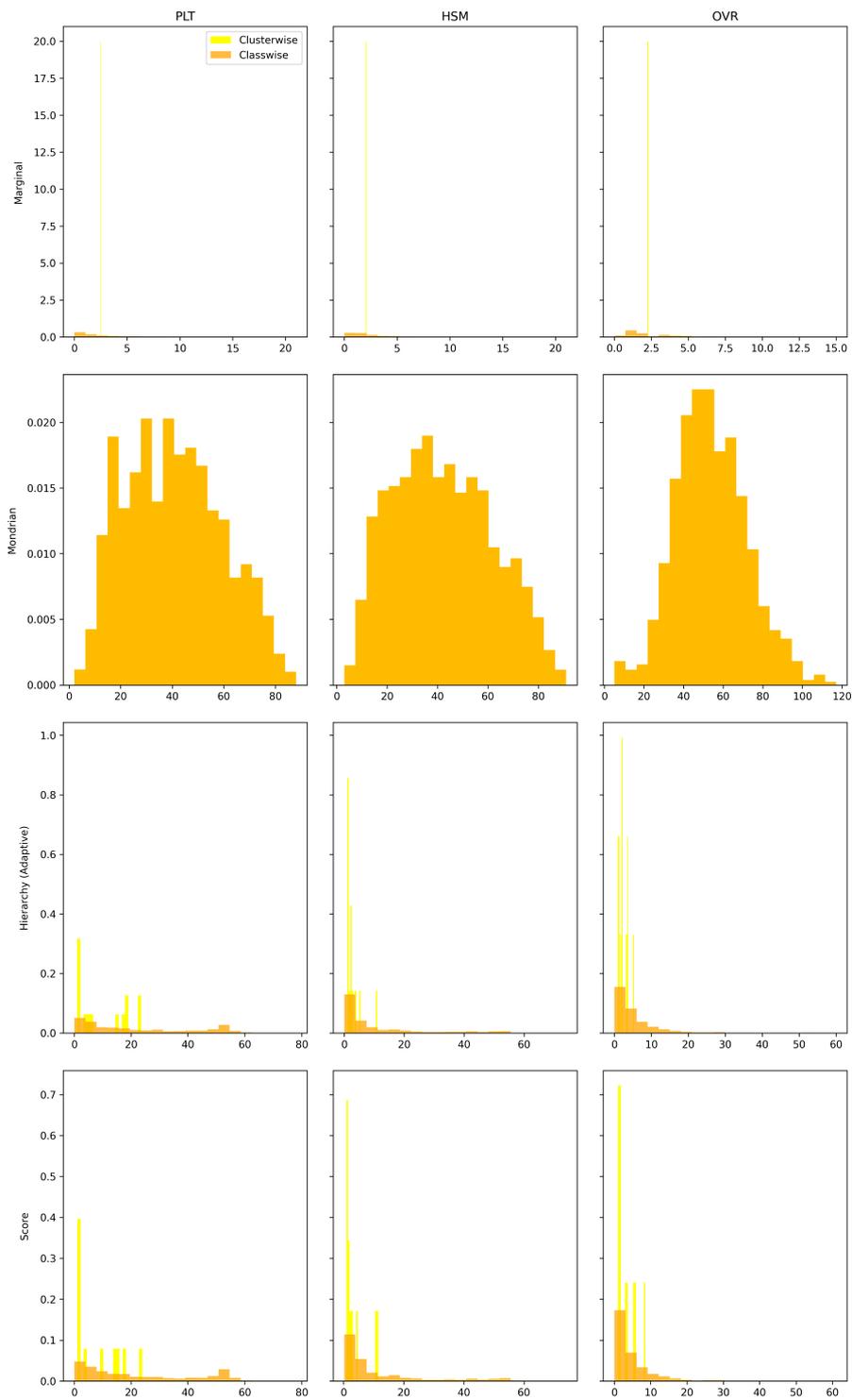

Figure 5.11: Histogram of classwise and clusterwise average prediction set sizes for the softmax nonconformity measure for three models (PLT, HSM and OVR) and four clustering methods (marginal, Mondrian, hierarchy-based and score-based) on the `Proteins` data set.

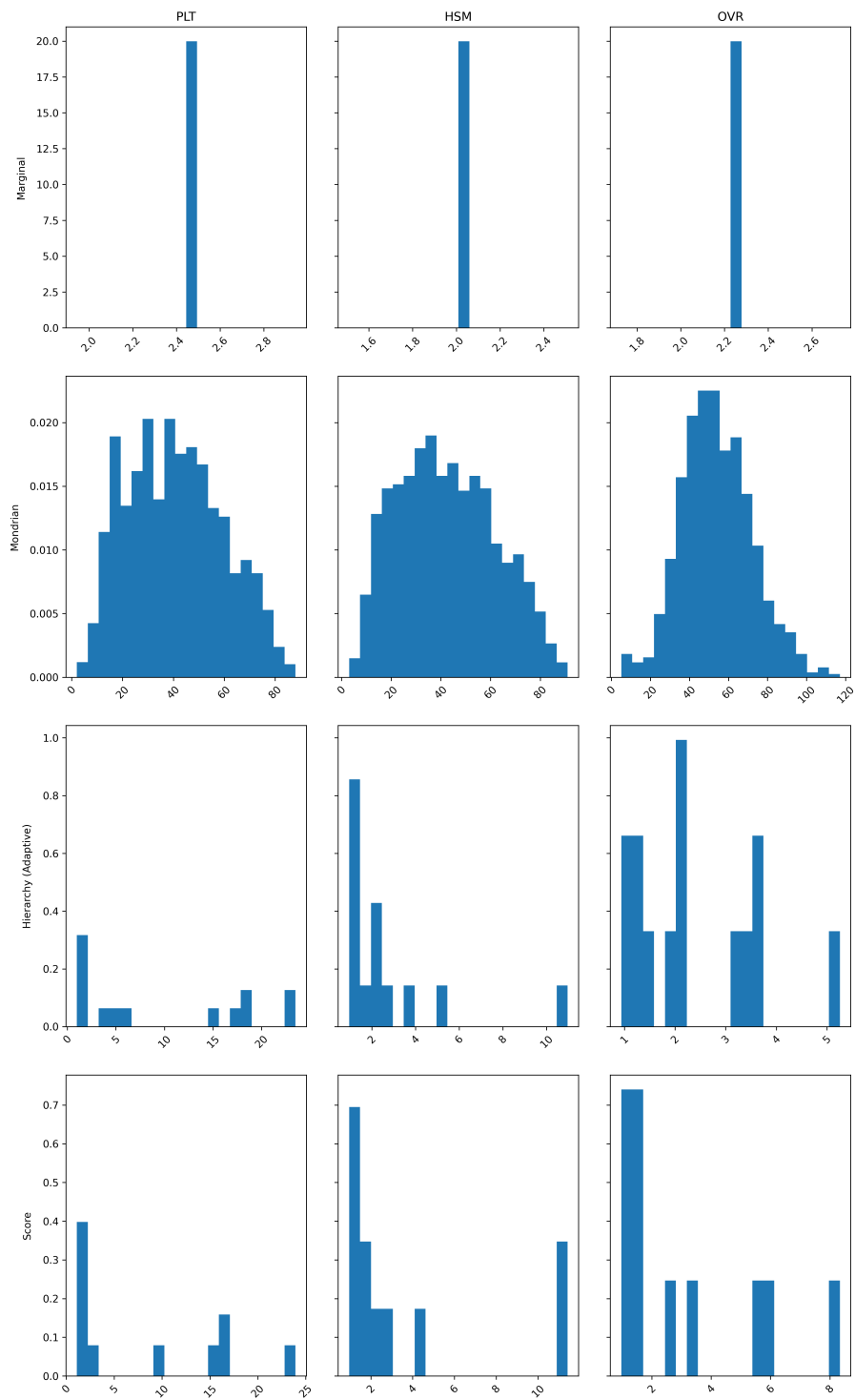

Figure 5.12: Histogram of clusterwise representation complexities of the prediction sets for the softmax nonconformity measure for three models (PLT, HSM and OVR) and four clustering methods (marginal, Mondrian, hierarchy-based and score-based) on the `Proteins` data set.

# Epilogue  6

## 6.1  Conclusion

Driven by ideas about and experiences with uncertainty from my physics background and a profound disapproval of black box modelling without trying to understand the underlying dynamics, this dissertation analysed how set-valued predictors can be used to get useful uncertainty estimates with valid statistical guarantees. The framework of conformal prediction provides exactly that. By now, many different conformal algorithms or methods exist and some specific classes were studied in detail, both using theoretical considerations and real-world data sets.

In Chapter 2, we got acquainted with the general theory of uncertainty, uncertainty quantification and, more importantly, conformal prediction. Some fundamental properties were highlighted and some interesting extensions were briefly covered. The essential idea behind conformal prediction is the robustness of rank statistics of a general class of distributions, those that satisfy the exchangeability property. Although the original formulation was not very computationally attractive, the methods that are currently in use admit very efficient implementations without sacrificing too much of their theoretical power.

After having introduced our beloved framework, Chapter 3 was the right place to study the main problem of interest: (univariate) regression. Regression problems with real-valued functions sadly have an important downside. The responses are continuous, which implies that, in general, the prediction regions will be infinite. However, one of the main characteristics that sets regression problems apart from other problems such as classification tasks is the ordering of the data. This allowed us to only predict the boundaries of prediction regions without having to bother with predicting every individual element. In this chapter, we compared different approaches to constructing prediction intervals and studied how they could be enhanced with conformal prediction as to obtain valid guarantees. Although some methods, mainly the ensemble-based ones, exhibited a strong performance by default, conformal prediction still emerged as the superior framework.

The main issue that remained was the fact that, on the one hand, one of the most widely used conformal predictors gives homoskedastic results and, on the other hand, standard conformal prediction only gives global or marginal guarantees. However, this leads to the model sacrificing or ignoring the difficult points in favour of the easily predictable points. In practice, this is exactly what we do not want, since the data points that are difficult to predict are usually the points that are of major interest. For this reason, Chapter 4 was concerned with the conditional and heteroskedastic incarnations of conformal prediction. The former approach — dubbed Mondrian or conditional conformal prediction — is a straightforward modification, where a separate conformal predictor is constructed for every class. A comparison between the normalized and Mondrian versions was made and a theoretical result was proven for when it is sufficient to simply use a single conformal predictor instead of constructing separate predictors. The main idea behind this theorem was Fisher's notion of pivotal quantities.

Finally, in Chapter 5, one of the shortcomings of the Mondrian approach was analysed. Conditional conformal prediction has very strong guarantees (the strongest for a given taxonomy function), but it does require a lot of data, since we have to construct a calibration set for every class. Clusterwise conformal prediction tries to overcome this data issue by grouping together similar or related classes. One possible way is by clustering the nonconformity distributions as this is optimal with respect to conformal prediction, while another possibility is to use side information to induce a notion of similarity. These two approaches can be related in certain situations, which led to the study of the Lispchitz continuity of the model and data-generating distribution. However, the approach using side information is preferable, whenever possible, since this allows us to incorporate an inductive bias and leverage the underlying structure. One interesting example is the existence of hierarchies over the class space. A small exploratory experiment was performed for (extreme) multiclass classification where functional hierarchies were provided. Preliminary results showed two things. First of all, that extreme classification is a rather difficult problem, especially for conditional uncertainty quantification. Secondly, that the clusterwise approach can improve over the conditional approach due to the increase in available data, but only marginally without further model improvements.

## 6.2 Future Perspectives

### 6.2.1 Clustering and hierarchies

In Chapter 5, the middle ground between marginal and classwise conformal prediction was treated, at least in an exploratory way. Instead of working with the initial taxonomy, we considered a clustered or coarsened one. Some general aspects were studied or highlighted already:

1. How this can be done, either in the nonconformity space (optimal with respect to conformal prediction) or in the feature/instance space (more informative and allows for inductive biases).

2. How these approaches are related, leading to studying the Lipschitz continuity of the predictive model.

3. How to incorporate side information in the form of hierarchies and how this allows to adaptively generate clusters.

4. How this can be used for (extreme) classification (and multitarget prediction).

Although all these points together make for an ambitious goal, the scope of this work did not allow for a detailed study of all these aspects (and of any other ones not listed above). Moreover, the experimental results showed that with the standard techniques, e.g. the softmax or APS measures, and the available hierarchy, the difficult, yet highly important, setting of extreme classification is not easily tackled. Due to the strong class imbalance, the marginal conformal predictors came out on top, especially in terms of efficiency. Finding a method, such as the APS measure (5.44), but which can handle the permutation problem, would be a major step forward.

### 6.2.2 General orders

The definition of conformal prediction in Section 2.3 uses nonconformity measures taking values in the set of real numbers $\mathbb{R}$. However, the only structure that is needed for Property 2.18 — the reason why conformal prediction works — is the order structure on $\mathbb{R}$. Consequently, it might feel unnecessarily restrictive to require that nonconformity measures take values in $\mathbb{R}$. It should be perfectly fine to work with nonconformity measures that take values in a total order (Definition A.2):

$$A : (\mathcal{X} \times \mathcal{Y})^* \to \mathbb{T} \tag{6.1}$$

for transductive models or

$$A : \mathcal{X} \times \mathcal{Y} \to \mathbb{T} \tag{6.2}$$

for inductive models. This makes essentially no difference, since any two finite total orders of the same cardinality are (order-)isomorphic. It follows that for the purpose of (finite) data sequences, the precise total order in which $A$ takes values does not matter.

Consider for example the case of hierarchical classification, where the hierarchy $\mathcal{H}$ is given by a tree (Definition A.3). When using a flat classifier, the most widely used nonconformity measure would be the softmax measure (5.39):

$$A_{\text{flat}}(x, c) := 1 - \widehat{P}(c \mid x), \tag{6.3}$$

where $\widehat{P} : \mathcal{H} \times \mathcal{X} \to [0, 1]$ is a probabilistic classifier. For a hierarchical classifier, however, we have access to other possibilities. The first one could be, for example, the length of the path from the true class to the predicted class:

$$A_{\text{graph}}(x, c) := d_{\text{graph}}\big(c, \hat{\rho}(x)\big), \tag{6.4}$$

where $\hat{\rho}$ is a class predictor $\hat{\rho} : \mathcal{X} \to \mathcal{H}$ and $d_{\text{graph}}$ is the *graph metric* induced by $\mathcal{H}$. (This essentially makes this method a particular instance of the general metric nonconformity measures defined in Eq. (3.37).) Note that for a tree-based hierarchy, $A_{\text{graph}}$ gives the sum of the distances from both the predicted class and the true class to their closest common ancestor. Since $\mathbb{N} \subset \mathbb{R}$, this is still a specific subcase of the class of real-valued nonconformity measures. A related function is obtained by taking the Cartesian product of Eqs. (6.3) and (6.4):

$$A_{\text{hier}}(x, c) := \big(d_{\text{graph}}\big(\xi(c, x), \hat{\rho}(x)\big), 1 - \widehat{P}(c \mid x)\big). \tag{6.5}$$

**Example** 6.1 (**Lexicographic order**). Let $(S, \leq)$ and $(T, \leqslant)$ be two posets. The lexicographic order on the product $S \times T$ is defined as follows:

$$s \leq s' \implies (s, t) \trianglelefteq (s', t') \qquad t \leqslant t' \implies (s, t) \trianglelefteq (s, t'), \tag{6.6}$$

for all $s, s' \in S$ and $t, t' \in T$. If $(S, \leq)$ and $(T, \leqslant)$ are totally ordered sets,

so is $(S \times T, \trianglelefteq)$.

By endowing $\mathbb{N} \times \mathbb{R}$ with the lexicographic order, we can apply conformal prediction directly with the nonconformity measure in Eq. (6.5). The idea is the following. If only the standard (flat) classification score is used, the conformal predictor might produce prediction sets that combine classes from very distant nodes. However, by first taking into account the graph distance within the hierarchy, the conformal predictor is forced to look at nearby nodes. This is similar to how the representation complexity (Definition 5.17) was used as a performance metric in Chapter 5.

> **Remark** 6.2 (**Branching factor**). Because of the definition of the lexicographic order, if the hierarchy has a high *branching factor*, this procedure might produce very large sets as soon as the critical score has a first component different from 0.

Another example, one in the regression setting, is where two (dependent) variables $Y_1, Y_2 : \Omega \to \prime Y$ are modelled at the same time. For example, the number of items and the total price of the items in a shopping bag. In this case, one of the two variables might be more important than the other and the lexicographic order could make more sense than a simple distance metric.[1]

We could go even further. Instead of the codomain $\mathbb{T}$ of the nonconformity measure being a total order, we could try to work with partial orders (Definition A.2), which are much more ubiquitous, but this is also where we run into trouble. Whereas the construction of quantiles gives unique values for total orders, this is not the case for partial orders, since not all elements are comparable. To this end, consider a measurable poset $(Q, \Sigma, \leq)$ equipped with a probability measure $P : \Sigma \to [0, 1]$ such that all sets of the form $y^\downarrow := \{x \in Q \mid x \leq y\}$ are measurable. If we would mirror the definition on $\mathbb{R}$, the $\alpha$-quantile would be the smallest element $q_\alpha \in Q$ such that $P(q_\alpha^\downarrow) \geq \alpha$. However, since there might be multiple incomparable elements satisfying this condition, we end up with a collection $B_\alpha$ of '$\alpha$-prequantiles' which then have to be aggregated in a meaningful way (Schollmeyer, 2017a, 2017b).

---

[1] Because of measure-zero issues, this is only relevant in the ordinal regression setting. More specifically, it is only relevant in the situation where the total order has a countable first factor.

### 6.2.3  Symmetries

An interesting concept in many fields of science, especially from the perspective of physics, is that of symmetries. Physics is famous for the construction of an almost complete theory of nature simply by starting from the notion of symmetry.[2] In machine learning, people have also started to realize the importance of symmetries. For example, CNNs have been shown to improve significantly when making the model invariant (or rather *equivariant*) with respect to symmetry groups, see e.g. Cohen and Welling (2016). Instead of augmenting the training data by adding rotated images, which has an impact on the training speed, the model architecture is modified in such a way that the orientation of the images becomes irrelevant.

The question of whether symmetries are relevant and, if so, how they can be implemented in conformal prediction can also be an interesting one. For example, recall Remark 2.1. Depending on the data-generating distribution, certain constructions of prediction sets are more natural than others and, moreover, choosing symmetric ones might lead to more efficient sets. This can also extend to symmetries in the feature space, i.e. when $P(Y \mid X) = P(Y \mid g \triangleright X)$ for some symmetry transformation $g$ (Definition A.6). It could be interesting to see whether nonconformity measures that take such symmetries into account, lead to more efficient prediction sets than those that do not.

Related to the idea of incorporating symmetries in confidence predictors and the introduction of Chapter 3 about how the structure of the target space can be leveraged, we also have the following remark.

> **Extra** 6.3 (**Measurement scales**). In statistics, data is often divided into four categories based on the structure present in the data (here shown in order of increasing structure):
>
> 1. nominal or categorical scale,
>
> 2. ordinal scale,
>
> 3. interval scale, and

---

[2] Even the notoriously annoying gravitational sector admits a symmetry-based description.

4. ratio scale.

In the first category, the target space has no real structure besides being a set and simply comes equipped with the discrete $\sigma$-algebra (Example A.10). For ordinal data, the target space is equipped with a (linear) order-theoretic structure (Definition A.2). Going one step further, the interval scale is (in a generalized sense) defined by an order structure together with the property that the difference between two values makes sense, but where no absolute zero exists. This algebraic structure corresponds to linearly ordered *affine spaces* (or *torsors* over linearly ordered groups). At last, the ratio scale is obtained by fixing the zero element, thereby forcing the *torsor* underlying the associated interval scale to actually become a group. [a]

---

[a] In the multivariate setting, these properties are usually only considered for each dimension independently, algebraic structures generally carry over to product spaces, but linear orders do not.

It could be an interesting venture to see how, or even whether, these algebraic properties can be leveraged to improve confidence prediction algorithms. In the next section, this idea of approaching confidence prediction from an algebraic point of view will be generalized even more.

### 6.2.4 Abstract nonsense

Given the content of the chapters in this dissertation and the vast literature on conformal prediction, it is clear that most of the research focuses on specializing the methods and ideas. However, in many other areas of science it has proven useful to study topics from a more abstract level. Especially given the intricate relations between conformal predictors — e.g. orders and inclusions, products and multivariate regression, marginalization, conditioning, ... — it could be interesting to study the more abstract structures behind conformal prediction. (A similar approach for probability theory was explored in Fritz (2020); Heunen et al. (2017).)

Some examples might help to see why this could be interesting:

- Just as functions map sets to sets and measurable functions map probability distributions to probability distributions (cf. Extra A.20), we could ask ourselves the question what the suitable maps are between conformal predictors and, by extension, general confidence predictors. Simply postcomposing (in an elementwise manner) with an ordinary function will not suffice, since this might ruin the validity of the model. For probability distributions or, more precisely, Markov kernels (Definition A.38), the solution is given by the Chapman–Kolmogorov equation:

$$(g \circ f)(A \mid x) = \int_{\mathcal{Y}} g(A \mid y) \, df(y \mid x). \tag{6.7}$$

Given the structure of the Giry monad (Extra A.20), this equation can be interpreted as the *Kleisli composition* of Markov kernels, which in turn form the *Kleisli morphisms* for the Giry monad (Fritz, 2020; Heunen et al., 2017). One possibility would be to interpret conformal predictors or, more generally, confidence predictors as (a subset of) set-valued functions, which are the *Kleisli morphisms* of the *powerset monad* and which compose as follows:

$$(g \circ f)(x) = \bigcup_{y \in f(x)} g(y). \tag{6.8}$$

- One of the aspects of conformal prediction that is often left aside is its application to multivariate regression (Diquigiovanni, Fontana, & Vantini, 2022; Messoudi, Destercke, & Rousseau, 2021). Due to the complex shapes prediction regions for such problems can have, efficient algorithms are hard to come up with. For this reason, we could try to decompose them into solutions of univariate problems and the resulting prediction regions might be reassembled. However, this will not work without modifying the significance levels of the individual predictors due to the problem of multiple hypothesis testing (Section 2.4). More generally, the question could be asked of how to aggregate/combine individual confidence predictors. Abstractly, this could be partially characterized by studying the right definitions of (co)products and/or (co)limits of conformal predictors.

- State-of-the-art research questions such as the incorporation of symmetries also factor in here, since symmetries are generally characterized by transformations that leave an object invariant (possibly up to some relaxations).

### 6.2.5   Multivariate conformal prediction

As noted in the preceding section, conformal prediction in the multivariate setting is computationally inefficient. Instead of seeing the problem of multiple hypothesis testing as a true issue, we could opt to treat it as a solution.[3] To this end, we could train a collection of univariate regressors and perform separate univariate hypothesis tests, i.e. for a target space of the form $\prod_{i=1}^{n} \mathcal{Y}_i$, we would perform $n \in \mathbb{N}$ hypothesis tests. However, as is well known in statistics, performing multiple tests at the same significance level $\alpha \in [0,1]$ would lead to a familywise error greater than $\alpha$ due to the multiple comparisons problem (Miller, 1981). To resolve this issue, we can, for example, apply a simple Bonferroni correction, where for every dimension, a conformal predictor at significance level $\alpha/n$ is constructed. However, it is also common knowledge that the Bonferroni correction strongly decreases the statistical power and more powerful tests, such as the Holm correction, where the $p$-values are sorted in increasing order and the $k^{\text{th}}$ value is compared to the modified significance level $\frac{\alpha}{n-k+1}$, are preferred. Moreover, such a sequential method also allows to assign a meaning or importance to the dimensions.

Besides the existing approaches and the possible approaches mentioned in this section, many other possibilities could be devised. Finding an efficient algorithm, in the simple spirit of univariate conformal regression, would be a major step forward in the field of multivariate confidence prediction.

---

[3]  Here, we crucially ignore the possible dependence between target dimensions since this would involve modelling the dependence structure as in Messoudi et al. (2021).

# Mathematical Preliminaries A

Although we often do not really care about the formalities of a framework in practice, from a theoretical point of view it is important to carefully treat all objects. For this reason, this appendix will (briefly[1]) cover the main notions and results that will be used throughout this dissertation. This mainly consists of the foundations of probability theory and machine learning.

## A.1 Set theory & Algebra

### A.1.1 Sets

For the purpose of data science, it will be important to consider collections of elements where duplicates are allowed. These will be called **multisets** or **bags** as is common in the conformal prediction literature (Vovk et al., 2022). The set of countable, meaning finite or countably infinite, multisets (or sequences) whose elements take value in a set $\mathcal{X}$ will be denoted by $\mathcal{X}^*$, where $*$ is called the **Kleene star**. It is the (disjoint) union of all finite (Cartesian) powers of $\mathcal{X}$:

$$\mathcal{X}^* := \bigcup_{i=1}^{+\infty} \mathcal{X}^i. \tag{A.1}$$

Note that what are called data *sets*, training *sets*, calibration *sets*, ... in data analysis are actually not sets but multisets, since we allow elements to be present multiple times.

Since a distinction is being made between sets and multisets, we also have to make a distinction between unions and disjoint unions. Taking the union

$$A \cup B := \{x \mid x \in A \lor x \in B\} \tag{A.2}$$

---

[1] Or, perhaps, not so briefly.

of two sets[2] $A, B$ would result in a set where elements that occur in both $A$ and $B$ are sent to the same element, i.e. copies are discarded. However, when taking unions of data sets this is not necessarily what we want. For this reason, a disjoint union should be used:

$$A \sqcup B := \big\{ (1, a) \mid a \in A \big\} \cup \big\{ (2, b) \mid b \in B \big\}. \tag{A.3}$$

When $A \cap B = \emptyset$, the disjoint union $A \sqcup B$ is easily seen to be isomorphic to the ordinary union $A \cup B$. However, to not overly complicate the notation throughout this dissertation, all unions will be denoted by $\cup$.

Aside from combining multiple sets, we will also sometimes need to decompose a set into smaller parts. One particular such way is of special importance.

**Definition** A.1 (**Partition**). A partition of a set $\mathcal{X}$ is a collection $\mathfrak{P} \subset 2^{\mathcal{X}}$ such that

$$\bigcup_{B \in \mathfrak{P}} B = \mathcal{X} \tag{A.4}$$

and

$$\forall A, B \in \mathfrak{P} : A \cap B = \emptyset. \tag{A.5}$$

## A.1.2    Order relations

**Definition** A.2 (**Order**). Let $S$ be a set equipped with a binary relation $\leq$. This relation can have different properties:

- $(S, \leq)$ is called a **partially ordered set** or **poset** if it satisfies:

    1. **Reflexivity**: $s \leq s$ for all $s \in S$.

    2. **Transitivity**: $r \leq s \wedge s \leq t \implies r \leq t$ for all $r, s, t \in S$.

    3. **Antisymmetry**: $r \leq s \wedge s \leq r \implies r = s$ for all $r, s \in S$.

- $(S, \leq)$ is called a **totally ordered set** (or **linearly ordered set**) if it is a

---

[2] Note that these should be subsets of some *universe* to avoid unbounded quantification in the above definition (Lavrov & Maksimova, 2012).

poset in which every two elements are comparable, i.e. $r \leq s \vee s \leq r$ for all $r, s \in S$.

**Definition** A.3 (**Tree**). Consider a set $S$. A tree with elements in $S$ is a set $T$ consisting of finite tuples in $S$ that is closed under taking initial segments (Kechris, 1995), i.e.

$$(s_1, \ldots, s_{n-1}, s_n) \in T \implies (s_1, \ldots, s_{n-1}) \in T \tag{A.6}$$

for all $s_1, \ldots, s_n \in S$. Equivalently, this is a poset with a minimal element in which the predecessors of any given element form are totally ordered.

### A.1.3 Algebra

**Definition** A.4 (**Group**). A set $G$ with a binary operation $\cdot : G \times G \to G$ is called a group if it satisfies the following conditions:

1. **Associativity**:

$$(a \cdot b) \cdot c = a \cdot (b \cdot c) \tag{A.7}$$

   for all $a, b, c \in G$.

2. **Unitality**: There exists a **neutral element** $e \in G$ such that

$$e \cdot g = g \cdot e = g \tag{A.8}$$

   for all $g \in G$.

3. **Inverses**: For every $g \in G$, there exists a $g^{-1} \in G$ such that

$$g \cdot g^{-1} = g^{-1} \cdot g = e. \tag{A.9}$$

**Example** A.5 (**Symmetric group**). The permutation group $S_n$ on $n \in \mathbb{N}$ elements is the group consisting of all bijections (or permutations) of the ordered $n$-element set $(1, \ldots, n)$.

To formalize the notion of a symmetry group, we need to express what it means for a group to act on a set. The simplest approach is to say that for every group element there is a function such that these interact according to the group laws.

**Definition** A.6 (**Group action**).  An action of a group $(G, \cdot, e)$ on a set $S$ is a binary operation $\rhd : G \times S \to S$ satisfying the following conditions:

1. **Compatibility**:

$$(g \cdot g') \rhd s = g \rhd (g' \rhd s) \qquad \text{(A.10)}$$

   for all $g, g' \in G$ and $s \in S$.

2. **Unitality**:

$$e \rhd s = s \qquad \text{(A.11)}$$

   for all $s \in S$.

Because all elements of a group are invertible, group actions also act through invertible functions.

When working with nested or sequential objects, each admitting a certain symmetry group, it is often useful to combine all of these groups together. One way to achieve this is the notion of a 'limit'.

**Definition** A.7 (**Direct limit**).  Consider a sequence of groups $(G_n)_{n \in \mathbb{N}}$ with group morphisms[a] $f_{ij} : G_i \to G_j$ for all $i \leq j \in \mathbb{N}$ such that:

1. **Identity**: $f_{ii} = \mathbb{1}_{G_i}$ for all $i \in \mathbb{N}$, where $\mathbb{1}_{G_i}$ denotes the identity function on $G_i$.

2. **Associativity**: $f_{jk} \circ f_{ij} = f_{ik}$ for all $i \leq j \leq k$.

The direct limit $\varinjlim G_n$ of this sequence is defined as the disjoint union $\bigsqcup_{n \in \mathbb{N}} G_n$ modulo the equivalence relation

$$g \in G_i \sim g' \in G_j \iff \exists f_{ik}, f_{jk} : f_{ik}(g) = f_{jk}(g'). \qquad \text{(A.12)}$$

This can be intuitively understood in the case where $(G_n)_{n \in \mathbb{N}}$ is a sequence of nested groups. In this case, two elements are identified if they 'eventually' become the same. For example, in the case of the symmetric groups $(S_n)_{n \in \mathbb{N}}$, the permutations that only interchange the first two elements in an ordered set are identified at every order.

---

[a] These are functions that preserve the group properties.

# A.2    Probability theory

## A.2.1    Measurability

In the beginning of the previous century, people realized that probability theory could be formally founded in the same mathematical theory that was being used to formalize notions of area and volume (Capiński & Kopp, 2004): 'measure theory'. The problem with assigning a volume to all subsets of a Euclidean space $\mathbb{R}^n$ was that, given the axioms of set theory, mathematicians could not consistently define such an operation. No matter how crazy this might sound, there is no way to consistently define a notion of volume for all subsets of Euclidean space, i.e. there is no way to 'measure' all these sets. A famous example is given by the *Vitali sets*.[3]

The relation between measuring sets and assigning probabilities should, after all, not come as a surprise. Consider, for example, the set $[n]$ of the first $n \in \mathbb{N}$ integers. What is the probability that a point, uniformly sampled from $[n]$, lies in a subset $S \subseteq [n]$? This is simply the size (or cardinality) of the subset divided by $n$:

$$\mathrm{Prob}(S) = \frac{|S|}{|[n]|} = \frac{|S|}{n}. \tag{A.13}$$

As such, calculating this probability is equivalent to determining the size of $S$, i.e. 'measuring' $S$. This uniform probability distribution corresponds to what is also called the **counting measure**, since it simply counts the number of elements in a set.

Sadly, in the case of infinite sets, the *axiom of choice* makes our lives slightly more miserable. For example, as Wikipedia so beautifully explains (Contributors, 2023), the *Banach–Tarski paradox* shows that there is no way to consistently define a notion of volume in three dimensions unless one of the following five concessions is made:

- The volume of a set changes when it is rotated.

- The volume of the union of two disjoint sets is different from the sum of their individual volumes.

---

[3]  The existence of nonmeasurable sets crucially depends on the *axiom of choice*, one of the most important, yet controversial axioms of set theory. As a consequence, they do not exist in *constructive mathematics*.

- Some sets are deemed 'nonmeasurable', and we need to check whether a set is 'measurable' before being able to talk about its volume.

- The axioms of ZFC[4] have to be altered.

- The volume of $[0, 1]^3$ is either 0 or $+\infty$.

In the case of measure theory, the third option is chosen, i.e. the whole procedure is simply turned around. Instead of starting from a measure and finding out that that not all sets are measurable, we start with a collection of sets that we would like to be measurable and study all measures consistent with this collection. This leads to the following notion.

**Definition** A.8 ($\sigma$-**algebra**). A $\sigma$-algebra on a set $\mathcal{X}$ is a collection of subsets $\Sigma \subseteq 2^{\mathcal{X}}$ that satisfies the following conditions:

1. **Triviality**: $\emptyset \in \Sigma$.

2. **Complements**: $A \in \Sigma \implies \mathcal{X} \backslash A \in \Sigma$.

3. **Countable unions**: $(A_n)_{n \in \mathbb{N}} \subseteq \Sigma \implies \bigcup\limits_{n=1}^{+\infty} A_n \in \Sigma$.

The elements of a $\sigma$-algebra are called **measurable sets** and the pair $(\mathcal{X}, \Sigma)$ is called a **measurable space**. (If the choice of $\sigma$-algebra is irrelevant, we will simply write $\mathcal{X}$.)

Before continuing the discussion about measure theory, let us first see why these conditions make sense. The triviality condition should be clear. If any set should be measurable, then let it at least be the empty one. It has no internal structure and it contains no elements, so measuring it should be trivial. The second condition is also quite straightforward. If we can measure $\mathcal{X}$ and a subset of it, then we should also be able to measure the complement. The fact that we can measure $\mathcal{X}$ itself is just a simple consequence of the first two conditions. For the third condition, we can argue that if every set $A_n$ can be measured and the results can be enumerated, we can iteratively combine the results to measure the union. However, what if someone asks how to measure intersections of measurable sets? The solution is given by a basic result from set theory that relates these operations.

---

**Property** A.9 (**De Morgan's laws**). De Morgan's laws state that

$$\mathcal{X} \setminus \bigcup_{n=1}^{+\infty} A_n = \bigcap_{n=1}^{+\infty} \mathcal{X} \setminus A_n \qquad (A.14)$$

and

$$\mathcal{X} \setminus \bigcap_{n=1}^{+\infty} A_n = \bigcup_{n=1}^{+\infty} \mathcal{X} \setminus A_n. \qquad (A.15)$$

Using one of these equalities, together with the second condition on $\sigma$-algebras, the intersection of measurable sets can be rewritten as a union of measurable sets and, hence, this intersection is itself measurable by virtue of the third condition above.

Now that we have seen the definition of measurable sets, it might also be a good idea to consider some examples to get a feeling of what $\sigma$-algebras might entail.

**Example** A.10 (**Trivial $\sigma$-algebras**). Two trivial examples of $\sigma$-algebras can be defined on any set $\mathcal{X}$, no matter the size or structure. These are the **trivial** (or **codiscrete**) $\sigma$-algebra

$$\Sigma_{\mathsf{codisc}}(\mathcal{X}) := \{\emptyset, \mathcal{X}\} \qquad (A.16)$$

and the **discrete** $\sigma$-algebra

$$\Sigma_{\mathsf{disc}}(X) := 2^{\mathcal{X}}. \qquad (A.17)$$

The first example can be interpreted as the situation where we are trying to measure the objects in a sealed box. The individual elements cannot be measured, only the box as a whole can be measured. The second example is the situation where we have perfect control over or knowledge about the whole set.

To obtain more interesting examples, we could pass to $\mathbb{R}^n$. However, just for fun (and because it is sometimes also of interest in practice), we can pass to an even more general setting, namely that of *topological spaces* (Bourbaki, 1995). Formally introducing these objects would lead us too far astray, so let it suffice to say that these allow to formalize what it means for a set to be 'open', 'closed', 'compact', etc. The definition of a *topology* is stated in terms of unions and intersections of sets and, hence, this structure always induces that of a measurable space.

> **Definition** A.11 (**Borel $\sigma$-algebra**). Consider a *topological space* $\mathcal{X}$ (Bourbaki, 1995). The Borel $\sigma$-algebra on $\mathcal{X}$ is the smallest $\sigma$-algebra containing all open sets of $\mathcal{X}$.

As usual, the prototypical example is the real line $\mathbb{R}$. In this setting, we know that every open set can be written as a (countable) union of open intervals. So, the Borel $\sigma$-algebra on $\mathbb{R}$ is the one 'generated' (in the sense of applying the operations defining a $\sigma$-algebra) by the open intervals. However, an open set is always the complement of closed set[5], so we also obtain that the Borel $\sigma$-algebra of $\mathbb{R}$ and, in fact, of any topological space also contains all closed sets.

> **Extra** A.12 (**Trivial topologies**). The trivial and discrete $\sigma$-algebras are actually the Borel $\sigma$-algebras of *codiscrete* and *discrete topologies*, respectively (Bourbaki, 1995).

As a last example, we consider the multisets in a measurable space. For the construction of (transductive) conformal predictors, we need to have a measurable structure on multisets. (Casual readers might prefer to skip this example.)

> **Example** A.13 (**Multisets**). Consider a measurable space $(\mathcal{X}, \Sigma)$. The set $\mathcal{X}^*$ can also be turned into a measurable space as follows. For every $n \in \mathbb{N}$, the product $\sigma$-algebra on $\mathcal{X}^n$ is defined to be the smallest $\sigma$-algebra such that all Cartesian products $\prod_{i=1}^{n} A_i$, where $A_i \in \Sigma$ for all $i \leq n$, are measurable.[a] Given these measurable spaces $(\mathcal{X}^n, \Sigma_n)$, we then take the (countable) disjoint union as in Eq. (A.1). The $\sigma$-algebra $\Sigma_*$ on $\mathcal{X}^*$ is defined such that[b]
>
> $$B \in \Sigma_* \iff B \cap \mathcal{X}^n \in \Sigma_n$$
>
> for all $B \subseteq \mathcal{X}^*$.
>
> ---
>
> [a] This is the smallest $\sigma$-algebra for which the projections $\pi_i : \mathcal{X}^n \to \mathcal{X}$ are measurable (see Definition A.18 further below).
>
> [b] This is the largest $\sigma$-algebra for which the inclusions $\iota_n : \mathcal{X}^n \hookrightarrow \mathcal{X}^*$ are measurable.

---

[5] This is how closed sets are defined in general topological spaces.

## A.2.2   Probability measures

Now that the notion of a measurable space has been introduced, it is time to move on. The first step as a true mathematician would be to ask which functions preserve the structure of a measurable space. However, to avoid having to immediately delve into the technicalities of set theory, it is better to first introduce the notion of a measure, which will serve as a motivation for further concepts.

---

**Definition** A.14 (**Measure**). Let $(\mathcal{X}, \Sigma)$ be a measurable space. A measure on $(\mathcal{X}, \Sigma)$ is a set function $\mu : \Sigma \to \overline{\mathbb{R}}$ satisfying the **Kolmogorov axioms**:

1. **Nonnegativity**: $\mu(A) \geq 0$ for all $A \in \Sigma$.

2. **Emptiness**: $\mu(\emptyset) = 0$.

3. **Countable additivity** (or $\sigma$-**additivity**): If $(A_n)_{n \in \mathbb{N}} \subset \Sigma$ are disjoint, then

$$\mu\left(\bigsqcup_{n=0}^{+\infty} A_n\right) = \sum_{n=0}^{+\infty} \mu(A_n). \tag{A.18}$$

The triple $(\mathcal{X}, \Sigma, \mu)$ is called a **measure space**. If $\mu(\mathcal{X}) = 1$, the measure is called a **probability measure** or **(probability) distribution**. It should be clear that any measure space for which $\mu(\mathcal{X}) < +\infty$ can be turned into a probability space by a suitable normalization (these are also said to be **finite**). If a measure space is not finite, but admits a countable cover by finite measures space, it is said to be $\sigma$-**finite**. The set of all probability measures on a set $\mathcal{X}$ will be denoted by $\mathbb{P}(\mathcal{X})$.

---

Requiring nonnegativity is simply a matter of convenience. There exist generalizations to so-called *signed measures*, but for most practical purposes, especially those of probability theory, the nonnegative ones suffice. The motivation for the third condition is the same as the one for the definition of $\sigma$-algebras. We want to be able to measure complex sets by decomposing them into smaller parts. This condition also has a more important consequence.

**Property** A.15. All measures are monotonic functions:

$$A \subseteq B \implies \mu(A) \leq \mu(B).  \tag{A.19}$$

At last, we come to the emptiness condition. For ordinary set functions $\kappa : \Sigma \to \mathbb{R}$, the $\sigma$-additivity condition would allow us to perform the following deduction:

$$\kappa(\emptyset) = \kappa(\emptyset \cup \emptyset) = 2\kappa(\emptyset) \implies \kappa(\emptyset) = 0.$$

However, a measure is allowed to take on the value $+\infty$, which makes this argument invalid. The second condition is simply there to exclude the highly degenerate possibility where the measure is identically $+\infty$.

As in the previous subsection, before introducing even more exotic concepts, we first give some examples of (probability) measures. The first one is simply the volume or **Lebesgue measure** on $\mathbb{R}^n$. It formalizes the way we measure the volume of everyday objects. On intervals, it is defined as follows (the choice of open, half-open or closed intervals does not matter):

$$\lambda\big(]a, b]\big) := b - a.  \tag{A.20}$$

To measure arbitrary Borel subsets of $\mathbb{R}$, we consider covers by such intervals and define the Lebesgue measure of a subset to be the infimum of the measures of its covers. Similarly, for higher-dimensional Euclidean spaces, we first define the Lebesgue measure of hyperrectangles $\mathbf{I} := I_1 \times I_2 \times \cdots I_n$ as the product of the lengths of the intervals:

$$\lambda(\mathbf{I}) := \lambda(I_1) \cdots \lambda(I_n),  \tag{A.21}$$

and again define the Lebesgue measure of arbitrary Borel subsets as the infimum of the measures of covers by such hyperrectangles. This procedure can be generalized to arbitrary ($\sigma$-**finite**) measure spaces to obtain a (**tensor**) **product** of measure spaces.

Equation (A.20) has the interpretation of the length of the interval. But what if instead of simply taking the values of the endpoints, we first apply a function $F : \mathbb{R} \to \mathbb{R}$? In this case, the expression becomes:

$$\lambda_F\big(]a, b]\big) := F(b) - F(a).  \tag{A.22}$$

To make this formula well-defined, we need $F$ to be right-continuous, i.e.

$$\lim_{\varepsilon \to 0^+} F(x + \varepsilon) = F(x)  \tag{A.23}$$

for all $x \in \mathbb{R}$. Moreover, to obtain a nonnegative and monotonic measure, we also need to require that $F$ is itself nonnegative and monotonic. But wait a minute! Does Eq. (A.22), together with the properties of $F$, not remind us of a well-known object from probability theory? Indeed, all **cumulative distribution functions** (CDFs) are exactly of this form. In the context of measure theory these are called **Lebesgue–Stieltjes measures**.

> **Extra** A.16 (**Càdlàg**). Functions that are right-continuous and that admit all left limits are also said to be càdlàg (abbreviation of the French expression "*continue à droite, limite à gauche*"). It can be shown that càdlàg functions $F : \mathbb{R} \to [0, 1]$ satisfying
>
> $$\lim_{x \to -\infty} F(x) = 0 \qquad \text{and} \qquad \lim_{x \to \infty} F(x) = 1 \qquad (A.24)$$
>
> are equivalent to cumulative distribution functions on $\mathbb{R}$.

The fact that CDFs only need to be right-continuous, allows them to have jump discontinuities, i.e. isolated points where the value of the function suddenly changes in a discontinuous way. The measure of a singleton can be shown to be determined by exactly that jump:

$$\lambda_F(\{x_0\}) = F(x_0) - \lim_{\varepsilon \to 0^+} F(x_0 - \varepsilon). \qquad (A.25)$$

Singletons with nonzero measure are examples of **atoms**, sets with nonzero measure for which every proper subset has vanishing measure. (By classical arguments from calculus, it can be shown that all atoms of a Lebesgue–Stieltjes measure are necessarily singletons.) The Lebesgue measure does not have any atoms since it is induced by the identity function. More generally, for continuous CDFs, all singletons are **null sets**, i.e. sets with measure zero. Null sets are in some sense the subsets that we can forget about when talking about a property in probabilistic terms. If a property holds everywhere, except for some null set, it is said to hold **almost everywhere** (**a.e.**) in measure theory or **almost surely** (**a.s.**) in probability theory.

Now that the notion of an atom has been introduced, it is time to move to another class of examples, namely the **discrete probability measures**. Discrete probability measures are **atomic** in the sense that any measurable set of nonzero measure contains an atom. More specifically, discrete measures are of the form

$$\mu = \sum_{i=1}^{+\infty} \lambda_i \delta_{x_i}, \qquad (A.26)$$

where all $\lambda_i \in \mathbb{R}^+$ and

$$\delta_{x_i}(A) := \mathbb{1}_A(x_i) = \begin{cases} 1 & \text{if } x_i \in A, \\ 0 & \text{if } x_i \notin A \end{cases} \tag{A.27}$$

are so-called **Dirac measures**. Common examples of such measures are the Poisson and binomial distributions.

> **Property** A.17 (**Simplex**).  By the normalization property of probability measures, every distribution on a (discrete) finite space $[k]$ can be represented as an element of the $(k-1)$-simplex
>
> $$\Delta^{k-1} := \left\{ x \in \mathbb{R}^k \,\middle|\, \sum_{i=1}^{k} x^i = 1 \right\}. \tag{A.28}$$

As promised, there is one last important notion to be introduced, namely the functions between measurable spaces. To this end, consider two measurable spaces $(\mathcal{X}, \Sigma_\mathcal{X})$ and $(\mathcal{Y}, \Sigma_\mathcal{Y})$. Given a function $f : \mathcal{X} \to \mathcal{Y}$, it might be tempting to induce a $\sigma$-algebra on $\mathcal{Y}$ generated by the images $\{ f(A) \mid A \in \Sigma_\mathcal{X} \}$. However, a set-theoretic problem arises at this point. To make this construction work, functions should preserve the operations defining a $\sigma$-algebra and, although empty sets and unions are preserved, complements are not: $f(A \backslash B) \neq f(A) \backslash f(B)$. This has as a consequence that we cannot 'pull back' measures from $\mathcal{Y}$ to $\mathcal{X}$ by the would-be definition

$$f^* \mu(A) := \mu\big(f(A)\big).$$

Luckily, there is another possibility.  The image might not preserve all required operations, but the preimage does, i.e. $f^* \Sigma_\mathcal{Y} := \{ f^{-1}(A) \mid A \in \Sigma_\mathcal{Y} \}$ is a $\sigma$-algebra, where the **preimage** is defined as follows:

$$f^{-1}(A) := \{ x \in \mathcal{X} \mid f(x) \in A \}. \tag{A.29}$$

So, instead of pushing forward measurable sets and pulling back measures, we should work the other way around. This leads to the following definition.

> **Definition** A.18 (**Measurable function**).  Consider a function
>
> $$f : (\mathcal{X}, \Sigma_\mathcal{X}) \to (\mathcal{Y}, \Sigma_\mathcal{Y}) \tag{A.30}$$
>
> between measurable spaces. This function is itself said to be measurable if and only if the pullback $\sigma$-algebra $f^*\Sigma_{\mathcal{Y}}$ is a sub-$\sigma$-algebra of $\Sigma_{\mathcal{X}}$, i.e. if

$$f^{-1}(A) \in \Sigma_{\mathcal{X}} \qquad \text{for all} \qquad A \in \Sigma_{\mathcal{Y}}. \tag{A.31}$$

Equipped with the measurable functions, we can now also transport measures between spaces.

**Definition** A.19 (**Pushforward**). The pushforward of a measure $\mu$ on $\mathcal{X}$ along a measurable function $f : \mathcal{X} \to \mathcal{Y}$ is defined by

$$f_*\mu(A) := \mu\big(f^{-1}(A)\big). \tag{A.32}$$

This is well defined, since by definition of measurability, $f^{-1}(A)$ is measurable in $\mathcal{X}$.

The following remark is of interest in the setting of ensemble methods and generalized probability frameworks such as *credal sets*. See Section A.4.2 and Augustin et al. (2014), respectively.

**Extra** A.20 (**Giry monad**). By passing to second-order probability distributions, i.e. distributions over distributions, we obtain even more structure. Consider a distribution $\mathfrak{p} \in \mathbb{P}^2(\mathcal{X})$ over the set of probability distributions on a measurable space $\mathcal{X}$. Integration (see the next section) induces the following mapping $\mu : \mathbb{P}^2(\mathcal{X}) \to \mathbb{P}(\mathcal{X})$:

$$\mu : \mathfrak{p} \mapsto \int_{\mathbb{P}(\mathcal{X})} P \, \mathrm{d}\mathfrak{p}(P), \tag{A.33}$$

where $\mathbb{P}(\mathcal{X})$ is equipped with the $\sigma$-algebra generated by the evaluation functionals $\mathrm{ev}_A : P \mapsto P(A)$, with $A$ ranging over all measurable subsets of $\mathcal{X}$. This structure of iterated probability spaces is sometimes called the **Giry monad** (Giry, 1982; Lawvere, 1962).

For completeness' sake, it is also worth mentioning that what is called a **random variable** in probability theory, is simply a measurable function from a probability space into an arbitrary measurable space (e.g. $\mathbb{R}$ in the case of univariate regression). The **distribution** of a random variable $X : (\Omega, \Sigma, P) \to (\mathcal{X}, \Sigma)$ is then defined as the pushforward of $P$ along this random variable:

$$P_X := X_*P. \tag{A.34}$$

Equipped with the notions of measurable functions and group actions (Section A.1.3), it is now possible to state the definition of invariant (probability) measures.

**Definition** A.21 (**Invariance**).  A measure space $(\mathcal{X}, \Sigma, \mu)$ is said to be **invariant** with respect to a measurable function $f : \mathcal{X} \to \mathcal{X}$ if the pushforward along $f$ leaves the measure invariant:

$$f_* \mu = \mu \, . \tag{A.35}$$

The function $f$ is also said to be **measure-preserving**.

If $\mathcal{X}$ comes equipped with a group action $\triangleright : G \times \mathcal{X} \to \mathcal{X}$ such that

$$g \triangleright \cdot : \mathcal{X} \to \mathcal{X}$$

is measure-preserving for all $g \in G$, then $\mu$ is also said to be $G$-**invariant**.

## A.2.3    Integration

One of the most powerful benefits of measure theory is that it allows for a solid theory of integration that can even be applied to situations where the ordinary Riemann integral breaks down.  We will not delve too deep into this subject, but some core ideas and notions are of importance to us.  The idea behind Lebesgue integration is to approximate functions by so-called simple functions (this is the approach introduced by Daniell (1918)).

**Definition** A.22 (**Simple function**).  A function $f : \mathcal{X} \to \mathbb{R}^+$ of the form

$$f(x) = \sum_{i=1}^{n} a_i \mathbb{1}_{A_i}(x) \tag{A.36}$$

for positive numbers $a_1, \dots, a_n \in \mathbb{R}^+$ and disjoint (measurable) subsets $A_1, \dots, A_n \in \Sigma_{\mathcal{X}}$.  The **Lebesgue integral** of the simple function $f$ with respect to a measure $\mu$ on $(\mathcal{X}, \Sigma_{\mathcal{X}})$ is defined as

$$\int_{\mathcal{X}} f \, \mathrm{d}\mu := \sum_{i=1}^{n} a_i \mu(A_i) \, . \tag{A.37}$$

To define the integral for general measurable functions, we should show that any nonnegative measurable function can be approximated (pointwisely) by a sequence of simple functions and define the integral as the supremum of the integrals over all simple functions bounded from above by it.  For more information, see Capiński and Kopp (2004).

**Definition** A.23 (**Integrable function**). A measurable function $f : \mathcal{X} \to \mathbb{R}$ is said to be (Lebesgue-)integrable with respect to a measure $\mu$ on $\mathcal{X}$ if both

$$\int_{\mathcal{X}} f^+ \, \mathrm{d}\mu < +\infty \qquad\qquad (\text{A.38})$$

and

$$\int_{\mathcal{X}} f^- \, \mathrm{d}\mu < +\infty \qquad\qquad (\text{A.39})$$

hold, where $f^+ := \max(f, 0)$ and $f^- := \max(0, -f)$. If $f$ is integrable, its (**Lebesgue-)integral** is defined as

$$\int_{\mathcal{X}} f \, \mathrm{d}\mu := \int_{\mathcal{X}} f^+ \, \mathrm{d}\mu - \int_{\mathcal{X}} f^- \, \mathrm{d}\mu . \qquad\qquad (\text{A.40})$$

If only one of the two conditions holds, $f$ is said to be **quasiintegrable**.

**Remark** A.24 (**Absolute integrability**). A crucial difference exists with the Riemannian case, where these conditions would imply *absolute integrability*, i.e.

$$\int_{\mathcal{X}} |f| \, \mathrm{d}\mu < +\infty . \qquad\qquad (\text{A.41})$$

With Lebesgue integrals, positive and negative infinity cannot cancel each other out (even in the limiting sense). Measurable functions are integrable if and only if they are absolutely integrable.

However, on a bounded interval, every Riemann-integrable function is also (Lebesgue-)integrable and the integrals coincide. Moreover, if the improper Riemann integral of a nonnegative function exists, it equals the Lebesgue integral of that function. [a]

---

[a] The nonnegativity condition is crucial in this statement. There exist improper Riemann integrals for which there exists no Lebesgue counterpart.

We give a simple example of a situation where Riemann integration does not suffice.

**Example** A.25 (**Atomic measures**). A theorem by Lebesgue (Capiński & Kopp, 2004) says that a bounded function is Riemann-integrable ex-actly when its set of discontinuities is a null set, i.e. when it is almost everywhere continuous.  The reason for this is that the Lebesgue measure is nonatomic: $\lambda(\{x\}) = 0$ for all $x \in \mathbb{R}$.  We can change the value of a function at any countable collection of points without changing the value of its integral.

However, many measures, such as the discrete probability distributions, are atomic, the simplest one being the Dirac measure (A.27).  The integral with respect to this measure is particularly straightforward to calculate:

$$\int_{\mathbb{R}} f \, d\delta_x = f(x). \tag{A.42}$$

This is one of the formulas that a physics student learns to accept without proper formal reasoning and where physicists have found all kinds of informal arguments and approximation methods because most refuse to accept any other integral besides the Riemann integral.

Integration theory is also important to define various notions in probability theory, one of the most common ones being the (central) moments.

**Definition** A.26 (**Moment**).  Consider a probability distribution $P_X \in \mathbb{P}(\mathbb{R})$.  For every $n \in \mathbb{N}$, its $n^{\text{th}}$ **moment** is defined as follows:

$$\mu_n(X) \equiv \langle X^n \rangle \equiv \mathsf{E}[X^n] := \int_{\mathbb{R}} x^n \, dP_X. \tag{A.43}$$

The first moment $\mathsf{E}[X]$ is called the **expectation** (**value**) or **mean**.  The $n^{\text{th}}$ **central moment** is defined by first subtracting the mean:

$$m_n(X) := \mathsf{E}\Big[\big(X - \mathsf{E}[X]\big)^n\Big] = \int_{\mathbb{R}} \big(x - \mathsf{E}[X]\big)^n \, dP_X. \tag{A.44}$$

The second central moment is called the **variance**.

**Method** A.27 (**Error propagation**).  Central moments transform nicely under affine transformations:

$$m_n(\alpha X + \beta) = \alpha^n m_n(X). \tag{A.45}$$

It follows that by approximating an arbitrary differentiable function by its first-order Taylor expansion, the variance of a transformed random variable can be approximated as follows (this is for example used in the *delta method* (Doob, 1935)):

$$\text{Var}\big[f(X)\big] \approx \big|f'\big(\text{E}[X]\big)\big|^2 \text{Var}[X]. \tag{A.46}$$

Such an approximation will be justified when the distribution of $X$ is peaked around its mean $\text{E}[X]$.

## A.2.4 Densities

On many occasions, especially in probability theory, measures are given by summing a sequence or integrating a function. For example, the standard normal distribution admits such a **probability density function** (PDF):

$$\Phi(x) = \int_{-\infty}^{x} \frac{1}{\sqrt{2\pi}} \exp\big(-t^2/2\big)\, dt. \tag{A.47}$$

This situation is an example of a more general concept.

**Definition** A.28 (**Absolutely continuous measure**). A measure $\nu$ on a measurable space $(\mathcal{X}, \Sigma)$ is said to be absolutely continuous with respect to a measure $\mu$ on $(\mathcal{X}, \Sigma)$ if

$$\mu(A) = 0 \implies \nu(A) = 0 \tag{A.48}$$

for all $A \in \Sigma$. This is often denoted by $\mu \gg \nu$.

The following very important result states that absolutely continuous measures admit density functions.

**Theorem** A.29 (**Radon–Nikodym**). If $\mu$ and $\nu$ are $(\sigma\text{-})$finite measures on a measurable space $(\mathcal{X}, \Sigma)$ such that $\nu$ is absolutely continuous with respect to $\mu$, there exists a $\mu$-a.e. unique measurable function $f : \mathcal{X} \to [0, +\infty[$ such that

$$\nu(A) = \int_A f\, d\mu \tag{A.49}$$

for all $A \in \Sigma$. The function $f$ is called the **Radon–Nikodym derivative**

and is sometimes denoted by $\frac{d\nu}{d\mu}$ in analogy to the ordinary derivative from calculus.

This generalized notion of density function also allows us to treat **probability mass functions** (PMFs) on equal footing with their continuous counterparts. PMFs are simply the Radon–Nikodym derivatives of discrete probability measures with respect to the counting measure (on, for example, $\mathbb{Z}$), while PDFs are the derivatives with respect to the Lebesgue measure.

**Property A.30 (Change of variables).** Let $f : (\mathcal{X}, \Sigma_{\mathcal{X}}) \to (\mathcal{Y}, \Sigma_{\mathcal{Y}})$ be a measurable function and consider a measure $\mu$ on $\mathcal{X}$. The following equality holds for all integrable functions $g : \mathcal{Y} \to \mathbb{R}$:

$$\int_{f^{-1}(\mathcal{Y})} (g \circ f) \, d\mu = \int_{\mathcal{Y}} g \, d(f_* \mu). \tag{A.50}$$

**Corollary A.31 (Density functions).** The change-of-variables formula for absolutely continuous measures on $\mathbb{R}$ with Radon–Nikodym derivative $f_X : \mathbb{R} \to \mathbb{R}$ implies

$$f_{g_* X}(y) = f_X\big(g^{-1}(y)\big) \left| \frac{dg^{-1}}{dy}(y) \right| = \frac{f_X\big(g^{-1}(y)\big)}{\big|g'\big(g^{-1}(y)\big)\big|} \tag{A.51}$$

when $g : \mathbb{R} \to \mathbb{R}$ is invertible and

$$f_{g_* X}(y) = \sum_{x \in g^{-1}(y)} \frac{f_X(x)}{|g'(x)|} \tag{A.52}$$

in general.

**Example A.32 (Dirac measure).** Recall the Dirac measure (A.27). Under a pushforward along the function $f : \mathbb{R} \to \mathbb{R}$, it transforms as

$$\delta\big(f(x)\big) = \sum_{y \in f^{-1}(0)} \frac{\delta(x - y)}{|f'(y)|}. \tag{A.53}$$

Another consequence of the change-of-variables formula is the following infamous result.

**Corollary A.33 (Law of the unconscious statistician).** Let $g : \mathcal{X} \to \mathbb{R}$

be an integrable function and $X$ a random variable on $\mathcal{X}$.

$$E\big[g(X)\big] = \int_{\mathcal{X}} g \, \mathrm{d}P_X \qquad (A.54)$$

Because of their relevance for this dissertation, some examples of probability density functions are given.

**Example** A.34 (**Pareto distribution**). The Pareto distribution with scale parameter $\lambda > 0$ and shape parameter $\xi > 0$ is given by the following CDF:

$$F_{\mathrm{Pareto}}(x; \lambda, \xi) = 1 - \left(\frac{\lambda}{x}\right)^{\alpha} \mathbb{1}_{[\lambda, +\infty[}(x)\,. \qquad (A.55)$$

The associated PDF is given by

$$f_{\mathrm{Pareto}}(x; \lambda, \xi) = \frac{\xi \lambda^{\xi}}{x^{\xi+1}} \mathbb{1}_{[\lambda, +\infty[}(x)\,. \qquad (A.56)$$

Note that this distribution is supported on the interval $[\lambda, +\infty[$.

**Example** A.35 (**Beta distribution**). The Beta distribution with shape parameters $\alpha, \beta > 0$ is given by the following PDF:

$$f_{\mathrm{Beta}}(x; \alpha, \beta) := \frac{\Gamma(\alpha + \beta)}{\Gamma(\alpha)\Gamma(\beta)} x^{\alpha-1} (1-x)^{\beta-1}\,. \qquad (A.57)$$

## A.2.5   Conditional probabilities

Without going into too much detail, some things have to be said about the notion of conditional probabilities.[6]

**Definition** A.36 (**Conditional probability**). Let $(\Omega, \Sigma, P)$ be a probability space. The conditional probability with respect to an event $A \in \Sigma$ is defined as follows:

$$P(B \mid A) := \frac{P(A \cap B)}{P(A)}\,. \qquad (A.58)$$

Note that this formula is only well defined when $A$ is not $P$-null. A (par-tial) solution exists when it is possible to find sequences $(A_n)_{n \in \mathbb{N}}$ of measurable sets of strictly positive probability converging to $A$. In this case, we could define

$$P(B \mid A) := \lim_{n \to \infty} P(B \mid A_n).$$  (A.59)

---

[6] The notions of *conditional moments* and *regular conditional distributions* will be left aside here.

However, it can be shown that this does not lead to a well-defined probability distribution, since the resulting value will depend on the choice of sequence (cf. the *Borel–Kolmogorov paradox*).

Whenever conditional probabilities exist, this definition immediately implies one of the most famous theorems in probability theory.

**Theorem** A.37 (**Bayes' theorem**).

$$P(A \mid B)P(B) = P(B \mid A)P(A)$$  (A.60)

One way to express conditional probabilities, which closely follows our intuition, is by modelling them as parametrized probability distributions.

**Definition** A.38 (**Markov kernel**). A Markov kernel with source space $(\mathcal{X}, \Sigma_{\mathcal{X}})$ and target space $(\mathcal{Y}, \Sigma_{\mathcal{Y}})$ is a function $P : \mathcal{X} \to \mathbb{P}(\mathcal{Y})$ such that:

1. **Measurability**: For every $B \in \Sigma_{\mathcal{Y}}$, the map $x \mapsto P(B \mid x)$ is measurable.

2. **Probability**: For every $x \in \mathcal{X}$, the map $B \mapsto P(B \mid x)$ is a probability distribution.

## A.2.6    Probabilistic relations

Three types of relations between random variables are considered throughout this dissertation. Here, we briefly introduce them.

**Definition** A.39 (**Independence**). Two random variables $X, Y : \Omega \to \mathcal{X}$ are said to be independent if their joint distribution factorizes as a product distribution:

$$P_{X,Y} = P_X \otimes P_Y.$$  (A.61)

**Definition A.40 (Equality in distribution).** Two random variables (not necessarily on the same base space) are said to be equal in distribution if their induced distributions coincide. This is denoted by $X \stackrel{d}{=} Y$.

**Definition A.41 (Stochastic dominance).** A CDF $F$ is said to dominate a CDF $G$ if

$$F(x) \geq G(x) \tag{A.62}$$

for all $x \in \mathbb{R}$.

## A.2.7 Quantiles

**Definition A.42 (Quantiles).** To every cumulative distribution $F : \overline{\mathbb{R}} \to [0, 1]$ we can assign a quantile function $Q : [0, 1] \to \overline{\mathbb{R}}$. This function can be defined in two (equivalent) ways:

- As a lower bound:

$$Q(\alpha) := \inf\left\{ x \in \overline{\mathbb{R}} \mid \alpha \leq F(x) \right\}, \tag{A.63}$$

- or through a *Galois connection*:

$$Q(\alpha) \leq x \iff \alpha \leq F(x). \tag{A.64}$$

When $F$ is invertible, the quantile function is easily seen to be its inverse: $Q = F^{-1}$.

For every $\alpha \in [0, 1]$, we can also define the empirical quantile functions

$$q_\alpha : \mathbb{R}^* \to \mathbb{R} \tag{A.65}$$

that assign to every (multi)subset $S$ of $\mathbb{R}$ the value $Q_S(\alpha)$, where $Q_S$ is the quantile function of the empirical distribution of $S$. For a given sorted data set $\{x_1, \ldots, x_n\}$, the quantile function associated to the empirical distribution $\hat{F}_n$ satisfies the following simple rule:

$$\widehat{Q}_n(\alpha) := q_\alpha\left(\{x_1, \ldots, x_n\}\right) = x_{(\lceil n\alpha \rceil)}. \tag{A.66}$$

**Example** A.43 (**Median**).  The 0.5-quantile of a cumulative distribution function $F$ is, by definition, given by the point smallest $x \in \mathbb{R}$ such that $F(x) \geq 0.5$.  This also means that it is the smallest real number such that at least 50% of the points sampled from $F$ would be smaller than or equal to $x$.

Since conformal prediction relies heavily on the use of sample quantiles, the properties of this estimator are of relevance as well.  Here, we only consider the most important one.

**Definition** A.44 (**Consistency**).  Consider a parameter $\theta \in \Theta$ of a distribution $P \in \mathbb{P}(\mathcal{X})$.  An estimator $(T_n : \mathcal{X}^n \to \Theta)_{n \in \mathbb{N}}$ is said to be consistent for $\theta$ if $T_n$ **converges in probability** to $\theta$:

$$\lim_{n \to \infty} P_\theta^n \big( |T_n(X) - \theta| \geq \varepsilon \big) = 0 \qquad (A.67)$$

for every $\varepsilon > 0$.

**Property** A.45 (**Sample quantiles**).  Let $F$ be a cumulative distribution function and choose $\alpha \in [0, 1]$.  If $Q(\alpha)$ is unique in the sense that

$$F\big( Q(\alpha) + \varepsilon \big) > F\big( Q(\alpha) \big) \qquad (A.68)$$

for all $\varepsilon > 0$, then the sample quantile $\widehat{Q}_n(\alpha)$ is a consistent estimator (Zieliński, 1998).  Moreover, if $F$ admits a density function $f$ in a neighbourhood around $Q(\alpha)$, the following result holds:

$$\sqrt{n}\big( \widehat{Q}_n(\alpha) - Q(\alpha) \big) \xrightarrow{d} \mathcal{N}\left( 0, \frac{\alpha(1 - \alpha)}{\big[ f(Q(\alpha)) \big]^2} \right).$$

## A.2.8   Law of large numbers

**Theorem** A.46 (**Law of large numbers**[a]).  Let $(X_n)_{n \in \mathbb{N}}$ be a sequence of i.i.d. random variables with expectation $\mu$.

$$\lim_{n \to \infty} \frac{1}{n} \sum_{i=0}^{n} X_i = \mu \qquad \text{a.s.} \qquad (A.69)$$

---

[a] This is actually the 'strong law of large numbers', also known as Kolmogorov's law.

The *weak law* (or Kinchin's law) states that this convergence holds in distribution.

**Corollary** A.47. Consider a sequence of i.i.d. random variables $(X_n)_{n \in \mathbb{N}}$ with distribution $P \in \mathbb{P}(\mathcal{X})$ and, for any $n \in \mathbb{N}$ and $x \in \mathcal{X}$, consider the random variable $\mathbb{1}_{]-\infty,x]}(X_n)$. Since

$$\mathsf{E}\left[\mathbb{1}_{]-\infty,x]}(X_n)\right] = P(X_n \leq x),\qquad (A.70)$$

the law of large numbers implies that

$$\lim_{n \to \infty} \frac{1}{n} \sum_{i=0}^{n} \mathbb{1}_{]-\infty,x]}(X_i) = P(X \leq x) \qquad \text{a.s.}\qquad (A.71)$$

The average on the left-hand side is the empirical distribution function $\widehat{F}$. Hence, by the law of large numbers, the empirical distribution function converges to the true distribution function almost surely.

In a similar way, for every $P$-integrable function $f : \mathcal{X} \to \mathcal{Y}$, the average over a large number of i.i.d. draws converges to the expectation value almost surely:

$$\lim_{n \to \infty} \frac{1}{n} \sum_{i=0}^{n} f(X_i) = \mathsf{E}_P[f] \qquad \text{a.s.}\qquad (A.72)$$

This is the idea behind Monte Carlo methods.

## A.2.9 Stochastic processes

In ordinary statistics, we often consider a single random variable $X : \Omega \to \mathcal{X}$. The typical setting would be the i.i.d. setting, where multiple values $X(\omega_i)$ are observed, with $\omega_i$ sampled independently from some base measure on $\Omega$. However, there exists another important type of data in statistics and probability theory: sequences of random variables $\{X_n : \Omega \to \mathcal{X}\}_{n \in \mathbb{N}}$, where the distributions $P_{X_n} \in \mathbb{P}(\mathcal{X})$ are not necessarily identical. Such a sequence of random variables is often called a **discrete-time stochastic process**. (We do not consider continuous-time stochastic processes since we will not need them.)

**Definition** A.48 (**Adapted process**). Let $\{X_n : \Omega \to \mathcal{X}\}_{n \in \mathbb{N}}$ be a stochastic process and consider a $\sigma$-algebra $\Sigma$ on $\Omega$. A sequence $(\Sigma_n)_{n \in \mathbb{N}}$ of sub-

$\sigma$-algebras of $\Sigma$ is called a **filtration** (of $\Sigma$) if $\Sigma_m \subseteq \Sigma_n$ whenever $m \leq n$. The process is said to be adapted to this filtration if $X_n$ is $\Sigma_n$-measurable for all $n \in \mathbb{N}$.

A filtration represents the amount of information that is accessible at every time $n \in \mathbb{N}$. The monotonicity condition then represents the simple assumption that as time progresses, no information can be lost.

One of the most essential types of stochastic processes is the following one.

**Definition** A.49 (**Martingale**). An integrable stochastic process $(X_n)_{n \in \mathbb{N}}$ on a probability space $(\Omega, \Sigma, P)$, adapted to a filtration $(\Sigma_n)_{n \in \mathbb{N}}$, is called a $(P\text{-})$martingale if it satisfies

$$\mathsf{E}\big[X_{n+1} \,|\, \Sigma_n\big] = X_n \qquad P\text{-a.s.} \tag{A.73}$$

for all $n \in \mathbb{N}$. When the equals sign is replaced by a smaller-than-or-equals (resp. greater-than-or-equals) sign, the stochastic process is said to be a **supermartingale** (resp. **submartingale**).

Martingales represent the situation where the best guess of the future is simply given by the current state of the system. (It is not hard to see that the definition implies $\mathsf{E}\big[X_{n+k} \,|\, \Sigma_n\big] = X_n$ for any $k \in \mathbb{N}$.) A basic result in martingale theory gives a bound on how fast a supermartingale can increase. In fact, this result by Ville predates martingale theory as recalled in Doob (1940).

**Property** A.50 (**Doob–Ville inequality**). Let $(X_n)_{n \in \mathbb{N}}$ be a nonnegative supermartingale.

$$\mathsf{Prob}\bigg(\max_{0 \leq k \leq T} X_k \geq C\bigg) \leq \frac{\mathsf{E}\big[X_0\big]}{C} \tag{A.74}$$

for all $C > 0$ and $T \in \mathbb{N}$.

*Proof*. This follows, for example, from Theorem 3.8(ii) in Karatzas and Shreve (1991) when taking the submartingale to be $-X_t$.          $\square$

## A.3   Metric spaces

Although this topic will be of restricted interest in this dissertation, we do include it for completeness.

## A.3.1   General notions

**Definition** A.51 (**Metric space**)**.** Let $\mathcal{X}$ be a set. A metric (or **distance function**) on $\mathcal{X}$ is a function $d : \mathcal{X} \times \mathcal{X} \to \mathbb{R}$ satisfying the following conditions:

1. **Nondegeneracy**: $d(x, y) = 0 \iff x = y$ for all $x, y \in \mathcal{X}$.

2. **Symmetry**: $d(x, y) = d(y, x)$ for all $x, y \in \mathcal{X}$.

3. **Triangle inequality**: $d(x, z) \leq d(x, y) + d(y, z)$ for all $x, y, z \in \mathcal{X}$.

Note that these conditions automatically imply that the metric is always nonnegative. If the nondegeneracy condition is relaxed and only nonnegativity is required, the notion of a **pseudometric** is obtained. The set of all metrics on a set $\mathcal{X}$ will be denoted by $\mathrm{Met}(\mathcal{X})$.

Common examples of metric spaces are given by vector spaces equipped with a 'norm', as defined in the following example.

**Example** A.52 (**Normed spaces**)**.** A **normed space** is a vector space $\mathcal{X}$ equipped with a function $\|\cdot\| : \mathcal{X} \to \mathbb{R}$ satisfying the following conditions:

1. **Nondegeneracy**: $\|x\| = 0 \iff x = 0$ for all $x \in \mathcal{X}$.

2. (**Absolute**) **homogeneity**: $\|\lambda x\| = |\lambda|\,\|x\|$ for all $x \in \mathcal{X}$ and $\lambda \in \mathbb{R}$.

3. **Triangle inequality**: $\|x + y\| \leq \|x\| + \|y\|$ for all $x, y \in \mathcal{X}$.

Consider, for example, $\mathcal{X} = \mathbb{R}^n$ equipped with the standard $\ell^p$-norms

$$\|x\|_p := \left( \sum_{i=1}^{n} |x_i|^p \right)^{1/p}. \tag{A.75}$$

It is easily verified that this norm induces a metric as follows:

$$d_p(\boldsymbol{x}, \boldsymbol{y}) := \|\boldsymbol{x} - \boldsymbol{y}\|_p. \tag{A.76}$$

This holds more generally for all normed spaces such as the function spaces $L^p(\mathcal{X}, \mu)$ with the norm

$$\|f\|_p := \left( \int_{\mathcal{X}} |f(x)|^p \, \mathrm{d}\mu(x) \right)^{1/p}. \tag{A.77}$$

The *supremum distance* between CDFs is a common occurrence in statistics and, therefore, deserves its own name.

> **Example** A.53 (**Kolmogorov–Smirnov distance**). Let $F, G : \mathbb{R} \to [0, 1]$ be two cumulative distribution functions. Their Kolmogorov–Smirnov (KS) distance is defined as follows:
>
> $$d_{\mathrm{KS}}(F, G) := \sup_{x \in \mathbb{R}} |F(x) - G(x)|. \qquad (A.78)$$
>
> When drawing the graphs of $F$ and $G$, the KS distance is simply the greatest vertical distance between the graphs.

A related distance is the following one.

> **Example** A.54 (**Total variation distance**). Let $P, Q \in \mathbb{P}(\mathcal{X})$ be two probability distributions on a measurable space $(\mathcal{X}, \Sigma)$. Their total variation (TV) distance (or **statistical distance**) is defined as follows:
>
> $$d_{\mathrm{TV}}(P, Q) := \sup_{B \in \Sigma} |P(B) - Q(B)|. \qquad (A.79)$$

It is also interesting to see how the total variation distance relates to other common tools in statistics and machine learning. The following property bounds the TV distance by the KS distance and the KL divergence[7] (see Example A.61 below).

> **Property** A.55. An upper bound to the total variation distance is given by **Pinsker's inequality**:
>
> $$d_{\mathrm{TV}}(P, Q) \leq \sqrt{\frac{1}{2} D_{\mathrm{KL}}(P, Q)}, \qquad (A.80)$$
>
> where $D_{\mathrm{KL}}$ denotes the Kullback–Leibler divergence (Example A.61). Moreover, on $\mathbb{R}$ with its standard Borel $\sigma$-algebra, a lower bound is given by the Kolmogorov–Smirnov distance:
>
> $$d_{\mathrm{KS}}(P, Q) \leq d_{\mathrm{TV}}(P, Q). \qquad (A.81)$$
>
> *Proof*. For the upper bound, see Lemma 2.5 in Tsybakov (2008). The lower bound immediately follows from the properties of the supremum and the fact that intervals of the form $]-\infty, x]$ form a subset of the Borel $\sigma$-algebra of $\mathbb{R}$. $\qquad\square$

---

[7] Note that the Kullback–Leibler divergence is not a distance since it is not symmetric.

## A.3.2 Related notions

**Definition A.56** (**Metric equivalence**). Two metrics $d, d' : \mathcal{X} \times \mathcal{X} \to \mathbb{R}$ are said to be equivalent (Kelley, 2017) if $d' = f \circ d$ for some continuous function $f : \mathbb{R}^+ \to \mathbb{R}^+$ satisfying the following conditions:

1. **Nondegeneracy**: $f(x) = 0 \iff x = 0$ for all $x \in \mathbb{R}^+$.

2. **Monotonicity**: $x < x' \implies f(x) < f(x')$ for all $x, x' \in \mathbb{R}^+$.

3. **Subadditivity**: $f(x + y) \leq f(x) + f(x')$ for all $x, x' \in \mathbb{R}^+$.

**Definition A.57** (**Diameter**). Consider a metric space $(\mathcal{X}, d)$. The diameter of a subset $S \subseteq \mathcal{X}$ is defined as follows:

$$\mathrm{diam}(S) := \sup_{p,q \in S} d(p, q) \,. \tag{A.82}$$

Equivalently, it is the diameter of the smallest ball $B(p, r) := \{q \in \mathcal{X} \mid d(p, q) \leq r\}$ that contains $S$.

Note that metric spaces, in particular normed spaces, are examples of *topological spaces*, where the open sets are generated by open balls. Usually these spaces are equipped with the induced Borel $\sigma$-algebra. It is not hard to check that the standard topology and measurable structure on $\mathbb{R}^n$ are exactly those generated by the usual Euclidean metric induced by the $\ell^2$-norm.

For completeness, we define two further notions induced by a metric $d : \mathcal{X} \times \mathcal{X} \to \mathbb{R}$. The distance between subsets is defined as

$$d(A, B) := \inf_{\substack{p \in A \\ q \in B}} d(p, q) \,. \tag{A.83}$$

Then, given a subset $S$ of a metric space $(\mathcal{X}, d)$, we can define the **signed distance** between a point $p \in \mathcal{X}$ and $S$:

$$\underline{d}(p, S) := \begin{cases} d(\{p\}, \partial S) & \text{if } p \notin S \,, \\ -d(\{p\}, \partial S) & \text{if } p \in S \,, \end{cases} \tag{A.84}$$

where $\partial S$ denotes the *boundary* of $S$.

The metric spaces that are induced by norms are, amongst other things, interesting for the following properties.

**Property** A.58 (**Convex metrics**). A metric $d : \mathcal{X} \times \mathcal{X} \to \mathbb{R}$ on vector space $\mathcal{X}$ is induced by a norm if and only if it is convex in one, and hence both, arguments, i.e.

$$d\big(\lambda x + (1-\lambda)y, z\big) \leq \lambda d(x,z) + (1-\lambda)d(y,z) \qquad (A.85)$$

for all $\lambda \in [0,1]$ and $x, y, z \in \mathcal{X}$.

*Proof.* See Witzgall (1965).                                    □

**Corollary** A.59. Consider a convex metric $d : \mathcal{X} \times \mathcal{X} \to \mathbb{R}$ as in the previous property. Given any collection of points $S \subset \mathcal{X}$, the distance from the convex hull $\mathsf{Conv}(S)$ to any point $z \in \mathcal{X}$ is maximized by a point on the boundary of $\mathsf{Conv}(S)$.

*Proof.* By definition of the convex hull, it is sufficient to prove this for the case of two points $S = \{x, y\}$.

$$
\begin{aligned}
d\big(\lambda x + (1-\lambda)y, z\big) &\leq \lambda d(x,z) + (1-\lambda)d(y,z) \\
&\leq \lambda \max\big(d(x,z), d(y,z)\big) + (1-\lambda)\max\big(d(x,z), d(y,z)\big) \\
&= \max\big(d(x,z), d(y,z)\big).
\end{aligned}
$$

□

Although metrics are very useful and important to compare elements of a set, a more general concept is often used to compare probability distributions.

**Definition** A.60 (**Divergence**). Let $\mathcal{X}$ be a set. A divergence (measure) on $\mathcal{X}$ is a function $D(\cdot \,\|\, \cdot) : \mathcal{X} \times \mathcal{X} \to \mathbb{R}$ satisfying the following conditions (Amari, 2016):[a]

1. **Nonnegativity**: $D(x \,\|\, y) \geq 0$ for all $x, y \in \mathcal{X}$.

2. **Nondegeneracy**: $D(x \,\|\, y) = 0 \iff x = y$ for all $x, y \in \mathcal{X}$.

Note that in stark contrast to metrics, divergences are not symmetric.

---

[a] Note that in information geometry, divergences are often required to satisfy a certain positive-definiteness condition as to be able to derive geometric structure from them.

By far the most common divergence measure in statistics is the following one.

**Example** A.61 (**Kullback–Leibler divergence**). Let $P, Q \in \mathbb{P}(\mathcal{X})$ be two probability distributions on a set $\mathcal{X}$. The Kullback–Leibler[a] (KL) divergence is defined as follows:

$$D_{\mathrm{KL}}(P \,\|\, Q) := \mathsf{E}_P\left[\frac{P}{Q}\right] = \int_{\mathcal{X}} \ln\left(\frac{P(x)}{Q(x)}\right) \mathrm{d}P(x) \tag{A.86}$$

When both $P$ and $Q$ are absolutely continuous with respect to a probability distribution $\mu \in \mathbb{P}(\mathcal{X})$, with associated densities $f$ and $g$, this can be rewritten as follows:

$$D_{\mathrm{KL}}(P \,\|\, Q) = \int_{\mathcal{X}} f(x) \ln\left(\frac{f(x)}{g(x)}\right) \mathrm{d}\mu(x)\,. \tag{A.87}$$

————————
[a]  This is also called the **relative entropy**.

**Definition** A.62 (**Kernel**). A symmetric (integrable[a]) function $K : \mathbb{R}^n \times \mathbb{R}^n \to \mathbb{R}$. The induced one-parameter function $x \mapsto K(x-y)$, where $y \in \mathbb{R}^n$ is fixed, is often called a **convolution kernel**.

————————
[a]  With respect to the Lebesgue measure.

### A.3.3   Lipschitz continuity

**Definition** A.63 (**Lipschitz continuity**). A continuous function

$$f : (\mathcal{X}, d) \to (\mathcal{X}', d') \tag{A.88}$$

between metric spaces is said to be Lipschitz(-continuous) with **Lipschitz constant** $\zeta \in \mathbb{R}^+$ if

$$d'\big(f(x), f(y)\big) \leq \zeta d(x, y) \tag{A.89}$$

for all $x, y \in \mathcal{X}$.

**Property** A.64. The composition of two Lipschitz-continuous functions is again Lipschitz-continuous and, moreover, the Lipschitz constant of the composite is equal to the product of the individual Lipschitz constants.

*Proof.* Consider Lipschitz-continuous functions $f : (\mathcal{X}, d) \to (\mathcal{X}', d')$ and $g : (\mathcal{X}', d') \to (\mathcal{X}'', d'')$ with Lipschitz constants $\zeta, \zeta' \in \mathbb{R}^+$, respectively. By definition we have:

$$d''\big(g \circ f(x), g \circ f(y)\big) \leq \zeta' d'\big(f(x), f(y)\big)$$
$$\leq \zeta' \zeta d(x, y)$$

for all $x, y \in \mathcal{X}$.                                                                          □

**Property** A.65.  Consider a differentiable function $f : \mathbb{R} \to \mathbb{R}$. If the derivative $f'$ is everywhere bounded in absolute value by a constant $M \in \mathbb{R}^+$, then $f$ is Lipschitz-continuous with Lipschitz constant $M$.

*Proof.* For any two real numbers $a, b \in \mathbb{R}$, the mean value theorem says that there exists a value $c \in {]a, b[}$ such that

$$f'(c) = \frac{f(b) - f(a)}{b - a}. \tag{A.90}$$

By assumption, the left-hand side is bounded in absolute value by some $M \in \mathbb{R}^+$. This implies that

$$|f(b) - f(a)| \leq M|b - a| \tag{A.91}$$

for all $a, b \in \mathbb{R}$, proving the result.                                             □

**Remark** A.66.  This property extends to higher dimensions. A differentiable function $f : \mathbb{R}^n \to \mathbb{R}$ is Lipschitz-continuous if all of its partial derivatives are bounded.

## A.4    Function estimation

Most of (supervised) machine learning is concerned with modelling various types of processes. Two general problems are classification and regression tasks. In the former situation, we essentially try to decompose the feature space $\mathcal{X}$ in disjoint subsets and assign to every subset a label chosen from a discrete space $\mathcal{C}$. In the latter situation, the discrete label space is replaced by a measurable target space $\mathcal{Y}$ and we try to estimate a function $f : \mathcal{X} \to \mathcal{Y}$. (Note that classification problems are technically just regression problems for discrete target spaces.)

## A.4.1 Regression

If we have perfect information about the structure of both the feature space $\mathcal{X}$ and the response space $\mathcal{Y}$ (and their relation), it might be feasible to find a single function $f : \mathcal{X} \to \mathcal{Y}$. However, in practice, we only have partial information and, moreover, this information is usually noisy. Given this context, it does not really make sense to talk about a function that assigns to every tuple of features exactly one response value. To solve this issue, we should rephrase general regression problems as follows.

**Definition** A.67 (**Regression task**). Given a data set $\mathcal{D} \in (\mathcal{X} \times \mathcal{Y})^*$, find a joint probability distribution $P \in \mathbb{P}(\mathcal{X} \times \mathcal{Y})$ that is optimal with respect to a predetermined metric such as the likelihood.

What is often done to unify this probabilistic model $P$ with the point predictor $f : \mathcal{X} \to \mathcal{Y}$, is that the function $f$ is for example taken to be the conditional mean $\mathsf{E}_P[Y \mid X]$. So, point predictors act as estimates of conditional statistics of the distribution.

**Definition** A.68 ($R^2$-**coefficient**). Consider a point predictor $\hat{y} : \mathcal{X} \to \mathbb{R}$. The $R^2$-coefficient of $\hat{y}$ with respect to a data set $\mathcal{D} \in (\mathcal{X} \times \mathcal{Y})^*$ is defined as follows:

$$R^2 := 1 - \frac{\sum_{(x,y)\in\mathcal{D}}\big(y - \hat{y}(x)\big)^2}{\sum_{(x,y)\in\mathcal{D}}(y - \overline{y})^2}, \tag{A.92}$$

where

$$\overline{y} := \frac{1}{|\mathcal{D}|}\sum_{(x,y)\in\mathcal{D}} y. \tag{A.93}$$

This gives a measure of how large the residuals of the model are compared to the ones we would obtain after discarding the features.

## A.4.2 Ensemble methods

When working with a finite data set, it might happen that there does not exist a unique probability measure that optimizes the chosen metric. In such cases, we can choose to work with a set of probability measures that could all describe the ground truth. Such a set $\mathfrak{C}$ is also known as a *credal set* (Au-gustin et al., 2014). In fact, even when there does exist a unique optimizer, it might be better to work with a credal set, since the chosen metric might not be the optimal choice for the given problem or the data might be too noisy to completely trust the results. However, working with credal sets can be difficult in practice. There is a vast literature about this subject, but covering all of it would lead us too far astray. A more practical approach is to take the credal set as is and extract a probability measure from it. Essentially, we put a second-order distribution $\mathfrak{p} \in \mathbb{P}(\mathfrak{C})$ over the credal set and obtain a new measure $\mu_{\mathfrak{p}}$ by integration (cf. the Giry multiplication map (A.33)):

$$\mu_{\mathfrak{p}}(A) := \int_{\mathfrak{C}} \mu(A) \, d\mathfrak{p}(\mu) \, . \tag{A.94}$$

Of course, the question then becomes how to choose the probability distribution $\mathfrak{p}$. For computational ease, this measure is generally chosen to be the uniform distribution, i.e. we construct a uniform mixture model from $\mathfrak{C}$. If $\mathfrak{C}$ is finite, this gives

$$\mu_{\mathfrak{C}} := \frac{1}{|\mathfrak{C}|} \sum_{\mu \in \mathfrak{C}} \mu \, . \tag{A.95}$$

In fact, if $\mathfrak{C}$ is (closed and) convex, the *Krein–Milman theorem* (Grothendieck & Chaljub, 1973) states that every element can be written as a convex combination of the *extreme points* of $\mathfrak{C}$ (e.g. the corners of a polytope). So, these mixture models certainly have their value.

Moreover, it are exactly these mixture models where ensemble methods enter the story. For basic ensembles, multiple models $\widehat{f}_i : \mathcal{X} \to \mathcal{Y}$, each representing a different distribution, are trained and a mixture model is constructed from them. If, for example, these functions estimate the (conditional) mean of the distributions, their average estimates the (conditional) mean of the mixture model:

$$\mathsf{E}_{\mu_{\mathfrak{C}}}[Y \mid X] = \frac{1}{|\mathfrak{C}|} \sum_{i=1}^{|\mathfrak{C}|} \mathsf{E}_{\mu_i}[Y \mid X] \approx \frac{1}{|\mathfrak{C}|} \sum_{i=1}^{|\mathfrak{C}|} \widehat{f}_i(X) \, . \tag{A.96}$$

To be able to cope with arbitrary distributions, we usually approximate the mixture model by a simpler distribution. In practice, this will usually be a normal distribution, since further constructions such as those of confidence or prediction intervals are well-studied for this family of distributions. To this end, the predictive distribution is modelled as a normal distribution with the same mean and variance as the mixture model:

$$\mu_{\mathfrak{C}}(Y \mid X) \approx \mathcal{N}\left( \mathsf{E}_{\mu_{\mathfrak{C}}}[Y \mid X], \mathsf{Var}_{\mu_{\mathfrak{C}}}[Y \mid X] \right), \tag{A.97}$$

with

$$\mathsf{Var}_{\mu_{\mathfrak{C}}}[Y \mid X] = \mathsf{E}_{\mu_{\mathfrak{C}}}[Y^2 \mid X] - \mathsf{E}_{\mu_{\mathfrak{C}}}[Y \mid X]^2$$

$$= \frac{1}{|\mathfrak{C}|} \sum_{i=1}^{|\mathfrak{C}|} \mathsf{E}_{\mu_i}[Y^2 \mid X] - \mathsf{E}_{\mu_{\mathfrak{C}}}[Y \mid X]^2 \tag{A.98}$$

$$= \frac{1}{|\mathfrak{C}|} \sum_{i=1}^{|\mathfrak{C}|} \left( \mathsf{E}_{\mu_i}[Y \mid X]^2 + \mathsf{Var}_{\mu_i}[Y \mid X] \right) - \left( \frac{1}{|\mathfrak{C}|} \sum_{i=1}^{|\mathfrak{C}|} \mathsf{E}_{\mu_i}[Y \mid X] \right)^2 .$$

When the different distributions in the mixture are given by delta measures and the mixture model simply represents the empirical distribution of a data sample, such as for an ensemble of ordinary point predictors, the individual variance terms drop out.

**Example** A.69 (**Bimodal distribution**). Consider two probability distributions with standard deviation $\sigma \in \mathbb{R}^+$, centered at $\mu - \lambda$ and $\mu + \lambda$. By Eq. (A.98), the total variance of this bimodal distribution is given by:

$$\mathsf{Var}[Y \mid X] = \sigma^2 + \lambda^2 . \tag{A.99}$$

A specific approach to ensemble modelling that is often used in practice is 'bagging' (**bootstrap aggregating**) as introduced by Breiman (1996).

**Definition** A.70 (**Bagging**). Assume that a data set $\mathcal{D} \in (\mathcal{X} \times \mathcal{Y})^*$ and a model architecture $\mathcal{A}$ are given. From $\mathcal{D}$, we resample $n \in \mathbb{N}_0$ new data sets $\{\mathcal{D}_i\}_{i \leq n}$ by uniformly sampling $k \in \mathbb{N}_0$ data points with replacement from $\mathcal{D}$ for all $i \leq n$. Such a sample is also called a **bootstrap sample** (Efron, 1979). To obtain the ensemble, we then train a model $\hat{y}_i : \mathcal{X} \to \mathcal{Y}$ with architecture $\mathcal{A}$ on every bootstrap sample $\mathcal{D}_i$.

For every data point $(x_i, y_i) \in \mathcal{D}$, we can also construct a predictor $\hat{y}_{(i)}$ by aggregating the submodels in the ensemble that are not trained on $(x_i, y_i)$. This is called the **out-of-bag** (OOB) predictor of $(x_i, y_i)$.

## A.4.3 Classification

Although the bulk of this dissertation focuses on regression problems, Chapter 5 is rather aimed at classification problems. For this reason, we give a small overview of the most important features of this problem setting. Since multiclass classification will be the most important problem to us, we restrict to this case.

**Definition** A.71 (**Multiclass classification**).  Given a data set $\mathcal{D} \in (\mathcal{X} \times \mathcal{Y})^*$, where $\mathcal{Y} = [k]$ is finite, the task is to find a joint probability distribution $P \in \mathbb{P}(\mathcal{X} \times [k])$ that is optimal with respect to a predetermined metric such as the accuracy or $F_1$-score (to be defined below).

**Definition** A.72 (**Accuracy**).  Let TP, FP, TN and FN denote the number of true positives, false positives, true negatives and false negatives, respectively.  The accuracy is then defined as the ratio of correctly classified data points to the total number of data points:

$$\text{Acc} := \frac{\text{TP} + \text{TN}}{\text{TP} + \text{FP} + \text{TN} + \text{FN}} . \tag{A.100}$$

Although the accuracy gives a general idea of the performance, it has some downsides.  Mainly, in the case of imbalanced data, which will be the case of interest in Chapter 5, the minority classes will be dominated by the majority classes and, hence, the accuracy gives a biased result.  To this end, we will also consider two different metrics: the balanced accuracy and the (weighted) $F_1$-score.

**Definition** A.73 (**Balanced accuracy**).

$$\text{bAcc} := \frac{\text{sens} + \text{spec}}{2} , \tag{A.101}$$

where

- the **sensitivity** (or **recall**) is defined as

$$\text{sens} := \frac{\text{TP}}{\text{TP} + \text{FN}} , \tag{A.102}$$

- and the **specificity** is defined as

$$\text{spec} := \frac{\text{TN}}{\text{TN} + \text{FP}} . \tag{A.103}$$

**Definition** A.74 (**Weighted $F_1$-score**).  The standard (macro) $F_1$-score is calculated as follows:

$$F_1 := \frac{\text{prec} \cdot \text{sens}}{\text{prec} + \text{sens}} , \tag{A.104}$$

where the **precision** is defined as

$$\text{prec} := \frac{\text{TP}}{\text{TP} + \text{FP}} \, .$$

$$(A.105)$$

If, instead, we calculated this quantity for every class and then took a weighted average based on the number of true positives per class, the weighted $F_1$-score is obtained.

# Additional Data       B

This appendix contains additional tables and figures for the experimental sections in this dissertation. These have been moved to this appendix to improve the flow of the text.

## B.1    Tables for Chapter 3

This section contains tables with data (mean and standard deviation) corresponding to the figures in Section 3.4.3. The results for every metric (coverage, average width and $R^2$-coefficient) are each split into two tables for clarity, with five data sets each. The rows represent the various models and the columns correspond to the different data sets. For every combination, the average over 50 random train-test splits is reported. The standard deviation is shown between parentheses.

Table B.1: Summary of coverage degrees (part I). For the data sets in each column, the results are shown for all models (standard deviation between parentheses). For clarity, models that come in both a default and a conformalized version have been grouped together.

|        | concrete | naval | turbine | puma32H | residential |
|--------|----------|-------|---------|---------|-------------|
| NN-CP  | 0.901 (0.025) | 0.899 (0.006) | 0.900 (0.010) | 0.900 (0.011) | 0.902 (0.048) |
| RF-CP  | 0.900 (0.028) | 0.900 (0.008) | 0.901 (0.009) | 0.900 (0.009) | 0.890 (0.047) |
| QR     | 0.914 (0.020) | 0.997 (0.003) | 0.919 (0.011) | 0.908 (0.011) | 0.950 (0.029) |
| QR-CP  | 0.899 (0.026) | 0.900 (0.007) | 0.900 (0.010) | 0.899 (0.009) | 0.904 (0.046) |
| DE     | 0.928 (0.019) | 0.995 (0.002) | 0.940 (0.008) | 0.948 (0.008) | 0.959 (0.023) |
| DE-CP  | 0.893 (0.028) | 0.900 (0.007) | 0.900 (0.011) | 0.902 (0.010) | 0.909 (0.040) |
| Drop   | 0.580 (0.057) | 0.962 (0.047) | 0.352 (0.031) | 0.671 (0.033) | 0.872 (0.053) |
| Drop-CP| 0.897 (0.025) | 0.900 (0.007) | 0.901 (0.012) | 0.900 (0.009) | 0.902 (0.043) |
| MVE    | 0.942 (0.019) | 1.000 (0.001) | 0.939 (0.009) | 0.951 (0.018) | 0.974 (0.023) |
| MVE-CP | 0.901 (0.022) | 0.900 (0.007) | 0.901 (0.010) | 0.901 (0.008) | 0.908 (0.040) |
| GP     | 0.887 (0.023) | 0.782 (0.071) | 0.916 (0.007) | 0.890 (0.014) | 0.928 (0.032) |
| GP-CP  | 0.900 (0.026) | 0.859 (0.024) | 0.902 (0.009) | 0.901 (0.009) | 0.899 (0.041) |
| SVGP   | 0.918 (0.020) | 0.902 (0.132) | 0.923 (0.009) | 0.891 (0.008) | 0.967 (0.022) |
| SVGP-CP| 0.900 (0.025) | 0.900 (0.008) | 0.901 (0.010) | 0.901 (0.009) | 0.897 (0.043) |

Table B.2: Summary of coverage degrees (part II). For the data sets in each column, the results are shown for all models (standard deviation between parentheses). For clarity, models that come in both a default and a conformalized version have been grouped together. OoT stands for "out of time", for these combinations of data sets and models, the maximum runtime was exceeded.

|         | crime2        | fb1           | blog          | traffic       | star          |
|---------|---------------|---------------|---------------|---------------|---------------|
| NN-CP   | 0.901 (0.015) | 0.901 (0.004) | 0.900 (0.003) | 0.903 (0.082) | 0.906 (0.018) |
| RF-CP   | 0.903 (0.017) | 0.901 (0.004) | 0.901 (0.003) | 0.910 (0.078) | 0.905 (0.016) |
| QR      | 0.857 (0.047) | 0.916 (0.041) | 0.898 (0.041) | 0.807 (0.080) | 0.838 (0.039) |
| QR-CP   | 0.904 (0.016) | 0.902 (0.004) | 0.900 (0.004) | 0.900 (0.077) | 0.904 (0.018) |
| DE      | 0.940 (0.016) | 0.960 (0.011) | 0.959 (0.010) | 0.884 (0.097) | 0.901 (0.019) |
| DE-CP   | 0.908 (0.017) | 0.901 (0.004) | 0.900 (0.003) | 0.898 (0.075) | 0.902 (0.016) |
| Drop    | 0.546 (0.051) | 0.802 (0.062) | 0.612 (0.071) | 0.467 (0.145) | 0.309 (0.050) |
| Drop-CP | 0.904 (0.017) | 0.901 (0.004) | 0.900 (0.004) | 0.913 (0.056) | 0.905 (0.016) |
| MVE     | 0.925 (0.022) | 0.969 (0.012) | 0.965 (0.012) | 0.865 (0.107) | 0.884 (0.029) |
| MVE-CP  | 0.909 (0.017) | 0.901 (0.004) | 0.900 (0.003) | 0.903 (0.074) | 0.904 (0.017) |
| GP      | 0.894 (0.019) | OoT           | OoT           | 0.867 (0.077) | 0.899 (0.014) |
| GP-CP   | 0.900 (0.020) | OoT           | OoT           | 0.901 (0.066) | 0.902 (0.018) |
| SVGP    | 0.913 (0.015) | 0.979 (0.002) | 0.974 (0.002) | 0.940 (0.044) | 0.906 (0.015) |
| SVGP-CP | 0.903 (0.020) | 0.901 (0.004) | 0.900 (0.003) | 0.892 (0.074) | 0.904 (0.017) |

Table B.3: Summary of average widths (part I). For the data sets in each column, the results are shown for all models (standard deviation between parentheses). For clarity, models that come in both a default and a conformalized version have been grouped together.

|         | concrete      | naval         | turbine       | puma32H       | residential   |
|---------|---------------|---------------|---------------|---------------|---------------|
| NN-CP   | 1.284 (0.090) | 0.370 (0.057) | 0.754 (0.014) | 1.136 (0.052) | 0.915 (0.241) |
| RF-CP   | 1.139 (0.090) | 0.199 (0.010) | 0.663 (0.013) | 0.873 (0.014) | 0.700 (0.148) |
| QR      | 1.209 (0.173) | 0.535 (0.021) | 0.775 (0.018) | 0.862 (0.044) | 0.693 (0.087) |
| QR-CP   | 1.322 (0.179) | 0.374 (0.046) | 0.731 (0.011) | 1.083 (0.075) | 0.857 (0.195) |
| DE      | 1.188 (0.121) | 1.921 (0.061) | 0.820 (0.015) | 1.026 (0.053) | 0.775 (0.562) |
| DE-CP   | 1.322 (0.158) | 1.505 (0.033) | 0.730 (0.015) | 1.620 (0.182) | 0.838 (0.311) |
| Drop    | 0.527 (0.084) | 0.351 (0.010) | 0.204 (0.017) | 0.527 (0.035) | 0.493 (0.122) |
| Drop-CP | 1.354 (0.147) | 0.334 (0.047) | 0.755 (0.012) | 1.150 (0.047) | 0.893 (0.186) |
| MVE     | 1.209 (0.112) | 0.586 (0.022) | 0.832 (0.033) | 1.053 (0.072) | 0.700 (0.151) |
| MVE-CP  | 1.282 (0.118) | 0.346 (0.047) | 0.744 (0.020) | 1.157 (0.044) | 0.714 (0.155) |
| GP      | 1.006 (0.060) | 0.222 (0.025) | 0.757 (0.013) | 3.064 (0.265) | 0.542 (0.046) |
| GP-CP   | 1.234 (0.087) | 0.287 (0.049) | 0.739 (0.012) | 3.050 (0.263) | 0.679 (0.090) |
| SVGP    | 1.240 (0.038) | 0.217 (0.021) | 0.800 (0.015) | 2.939 (0.047) | 2.769 (0.224) |
| SVGP-CP | 1.356 (0.094) | 0.210 (0.087) | 0.754 (0.013) | 3.023 (0.056) | 2.256 (0.294) |

Table B.4: Summary of average widths (part II). For the data sets in each column, the results are shown for all models (standard deviation between parentheses). For clarity, models that come in both a default and a conformalized version have been grouped together. OoT stands for "out of time", for these combinations of data sets and models, the maximum runtime was exceeded. OoR stands for "out of range", these values exceeded reasonable ranges for the average interval width.

|         | crime2         | fb1            | blog           | traffic        | star           |
|---------|----------------|----------------|----------------|----------------|----------------|
| NN-CP   | 2.021 (0.130)  | 0.444 (0.059)  | 0.583 (0.118)  | 2.724 (0.711)  | 3.101 (0.117)  |
| RF-CP   | 2.009 (0.106)  | 0.377 (0.020)  | 0.487 (0.027)  | 2.314 (0.428)  | 3.023 (0.090)  |
| QR      | 1.567 (0.293)  | 0.394 (0.041)  | 0.327 (0.027)  | 2.067 (0.430)  | 2.648 (0.267)  |
| QR-CP   | 1.704 (0.153)  | 0.285 (0.026)  | 0.271 (0.016)  | 2.939 (0.729)  | 3.095 (0.136)  |
| DE      | 1.847 (0.183)  | OoR            | OoR            | 2.229 (0.535)  | 2.958 (0.150)  |
| DE-CP   | 1.649 (0.093)  | OoR            | OoR            | 2.565 (0.609)  | 3.027 (0.102)  |
| Drop    | 0.663 (0.077)  | 0.310 (0.062)  | 0.282 (0.058)  | 0.997 (0.301)  | 0.746 (0.106)  |
| Drop-CP | 1.807 (0.120)  | 0.405 (0.053)  | 0.438 (0.081)  | 2.873 (0.668)  | 3.100 (0.115)  |
| MVE     | 1.683 (0.213)  | OoR            | OoR            | 2.083 (0.503)  | 2.870 (0.189)  |
| MVE-CP  | 1.602 (0.109)  | OoR            | OoR            | 2.602 (0.727)  | 3.052 (0.114)  |
| GP      | 1.972 (0.266)  | OoT            | OoT            | 2.427 (0.343)  | 2.931 (0.063)  |
| GP-CP   | 2.251 (0.433)  | OoT            | OoT            | 2.845 (0.534)  | 3.014 (0.111)  |
| SVGP    | 1.986 (0.045)  | 2.567 (0.170)  | 2.556 (0.158)  | 3.449 (0.168)  | 3.012 (0.052)  |
| SVGP-CP | 2.590 (0.475)  | 0.787 (0.059)  | 0.695 (0.061)  | 3.164 (0.707)  | 3.012 (0.100)  |

Table B.5: Summary of $R^2$-coefficients (part I). For the data sets in each column, the results are shown for all models (standard deviation between parentheses). For clarity, models that come in both a default and a conformalized version have been grouped together.

|         | concrete      | naval         | turbine       | puma32H       | residential   |
|---------|---------------|---------------|---------------|---------------|---------------|
| NN-CP   | 0.833 (0.027) | 0.987 (0.004) | 0.942 (0.004) | 0.879 (0.011) | 0.907 (0.073) |
| RF-CP   | 0.872 (0.021) | 0.990 (0.002) | 0.955 (0.004) | 0.930 (0.003) | 0.923 (0.033) |
| QR      | 0.865 (0.035) | 0.992 (0.002) | 0.944 (0.004) | 0.936 (0.006) | 0.962 (0.018) |
| QR-CP   | 0.817 (0.038) | 0.986 (0.004) | 0.942 (0.004) | 0.889 (0.014) | 0.919 (0.062) |
| DE      | 0.870 (0.031) | 0.813 (0.009) | 0.947 (0.004) | 0.929 (0.007) | 0.946 (0.068) |
| DE-CP   | 0.809 (0.047) | 0.740 (0.010) | 0.945 (0.004) | 0.697 (0.085) | 0.899 (0.081) |
| Drop    | 0.875 (0.023) | 0.994 (0.003) | 0.945 (0.004) | 0.931 (0.005) | 0.957 (0.018) |
| Drop-CP | 0.813 (0.042) | 0.990 (0.003) | 0.943 (0.004) | 0.876 (0.010) | 0.905 (0.029) |
| MVE     | 0.874 (0.020) | 0.994 (0.002) | 0.945 (0.004) | 0.935 (0.004) | 0.967 (0.016) |
| MVE-CP  | 0.820 (0.031) | 0.990 (0.003) | 0.943 (0.004) | 0.883 (0.010) | 0.931 (0.036) |
| GP      | 0.884 (0.022) | 1.000 (0.000) | 0.952 (0.004) | 0.132 (0.169) | 0.903 (0.073) |
| GP-CP   | 0.844 (0.025) | 1.000 (0.000) | 0.945 (0.004) | 0.187 (0.107) | 0.866 (0.087) |
| SVGP    | 0.864 (0.021) | 0.995 (0.008) | 0.943 (0.004) | 0.216 (0.021) | 0.659 (0.094) |
| SVGP-CP | 0.813 (0.027) | 0.995 (0.003) | 0.941 (0.004) | 0.211 (0.020) | 0.254 (0.062) |

Table B.6: Summary of $R^2$-coefficients (part II). For the data sets in each column, the results are shown for all models (standard deviation between parentheses). For clarity, models that come in both a default and a conformalized version have been grouped together. OoT stands for "out of time", for these combinations of data sets and models, the maximum runtime was exceeded. OoR stands for "out of range", these values exceeded reasonable ranges for the $R^2$-coefficient.

|         | crime2        | fb1           | blog          | traffic       | star          |
|---------|---------------|---------------|---------------|---------------|---------------|
| NN-CP   | 0.609 (0.063) | 0.424 (0.067) | 0.010 (1.390) | 0.151 (0.755) | 0.116 (0.050) |
| RF-CP   | 0.629 (0.032) | 0.616 (0.052) | 0.554 (0.050) | 0.549 (0.164) | 0.160 (0.036) |
| QR      | 0.638 (0.037) | 0.540 (0.054) | 0.462 (0.054) | 0.485 (0.192) | 0.142 (0.044) |
| QR-CP   | 0.627 (0.046) | 0.495 (0.048) | 0.430 (0.052) | 0.277 (0.560) | 0.117 (0.045) |
| DE      | 0.660 (0.031) | 0.371 (0.047) | 0.289 (0.034) | 0.500 (0.173) | 0.179 (0.033) |
| DE-CP   | 0.642 (0.032) | 0.353 (0.048) | 0.287 (0.031) | 0.340 (0.295) | 0.148 (0.035) |
| Drop    | 0.656 (0.033) | 0.471 (0.097) | 0.463 (0.061) | 0.441 (0.191) | 0.177 (0.036) |
| Drop-CP | 0.635 (0.038) | 0.425 (0.064) | OoR           | 0.315 (0.318) | 0.131 (0.045) |
| MVE     | 0.658 (0.036) | 0.411 (0.065) | OoR           | 0.523 (0.164) | 0.173 (0.032) |
| MVE-CP  | 0.641 (0.039) | 0.379 (0.074) | OoR           | 0.314 (0.366) | 0.143 (0.041) |
| GP      | 0.562 (0.122) | OoT           | OoT           | 0.316 (0.197) | 0.205 (0.026) |
| GP-CP   | 0.460 (0.206) | OoT           | OoT           | 0.244 (0.191) | 0.169 (0.045) |
| SVGP    | 0.654 (0.031) | 0.385 (0.042) | 0.421 (0.035) | 0.182 (0.151) | 0.184 (0.025) |
| SVGP-CP | 0.258 (0.209) | 0.360 (0.038) | 0.393 (0.034) | 0.096 (0.141) | 0.168 (0.026) |

## B.2   Proteins

This sections contains the figures for the experiments from Chapter 5 on the `Bacteria` data set using the APS nonconformity measure (5.44).

In Fig. B.1, we can see histograms of the classwise and clusterwise coverage for the three models ('PLT', 'HSM' and 'OVR') for a single train-calibration-test split using the APS nonconformity measure. The rows indicate the four different ways to construct clusters: marginal (all classes together), Mondrian (one cluster per class), Hierarchy (use the predetermined hierarchy together with a size threshold) and Score (cluster quantile embeddings of the nonconformity distributions together with a size threshold). For each subfigure, two different histograms are shown. For the blue one (with the label 'Clusterwise'), the coverage is calculated on the level of the clusters and, hence, the standard coverage guarantee predicts that this histogram should be centered around the nominal confidence level of $1 - \alpha$ (with $\alpha = 0.1$). For the orange histogram (with the label 'Classwise'), the coverage is calculated on the level of the classes and, hence, gives the conditional coverage.

Since these histograms depend on the specific calibration-test split (especially in the setting of extreme classification), a boxplot of the clusterwise histogram values is also shown in Fig. B.2 and for the classwise values in Fig. B.3.

Histogram plots for the average prediction set size (Eq. (2.11)) and the representation complexity (Definition 5.17) are shown in Figs. B.4 to B.5.

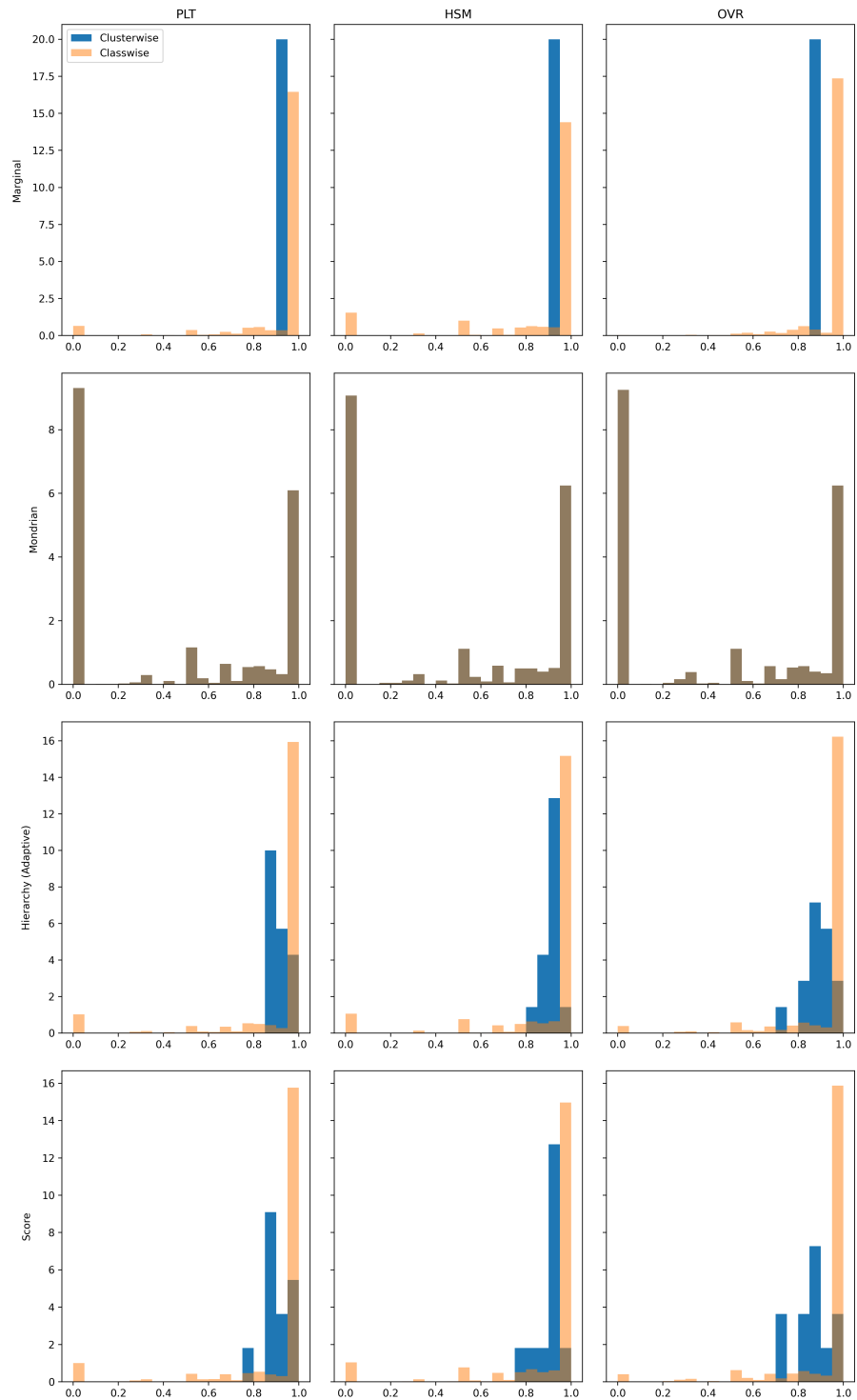

Figure B.1: Histogram of classwise and clusterwise coverage for the APS nonconformity measure for three models (PLT, HSM and OVR) and four clustering methods (marginal, Mondrian, hierarchy-based and score-based) on the `Proteins` data set.

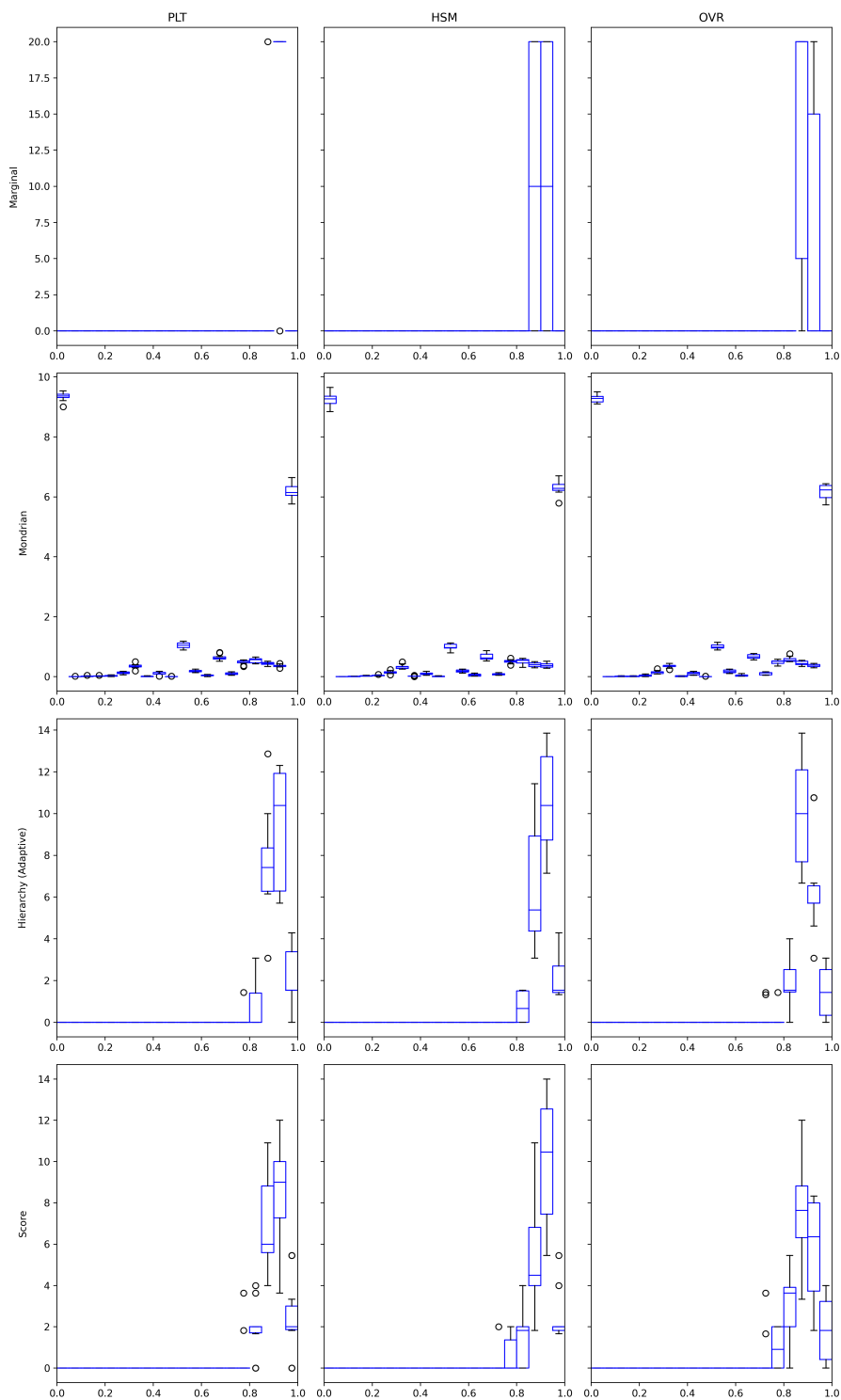

Figure B.2: Boxplot of clusterwise histogram densities for the APS noncon-formity measure for three models (PLT, HSM and OVR) and four clustering methods (marginal, Mondrian, hierarchy-based and score-based) over all calibration-test splits on the `Proteins` data set.

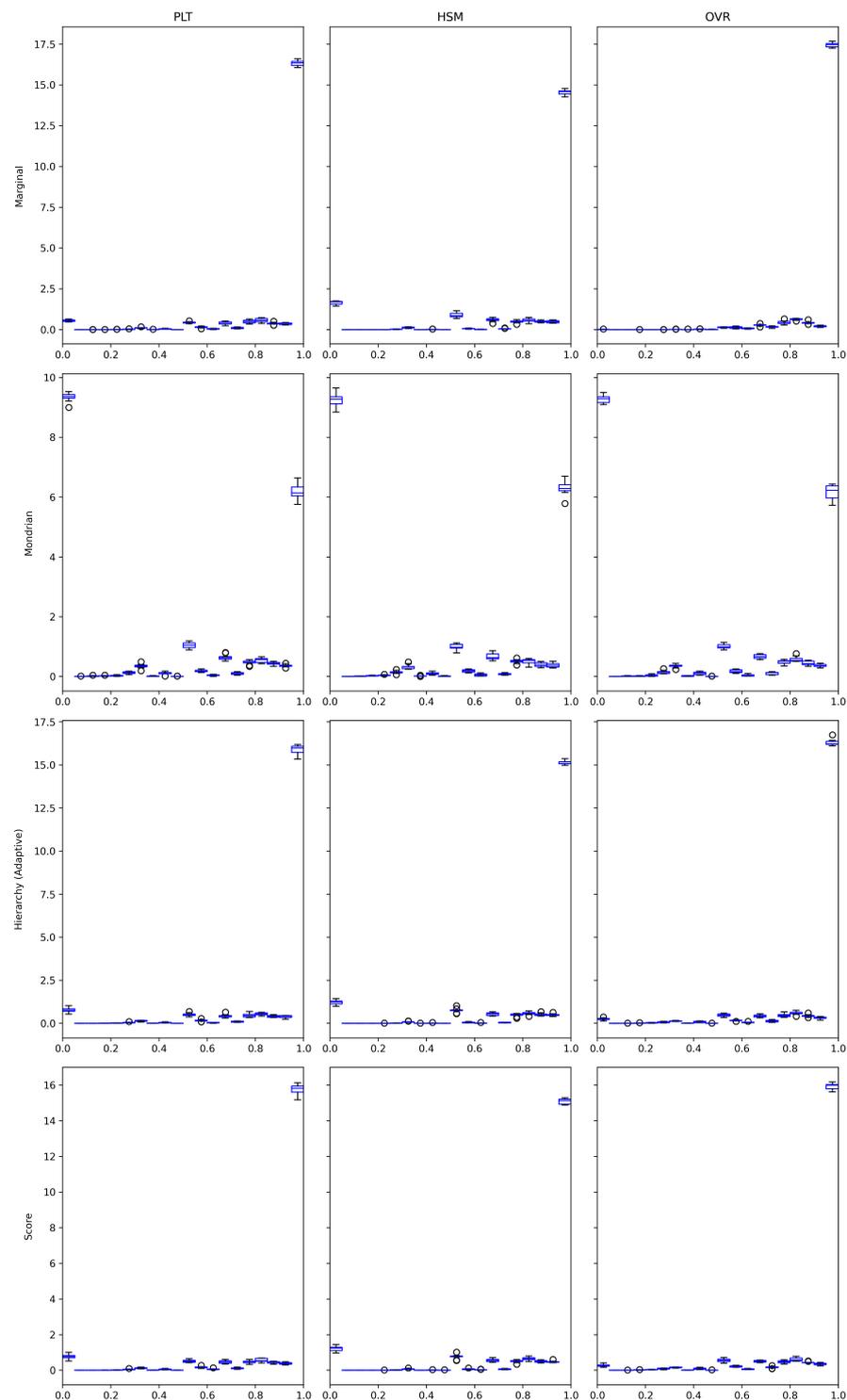

Figure B.3: Boxplot of classwise histogram densities for the softmax noncon-formity measure for three models (PLT, HSM and OVR) and four clustering methods (marginal, Mondrian, hierarchy-based and score-based) over all calibration-test splits on the `Proteins` data set.

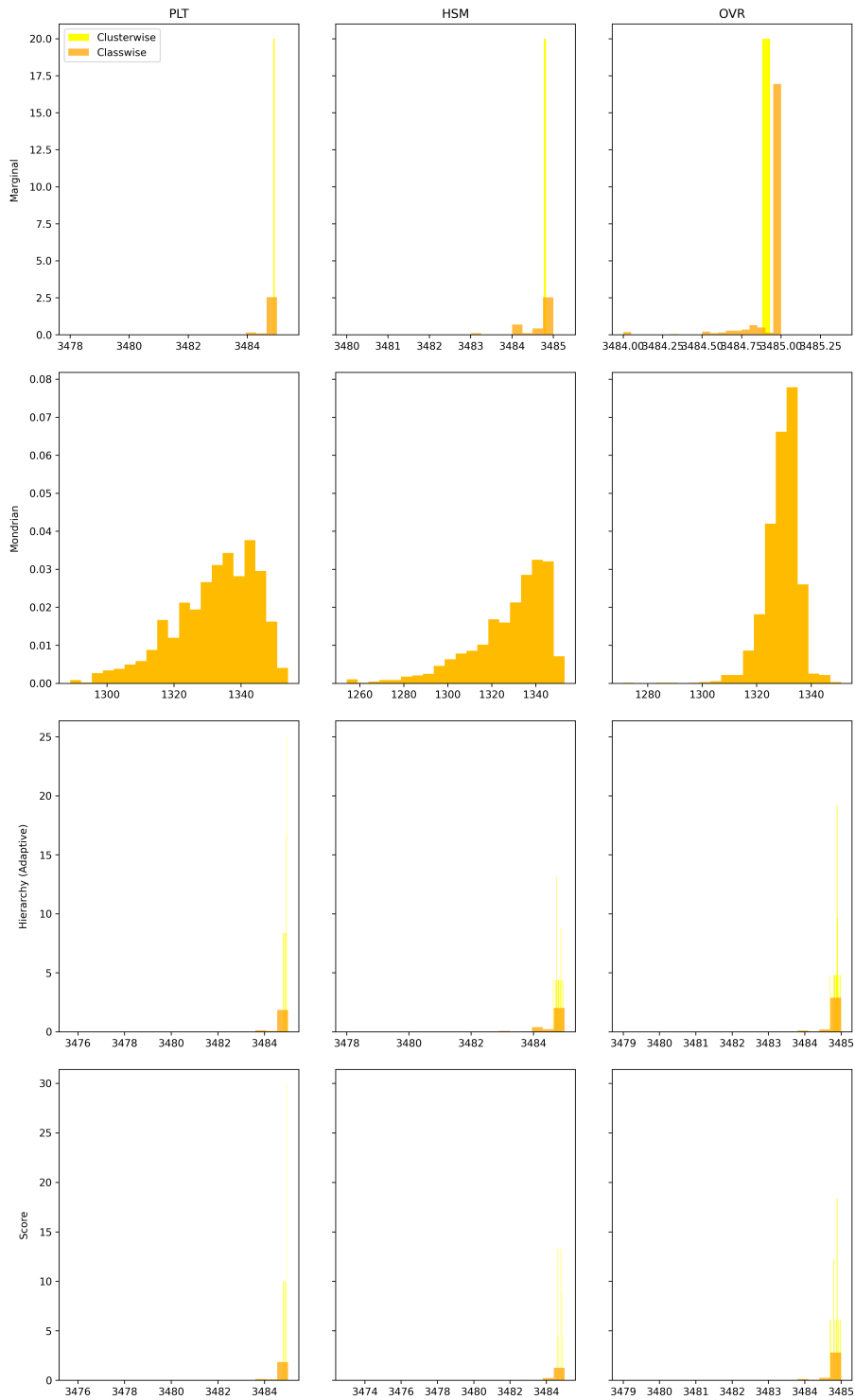

Figure B.4: Histogram of classwise and clusterwise average prediction set sizes for the APS nonconformity measure for three models (PLT, HSM and OVR) and four clustering methods (marginal, Mondrian, hierarchy-based and score-based) on the `Proteins` data set.

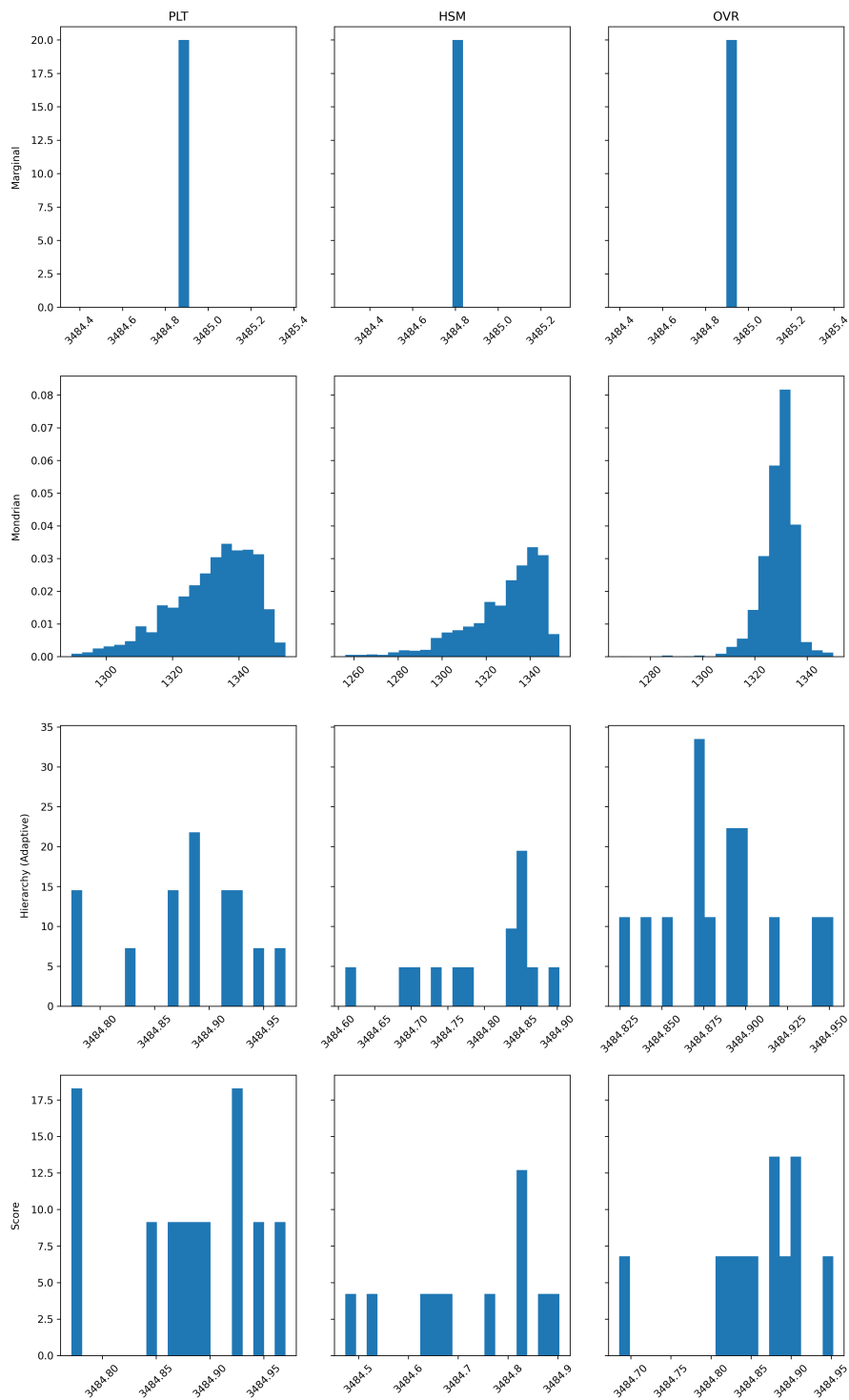

Figure B.5: Histogram of clusterwise representation complexities of the prediction sets for the APS nonconformity measure for three models (PLT, HSM and OVR) and four clustering methods (marginal, Mondrian, hierarchy-based and score-based) on the `Proteins` data set.

# B.3   `Bacteria`

This sections contains all figures for the experiments from Chapter 5 on the `Bacteria` data set.

In Fig. B.6, we can see histograms of the classwise and clusterwise coverage for the three models ('PLT', 'HSM' and 'OVR') for a single train-calibration-test split using the softmax nonconformity measure (see Fig. B.7 for the APS measure). The rows indicate the four different ways to construct clusters: marginal (all classes together), Mondrian (one cluster per class), Hierarchy (use the predetermined hierarchy together with a size threshold) and Score (cluster quantile embeddings of the nonconformity distributions together with a size threshold). For each subfigure, two different histograms are shown. For the blue one (with the label 'Clusterwise'), the coverage is calculated on the level of the clusters and, hence, the standard coverage guarantee predicts that this histogram should be centered around the nominal confidence level of $1 - \alpha$ (with $\alpha = 0.1$). For the orange histogram (with the label 'Classwise'), the coverage is calculated on the level of the classes and, hence, gives the conditional coverage.

Since these histograms depend on the specific calibration-test split (especially in the setting of extreme classification), a boxplot of the clusterwise histogram values is also shown in Figs. B.8 and B.10 and for the classwise values in Figs. B.9 and B.11.

Histogram plots for the average prediction set size (Eq. (2.11)) and the representation complexity (Definition 5.17) are shown in Figs. B.12 to B.15.

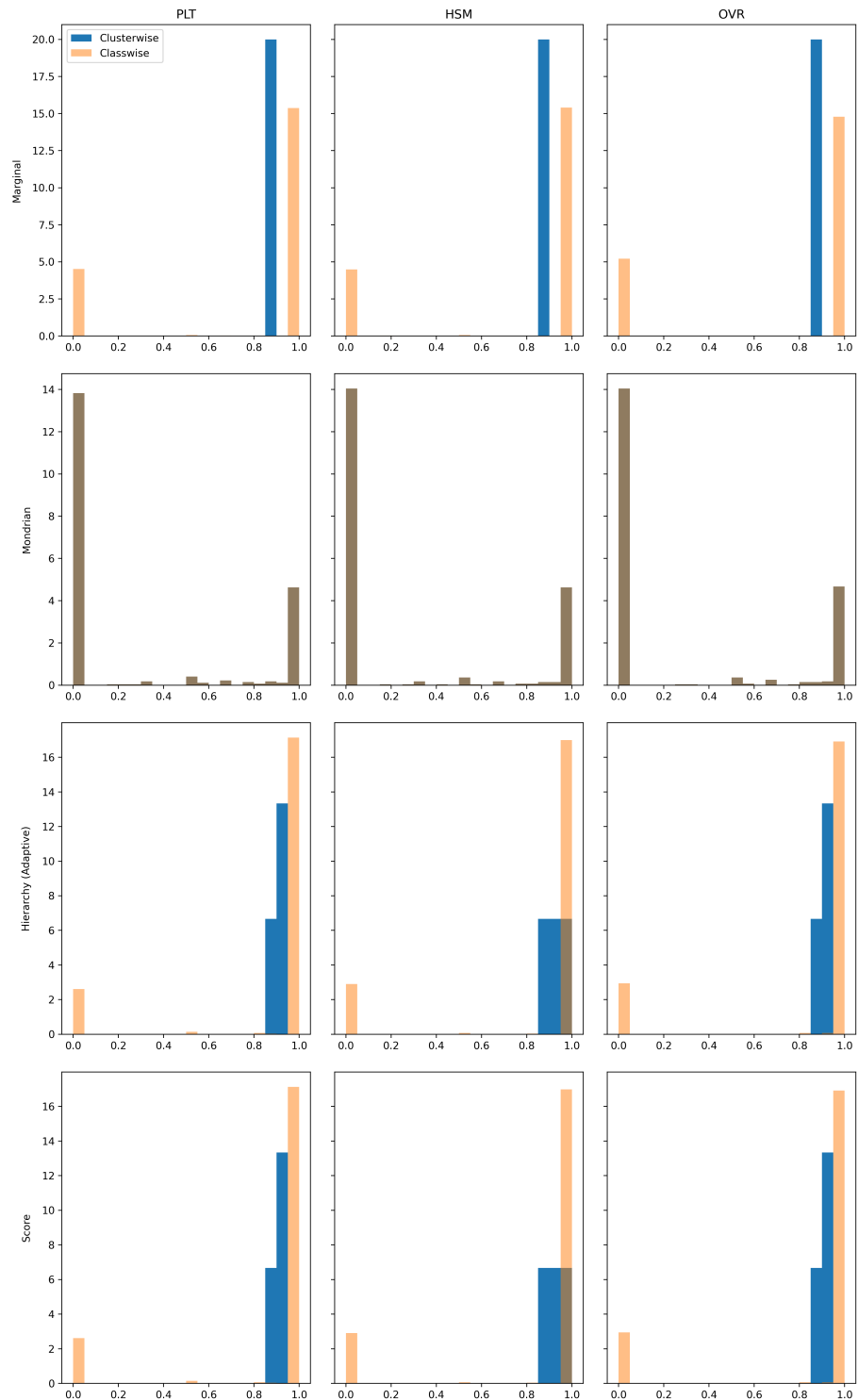

Figure B.6: Histogram of classwise and clusterwise coverage for the softmax nonconformity measure for three models (PLT, HSM and OVR) and four clustering methods (marginal, Mondrian, hierarchy-based and score-based) on the `Bacteria` data set.

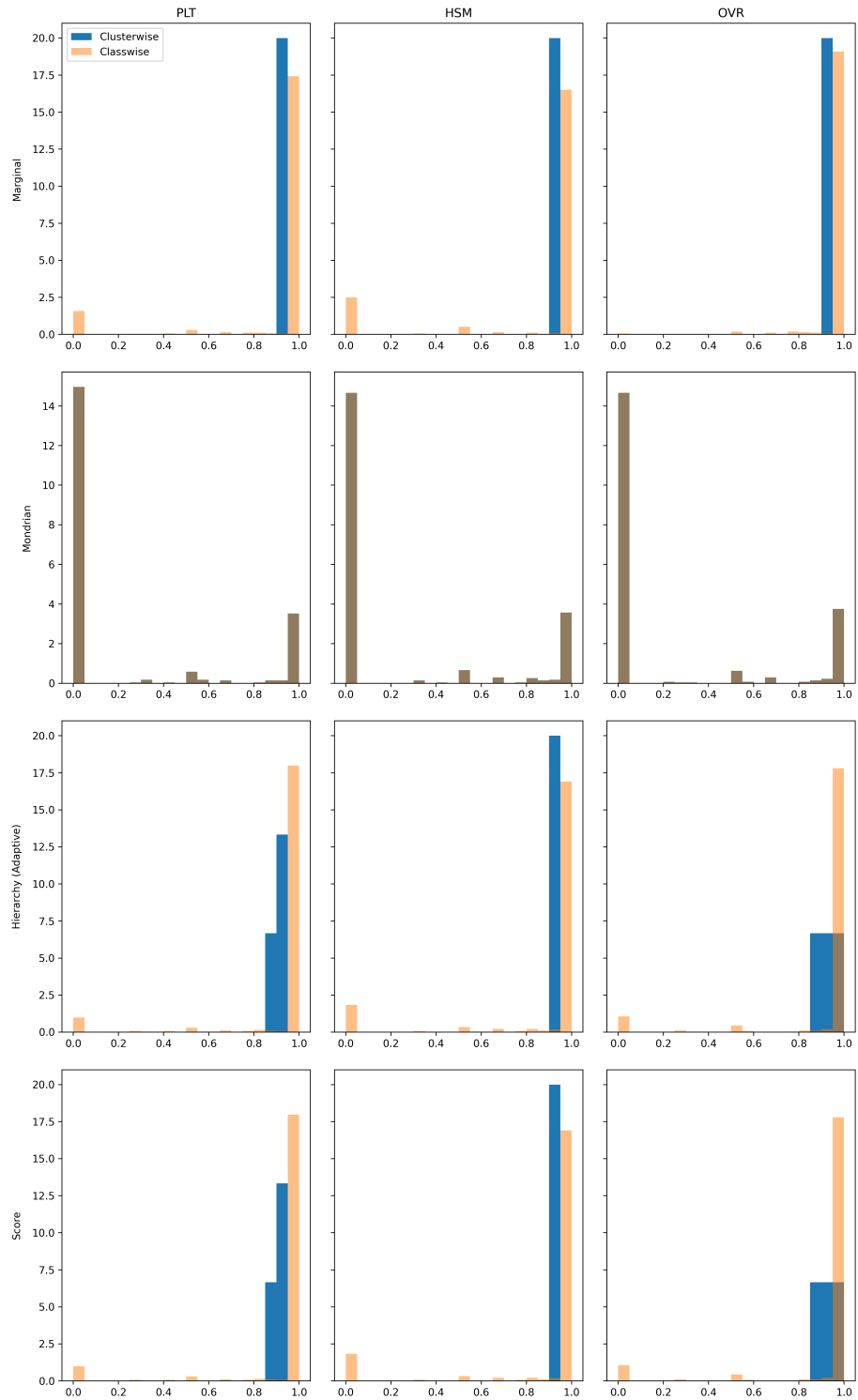

Figure B.7: Histogram of classwise and clusterwise coverage for the APS nonconformity measure for three models (PLT, HSM and OVR) and four clustering methods (marginal, Mondrian, hierarchy-based and score-based) on the Bacteria data set.

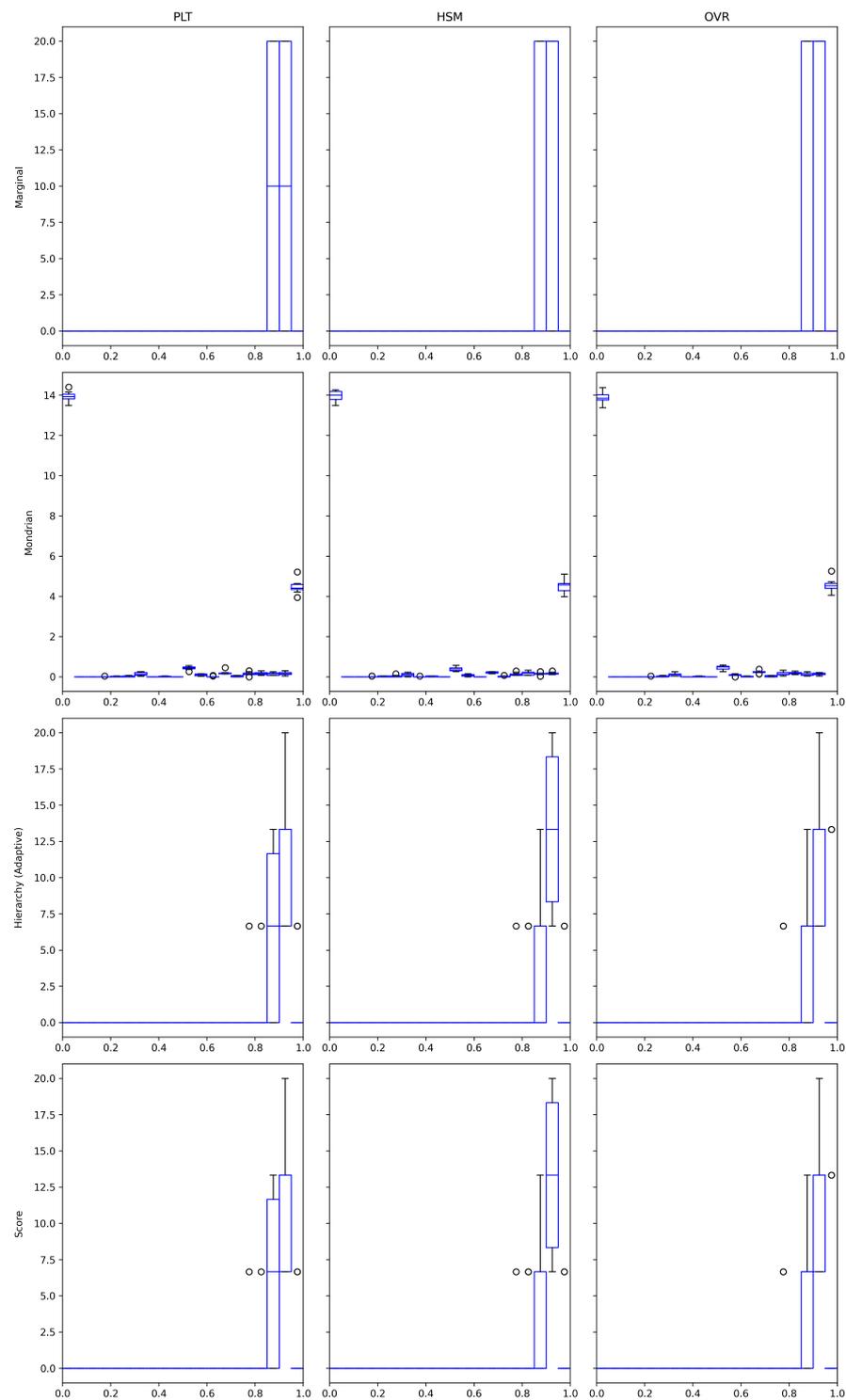

Figure B.8: Boxplot of clusterwise histogram densities for the softmax non-conformity measure for three models (PLT, HSM and OVR) and four clustering methods (marginal, Mondrian, hierarchy-based and score-based) over all calibration-test splits on the `Bacteria` data set.

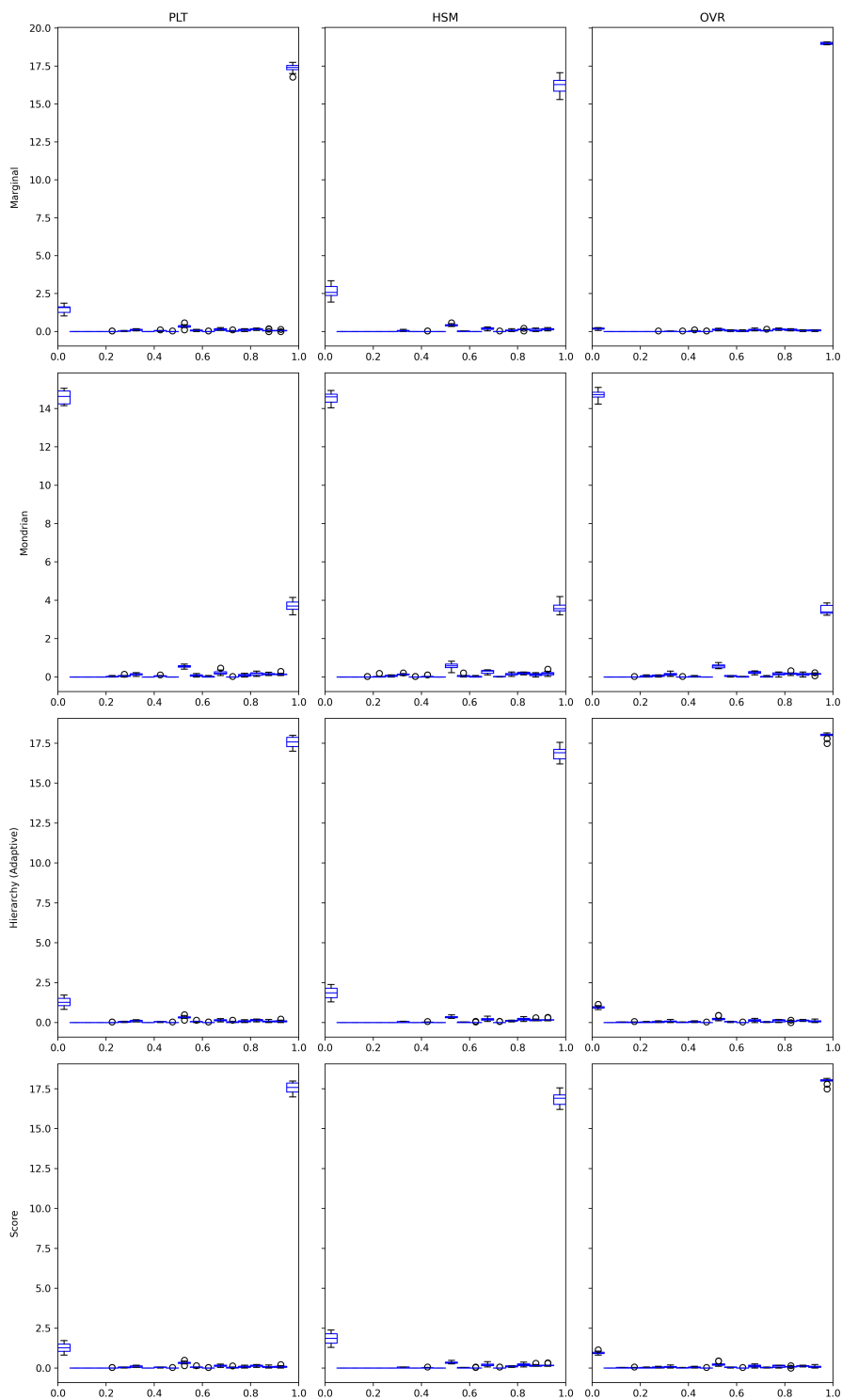

Figure B.9: Boxplot of classwise histogram densities for the softmax noncon-
formity measure for three models (PLT, HSM and OVR) and four clustering
methods (marginal, Mondrian, hierarchy-based and score-based) over all
calibration-test splits on the `Bacteria` data set.

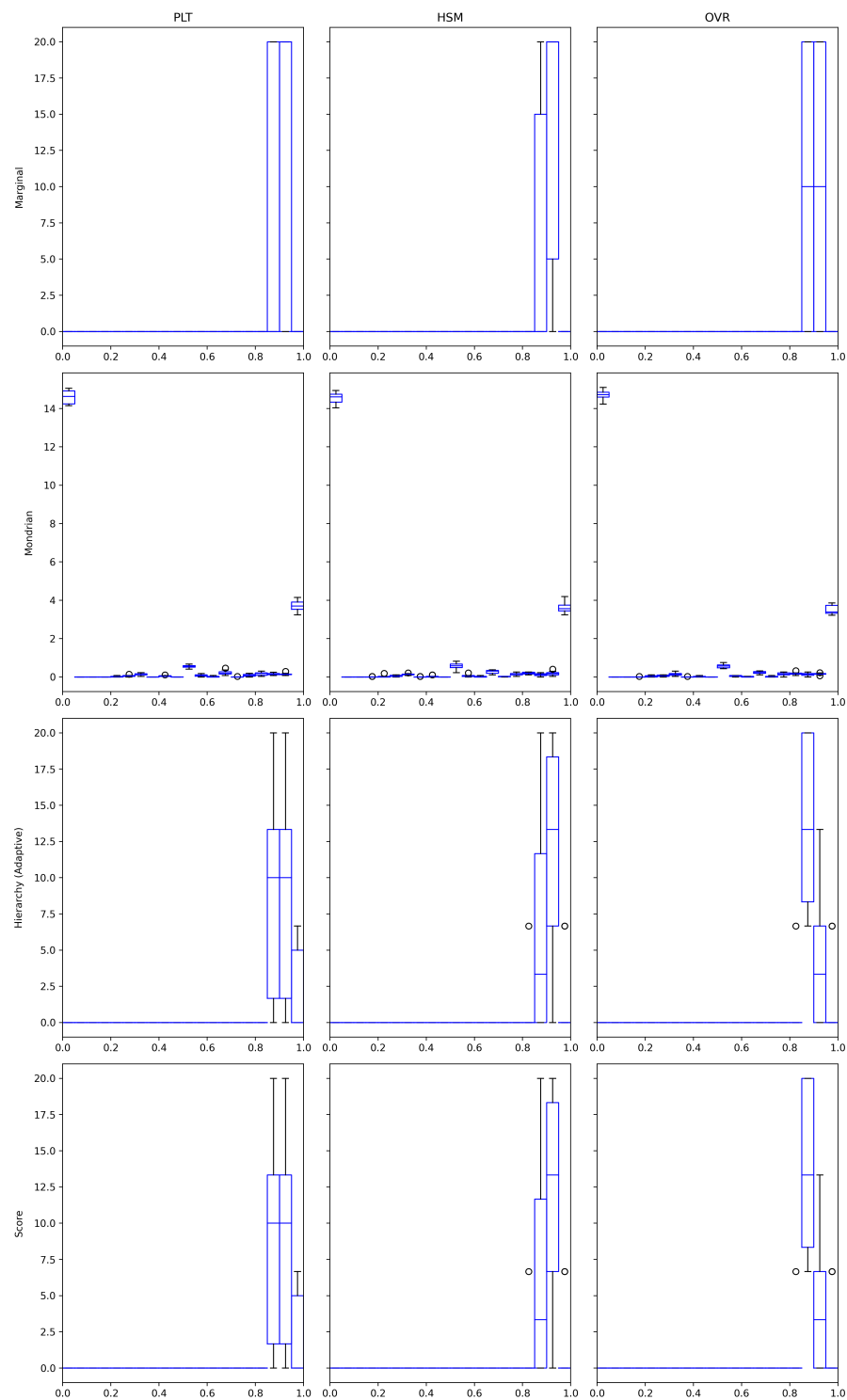

Figure B.10: Boxplot of clusterwise histogram densities for the APS noncon-formity measure for three models (PLT, HSM and OVR) and four clustering methods (marginal, Mondrian, hierarchy-based and score-based) over all calibration-test splits on the `Bacteria` data set.

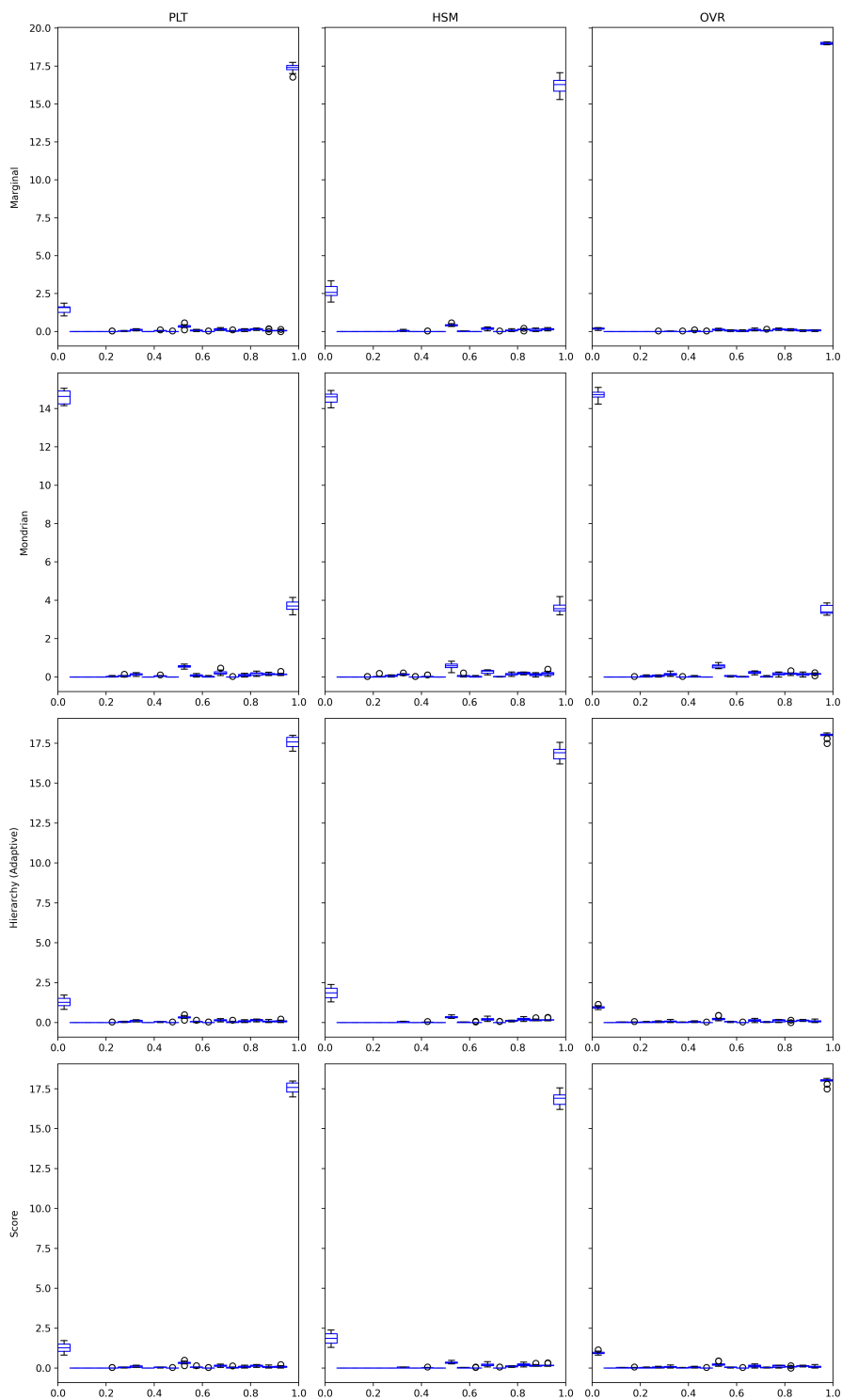

Figure B.11: Boxplot of classwise histogram densities for the softmax noncon-formity measure for three models (PLT, HSM and OVR) and four clustering methods (marginal, Mondrian, hierarchy-based and score-based) over all calibration-test splits on the `Bacteria` data set.

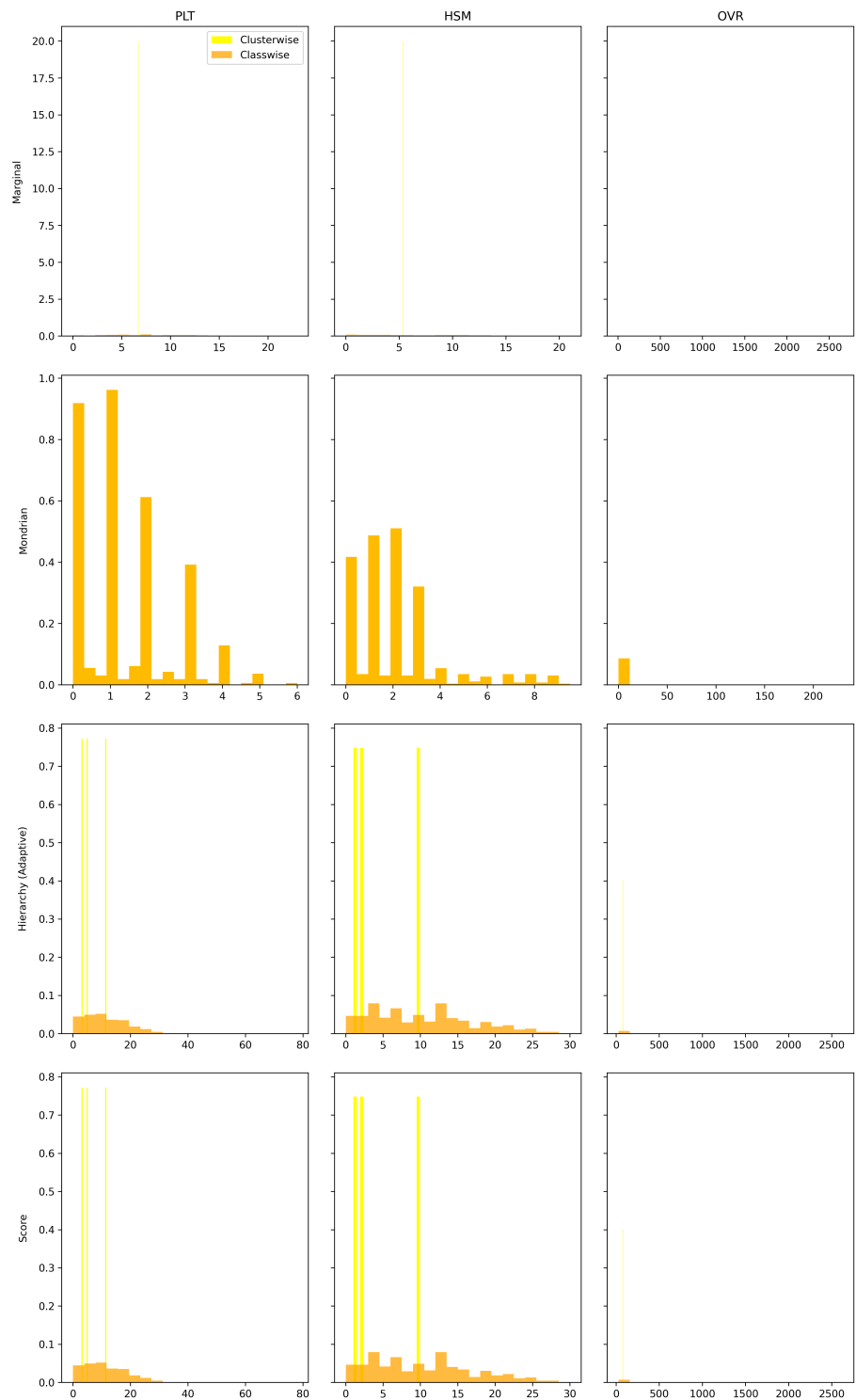

Figure B.12: Histogram of classwise and clusterwise average prediction set sizes for the softmax nonconformity measure for three models (PLT, HSM and OVR) and four clustering methods (marginal, Mondrian, hierarchy-based and score-based) on the `Bacteria` data set.

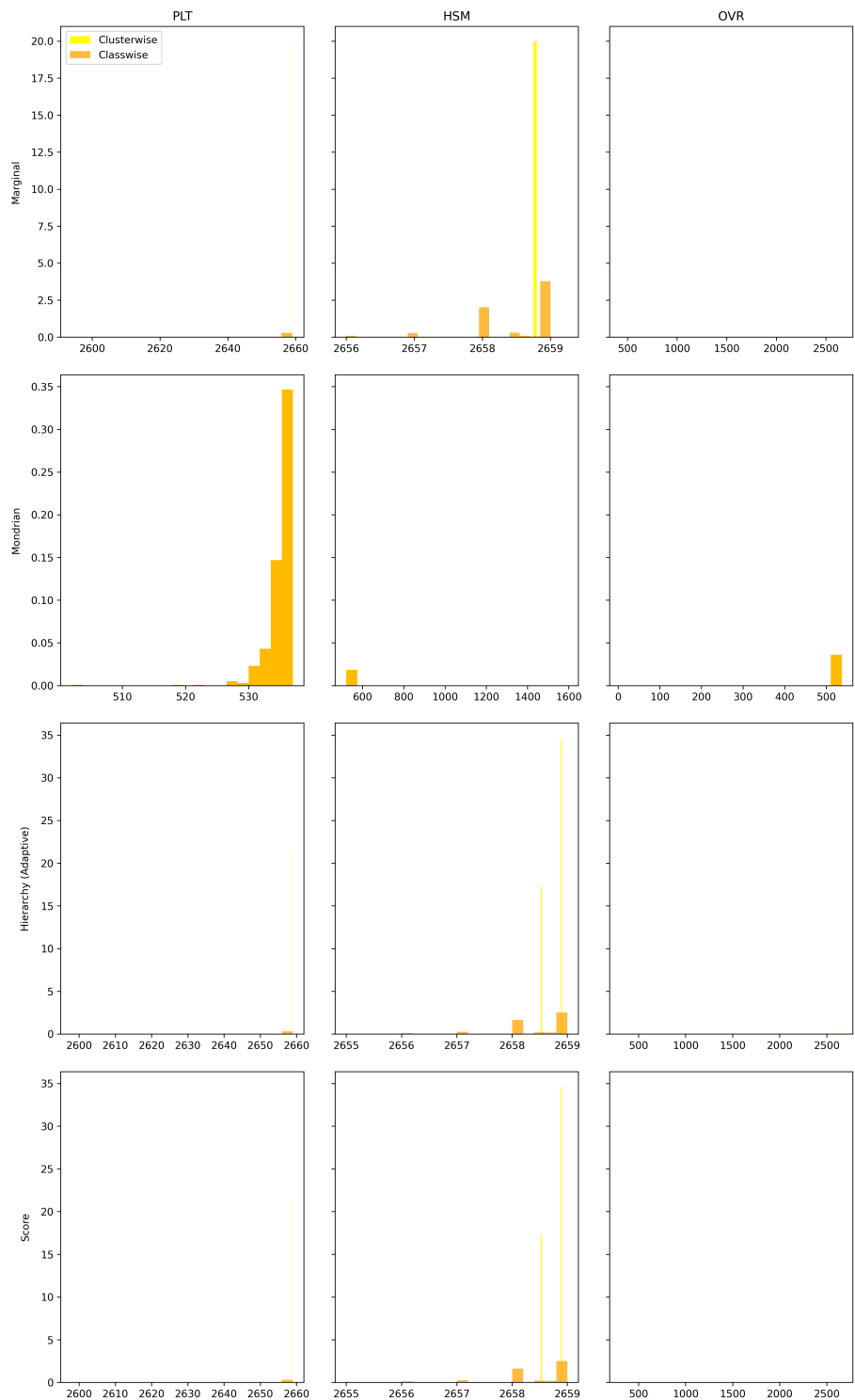

Figure B.13: Histogram of classwise and clusterwise average prediction set sizes for the APS nonconformity measure for three models (PLT, HSM and OVR) and four clustering methods (marginal, Mondrian, hierarchy-based and score-based) on the `Bacteria` data set.

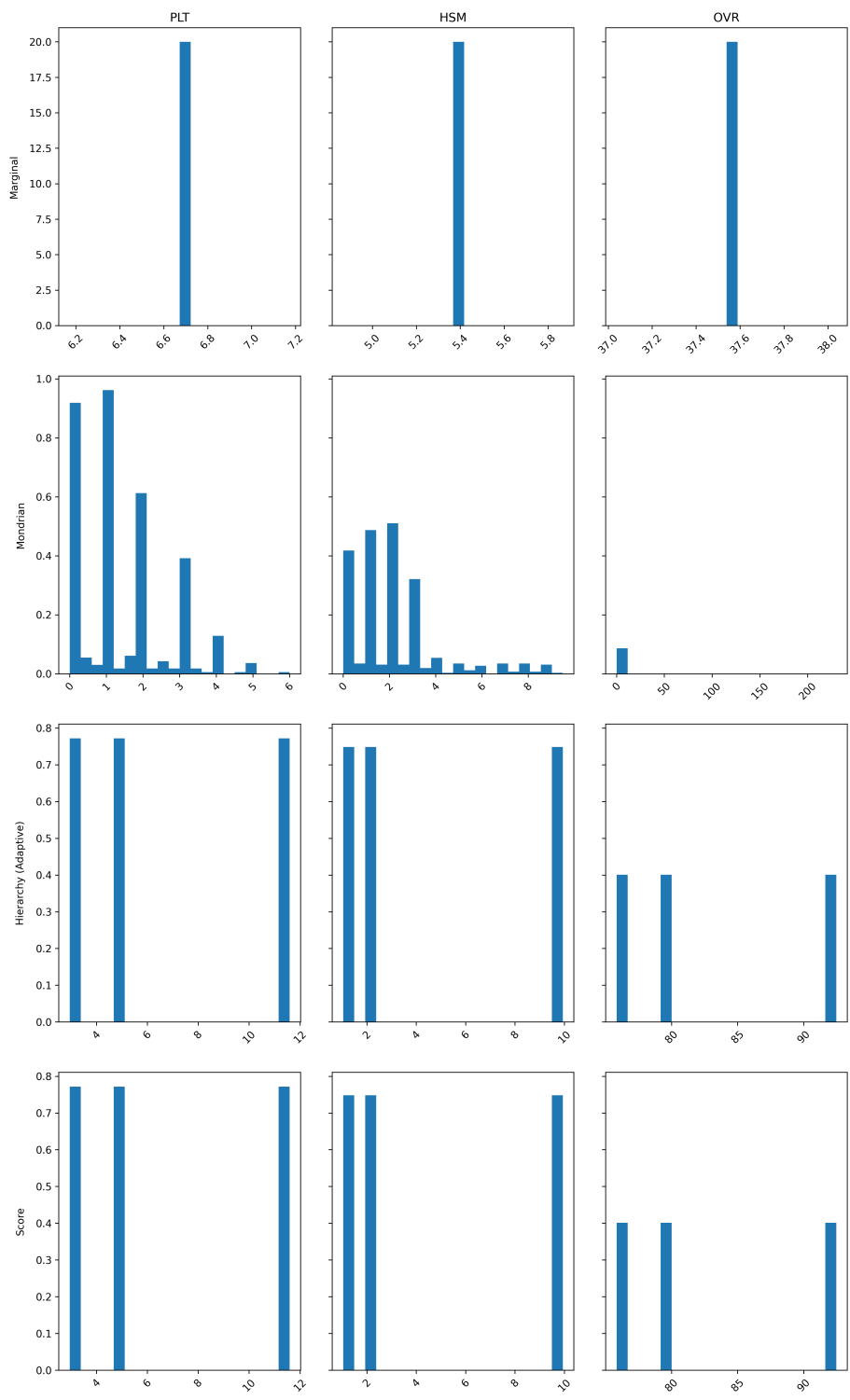

Figure B.14: Histogram of clusterwise representation complexities of the prediction sets for the softmax nonconformity measure for three models (PLT, HSM and OVR) and four clustering methods (marginal, Mondrian, hierarchy-based and score-based) on the `Bacteria` data set.

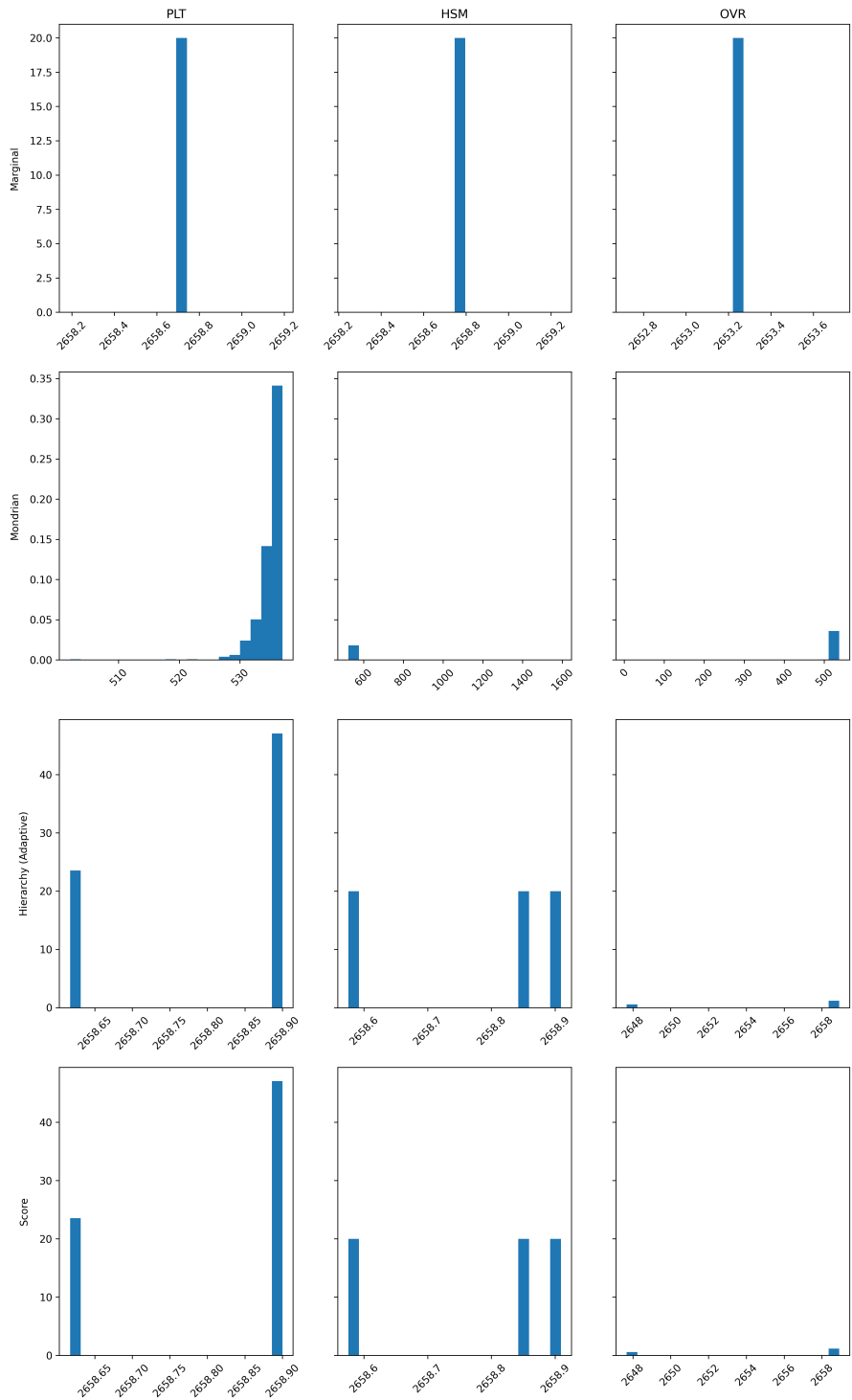

Figure B.15: Histogram of clusterwise representation complexities of the prediction sets for the APS nonconformity measure for three models (PLT, HSM and OVR) and four clustering methods (marginal, Mondrian, hierarchy-based and score-based) on the `Bacteria` data set.

# Artistic Explanations     C

In this appendix, the formulas appearing on the cover and at the beginning of every chapter are explained in some more detail. Each of the formulas on the cover are related to a subject that was highly interesting and intellectually motivating to me, either during my undergraduate, graduate or PhD years. Moreover, every chapter formula has a relation to the chapter in which it appears, even though that may not be entirely clear at first.

## C.1    Cover

The Mondrian-like sketch on the cover contains some formulas from physics or mathematics. To give some context, the motivation behind these formulas is presented below. The cover is also shown diagrammatically in Fig. C.1 for reference.

1. **Bayes' theorem**: This is probably the most well-known equation on the cover. It describes how conditional probabilities are related and is one of the most fundamental equations in statistics and probability theory.

2. **Bott periodicity**: A fundamental relation in the study of Clifford algebras, *operator algebras* and *vector bundles* through (complex) *K-theory*. Aside from a fundamental mathematical result, it also has important applications in physics. For example, it leads to a classification of *topological insulators* in condensed matter physics.

3. **Yang–Mills connection**: The Cartan structure equation gives an expression for the curvature of a *principal bundle* in terms of a chosen connection form. It also gives the Yang–Mills equations shown at the beginning of Chapter 4 when setting the curvature to zero.

4. **Einstein field equation**: The fundamental equation governing gravity and, being introduced in 1915, it is the oldest of the fundamental equations governing our universe. However, it still resists unification with the other equations.

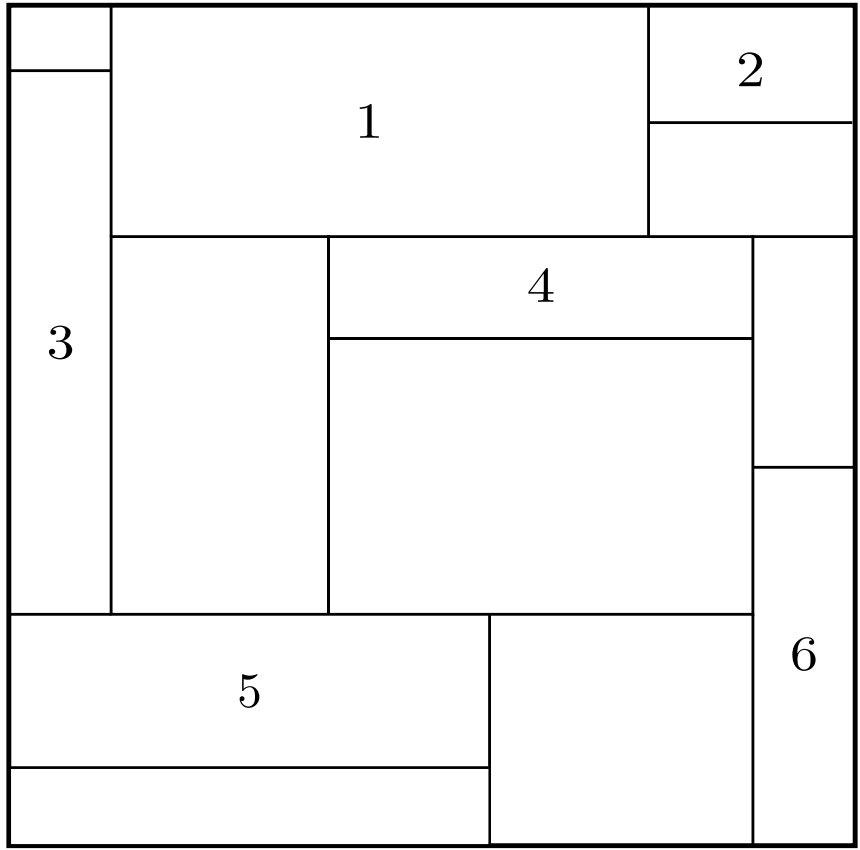

Figure C.1: Diagram of cover drawing.

5. **Validity guarantee**: Since the work in this dissertation focuses on conformal prediction, the main result of this framework ought to be present on the cover. This validity result (Theorem 2.27) says that the sets produced by conformal predictors are actually valid prediction sets in the statistical sense.

6. **Cobordism hypothesis**: To me, this is probably the most fascinating equation on the cover. This deep result in mathematics can be used to give a geometric formalization of *quantum field theory*.

## C.2    Uncertainty Estimation

In our daily lives, we are used to be able (at least in theory) to measure everything we want with arbitrary precision: the speed of a baseball, the height of the Eiffel tower, ... However, in quantum mechanics this is a whole other story. The intuitively mysterious ways of the quantum realm lead to the following equations:[1]

$$\Delta\hat{p}\,\Delta\hat{q} \geq \frac{\hbar}{2} \qquad\qquad \Delta\hat{A}\Delta\hat{B} \geq \left|\frac{1}{2}\langle[\hat{A},\hat{B}]\rangle\right|^2 \qquad\qquad \text{(C.1)}$$

Originally introduced by Heisenberg, and later refined by Robertson and Schrödinger (and others), these relations show that knowledge about non-commuting observables is not independent.

## C.3   Regression

Using the path integral formulation of quantum mechanics (and quantum field theory), expectations of random variables obtain a form that is very similar to the usual one in measure theory and probability theory. The measure over all possible realities or paths is given by a Gibbs-like measure $e^{iS[\phi]/\hbar}\,\mathcal{D}\phi$. The calligraphic $\mathcal{D}$ indicates that this measure should not be interpreted like the usual Lebesgue measure on $\mathbb{R}^n$. In fact, there still exists no formal definition for general physical systems.

## C.4   Conditional Validity

The Yang–Mills equation(s) are the equations of motion for the particles (or *fields*) that govern the forces of nature: the electromagnetic, weak and strong interactions. The relation to the chapter on conditional validity is given by the fact that these Yang–Mills fields only arise in the mathematical treatment of physics when we work in so-called 'local' coordinates, i.e. when we condition on the local environment so to speak.

## C.5   Clusterwise Validity

The *mirror symmetry conjecture* states that there exists a duality between two seemingly different objects. In its physical incarnation, it says that two specific types of *field theories* — the kind of objects that describe the standard model of particle physics — are actually two ways of looking at the same thing. In its mathematical incarnation, it says that two seemingly different

---

[1]   In fact, this is just linear algebra applied to the basic postulates of quantum mechanics.

*categories* of geometric data of a space are actually isomorphic in a precise sense. (The relation between these two points of view is given by the fact that the content, e.g. the objects, in such a field theory exactly constitute such a category.) The idea of duality, i.e. of looking at the same object from two different perspectives, is a central idea in both physics and mathematics and also features in Chapter 5. We were able to study the same problem from both the point of view of the features and that of the targets and we were able to cluster both in the score space and the feature space without losing any information (as long as the Lipschitz condition was satisfied).

# INDEX

# Nicolas Dewolf
*Curriculum Vitae*

✉ nicolas.dewolf@hotmail.com
in nicolas-dewolf-566924160
� nmdwolf

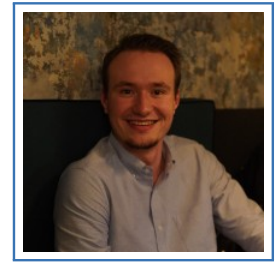

---

### ▬▬▬ Professional Interests

Working on the intersection of machine learning and statistics, with a strong focus on statistical guarantees for uncertainty modelling. Personally, I am interested in the interplay of probability theory, statistics and (the theoretical framework of) machine learning.

In line with my prior education, I am still interested in (theoretical) physics, in particular gauge theory and category theory, and all subjects related to this. More generally, I am also fascinated by the role these subjects play in other branches of science such as data science, biology, philosophy, etc.

### ▬▬▬ Education

2020 − 2024 **Ph.D. in Machine Learning − Universiteit Gent**, *Ghent (Belgium)*.
**Thesis title:** A comparative study of conformal prediction methods for valid uncertainty quantification in machine learning.
**Supervisors:** Willem Waegeman and Bernard De Baets.
**Status:** Ongoing.

2019 − 2020 **Ph.D. in Theoretical Physics − Universiteit Gent**, *Ghent (Belgium)*.
**Thesis title:** Classifying and implementing manifestly symmetric matrix product states.
**Supervisor:** Frank Verstraete.
**Status:** Terminated

2017 − 2019 **M.Sc. in Physics and Astronomy − Universiteit Gent**, *Ghent (Belgium)*.
**Thesis title:** Spatial symmetries and symmetry breaking with matrix product states.
**Supervisor:** Frank Verstraete
**Graduation grade:** Magna Cum Laude

**2014 – 2017**  **B.Sc. in Physics and Astronomy – Universiteit Gent**, *Ghent (Belgium)*.
**Graduation project:** The cosmophone: An auditory representation of muon Trajectories.
**Supervisor:** Dirk Ryckbosch
**Graduation grade:** Cum Laude

## Work Experience

**2024–**  **Senior Consultant**, *EY Belgium (FSO Risk)*.
Quantitative & Financial Risk

**2023**  **Co-organizer**, *WUML 2023*.
Organizing the "Workshop on Uncertainty in Machine Learning" in Ghent (`https://sites.google.com/view/wuml2023/home`).

**2022 – present**  **Webmaster & Co-founder**, *Tumaini Heart For Kids (VZW)*.
Responsible for social media, website and general organization.

**2020 – 2022**  **Teaching Assistant**, *UGent*.
Assisting with computer labs for the course "Probabilistische modellen" (Second Bachelor Bioscience Engineering).

**2019**  **Tutorial Assistant**, *TENSOR19*.
Assisting with tutorials at the "European Tensor School 2019" in San Sebastian (`http://tensor2019.dipc.org/`).

**2019**  **Teaching Assistant**, *UGent*.
Assisting with labs for the course "Experimenteren in de Fysica en de Sterrenkunde 1" (First Bachelor Physics & Astronomy).

**2016 – 2020**  **Tutor**, *BijlesHuis*.
Tutoring high school students in physics and mathematics (or related subjects).

**2016 – 2019**  **PR Extern & Webmaster**, *Vereniging voor Natuurkunde*.
Responsible for maintaining the website and social media, and finding funding for the student society of the Physics & Astronomy department.

**2017 & 2018**  **Administrative Student Job**, *Vinçotte*.
Maintaining and restructuring databases.

## Languages

Dutch  Native
English  Fluent
French  B2

Spanish    A2

# Publications

Journals
- ○ Dewolf, N., De Baets, B., & Waegeman, W. (2023). Valid prediction intervals for regression problems. *Artif Intell Rev*, 56(1), 577–613.

- ○ Vancraeynest-De Cuiper, B., Bridgeman, J. C., Dewolf, N., Haegeman, J., & Verstraete, F. (2023). One-dimensional symmetric phases protected by frieze symmetries. *Phys. Rev. B.*, 107, 115123.

Conferences
- ○ Nicolas Dewolf, Bernard De Baets, and Willem Waegeman. Conditional conformal prediction. Workshop on Uncertainty in Machine Learning. WUML, 2023.

- ○ Nicolas Dewolf, Bernard De Baets, and Willem Waegeman. Valid prediction intervals for regression problems. Joint International Scientific Conferences on AI and Machine Learning. BNAIC/BeNeLearn, 2022.

- ○ Nicolas Dewolf, Bernard De Baets, and Willem Waegeman. Regression problems in machine learning. Data Science, Statistics & Visualisation (DSSV) and the European Conference on Data Analysis (ECDA). DSSV-ECDA, 2021.

- ○ Nicolas Dewolf, Bernard De Baets, and Willem Waegeman. Calibrated prediction intervals. Workshop on Uncertainty in Machine Learning. WUML, 2021.